\documentclass[acmtog, authorversion, nonacm]{acmart}
\renewcommand\footnotetextcopyrightpermission[1]{} %
\acmJournal{FACMP}

\usepackage{booktabs} %

\usepackage{graphicx,dblfloatfix}
\usepackage[inkscapeformat=pdf]{svg} 
\usepackage{hhline}
\usepackage[ruled]{algorithm2e} %
\usepackage{subcaption}
\usepackage{pifont}
\usepackage{arydshln}
\usepackage{multirow}
\usepackage{multicol}
\usepackage{makecell}
\usepackage{xcolor}
\usepackage[export]{adjustbox}
\usepackage[normalem]{ulem}
\usepackage{xspace}
\usepackage{rotating}
\usepackage{xhfill}

\def\Bezier{B\'{e}zier\xspace}

\usepackage[ruled]{algorithm2e} %

\SetAlFnt{\small}
\SetAlCapFnt{\small}
\SetAlCapNameFnt{\small}
\SetAlCapHSkip{0pt}

\begin{document}

\title{Word-As-Image for Semantic Typography}

\begin{abstract}
A word-as-image is a semantic typography technique where a word illustration presents a visualization of the meaning of the word, while also preserving its readability.
We present a method to create word-as-image illustrations automatically. This task is highly challenging as it requires semantic understanding of the word and a creative idea of where and how to depict these semantics in a visually pleasing and legible manner. 
We rely on the remarkable ability of recent large pretrained language-vision models to distill textual concepts visually.
We target simple, concise, black-and-white designs that convey the semantics clearly. We deliberately do not change the color or texture of the letters and do not use embellishments. 
Our method optimizes the outline of each letter to convey the desired concept, guided by a pretrained Stable Diffusion model.
We incorporate additional loss terms to ensure the legibility of the text and the preservation of the style of the font.
We show high quality and engaging results on numerous examples and compare to alternative techniques.

Code will be available at \textcolor{magenta}{\href{http://WordAsImage.github.io/Word-As-Image-Page}{our project page.}}
\end{abstract}

\author{Shir Iluz}
\authornote{Denotes equal contribution.}
\affiliation{%
  \institution{Tel-Aviv University}
  \country{Israel}}
\author{Yael Vinker}
\authornotemark[1]
\affiliation{%
  \institution{Tel-Aviv University}
    \country{Israel}}
\author{Amir Hertz}
\affiliation{%
  \institution{Tel-Aviv University}
    \country{Israel}}
\author{Daniel Berio}
\affiliation{%
  \institution{Goldsmiths University}
    \country{London}}
\author{Daniel Cohen-Or}
\affiliation{%
  \institution{Tel-Aviv University}
    \country{Israel}}
\author{Ariel Shamir}
\affiliation{%
  \institution{Reichman University}
    \country{Israel}}

\begin{teaserfigure}
    \centering
    \includegraphics[width=0.95\linewidth]{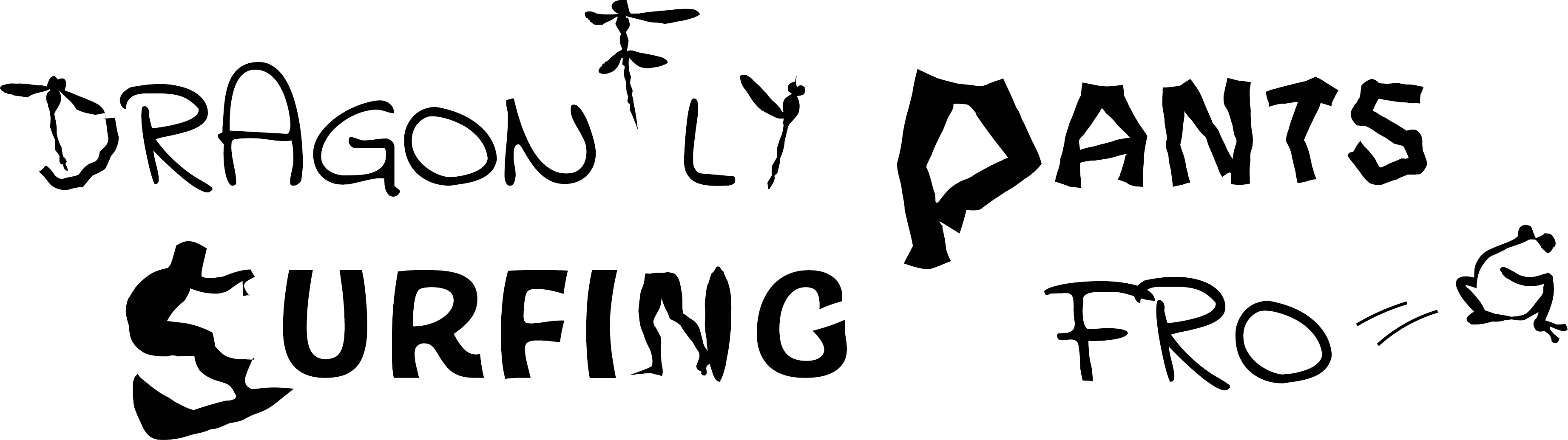} 
    \caption{A few examples of our word-as-image illustrations in various fonts and for different textual concept. The semantically adjusted letters are created completely automatically using our method, and can then be used for further creative design as we illustrate here.} \label{fig:teaser}
\end{teaserfigure}

\maketitle

\section{Introduction}
\label{sec:intro}
Semantic typography is the practice of using typography to visually reinforce the meaning of text. This can be achieved through the choice of typefaces, font sizes, font styles, and other typographic elements. A more elaborate and engaging technique for semantic typography is presented by word-as-image illustrations, where the semantics of a given word are illustrated using only the graphical elements of its letters. 
Such illustrations provide a visual representation of the meaning of the word, while also preserving the readability of the word as a whole.

\begin{figure}[t]
    \centering
    \includegraphics[width=\linewidth]{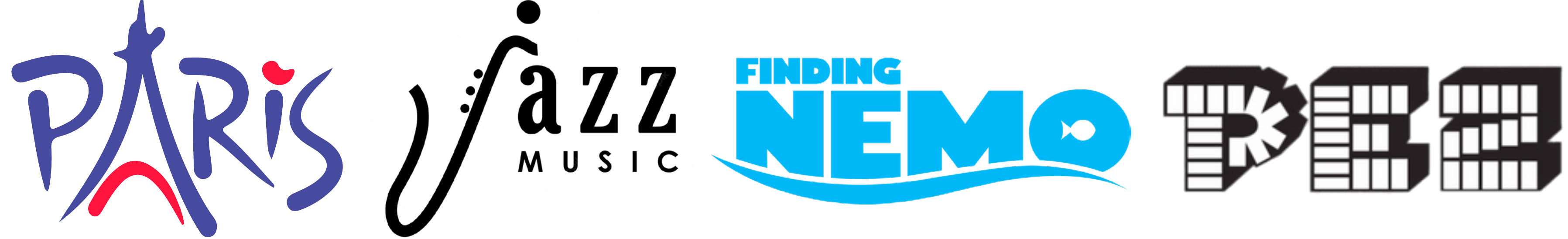}
    \caption{Manually created word-as-image illustrations.}
    \label{fig:logos_etc}
\end{figure}

The task of creating a word-as-image is highly challenging, as it requires the ability to understand and depict the visual characteristics of the given concept, and to convey them in a concise, aesthetic, and comprehensible manner without harming legibility. It requires a great deal of creativity and design skills to integrate the chosen visual concept into the letter's shape \cite{WordAsImage}. 
In Figure~\ref{fig:logos_etc} we show some word-as-image examples created manually. For example, to create the ``jazz'' depiction, the designer had to first choose the visual concept that would best fit the semantics of the text (a saxophone), consider the desired font characteristics, and then choose the most suitable letter to be replaced. Finding the right visual element to illustrate a concept is ill-defined as there are countless ways to illustrate any given concept. In addition, one cannot simply copy a selected visual element onto the word -- there is a need to find subtle modifications of the letters shape. 

Because of these complexities, the task of automatic creation of word-as-image illustrations was practically impossible to achieve using computers until recently.
In this paper, we define an algorithm for automatic creation of word-as-image illustrations based on recent advances in deep-learning and the availability of huge foundational models that combine language and visual understanding.
Our resulting illustrations (see Figure~\ref{fig:teaser}) could be used for logo design, for signs, in greeting cards and invitations, and simply for fun. They can be used as-is, or as inspiration for further refinement of the design.

Existing methods in the field of text stylization often rely on raster textures \cite{Yang_2018_Context}, place a manually created style on top of the strokes segmentation \cite{BerioStrokestyles2022}, or deform the text into a pre-defined target shape \cite{zouLegibleCompactCalligrams2016} (see Figure~\ref{fig:prev_work}). 
Only a few works \cite{tendulkarTrickTReATThematic2019, zhangSynthesizingOrnamentalTypefaces2017} deal with \textit{semantic} typography, and they often operate in the raster domain and use existing icons for replacement (see Figure~\ref{fig:prev_work}E).

Our word-as-image illustrations concentrate on changing only the \emph{geometry} of the letters to convey the meaning. We deliberately do not change color or texture and do not use embellishments. This allows simple, concise, black-and-white designs that convey the semantics clearly. In addition, since we preserve the vector-based representation of the letters, this allows smooth rasterization in any size, as well as applying additional style manipulations to the illustration using colors and texture, if desired.

Given an input word, our method is applied separately for each letter, allowing the user to later choose the most likeable combination for replacement.
We represent each letter as a closed vectorized shape, and optimize its parameters to reflect the \emph{meaning} of the word, while still preserving its original style and design.

We rely on the prior of a pretrained Stable Diffusion model ~\cite{stableDiffusion} to connect between text and images, and utilize the Score Distillation Sampling approach \cite{poole2022dreamfusion} (see Section~\ref{sec:background}) to encourage the appearance of the letter to reflect the provided textual concept. 
Since the Stable Diffusion model is trained on raster images, we use a differentiable rasterizer \cite{diffvg} that allows to backpropagate gradients from a raster-based loss to the shape's parameters.

To preserve the shape of the original letter and ensure legibility of the word, we utilize two additional loss functions. The first loss regulates the shape modification by constraining the deformation to be as-conformal-as-possible over a triangulation of the letter's shape. The second loss preserves the local tone and structure of the letter by comparing the low-pass filter of the resulting rasterized letter to the original one. 

We compare to several baselines, and present many results using various typefaces and a large number of concepts. Our word-as-image illustrations convey the intended concept while maintaining legibility and preserving the appearance of the font, demonstrating visual creativity.

\section{Related Work}
\label{sec:related}

\begin{figure}[t]
    \centering
    \includegraphics[width=1\linewidth]{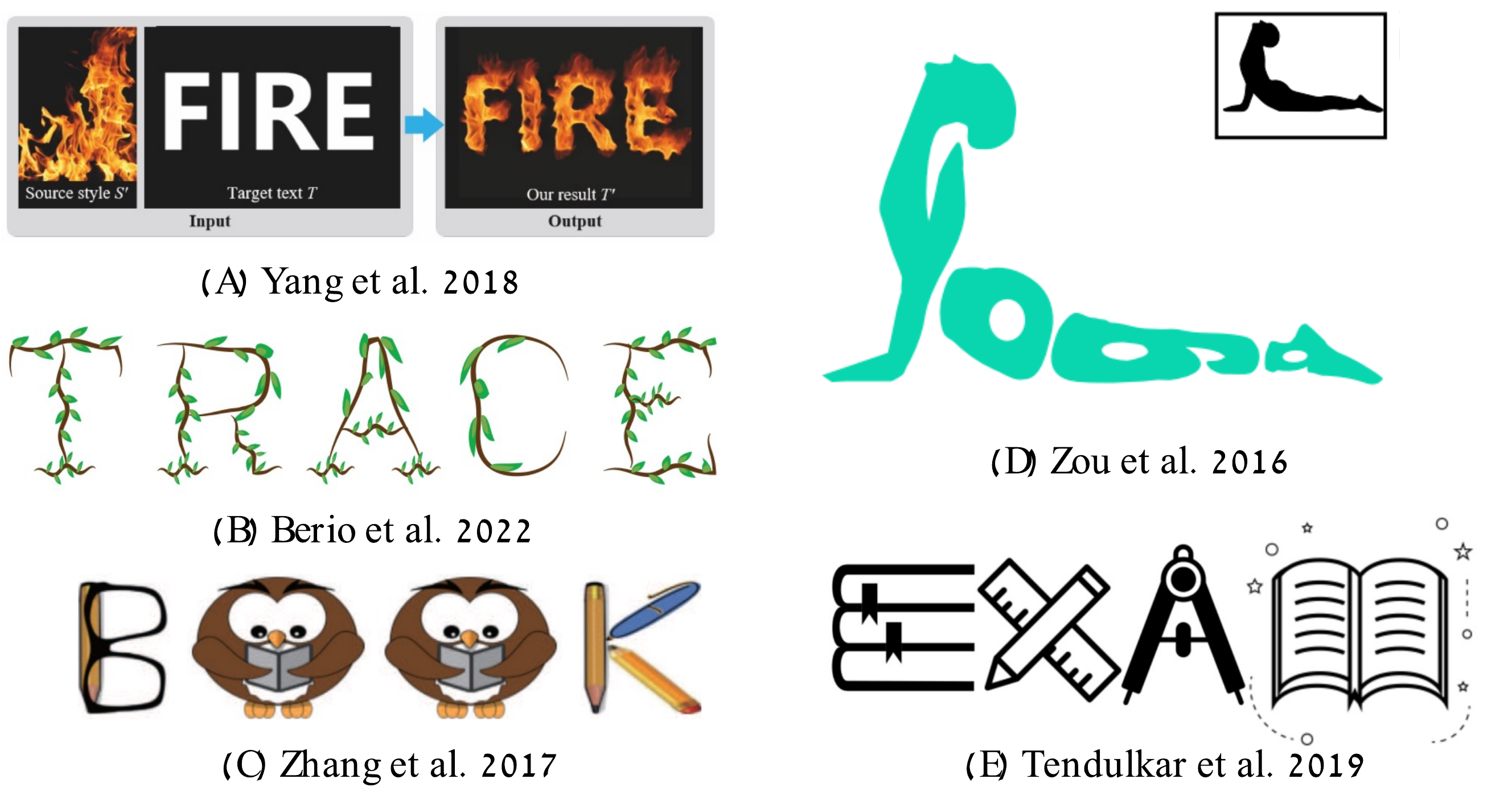}
    \caption{Examples of previous text stylization works -- (A) Yang et al. \shortcite{Yang_2018_Context}, (B) Berio et al. \shortcite{BerioStrokestyles2022}, (C) Zhang et al. \shortcite{zhangSynthesizingOrnamentalTypefaces2017}, (D) Zou et al. \shortcite{zouLegibleCompactCalligrams2016}, and (E) Tendulkar et al. \shortcite{tendulkarTrickTReATThematic2019}. Most use color and texture or copy icons onto the letters. Our work concentrates on subtle \emph{geometric} shape deformations of the letters to convey the \emph{semantic meaning} without color or texture (that can be added later).}
    \vspace{-0.2cm}
    \label{fig:prev_work}
\end{figure}

\paragraph{Text Stylization}
One approach to text stylization is artistic text style transfer, where the style from a given source image is migrated into the desired text (such as in Figure \ref{fig:prev_work}A).
To tackle this task, existing works incorporate patch-based texture synthesis \cite{Yang_2017_CVPR, SketchPatch2020} as well as variants of GANs \cite{Azadi_2018_CVPR, Wang_2019_CVPR, Jiang_SC_2019, YangGAN2022, Mao_2022_Intelligent}.
These works operate within the raster domain, a format that is undesirable for typographers since fonts must be scalable. In contrast, we operate on the \textit{parametric} outlines of the letters, and our glyph manipulation is guided by the semantic meaning of the word, rather than a pre-defined style image.

A number of works \cite{ha2017neural,Lopes_2019_ICCV,wangDeepVecFontSynthesizingHighquality2021} tackle the task of font generation and stylization in the vector domain. 
Commonly, a latent feature space of font's outlines is constructed, represented as outline samples \cite{Campbell2014, Balashova2018} or parametric curve segments \cite{ha2017neural,Lopes_2019_ICCV,wangDeepVecFontSynthesizingHighquality2021}. These approaches are often limited to mild deviations from the input data. Other methods rely on
templates \cite{Suveeranont2010, lian2018easyfont} or on user guided \cite{Phan2015} and automatic \cite{BerioStrokestyles2022} stroke segmentation to produce letter stylization (such as in Figure \ref{fig:prev_work}B). However, they rely on a manually defined style, while we rely on the expressiveness of Stable Diffusion to guide the modification of the letters' shape, to convey the \textit{meaning} of the provided word.
In the task of calligram generation \cite{zouLegibleCompactCalligrams2016,xuCalligraphicPacking2007} the entire word is deformed into a \textit{given} target shape. 
This task prioritises shape over the readability of the word (see Figure \ref{fig:prev_work}D), and is inherently different from ours, as we use the \textit{semantics} of the word to derive the deformation of individual letters.

Most related to our goal, are works that perform semantic stylization of text. \citet{tendulkarTrickTReATThematic2019} replace letters in a given word with clip-art icons describing a given theme (see Figure \ref{fig:prev_work}E). To choose the most suitable icon for replacement, an autoencoder is used to measure the distance between the letter and icons from the desired class. 
Similarly, \citet{zhangSynthesizingOrnamentalTypefaces2017} replace stroke-like parts of one or more letters with instances of clip art to generate ornamental stylizations. An example is shown in Figure \ref{fig:prev_work}C.
These approaches operate in the raster domain, and replace letters with existing icons, which limits them to a predefined set of classes present in the dataset. Our method, however, operates in the vector domain, and incorporates the expressiveness of large pretrained image-language models to create a new illustration that conveys the desired concept.

\paragraph{Large Language-Vision Models}
With the recent advancement of language-vision models \cite{clip} and diffusion models ~\cite{ramesh2022hierarchical, nichol2021glide, stableDiffusion}, the field of image generation and editing has undergone unprecedented evolution. 
Having been trained on millions of images and text pairs, these models have proven effective for performing challenging vision related tasks such as image segmentation \cite{SegDiff}, domain adaptation \cite{DomainAdaptationSong}, image editing \cite{Avrahami_2022_CVPR, hertz2022prompt, Plug-and-Play}, personalization \cite{ruiz2022dreambooth, PersonalizationGal23, gal2022textual}, and explainability \cite{Chefer_2021_CVPR}.
Despite being trained on raster images, their strong visual and semantic priors have also been shown to be successfully applied to other domains, such as motion \cite{tevet2022motionclip}, meshes \cite{text2mesh}, point cloud \cite{PointCLIP}, and vector graphics. 
CLIPDraw \cite{frans2021clipdraw} uses a differentiable rasterizer \cite{diffvg} to optimize a set of colorful curves w.r.t. a given text prompt, guided by CLIP's image-text similarity metric.
Tian and Ha \shortcite{tian2021modern} use evolutionary algorithms combined with CLIP guidance to create abstract visual concepts based on text. Other works \cite{vinker2022clipasso, clipascene} utilize the image encoder of CLIP to generate abstract vector sketches from images. 

Diffusion models have been used for the task of text guided image-to-image translation \cite{ILVR, plugplay}.
In SDEdit \cite{meng2022sdedit}, an adequate amount of noise is added to a reference image, such that its overall structure is preserved, and then the image is denoised in a reverse process with a guiding text. 
Pretrained diffusion models have also been used to generate 3D objects \cite{poole2022dreamfusion,Latent-NeRF}, or vector art \cite{jain2022vectorfusion} conditioned on text.

In our work we also utilize the strong visual and semantic prior induced by a pretrained Stable Diffusion model \cite{stableDiffusion}, however, for the task of \textit{semantic typography}. For that purpose we add new components to the optimization process to preserve the font's style and text legibility.

\begin{figure}[t]
\centering
\includegraphics[width=0.95\linewidth]{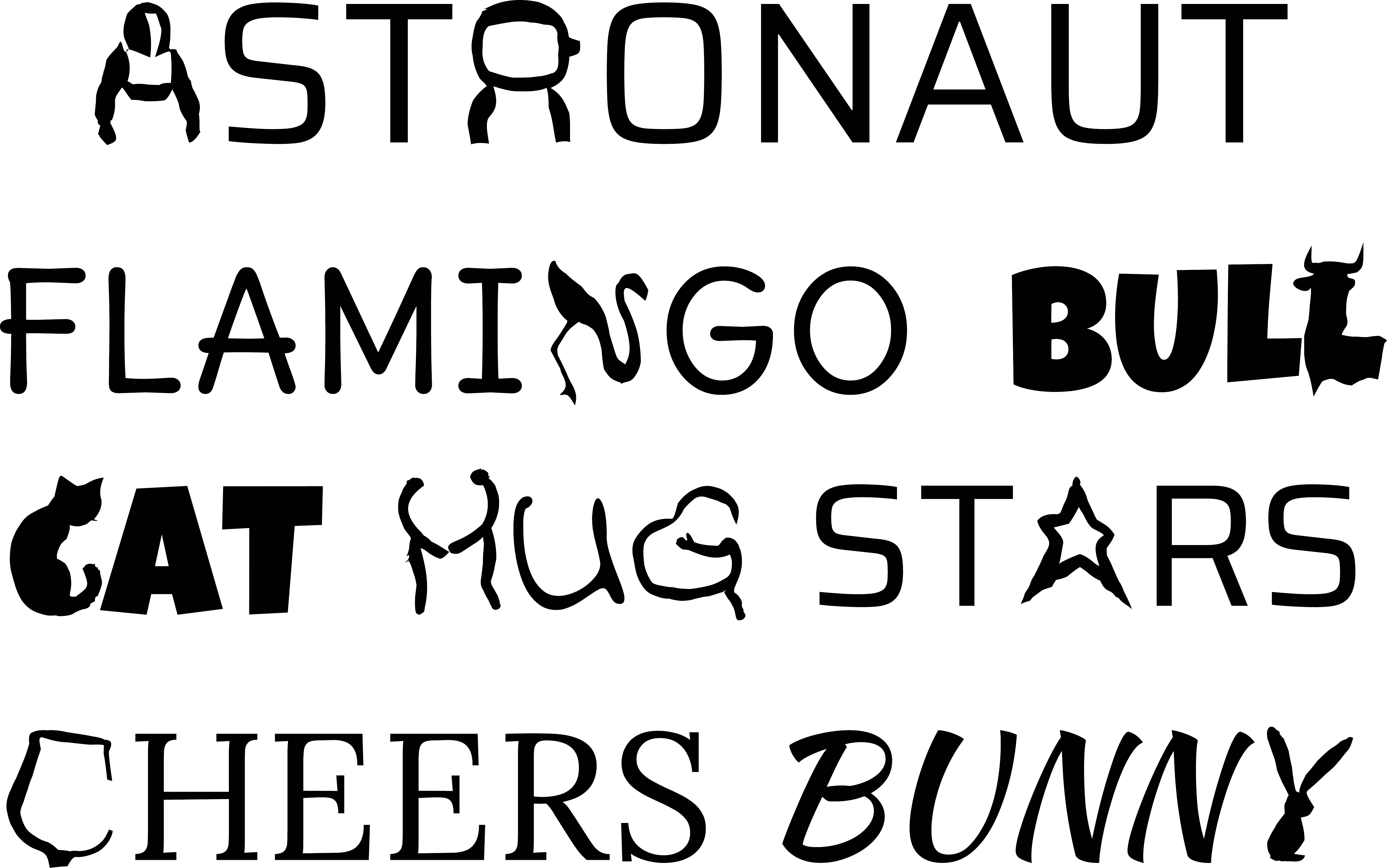}
    \caption{More word-as-images produced by our method. Note how styles of different fonts are preserved by the semantic modification.}
    \label{fig:intro_res}
\end{figure}

\begin{figure*}
    \centering
    \includegraphics[width=0.95\linewidth]{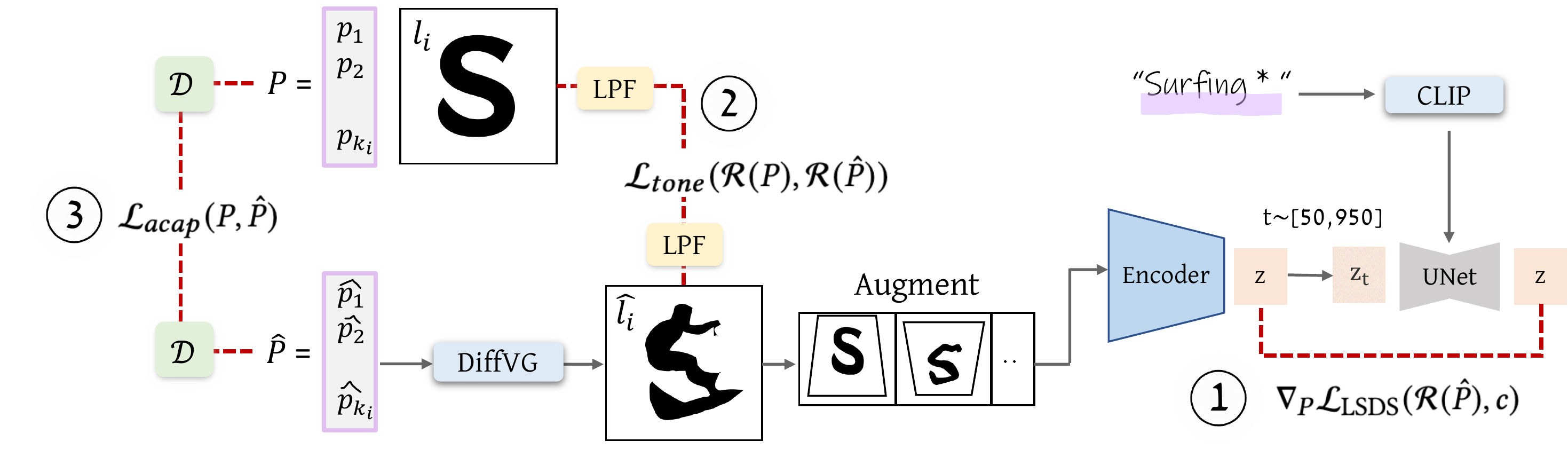} 
    \vspace{-0.2cm}
    \caption{An overview of our method. Given an input letter $l_i$ represented by a set of control points $P$, and a concept (shown in purple), we optimize the new positions $\hat{P}$ of the deformed letter $\hat{l_i}$ iteratively. At each iteration, the set $\hat{P}$ is fed into a differentiable rasterizer (DiffVG marked in blue) that outputs the rasterized deformed letter $\hat{l_i}$. $\hat{l_i}$ is then augmented and passed into a pretrained frozen Stable Diffusion model, that drives the letter shape to convey the semantic concept using the $\nabla_{\hat{P}} \mathcal{L}_\text{LSDS}$ loss (1). $l_i$ and $\hat{l_i}$ are also passed through a low pass filter (LPF marked in yellow) to compute $\mathcal{L}_{tone}$ (2) which encourages the preservation of the overall tone of the font style and also the local letter shape. Additionally, the sets $P$ and $\hat{P}$ are passed through a  Delaunay triangulation operator ($\mathcal{D}$ marked in green), defining $\mathcal{L}_{acap}$ (3) which encourages the preservation of the initial shape.}
    \label{fig:method}
\end{figure*}

\section{Background}
\label{sec:background}

\subsection{Fonts and Vector Representation}
Modern typeface formats such as TrueType~\cite{TrueTypeHist} and PostScript~\cite{AdobeType1} represent glyphs using a vectorized graphic representation of their outlines. Specifically, the outline contours are typically represented by a collection of lines and \Bezier or B-Spline curves.
This representation allows to scale the letters and rasterize them in any desired size similar to other vector representations. This property is preserved by our method as our output preserves the vectorized representations of the letters. 

\subsection{Latent Diffusion Models}
Diffusion models are generative models that are trained to learn a data distribution by the gradual denoising of a variable sampled from a Gaussian distribution.

In our work, we use the publicly available text-to-image Stable Diffusion model \cite{stableDiffusion}. 
Stable Diffusion is a type of a latent diffusion model (LDM), where the diffusion process is done over the latent space of a pretrained image autoencoder. The encoder $\mathcal{E}$ is tasked with mapping an input image $x$ into a latent vector $z$, and the decoder $\mathcal{D}$ is trained to decode $z$ such that $\mathcal{D}(z)\approx x$.

As a second stage, a denoising diffusion probabilistic model (DDPM) \cite{ddpm} is trained to generate codes within the learned latent space.
At each step during training, a scalar $t\in\{1,2,...T\}$ is uniformly sampled and used to define a noised latent code $z_t = \alpha_t z + \sigma_t \epsilon$, where $\epsilon \sim \mathcal{N}(0, I)$ and $\alpha_t, \sigma_t$ are terms that control the noise schedule, and are functions of the diffusion process time $t$.

The denoising network $\epsilon_{\theta}$ which is based on a UNet architecture \cite{unet2015}, receives as input the noised code $z_t$, the timestep $t$ and an optional condition vector $c(y)$, and is tasked with predicting the added noise $\epsilon$.
The LDM loss is defined by:
\begin{equation}
    \mathcal{L}_{LDM} = \mathbb{E}_{z\sim\mathcal{E}(x),y,\epsilon\sim\mathcal{N}(0,1),t} \left [ || \epsilon - \epsilon_\theta(z_t, t, c(y)) ||_2^2 \right ].
\end{equation}

In Stable Diffusion, for text-to-image generation, the condition vector is the text embedding produced by a pre-trained CLIP text encoder~\cite{clip}.
At inference time, a random latent code $z_T \sim \mathcal{N}(0, I)$ is sampled, and iteratively denoised by the trained $\epsilon_{\theta}$ until producing a clean $z_0$ latent code, which is passed through the decoder $D$ to produce the image $x$. 

\subsection{Score Distillation}
\label{subsec:ScoreDistillation}
It is desirable to utilize the strong prior of pretrained large text-image models for the generation of modalities beyond rasterized images.
In Stable Diffusion, text conditioning is performed via the cross-attention layers defined at different resolutions in the UNet network.
Thus, it is not trivial to guide an optimization process using the conditioned diffusion model. 

DreamFusion~\cite{poole2022dreamfusion} proposed a way to use the diffusion loss to optimize the parameters of a NeRF model for text-to-3D generation.
At each iteration, the radiance field is rendered from a random angle, forming the image $x$, which is then noised to form $x_t = \alpha_t x + \sigma_t \epsilon$.
The noised image is then passed to the pretrained UNet model of Imagen \cite{imagen}, that outputs the prediction of the noise $\epsilon$.
The score distillation loss is defined by the gradients of the original diffusion loss:
\begin{equation}
    \nabla_\phi \mathcal{L}_{SDS} = \left[ w(t)(\epsilon_\theta(x_t,t,y) - \epsilon) \frac{\partial x}{\partial \phi} \right]
\end{equation}
where $y$ is the condition text prompt, $\phi$ are the NeRF's parameters and $w(t)$ is a constant multiplier that depends on $\alpha_t$.
During training, the gradients are back-propagated to the NeRF parameters to gradually change the 3D object to fit the text prompt. 
Note that the gradients of the UNet are skipped, and the gradients to modify the Nerf's parameters are derived directly from the LDM loss.

\subsection{VectorFusion}
\label{subsec:vectorFusion}
Recently, VectorFusion \cite{jain2022vectorfusion} utilized the SDS loss for the task of text-to-SVG generation. The proposed generation pipeline involves two stages. Given a text prompt, first, an image is generated using Stable Diffusion (with an added suffix to the prompt), and is then vectorized automatically using LIVE \cite{live}.
This defines an initial set of parameters to be optimized in the second stage using the SDS loss.
At each iteration, a differentiable rasterizer \cite{diffvg} is used to produce a $600 \times 600$ image, which is then augmented as suggested in CLIPDraw \cite{frans2021clipdraw} to get a $512 \times 512$ image $x_{aug}$. 
Then $x_{aug}$ is fed into the pretrained encoder $\mathcal{E}$ of Stable Diffusion to produce the corresponding latent code $z=\mathcal{E}(x_{aug})$.
The SDS loss is then applied in this latent space, in a similar way to the one defined in DreamFusion:

\begin{equation}
    \nabla_\theta \mathcal{L}_\text{LSDS} = 
    \quad\mathbb{E}_{t, \epsilon} \left[ w(t) \Big(\hat{\epsilon}_\phi(\alpha_t z_t + \sigma_t \epsilon, y)  - \epsilon\Big) \frac{\partial z}{\partial z_{aug}} \frac{\partial x_{aug}}{\partial \theta} \right]   
\end{equation}

We find the SDS approach useful for our task of producing semantic glyphs, and we follow the technical steps proposed in VectorFusion (e.g. augmentations and the added suffix).

\section{Method}
\label{sec:method}
Given a word $W$ represented as a string with $n$ letters $\{l_1, ... l_n\}$, our method is applied to every letter $l_i$ separately to produce a semantic visual depiction of the letter. The user can then choose which letters to replace and which to keep in their original form.

\subsection{Letter Representation}
\label{subsec:letter_representation}
We begin by defining the parametric representation of the letters in $W$.
We use the FreeType font library \cite{freetype} to extract the outline of each letter. We then translate each outline into a set of cubic \Bezier curves, to have a consistent representation across different fonts and letters, and to facilitate the use of diffvg \cite{diffvg} for differentiable rasterization. 

\begin{figure}[t]
    \centering
    \includegraphics[width=0.95\linewidth]{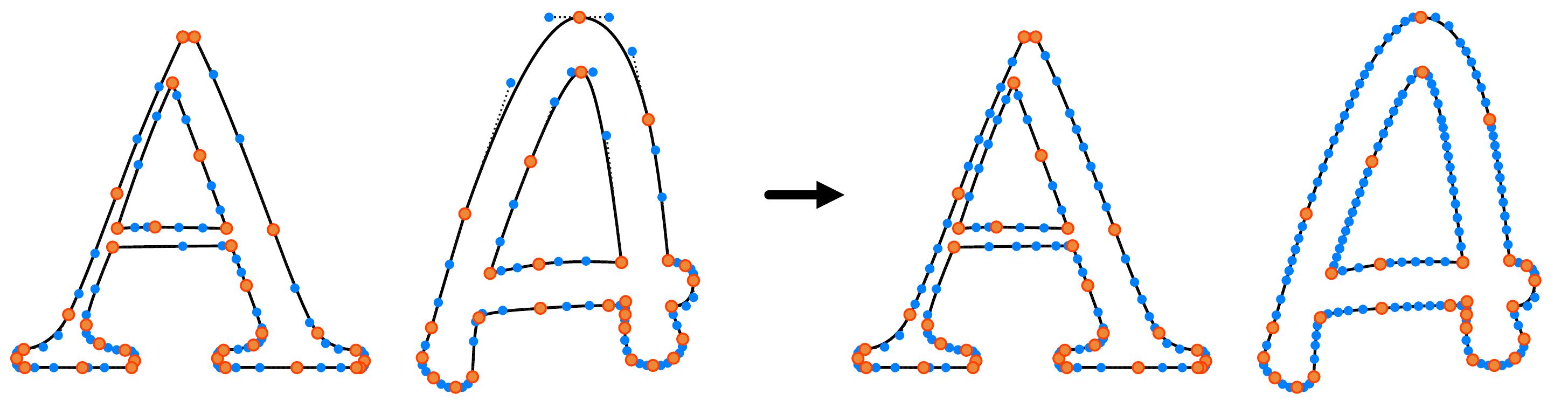}
    \caption{Illustration of the letter's outline and control points before (left) and after (right) the subdivision process. The orange dots are the initial \Bezier curve segment endpoints. The blue dots are the remaining control points respectively before and after subdivision. }
    \label{fig:initCC}
\end{figure}

Depending on the letter's complexity and the style of the font, the extracted outlines are defined by a different number of control points.
We have found that the initial number of control points affects the final appearance significantly: as the number of control points increases, there is more freedom for visual changes to occur.
Therefore, we additionally apply a subdivision procedure to letters containing a small number of control points. 
We define a desired number of control points for each letter of the alphabet (shared across different fonts), and then iteratively subdivide the \Bezier segments until reaching this target number. 
At each iteration, we compute the maximum arc length among all \Bezier segments and split each segment with this length into two (see Figure~\ref{fig:initCC}). We analyse the effect of the number of control points in Section \ref{sec:ablation}.

This procedure defines a set of $k_i$ control points $P_i = \{p_j\}_{j=1}^{k_i}$ representing the shape of the letter $l_i$.

\subsection{Optimization}
The pipeline of our method is provided in Figure \ref{fig:method}. Since we are optimizing each letter $l_i$ separately, for brevity, we will omit the letter index $i$ in the following text and define the set of control points for the input letter as $P$. 

Given $P$ and the desired textual concept $c$ (both marked in purple in Figure \ref{fig:method}), our goal is to produce a new set of control points, $\hat{P}$, defining an adjusted letter $\hat{l}$ that conveys the given concept, while maintaining the overall structure and characteristics of the initial letter $l$.

We initialize the learned set of control points $\hat{P}$ with $P$, and pass it through a differentiable rasterizer $\mathcal{R}$ \cite{diffvg} (marked in blue), which outputs the rasterized letter $\mathcal{R}(\hat{P})$.
The rasterized letter is then randomly augmented and passed into a pretrained Stable Diffusion \cite{stableDiffusion} model, conditioned on the CLIP's embedding of the given text $c$.
The SDS loss $\nabla_{\hat{P}} \mathcal{L}_\text{LSDS}$ is then 
 used as described in Section \ref{sec:background} to encourage $\mathcal{R}(\hat{P})$ to convey the given text prompt.

To preserve the shape of each individual letter and ensure the legibility of the word as a whole, we use two additional loss functions to guide the optimization process. The first loss limits the overall shape change by defining as-conformal-as-possible constraint on the shape deformation. The second loss preserves the overall shape and style of the font by constraining the tone (i.e. amount of dark vs. light areas in local parts of the shape) of the modified letter not to diverge too much from the original letter (see Section \ref{subsec:losses}). 

The gradients obtained from all the losses are then backpropagated, to update the parameters $\hat{P}$. We repeat this process for 500 steps, which takes $\sim 5$ minutes to produce a single letter illustration on RTX2080 GPU. %

\subsection{Loss Functions}
\label{subsec:losses}
Our primary objective of encouraging the resulting shape to convey the intended semantic concept, is utilized by $\nabla_{\hat{P}} \mathcal{L}_\text{LSDS}$ loss (described in Section \ref{sec:background}).
We observe that using $\nabla_{\hat{P}} \mathcal{L}_\text{LSDS}$ solely can cause large deviations from the initial letter appearance, which is undesired. Hence, our additional goal is to maintain the shape and legibility of the letter $\mathcal{R}(\hat{P})$, as well as to keep the original font's characteristics. For that purpose we use two additional losses.

\begin{figure}[t]
    \centering
    \includegraphics[width=0.9\linewidth]{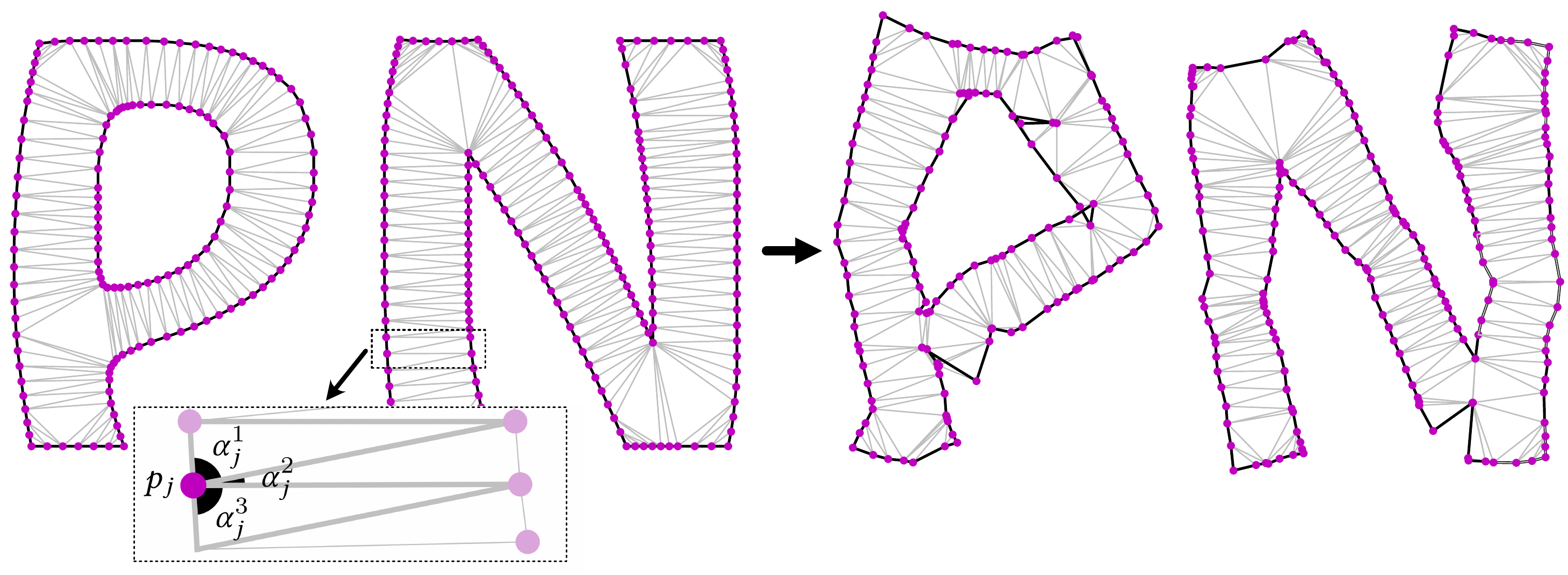}
    \caption{Visual illustration of the constraint Delaunay triangulation applied to the initial shapes (left) and the resulting ones (right), for the word ``pants''. The ACAP loss maintains the structure of the letter after the deformation. The zoomed rectangle shows the angles for a given control point $p_j$. }
    \label{fig:triangulation}
    \vspace{-0.2cm}
\end{figure}

\paragraph{As-Conformal-As-Possible Deformation Loss}
To prevent the final letter shape from diverging too much from the initial shape, we triangulate the inner part of the letter and constrain the deformation of the letter to be as conformal as possible (ACAP) \cite{hormann2000mips}. 
We use constrained Delaunay triangulation \cite{delaunay1934sphere, barber1995qhull} on the set of control points defining the glyph. It is known that Delaunay triangulation can be used to produce the skeleton of an outline \cite{prasad1997morphological,zou2001shape}, so the ACAP loss also implicitly captures a skeletal representation of the letter form.

The Delaunay triangulation $\mathcal{D}(P)$ splits the glyph represented by $P$ into a set of triangles. This defines a set of size $m_j$ of corresponding angles for each control point $p_j$ (see Figure \ref{fig:triangulation}). We denote this set of angles as $\{\alpha_{j}^i\}_{i=1}^{m_j}$. 
The ACAP loss encourages the induced angles of the optimized shape $\hat{P}$ not to deviate much from the angles of the original shape $P$, and is defined as the L2 distance between the corresponding angles:
\begin{equation}
\label{eqn:acap_loss}
    \mathcal{L}_{acap}(P, \hat{P}) = \frac{1}{k} \sum_{j=1}^{k} \left( \sum_{i=1}^{m_j} \big{(} \alpha_j^i - \hat{\alpha}_j^i \big{)}^2 \right)
\end{equation}
where $k=|P|$ and $\hat{\alpha}$ are the angles induced by $\mathcal{D}(\hat{P})$.

\begin{figure}
    \centering
    \includegraphics[width=0.7\linewidth]{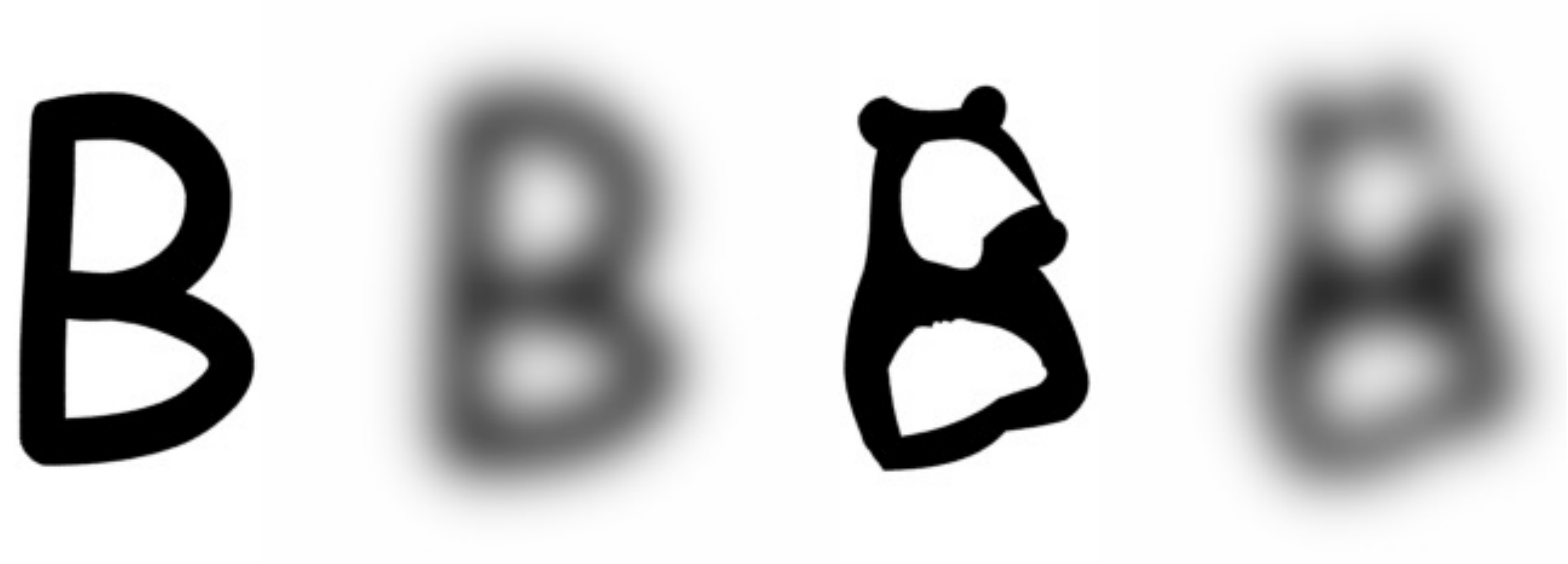}
    \caption{Our tone-preserving loss preserves the local tone of the font by comparing the low-pass filter of the letters images before (left) and after deformation (right). It constrains the adjusted letter not to deviate too much from the original.
    This example is of the letter B and the word ``Bear''.}
    \label{fig:LPF}
\end{figure}

\paragraph{Tone Preservation Loss}
To preserve the style of the font as well as the structure of the letter we add a local-tone preservation loss term. This term constrains the tone (amount of black vs. white in all regions of the shape) of the adjusted letter not to deviate too much from tone of the original font's letter. Towards this end, we apply a low pass filter (LPF) to the rasterized letter (before and after deformation) and compute the L2 distance between the resulting blurred letters:
\begin{equation}
\label{eqn:tone_loss}
    \mathcal{L}_{tone}= \big{\|} LPF(\mathcal{R}(P)) - LPF(\mathcal{R}(\hat{P})) \big{\|}_2^2
\end{equation}
An example of the blurred letters is shown in Figure \ref{fig:LPF}, as can be seen, we use a high value of standard deviation $\sigma$ in the blurring kernel to blur out small details such as the ears of bear.

Our final objective is then defined by the weighted average of the three terms:
\begin{equation}
\begin{aligned}
\label{eqn:final_loss}
    \min_{\hat{P}} \nabla_{\hat{P}} \mathcal{L}_\text{LSDS}(\mathcal{R}(\hat{P}), c) + \alpha\cdot\mathcal{L}_{acap}(P,\hat{P}) \\
    + \beta_t\cdot\mathcal{L}_{tone}(\mathcal{R}(P),\mathcal{R}(\hat{P}))    
\end{aligned}
\end{equation}
where $\alpha=0.5$ and $\beta_t$ depends on the step $t$ as described next.

\subsection{Weighting}
Choosing the relative weights of the three losses presented above is crucial to the appearance of the final letter. 
While the $\nabla_{\hat{P}} \mathcal{L}_\text{LSDS}$ loss encourages the shape to deviate from its original appearance to better fit the semantic concept, the two terms $\mathcal{L}_{tone}$ and $\mathcal{L}_{acap}$ are responsible for maintaining the original shape.
Hence, we have two competing parts in the formula, and would like to find a balance between them to maintain the legibility of the letter while allowing the desired semantic shape to change.

We find that $\mathcal{L}_{tone}$ can be very dominant. In some cases, if it is used from the beginning, no semantic deformation is performed.
Therefore, we adjust the weight of $\mathcal{L}_{tone}$ to kick-in only after some semantic deformation has occurred. We define $\beta_t$ as follows:
\begin{equation}
    \beta_t = a \cdot \exp \big{(}-\frac{(t-b)^2}{2c^2}\big{)}
\end{equation}
with $a = 100, b = 300, c = 30$.
We analyse the affect of various weighting in Section \ref{sec:ablation}.
Note that the same hyper-parameter choice works for various words, letters, and fonts. 

\begin{figure}[t]
\centering
\includegraphics[width=0.9\linewidth]{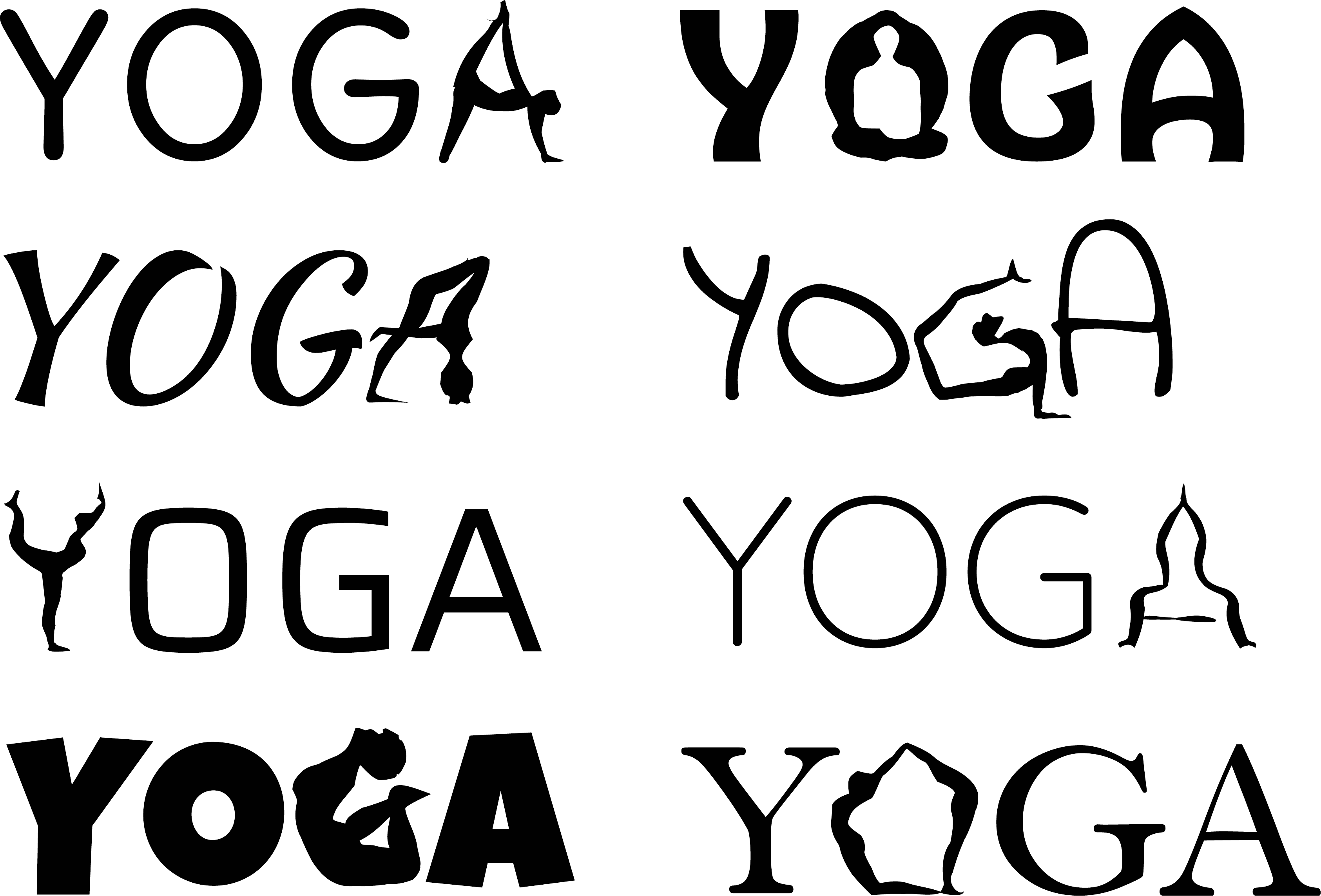}
    \caption{Word-as-images produced by our method for the word ``YOGA'', using eight different fonts.}
    \label{fig:many_fonts}
\end{figure}

\section{Results}
The robustness of our approach means it should be capable of handling a wide range of input concepts as well as supporting different font designs.
Figures \ref{fig:teaser}, \ref{fig:intro_res}, \ref{fig:all_res1}, \ref{fig:mayne_res}, and more results in the supplemental file demonstrate that our approach can handle inputs from many different categories and various fonts, and that the generated results are legible and creative.
Figure~\ref{fig:many_fonts} demonstrate how the illustrations created by our method for the same word follow the characteristics of different fonts.
Although the perceived aesthetics of a word-as-image illustration can be subjective, we define three objectives for an effective result: (1) it should visually capture the given semantic concept, (2) it should maintain readability, and (3) it should preserve the original font's characteristics.

We evaluate the performance of our method on a randomly selected set of inputs.
We select five common concept classes - animals, fruits, plants, sports, and professions. 
Using ChatGPT, we sample ten random instances for each class, resulting in 50 words in total.
Next, we select four fonts that have distinct visual characteristics, namely Quicksand, Bell MT, Noteworthy-Bold, and HobeauxRococeaux-Sherman.
For each word, we randomly sampled one of the four fonts, and applied our method to each letter.
For each word with $n$ letters we can generate $2^n$ possible word-as-images, which are all possible combinations of replacements of illustrated letters.
A selected subset of these results is presented in Figure \ref{fig:all_res1}.
The results of all letters and words are presented in the supplementary material.

As can be seen, the resulting word-as-image illustrations successfully convey the given semantic concept in most cases while still remaining legible.
In addition, our method successfully captures the font characteristics. For example, in Figure \ref{fig:all_res1}, the replacements for the ``DRESS'' and ``LION'' are thin and fit well with the rest of the word. In addition, observe the serifs of the letter A used for the fin of the shark in the ``SHARK'' example. We further use human evaluation to validate this as described below.

\subsection{Quantitative}
\label{subsec:quant}
We conduct a perceptual study to quantitatively assess the three objectives of our resulting word-as-images. 
We randomly select two instances from each of the resulting word-as-image illustrations for the five classes described above, and visually select one letter from each word, resulting in 10 letters in total. 
In each question we show an isolated letter illustration, without the context of the word.
To evaluate the ability of our method to visually depict the desired concept, we present four label options from the same class, and ask participants to choose the one that describes the letter illustration best.
To evaluate the legibility of the results, we ask participants to choose the most suitable letter from a random list of four letters.
To asses the preservation of the font style, we present the four fonts and ask participants to choose the most suitable font for the illustration.
We gathered answers from 40 participants, and the results are shown in Table \ref{tb:user_study}. As can be seen, the level of concept recognizability and letter legibility are very high, and the $51\%$ of style matching of the letter illustration to the original font is well above random, which is $25\%$. We also test our algorithm without the two additional structure and style preserving losses ($\mathcal{L}_{acap}$ and $\mathcal{L}_{tone}$) on the same words and letters (``Only SDS'' in the table). As expected, without the additional constraints, the letter deforms significantly resulting in higher concept recognizability but lower legibility and font style preservation. More details and examples are provided in the supplementary material.
 
\begin{table}
    \small
    \centering
    \setlength{\tabcolsep}{2pt}
    \caption{Perceptual study results. The level of concept recognizability and letter legibility are very high, and style matching of the font is well above random. The ``Only SDS'' results are created by removing our structure and style preserving losses.} 
    \begin{tabular}{l c c c} 
    \toprule
    Method & \begin{tabular}{c} Semantics \end{tabular} & \begin{tabular}{c} Legibility \end{tabular} & \begin{tabular}{c} Font \end{tabular} \\
    \midrule
    Ours    & 0.8 & 0.9 & 0.51 \\
    Only SDS  & 0.88 & 0.53 & 0.33 \\
    \bottomrule
    \end{tabular}
    \vspace{-0.2cm}
    \label{tb:user_study}
\end{table}

\begin{figure*}
    \centering
    \setlength{\tabcolsep}{1.5pt}
    {\small
    \begin{tabular}{l@{\hspace{0.2cm}} c@{\hspace{0.2cm}} | @{\hspace{0.2cm}}c c@{\hspace{0.2cm}} | c c @{\hspace{0.2cm}} | c @{\hspace{0.2cm}} | @{\hspace{0.2cm}}c l}

        \raisebox{0.5cm}{\makecell[l]{The word \\ BIRD and \\ the letter R}} &
        \hspace{0.1cm}
        \raisebox{0.25cm}{\includegraphics[height=0.042\textwidth]{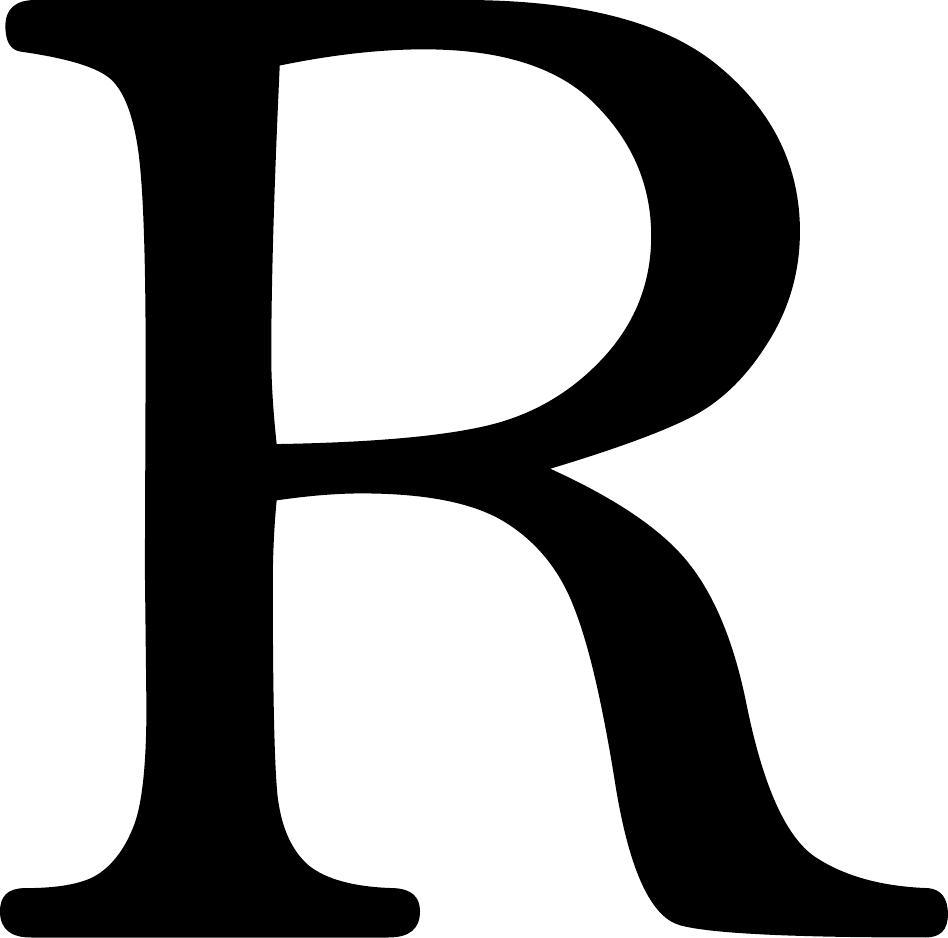}} &
        \hspace{0.1cm}
        \includegraphics[height=0.07\textwidth]{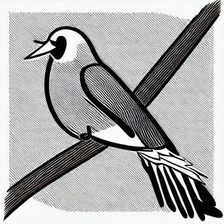} &
        \includegraphics[height=0.07\textwidth]{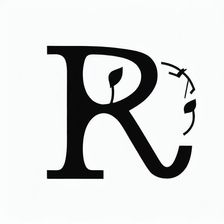} &
        \hspace{0.1cm}
        \includegraphics[height=0.07\textwidth]{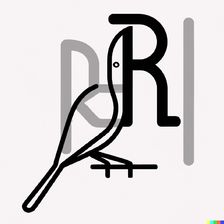} &
        \includegraphics[height=0.07\textwidth]{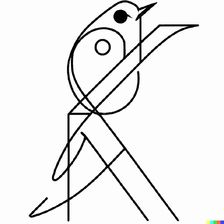} &
        \hspace{0.1cm}
        \includegraphics[height=0.07\textwidth]{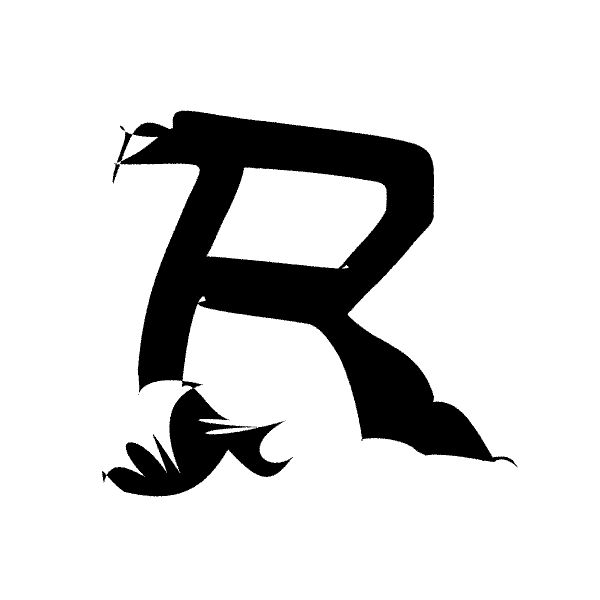} &
        \raisebox{0.23cm}{\includegraphics[height=0.048\textwidth]{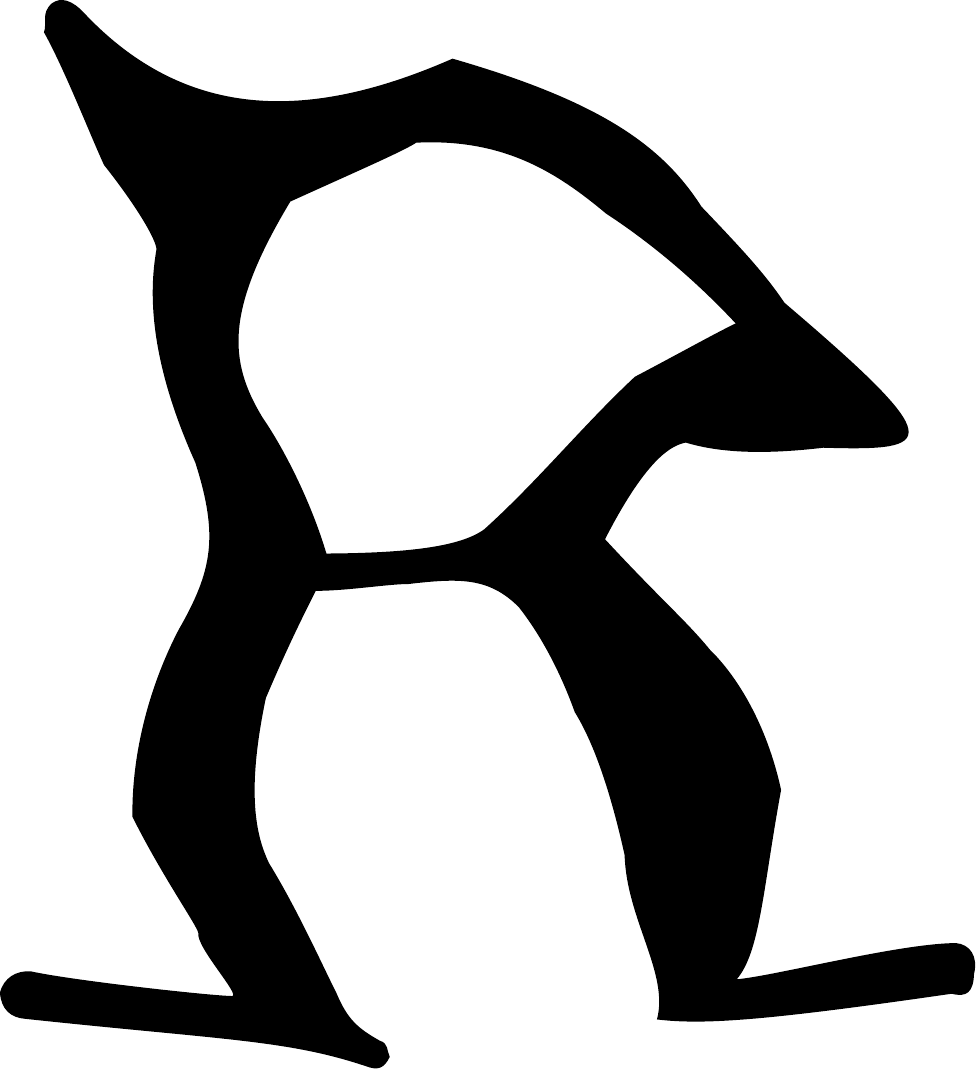}} &
        \hspace{0.1cm}
        \raisebox{0.23cm}{\includegraphics[height=0.048\textwidth]{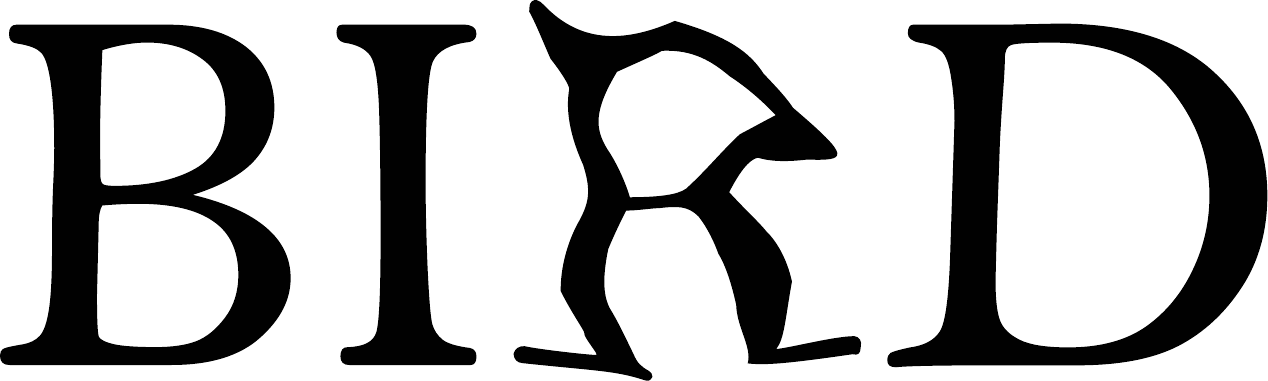}} \\

        \raisebox{0.5cm}{\makecell[l]{The word \\ DRESS and \\ the letter E}} &
        \hspace{0.1cm}
        \raisebox{0.25cm}{\includegraphics[height=0.042\textwidth]{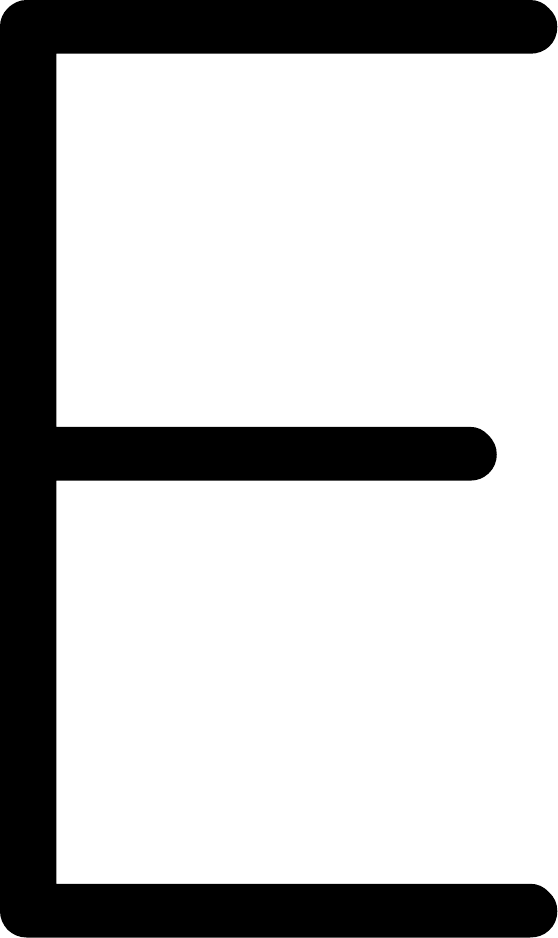}} &
        \hspace{0.1cm}
        \includegraphics[height=0.07\textwidth]{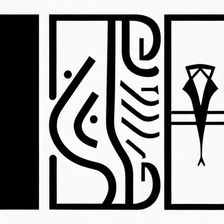} &
        \includegraphics[height=0.07\textwidth]{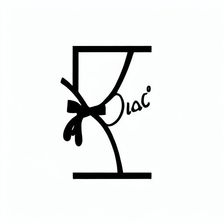} &
        \hspace{0.1cm}
        \includegraphics[height=0.07\textwidth]{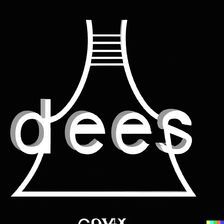} &
        \includegraphics[height=0.07\textwidth]{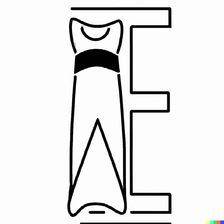} &
        \hspace{0.1cm}
        \includegraphics[height=0.07\textwidth]{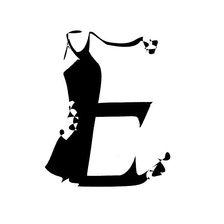} &
        \raisebox{0.2cm}{\includegraphics[height=0.048\textwidth]{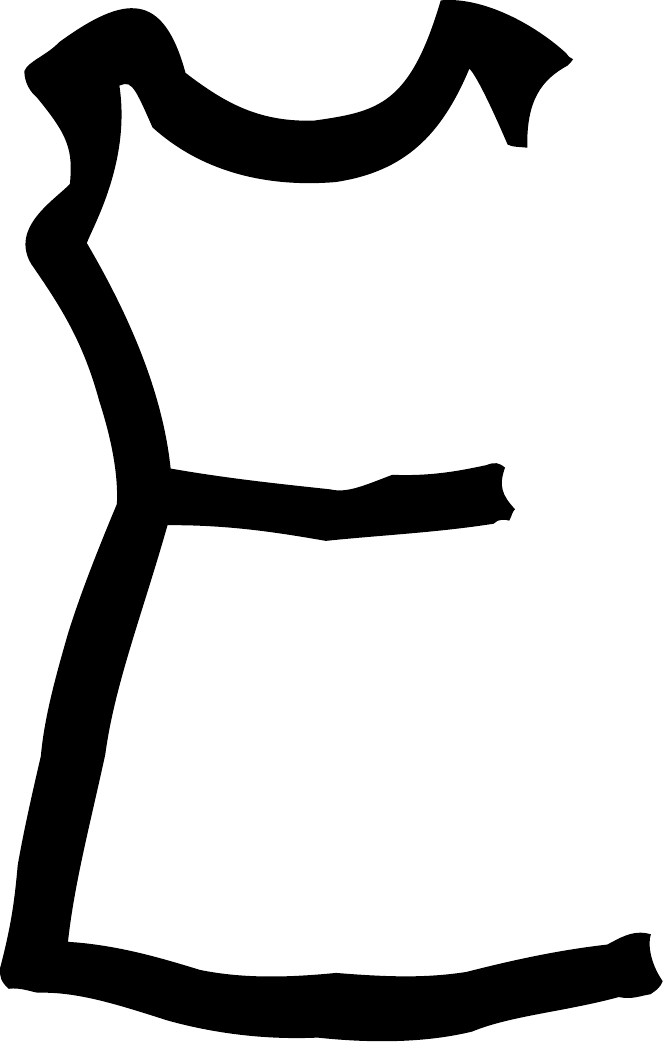}} &
        \hspace{0.1cm}
        \raisebox{0.2cm}{\includegraphics[height=0.048\textwidth]{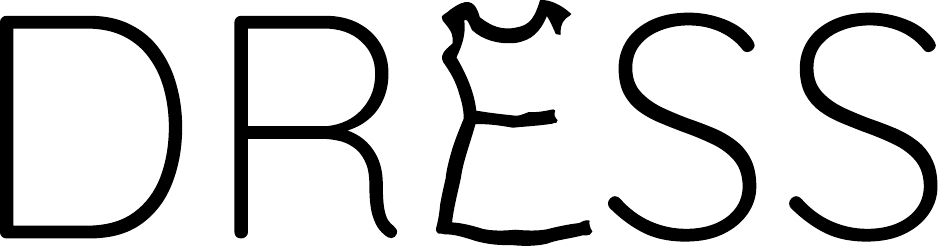}} \\

        \raisebox{0.5cm}{\makecell[l]{The word \\ TULIP and \\ the letter U}} &
        \hspace{0.1cm}
        \raisebox{0.25cm}{\includegraphics[height=0.042\textwidth]{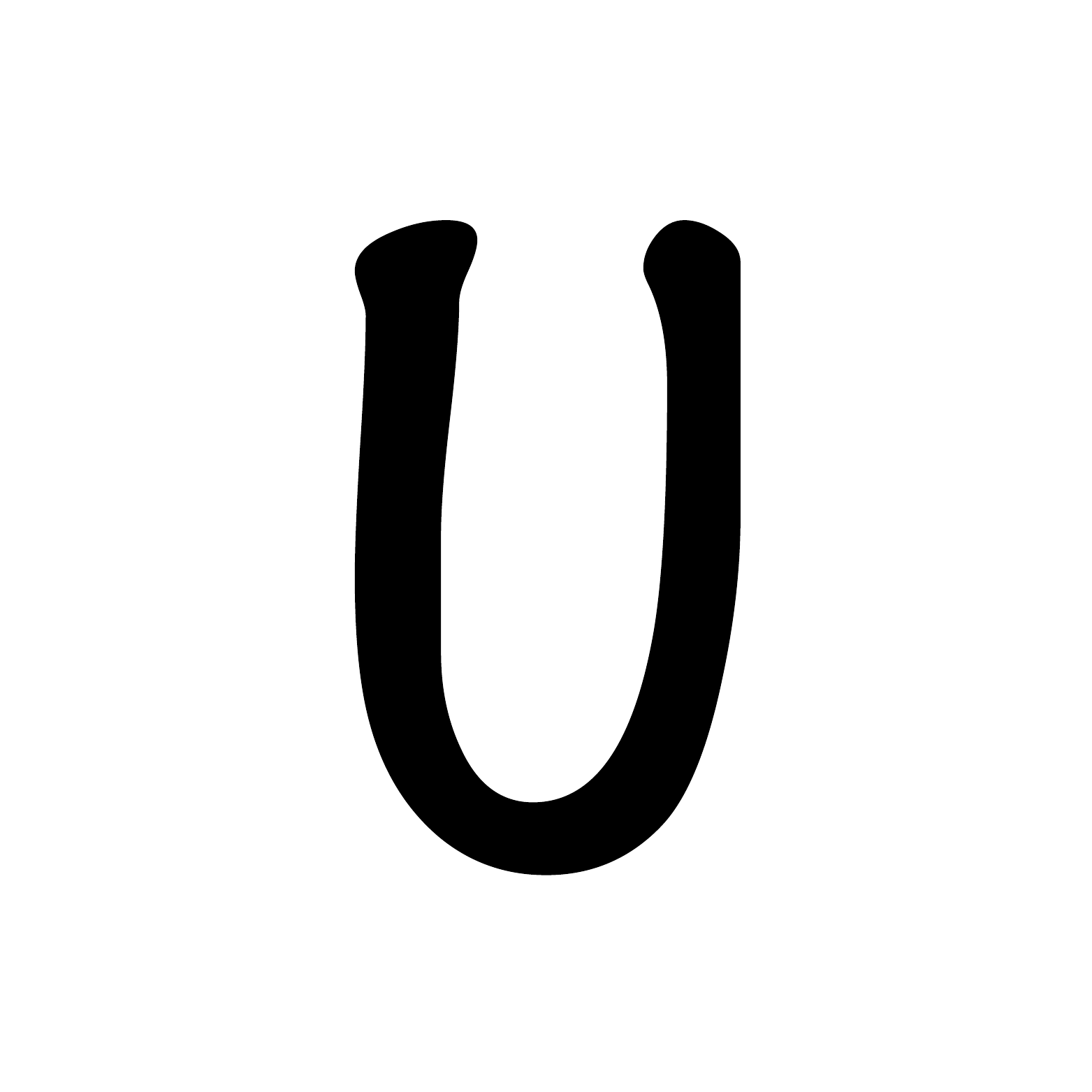}} &
        \hspace{0.1cm}
        \includegraphics[height=0.07\textwidth]{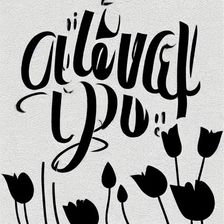} &
        \includegraphics[height=0.07\textwidth]{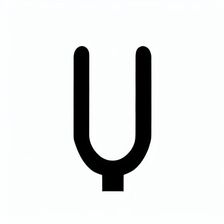} &
        \hspace{0.1cm}
        \includegraphics[height=0.07\textwidth]{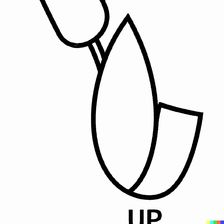} &
        \includegraphics[height=0.07\textwidth]{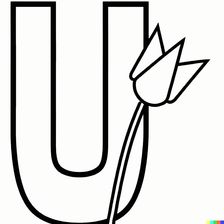} &
        \hspace{0.1cm}
        \includegraphics[height=0.07\textwidth]{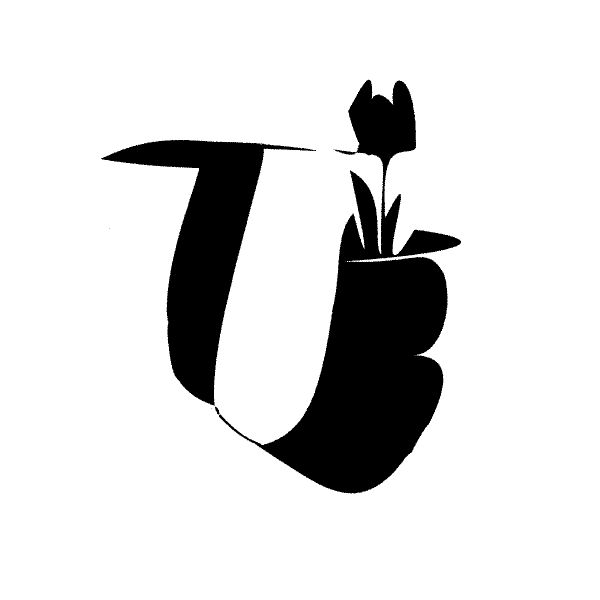} &
        \raisebox{0.16cm}{\includegraphics[height=0.048\textwidth]{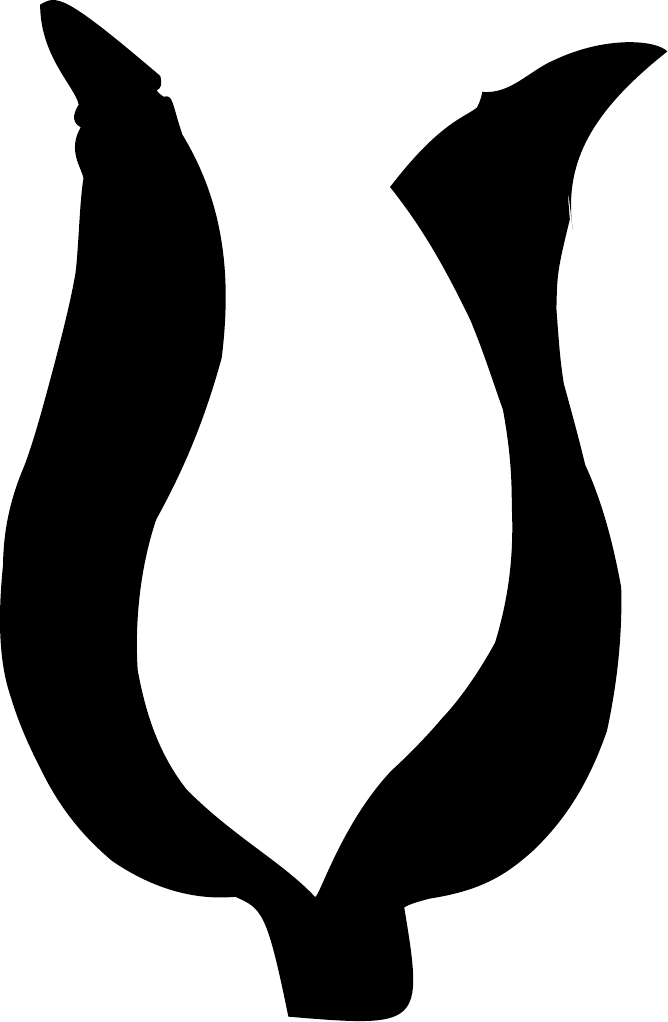}} &
        \hspace{0.1cm}
        \raisebox{0.16cm}{\includegraphics[height=0.048\textwidth]{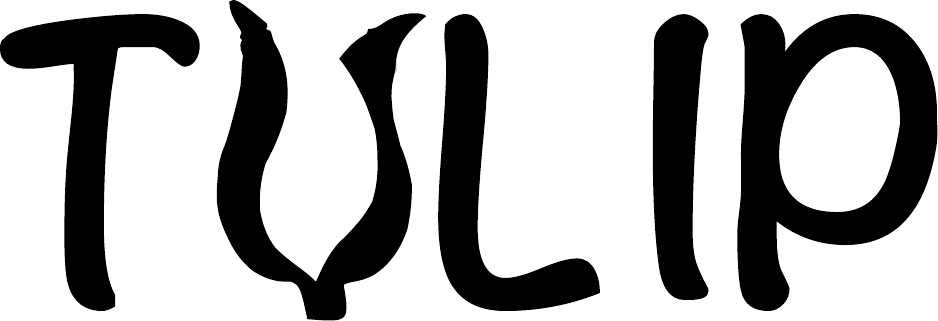}} \\

        \multicolumn{2}{c@{\hspace{0.2cm}}}{Input}&\hspace{0.1cm} SD & SDEdit &\hspace{0.1cm} DallE2 & DallE2+letter &\hspace{0.1cm} CLIPDraw & \multicolumn{2}{c}{Ours}

    \end{tabular}
    }
    \caption{Comparison to alternative methods based on large scale text-to-image models. On the left are the letters used as input (only for SDEdit, CLIPDraw, and ours), as well as the desired object of interest. The results from left to right obtained using Stable Diffusion \cite{stableDiffusion}, SDEdit \cite{meng2022sdedit}, DallE2 \cite{ramesh2022hierarchical}, DallE2 with a letter specific prompt, CLIPDraw \cite{frans2021clipdraw}, and our single-letter results, as well as the final word-as-image.}
    \label{fig:comp_diffusion}
\end{figure*}

\subsection{Comparison}
\label{subsec:comparisons}
In the absence of a relevant baseline for comparison, we define baselines based on large popular text-to-image models. Specifically, we use \textbf{(1) SD} Stable Diffusion \cite{stableDiffusion}, \textbf{(2) SDEdit} \cite{meng2022sdedit}, \textbf{(3) DallE2} \cite{ramesh2022hierarchical} illustrating the word, \textbf{(4) DallE2+letter} illustrating only the letter, and \textbf{(5) CLIPDraw} \cite{frans2021clipdraw}. We applied the methods above (details can be found in supplemental material) to three representative words -- ``bird'', ``dress'', and ``tulip'', with the fonts Bell MT, Quicksand, and Noteworthy-Bold, respectively. The results can be seen in Figure \ref{fig:comp_diffusion}.

In some cases Stable Diffusion (SD) did not manage to produce text at all (such as for the bird) and when text is produced, it is often not legible.
The results obtained by SDEdit preserve the font's characteristics and the letter's legibility, but often fail to reflect the desired concept, such as in the case of the bird and the dress. Additionally, it operates in the raster domain and tends to \textit{add} details on top of the letter, while our method operates directly on the vector representation of the letters with the objective of modifying their \textit{shape}.
DallE2 manages to reflect the visual concept, however it often fails to produce legible text. When applied with a dedicated prompt to produce the word-as-image of only one letter (fifth column), it manages to produce a legible letter, but there is less control over the output -- it is impossible to specify the desired font or to control the size, position, and shape of the generated letter. Therefore, it is not clear how to combine these output illustrations into the entire word to create a word-as-image.

CLIPDraw produces reasonable results conveying the semantics of the input word. However, the results are non-smooth and the characteristics of the font are not preserved (for example observe how the letter "E" differs from the input letter). We further examine CLIPDraw with our shape preservation losses in the next Section.

\subsection{Ablation}
\label{sec:ablation}
\begin{figure}[ht]
    \centering
    \setlength{\tabcolsep}{5pt}
    \renewcommand{\arraystretch}{1.5} 
    \begin{tabular}{c c | c c c}
        \raisebox{0.4cm}{"Ballet"} &
        \includegraphics[height=0.1\linewidth]{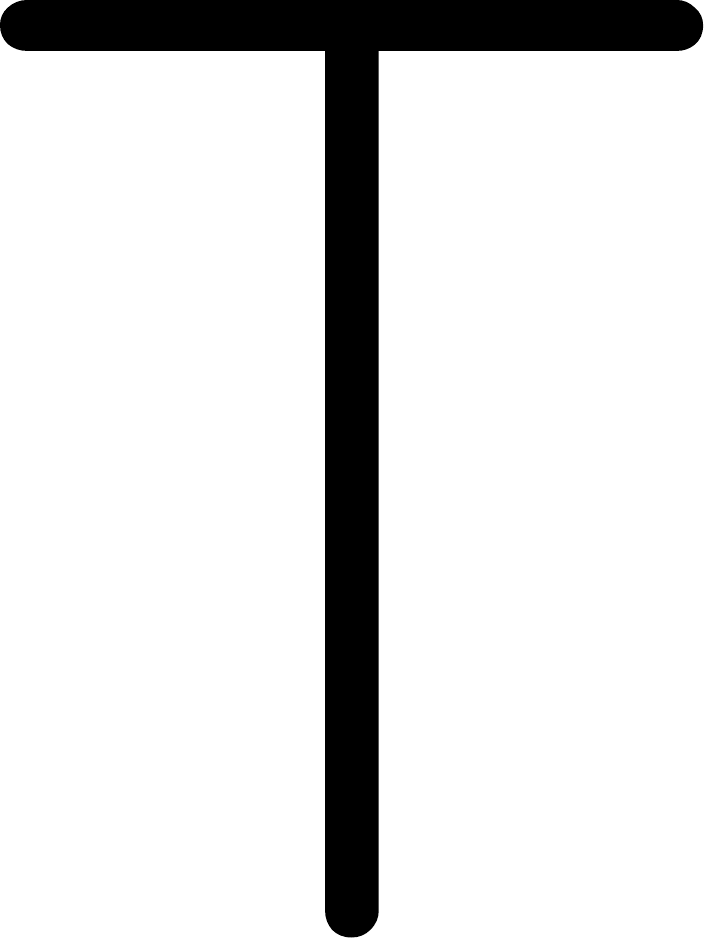} &
        \hspace{0.1cm}
        \includegraphics[height=0.1\linewidth]{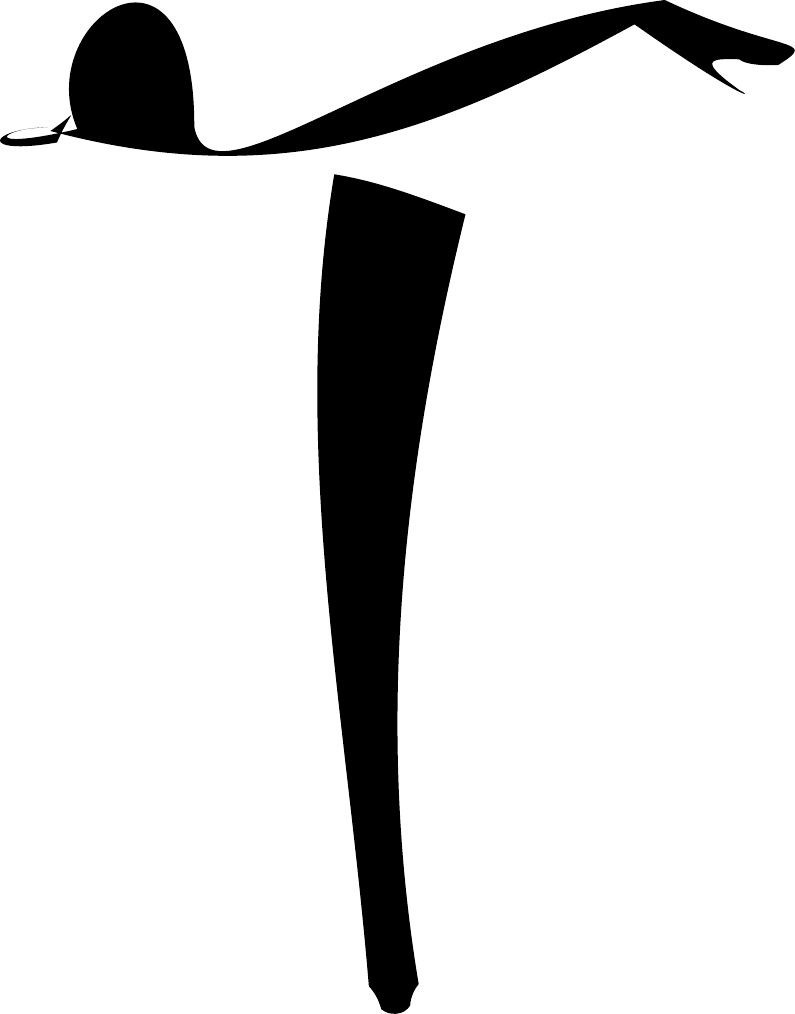} &
        \includegraphics[height=0.1\linewidth]{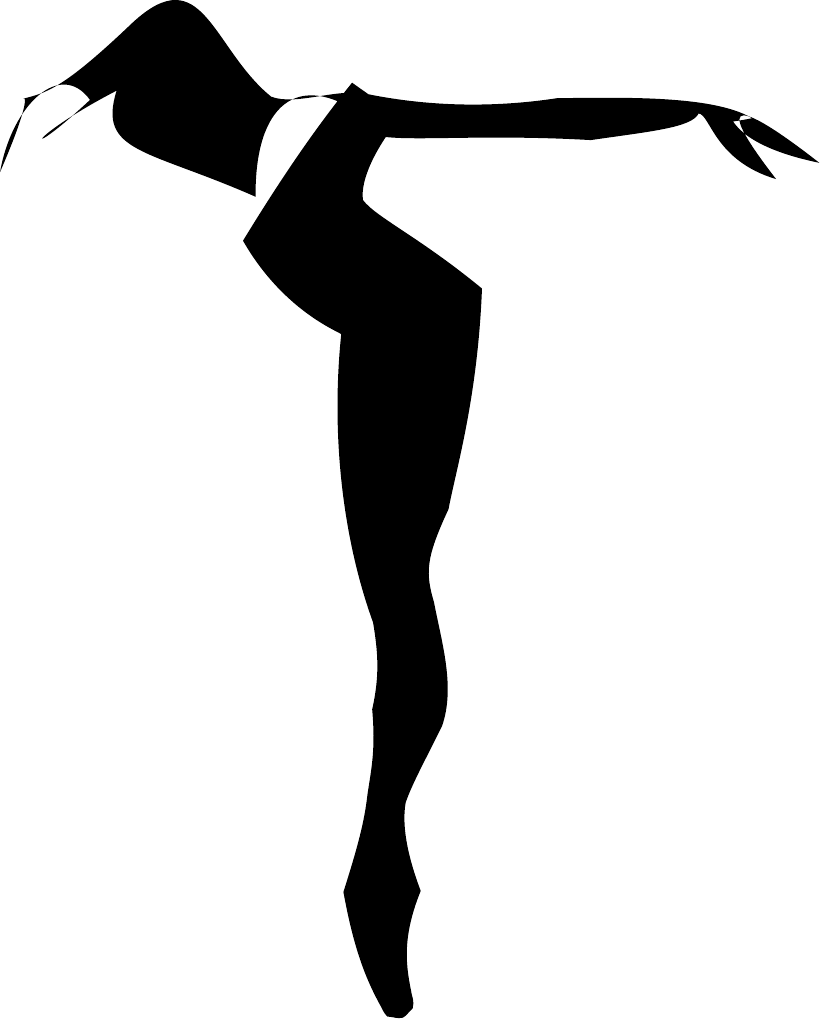} &
        \includegraphics[height=0.1\linewidth]{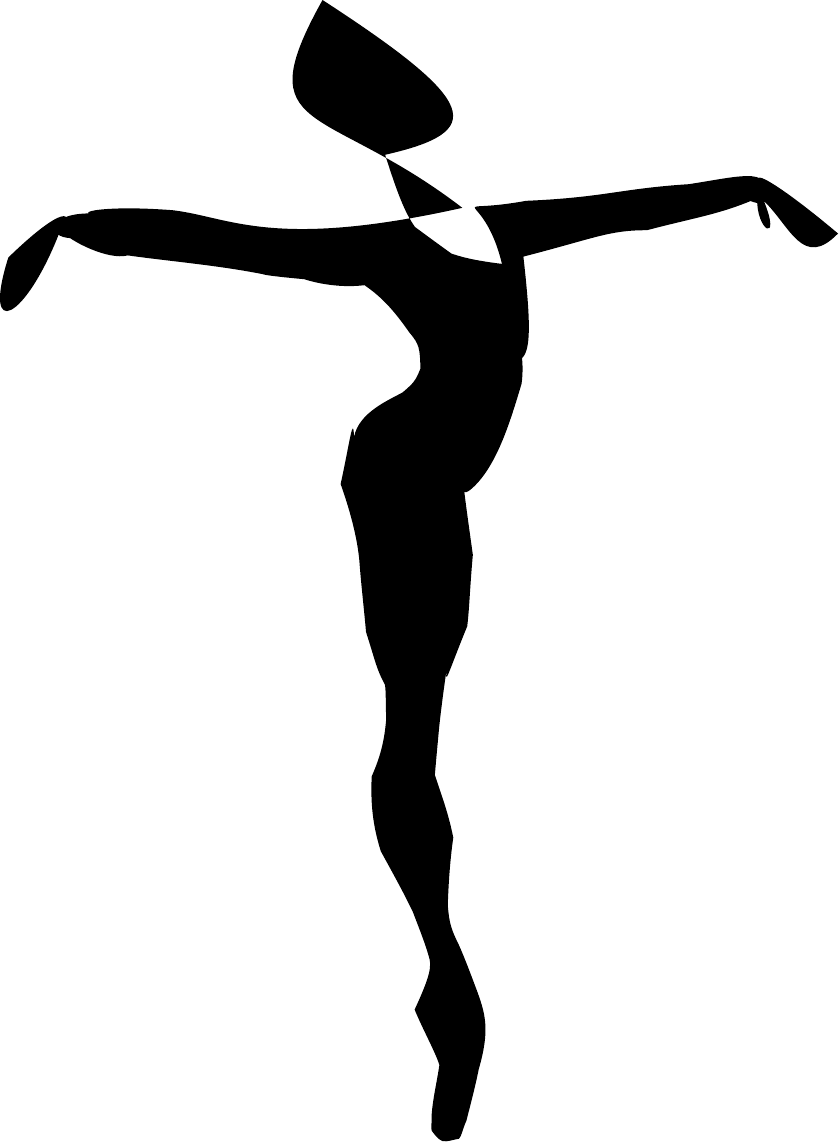} \\
        
        \raisebox{0.4cm}{"Gorilla"} &
        \includegraphics[height=0.1\linewidth]{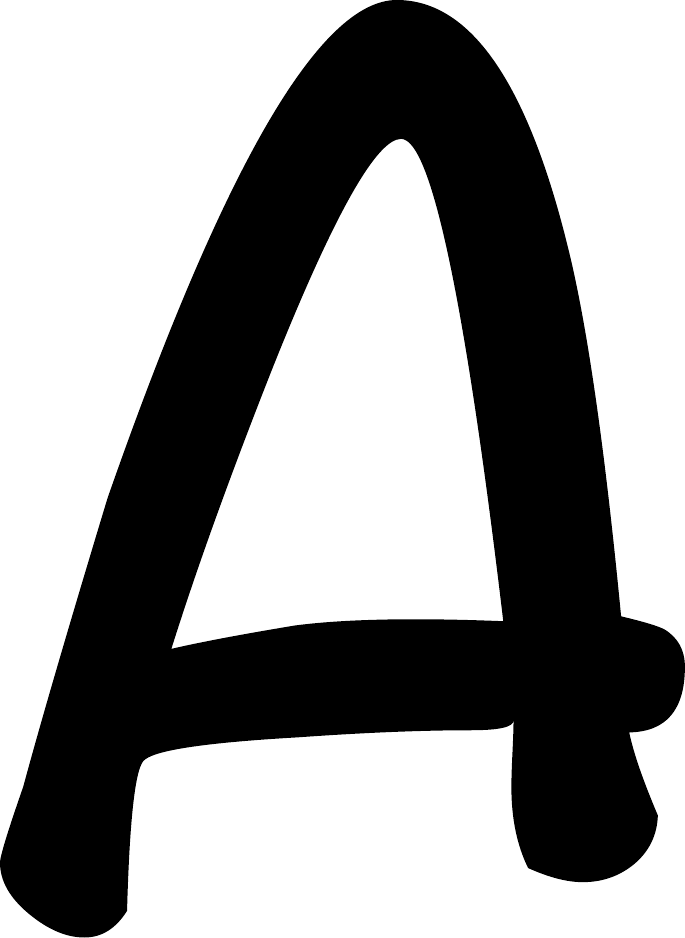} &
        \hspace{0.1cm}
        \includegraphics[height=0.1\linewidth]{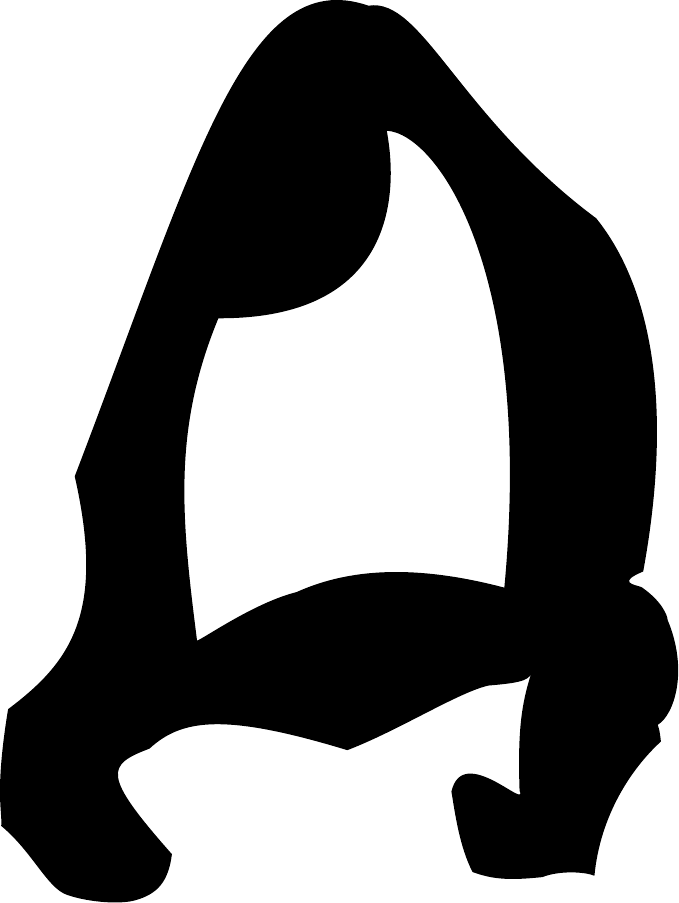} &
        \includegraphics[height=0.1\linewidth]{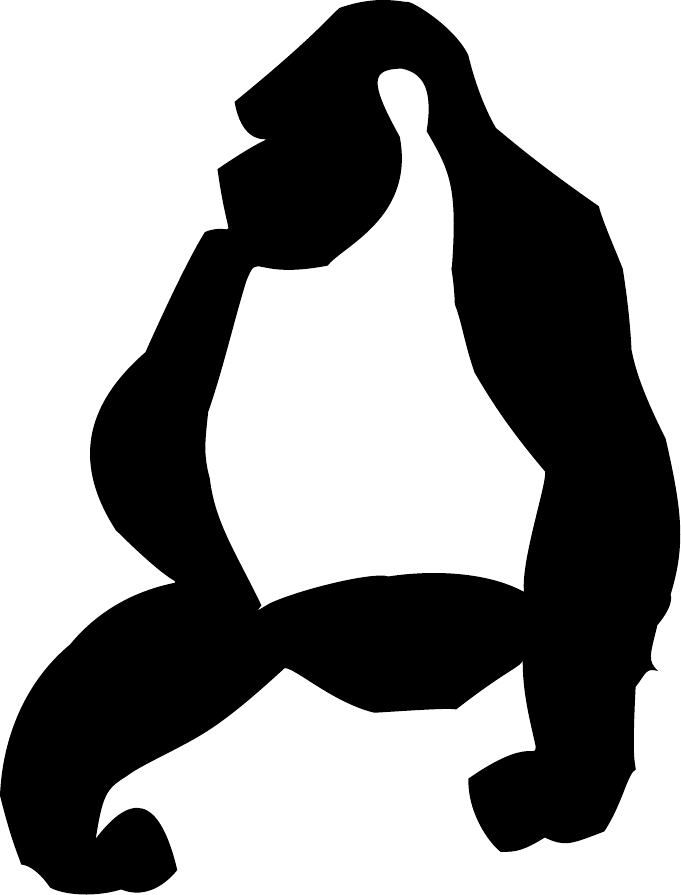} &
        \includegraphics[height=0.1\linewidth]{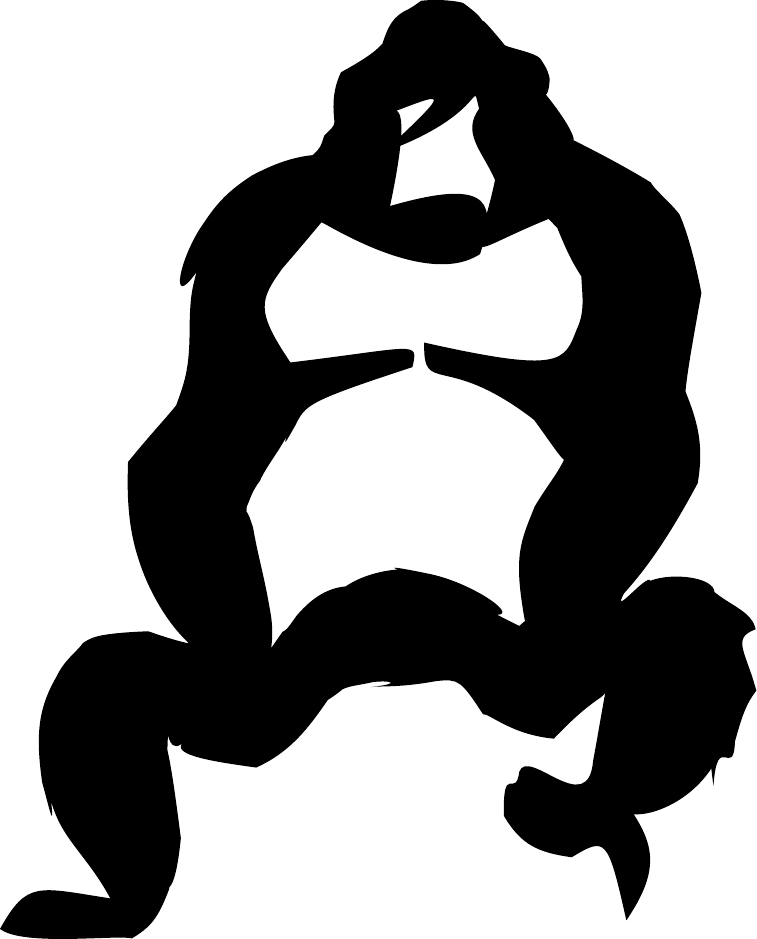} \\

        \raisebox{0.4cm}{"Gym"} &
        \includegraphics[height=0.1\linewidth]{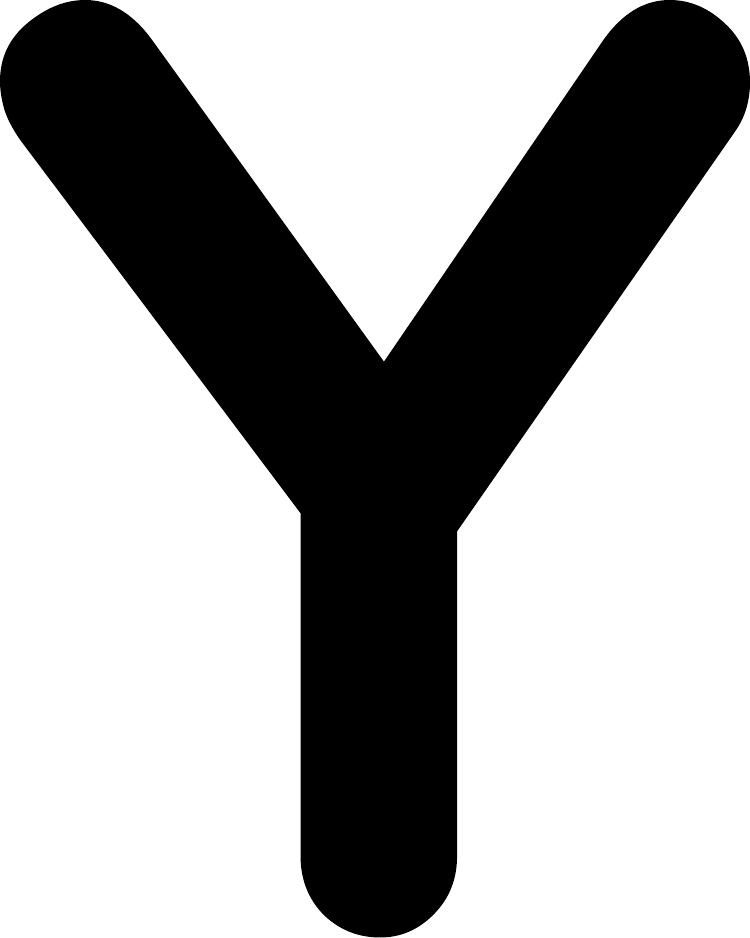} &
        \hspace{0.1cm}
        \includegraphics[height=0.1\linewidth]{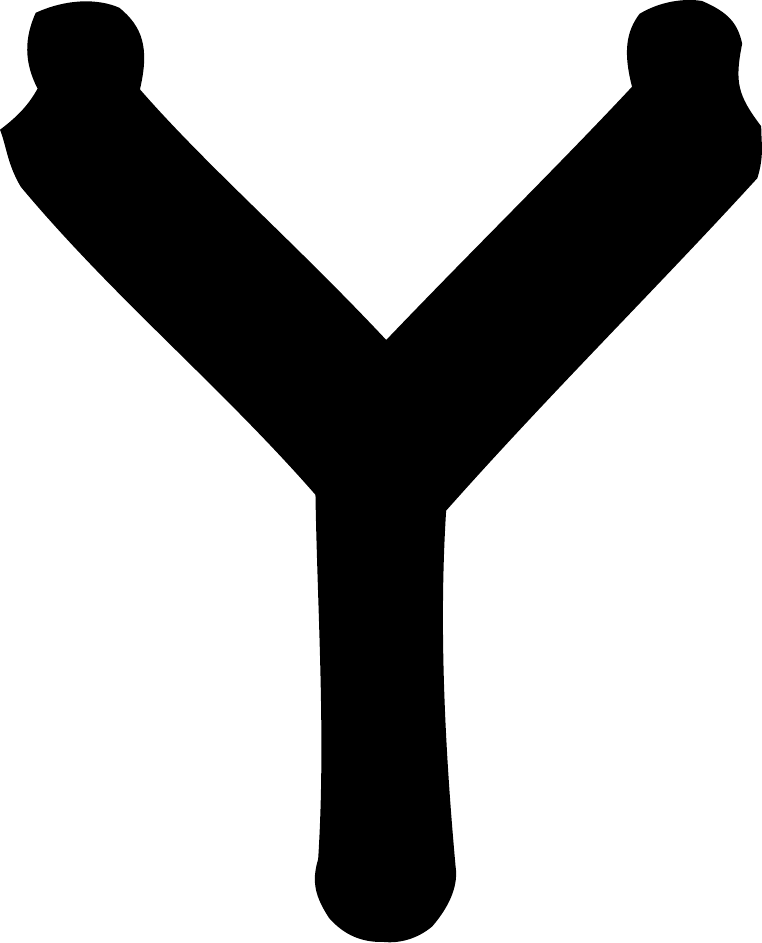} &
        \includegraphics[height=0.1\linewidth]{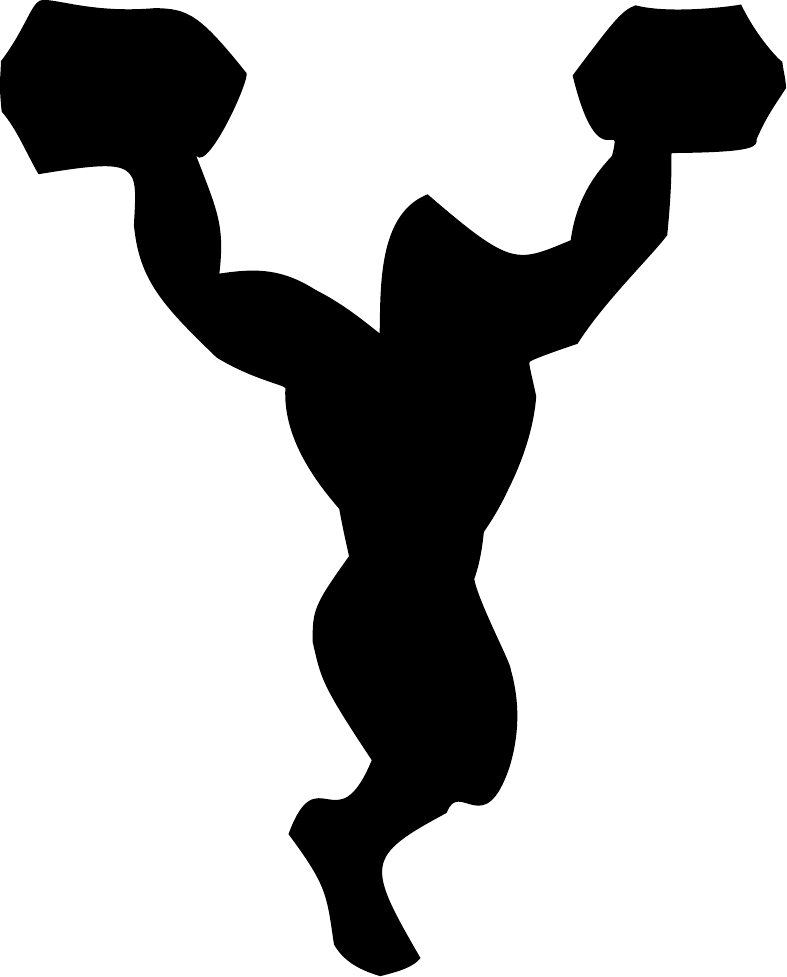} &
        \includegraphics[height=0.1\linewidth]{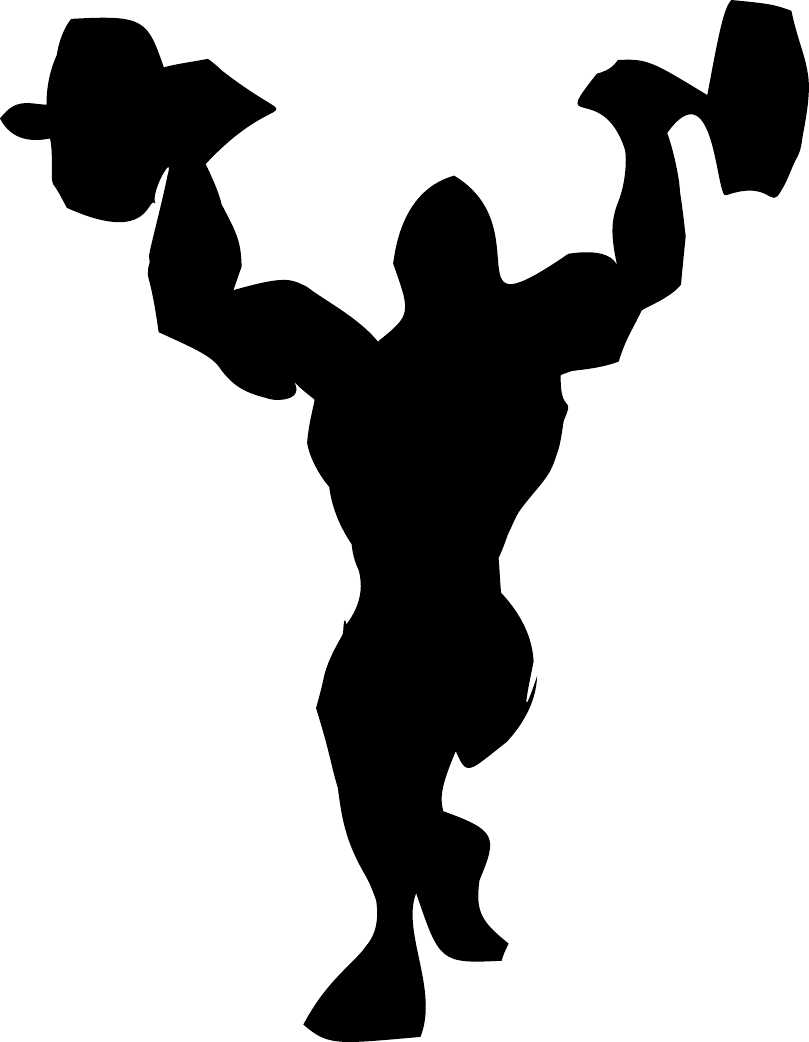} \\
        
        \multicolumn{2}{c}{Input} & $P_o$ & $P$ & $2\times{P}$\\
    \end{tabular}
    \vspace{-0.2cm}
    \caption{The effect of the initial number of control points on outputs. On the left are the input letters and the target concepts used to generate the results on the right. $P_o$ indicates the original number of control points as extracted from the font, $P$ is the input letter with our chosen hyperparameters, and for $2\times{P}$ we increase the number of control points in $P$ by two.}
    \label{fig:cc_effect}
\end{figure}

\begin{figure}[!t]
\centering
\setlength{\tabcolsep}{5pt}
\renewcommand{\arraystretch}{1} 
\begin{tabular}{c c | c c c c c}
    \raisebox{0.4cm}{"Bear"} &
    \raisebox{0.01cm}{\includegraphics[height=0.09\linewidth]{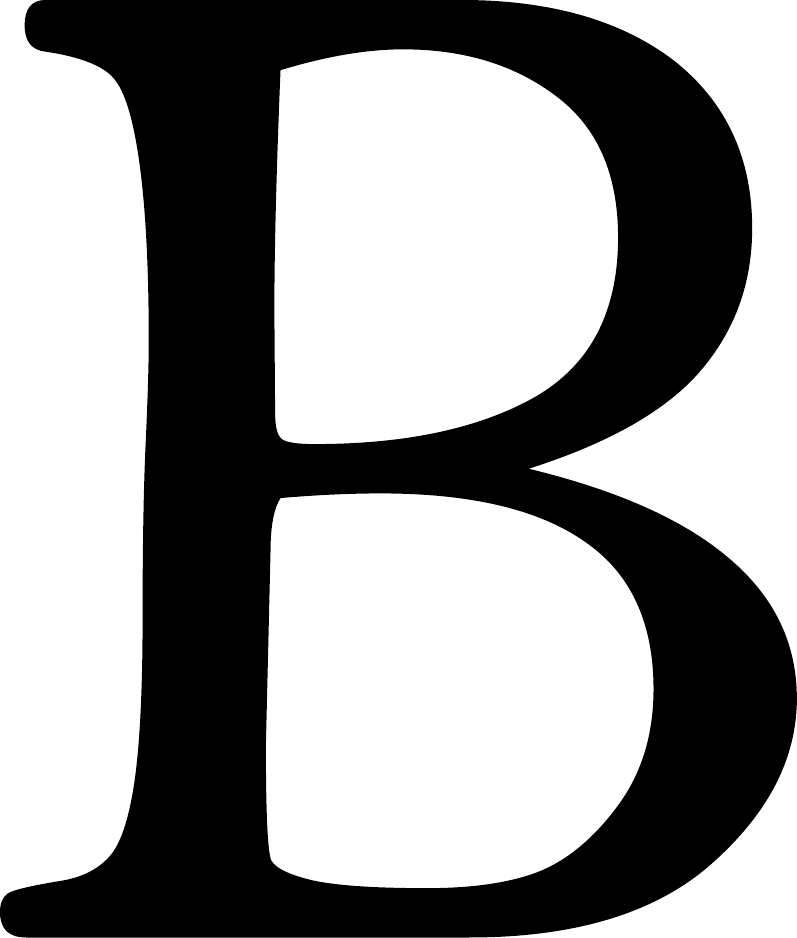}} &
    \hspace{0.2cm}
    \includegraphics[height=0.1\linewidth]{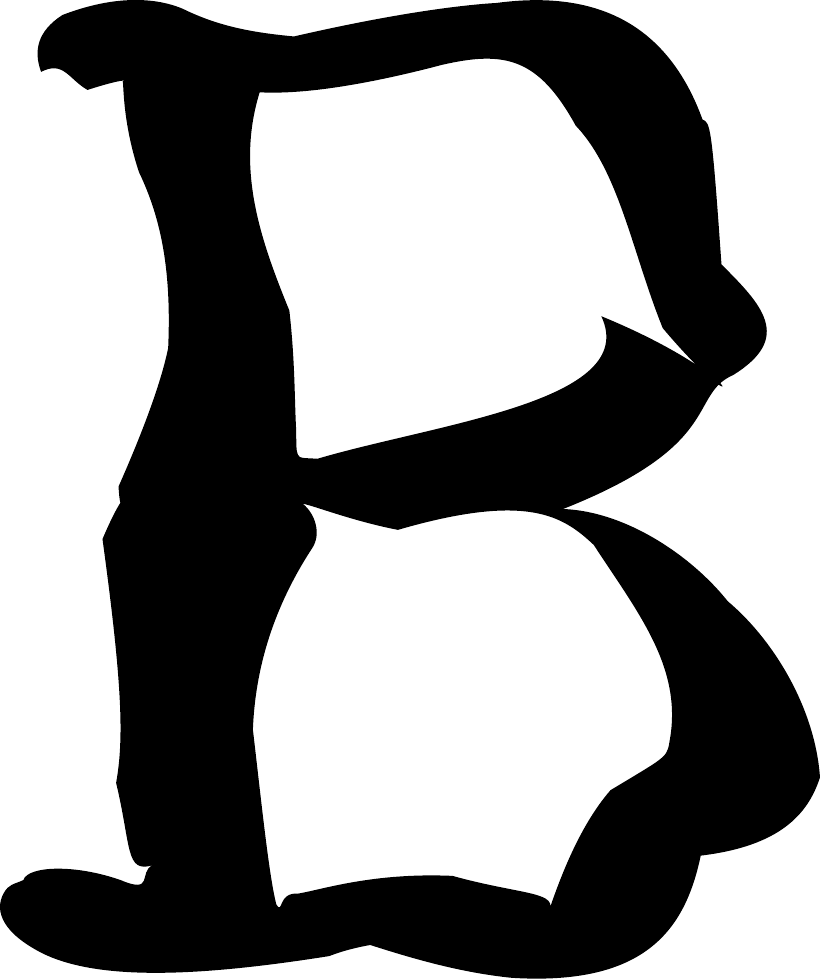} &
    \includegraphics[height=0.1\linewidth]{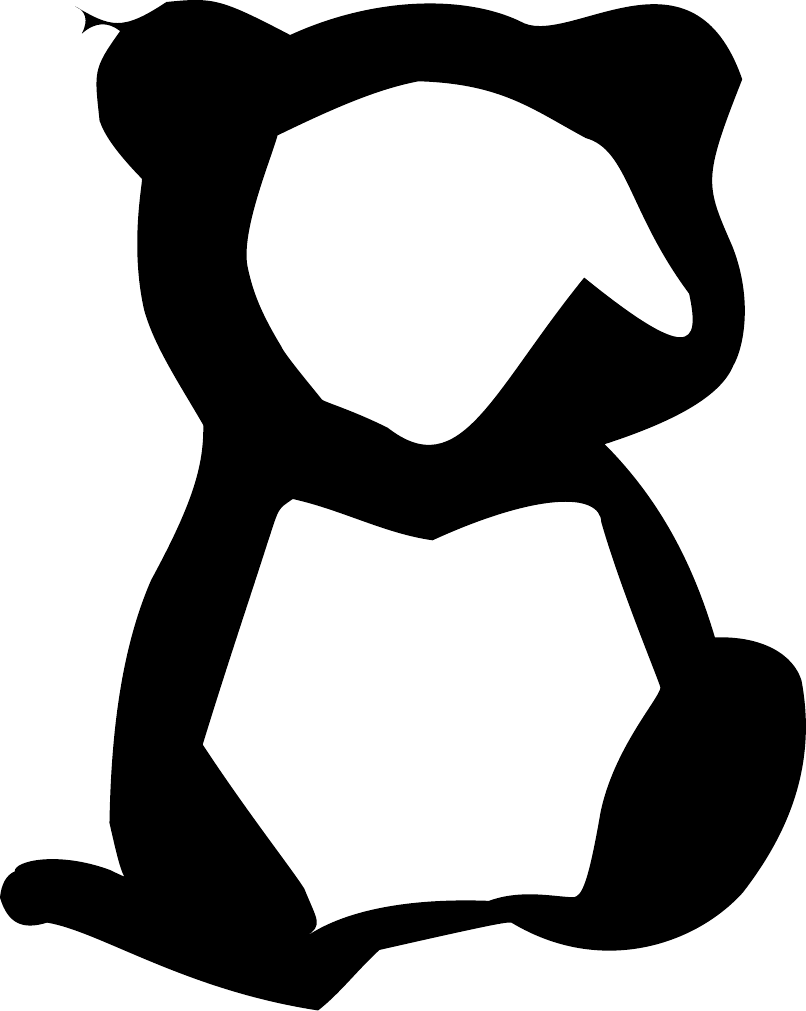} &
    \includegraphics[height=0.1\linewidth]{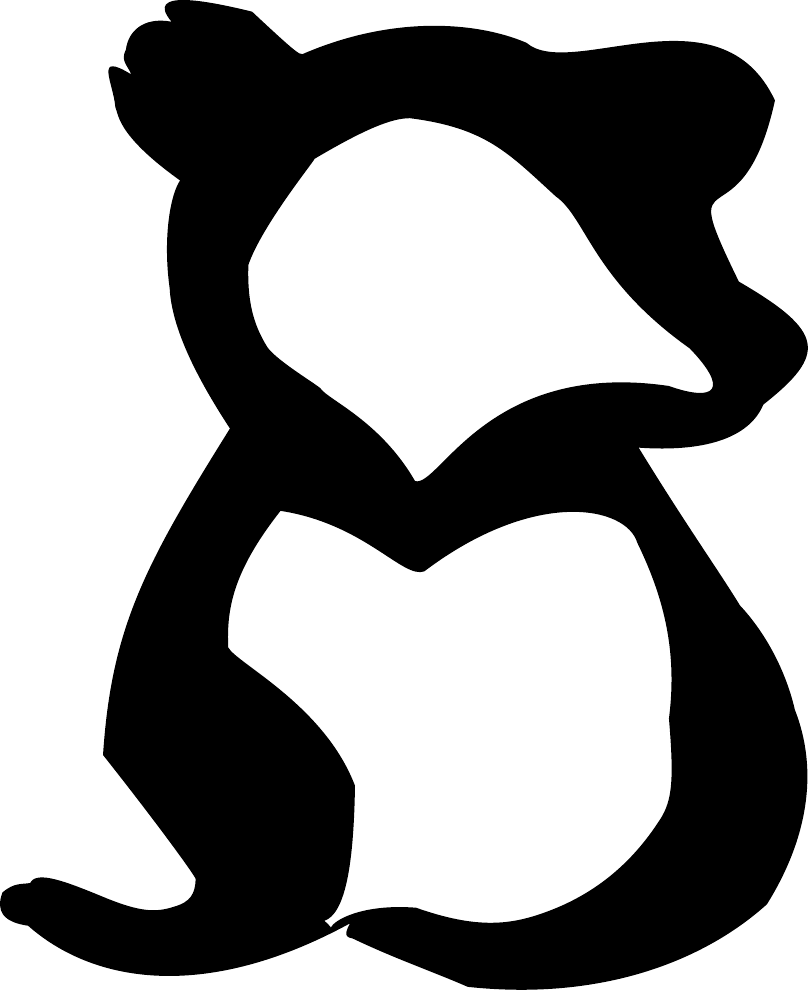} &
    \includegraphics[height=0.1\linewidth]{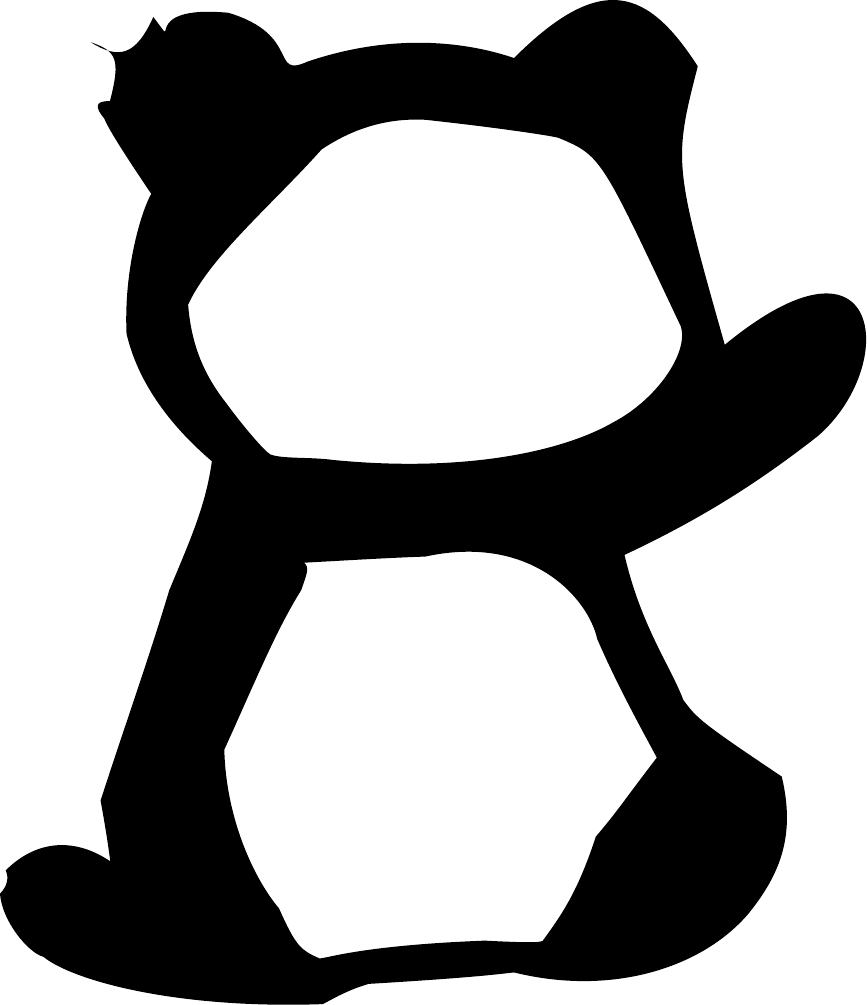} &
    \includegraphics[height=0.1\linewidth]{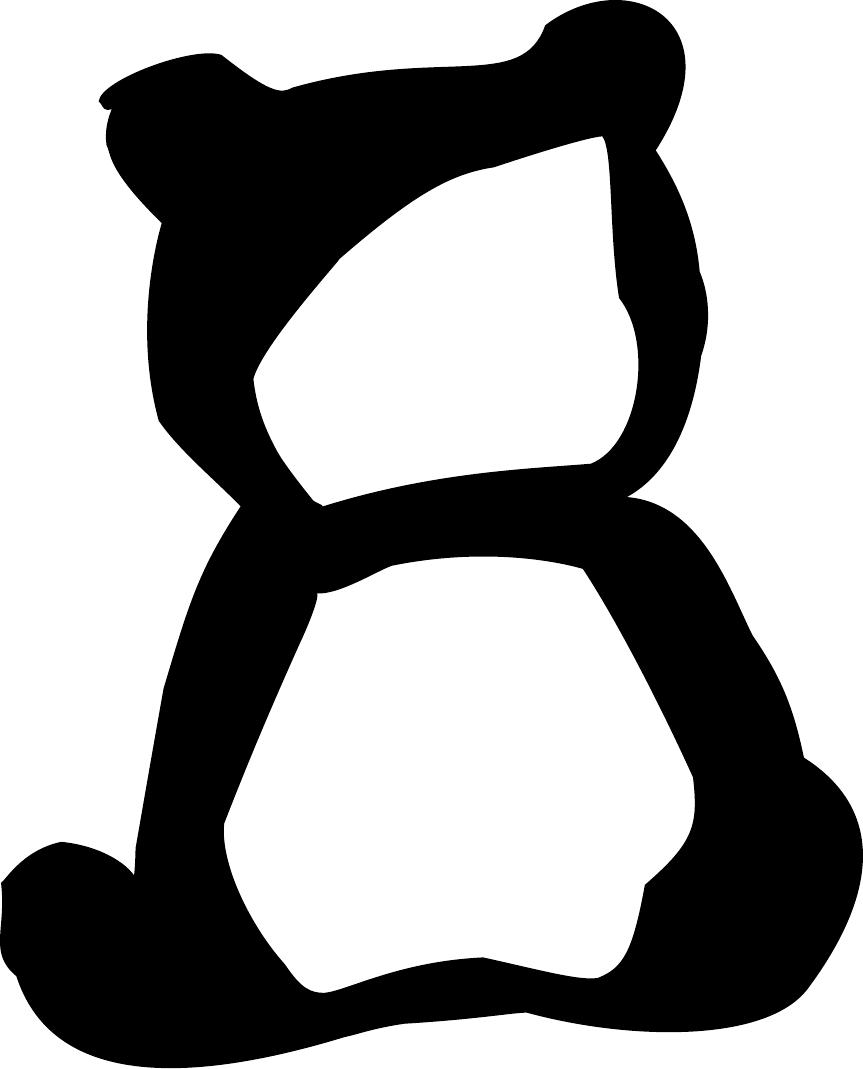}\\
    
    \raisebox{0.4cm}{"Singer"} &
    \includegraphics[height=0.09\linewidth]{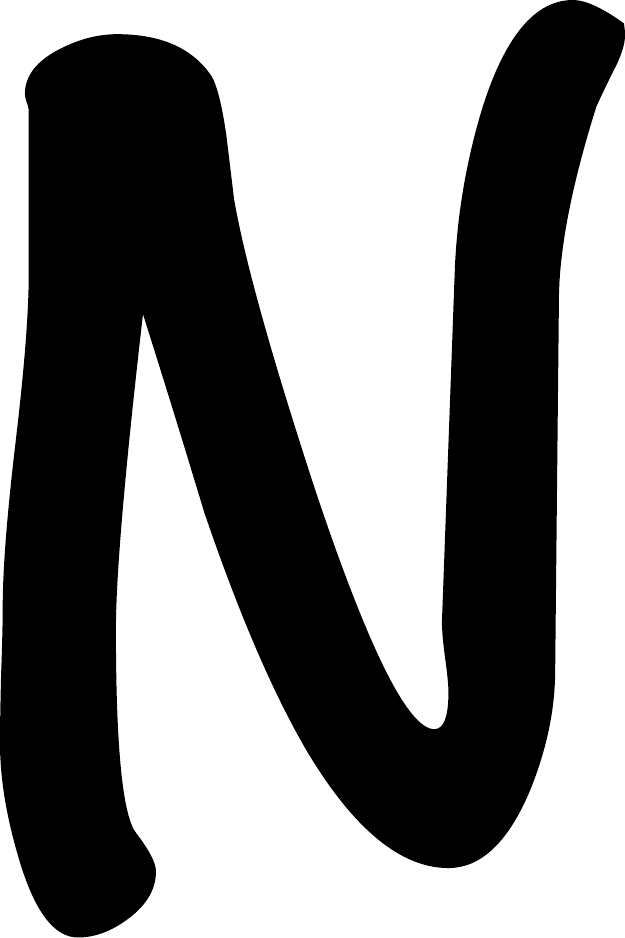} &
    \hspace{0.2cm}
    \includegraphics[height=0.1\linewidth]{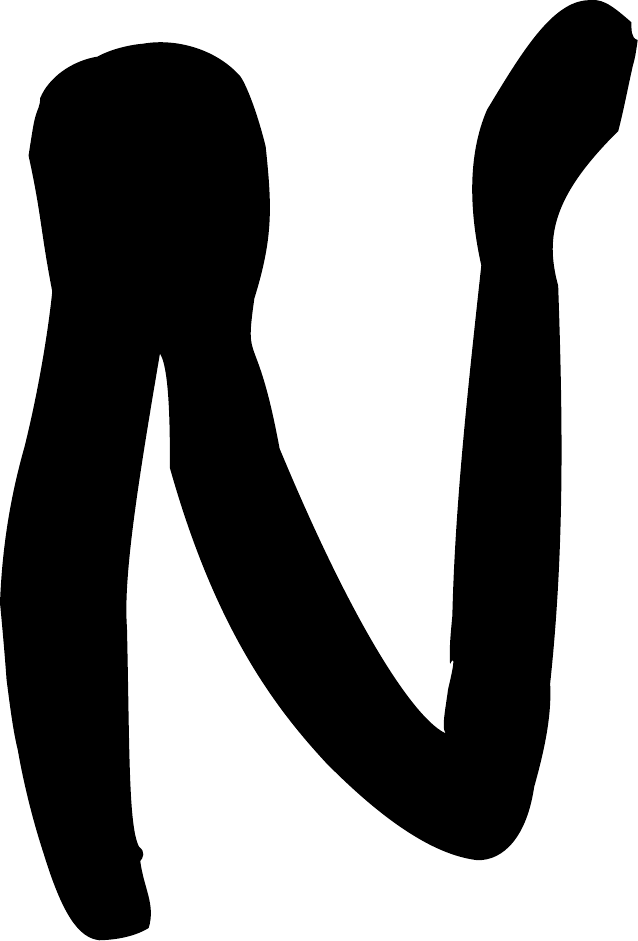} &
    \includegraphics[height=0.1\linewidth]{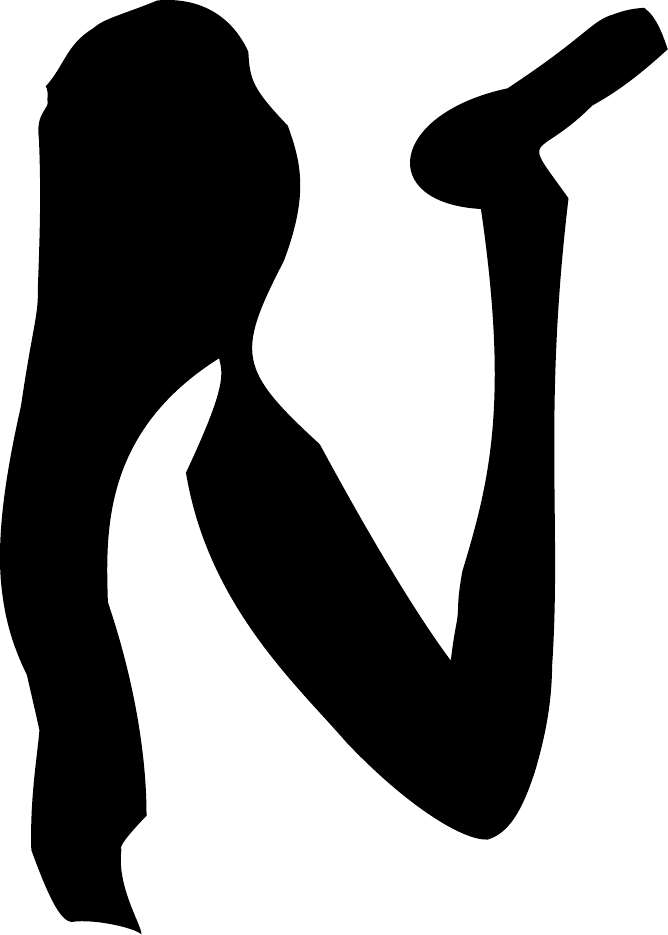} &
    \includegraphics[height=0.1\linewidth]{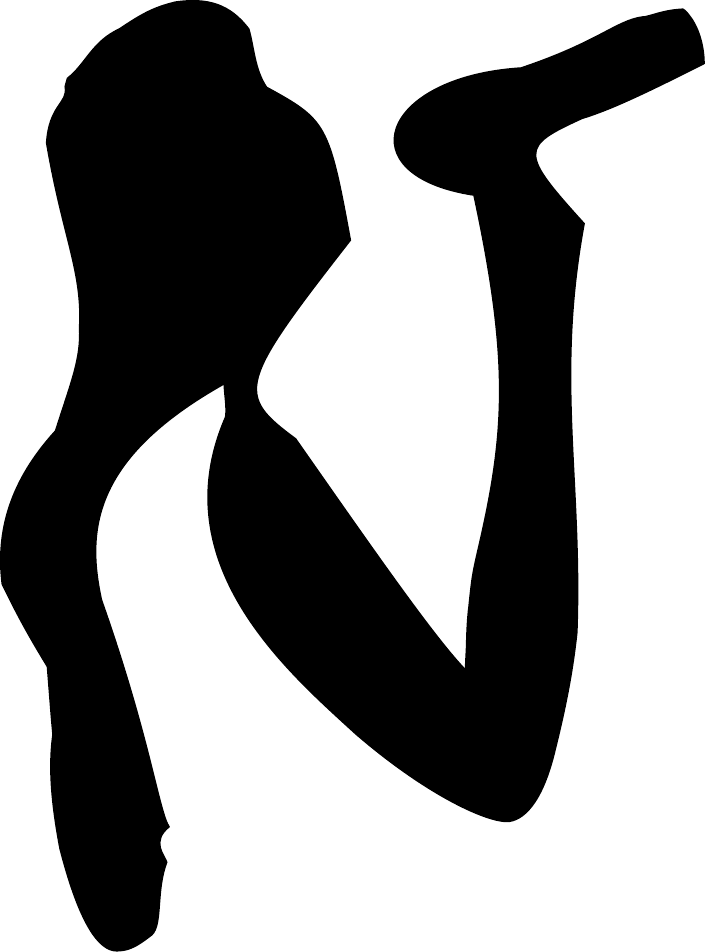} &
    \includegraphics[height=0.1\linewidth]{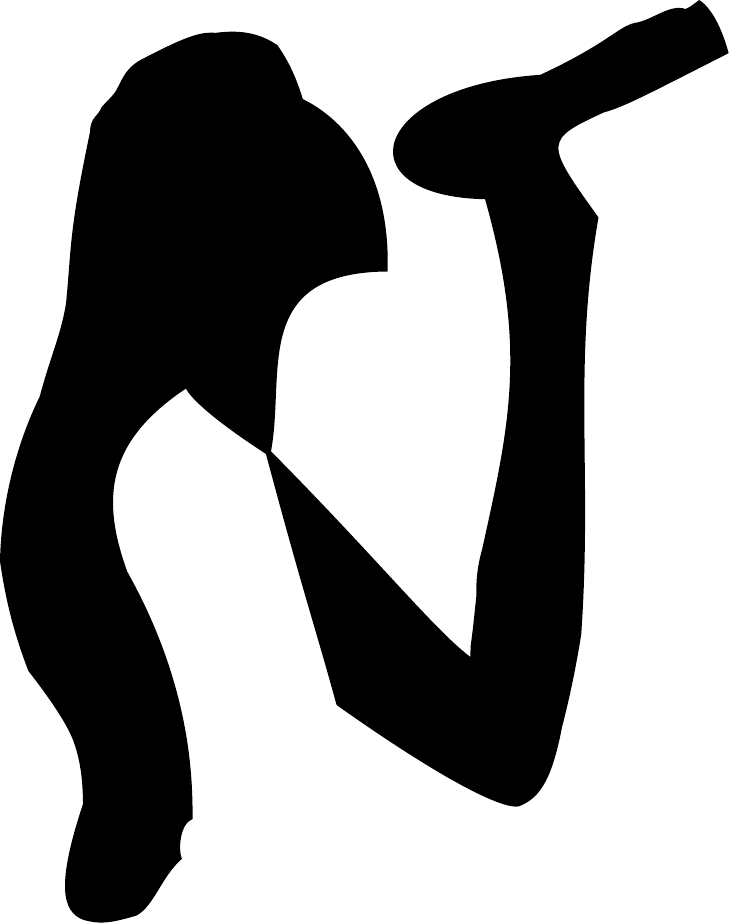} &
    \includegraphics[height=0.1\linewidth]{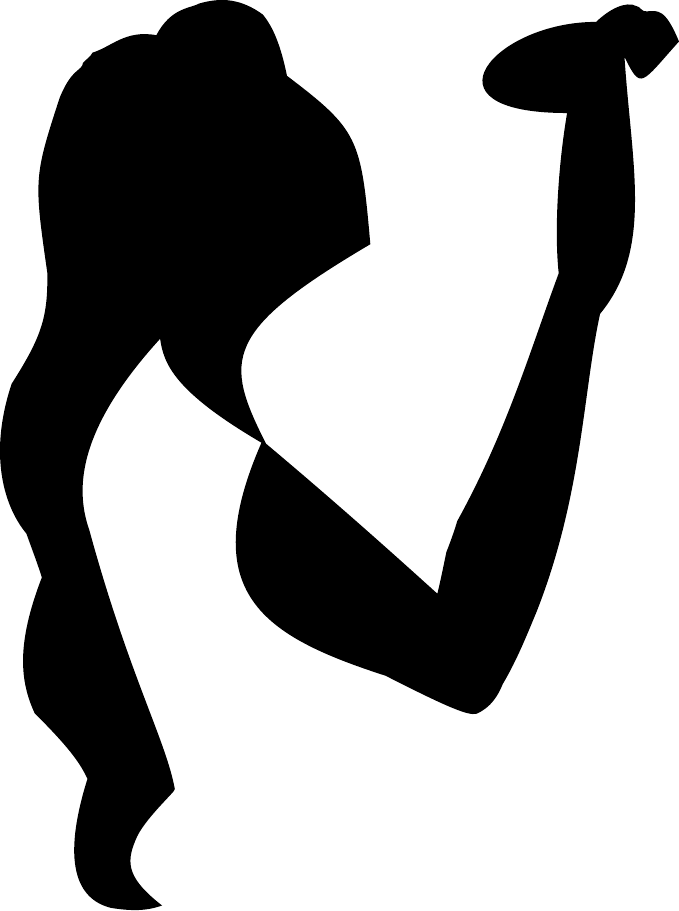} \\

    \raisebox{0.4cm}{"Giraffe"} &
    \includegraphics[height=0.09\linewidth]{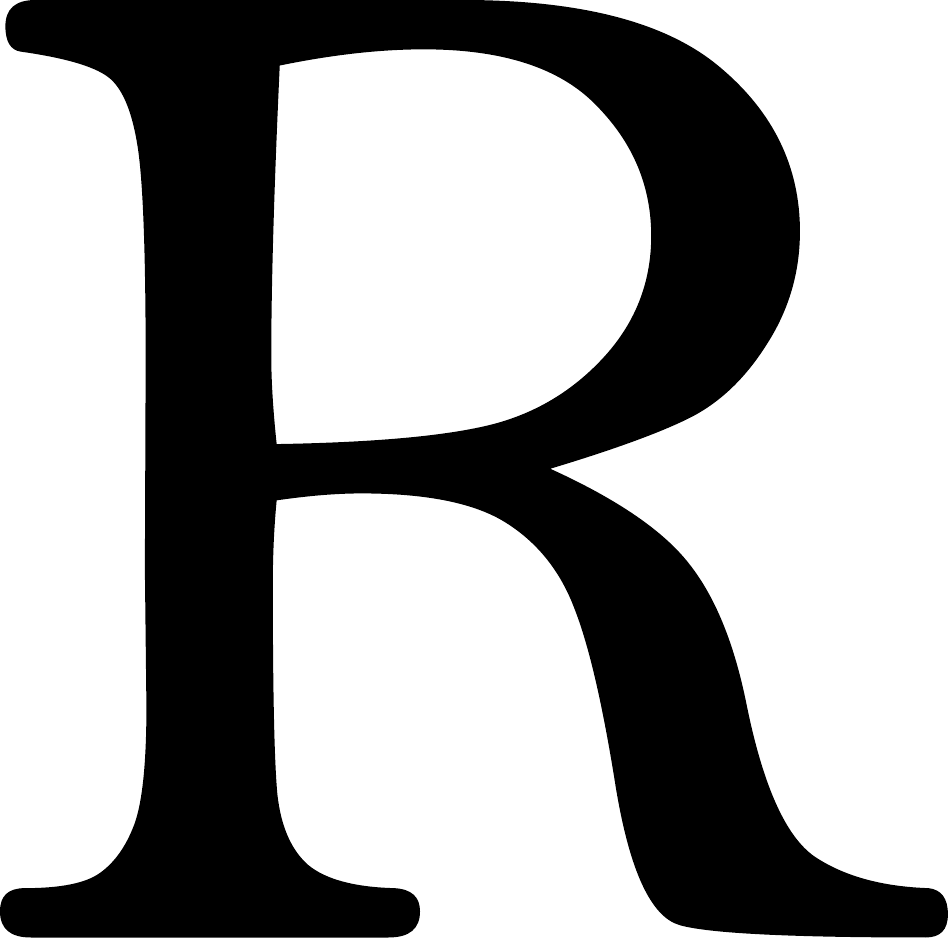} &
    \hspace{0.2cm}
    \includegraphics[height=0.1\linewidth]{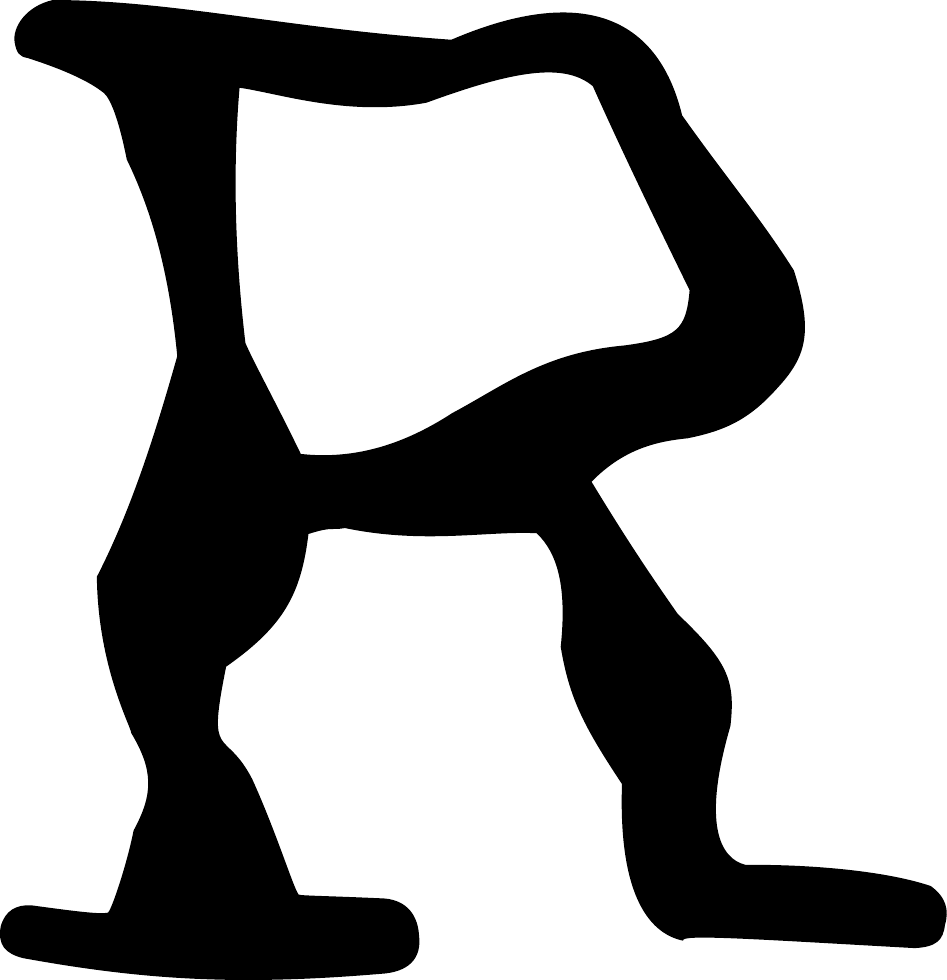} &
    \includegraphics[height=0.1\linewidth]{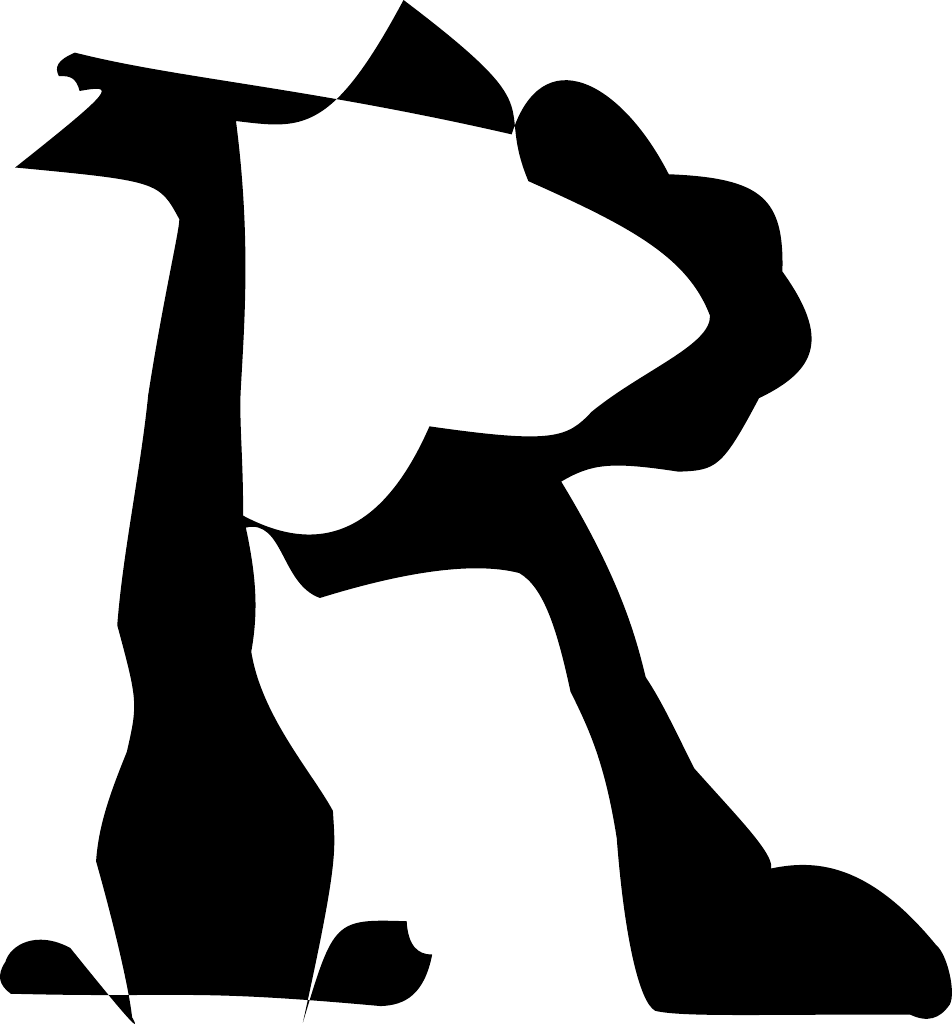} &
    \includegraphics[height=0.1\linewidth]{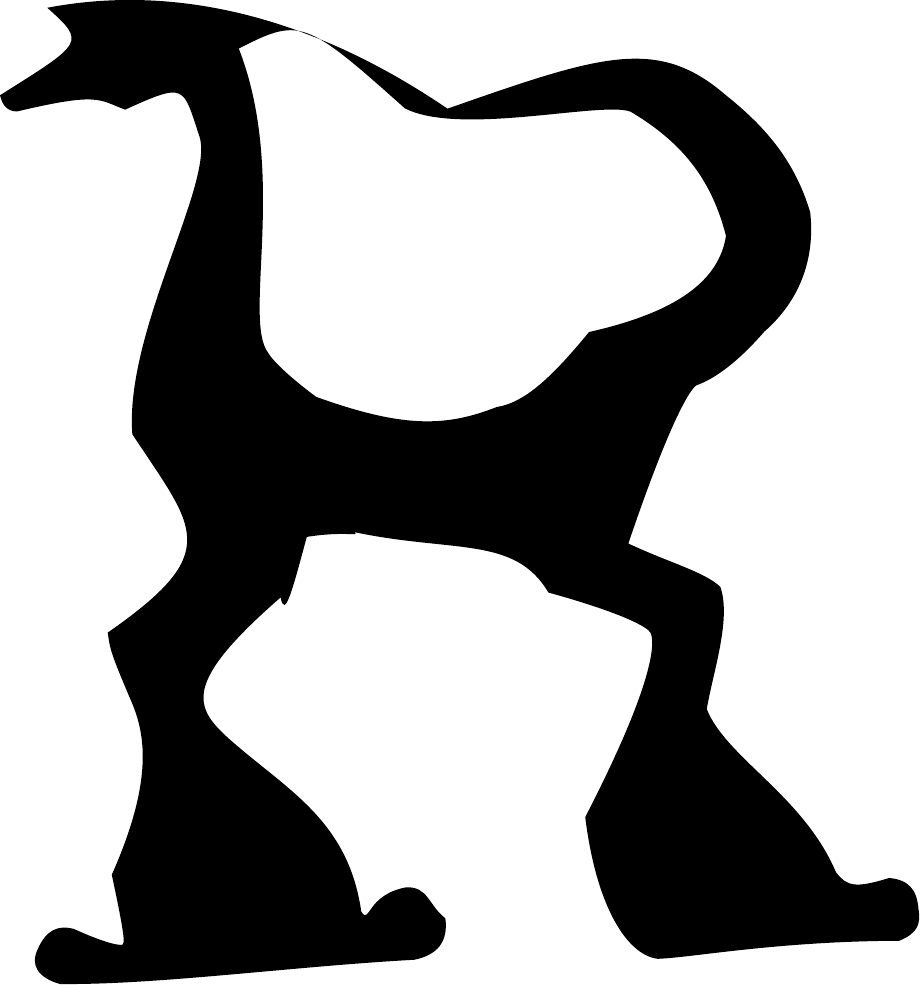} &
    \includegraphics[height=0.1\linewidth]{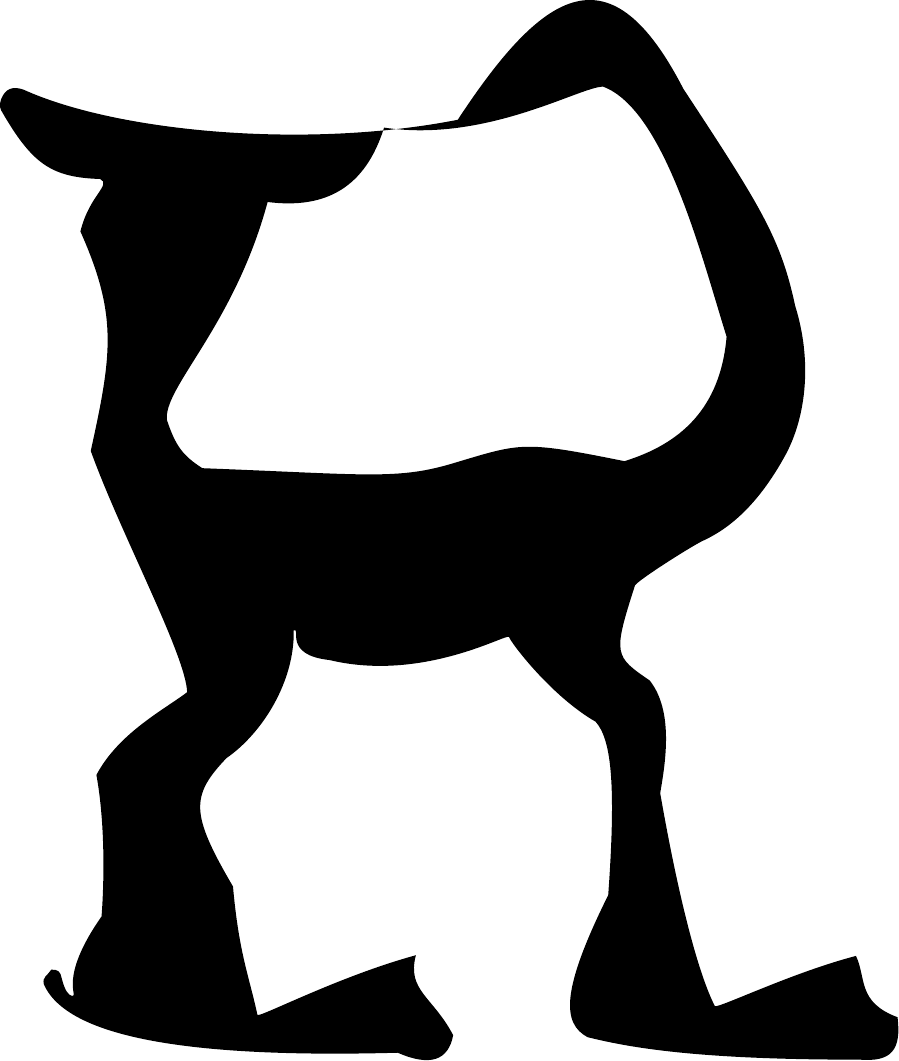} &
    \includegraphics[height=0.1\linewidth]{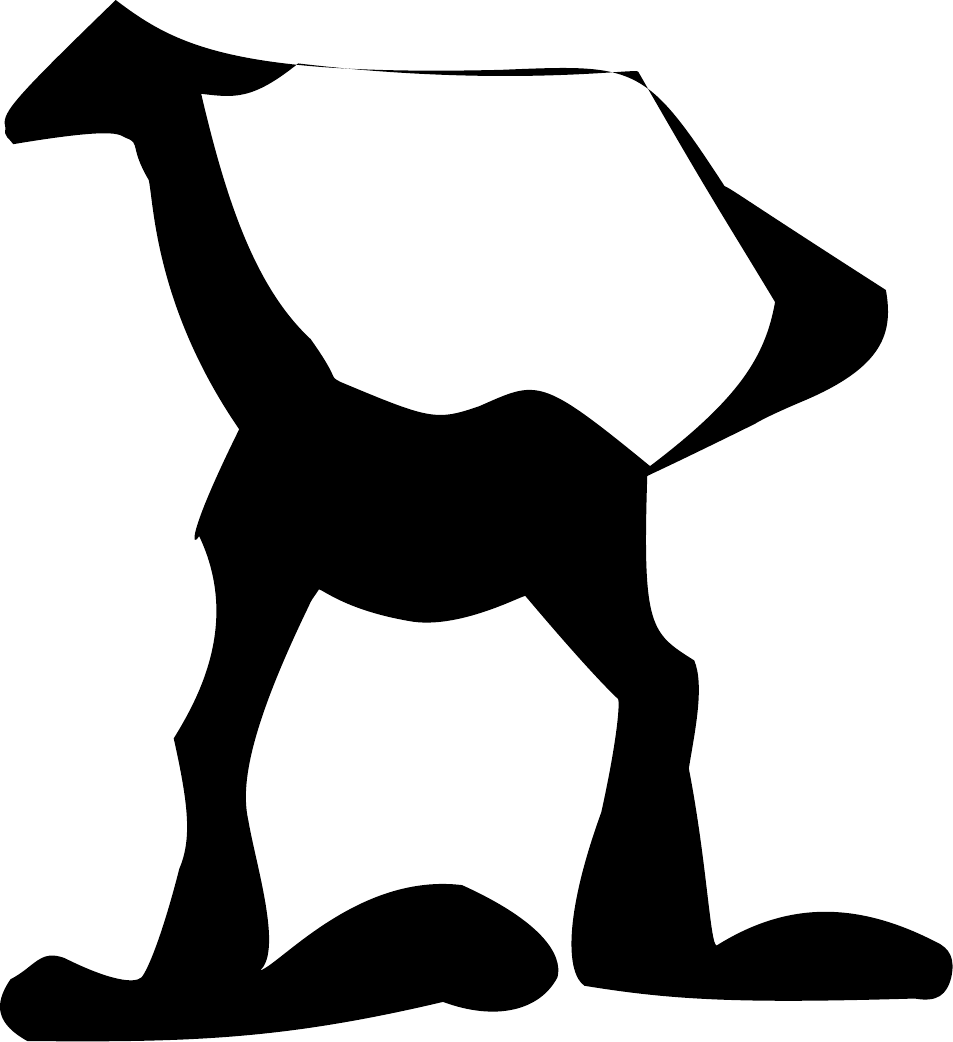} \\

    \multicolumn{2}{c}{Input} & 1 & 5 & 30 & 200 & \makecell[c]{Without \\ $\mathcal{L}_{tone}$}\\
        
\end{tabular}
\caption{Altering the $\sigma$ parameter of the low pass filter using in the $\mathcal{L}_{tone}$ loss. On the leftmost column are the original letters and concepts used, then from left to right are the results obtained when using $\sigma\in\{1, 5, 30, 200\}$, and without $\mathcal{L}_{tone}$.}
\label{fig:sigma_L2}
\end{figure}

\begin{figure}[t!]
\centering
    \setlength{\tabcolsep}{5pt}
    \renewcommand{\arraystretch}{1}
    \begin{tabular}{c c c c c c}
        
        \raisebox{0.4cm}{\makecell[l]{Input \\ Letter}} & 
        \includegraphics[height=0.098\linewidth]{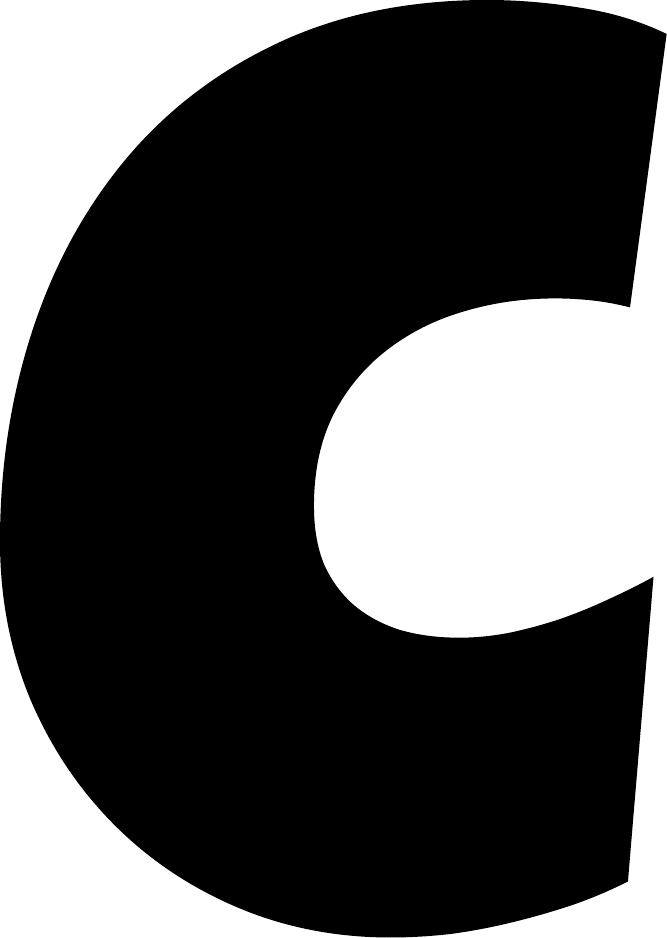} &
        \includegraphics[height=0.098\linewidth]{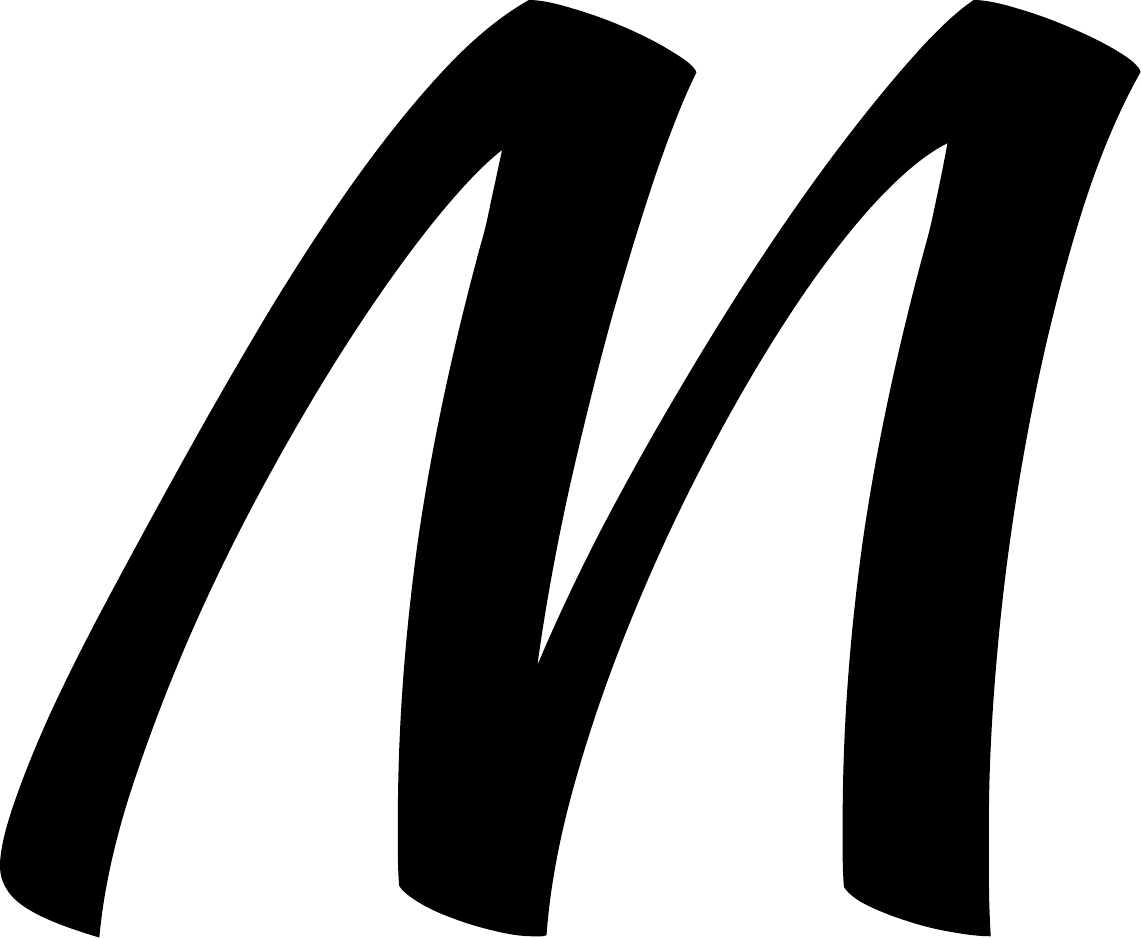} &
        \includegraphics[height=0.098\linewidth]{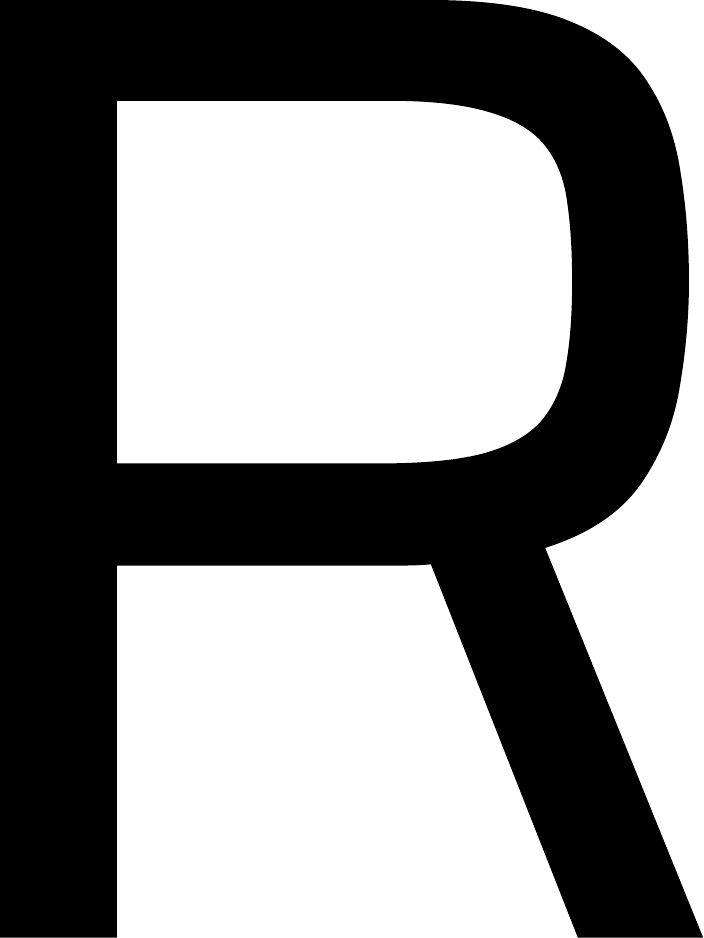} &
        \includegraphics[height=0.098\linewidth]{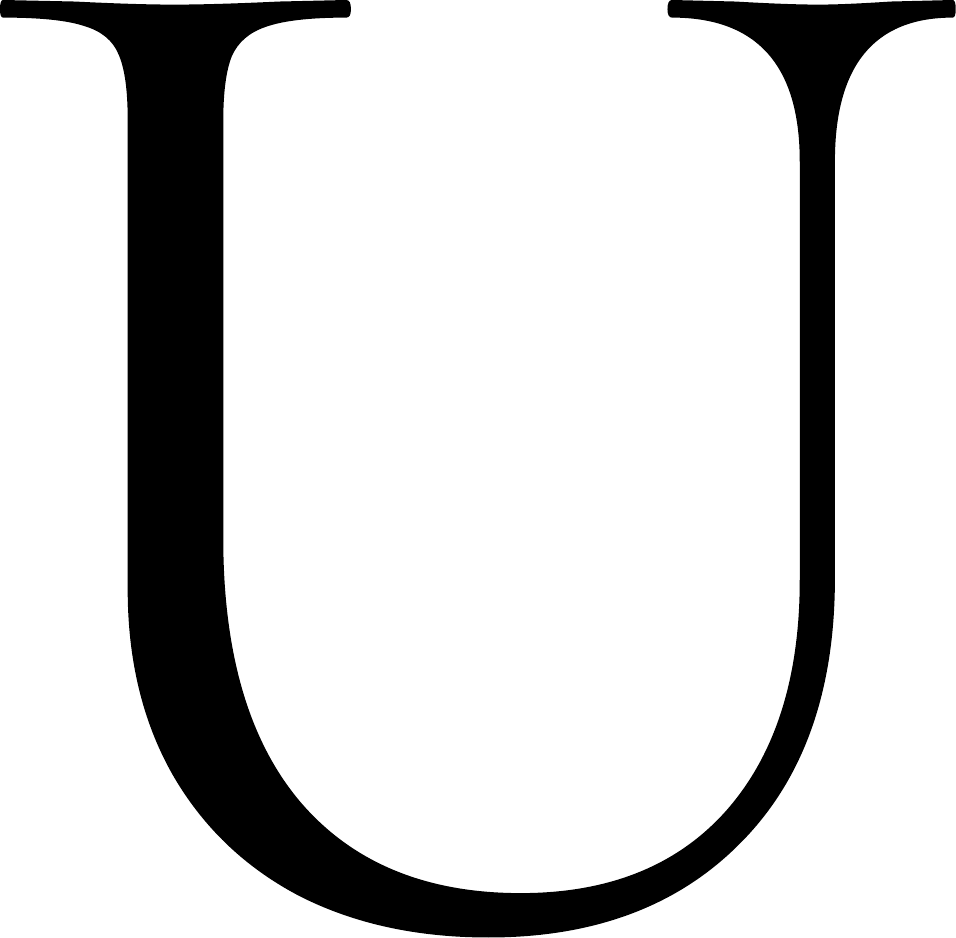} &
        \includegraphics[height=0.1\linewidth]{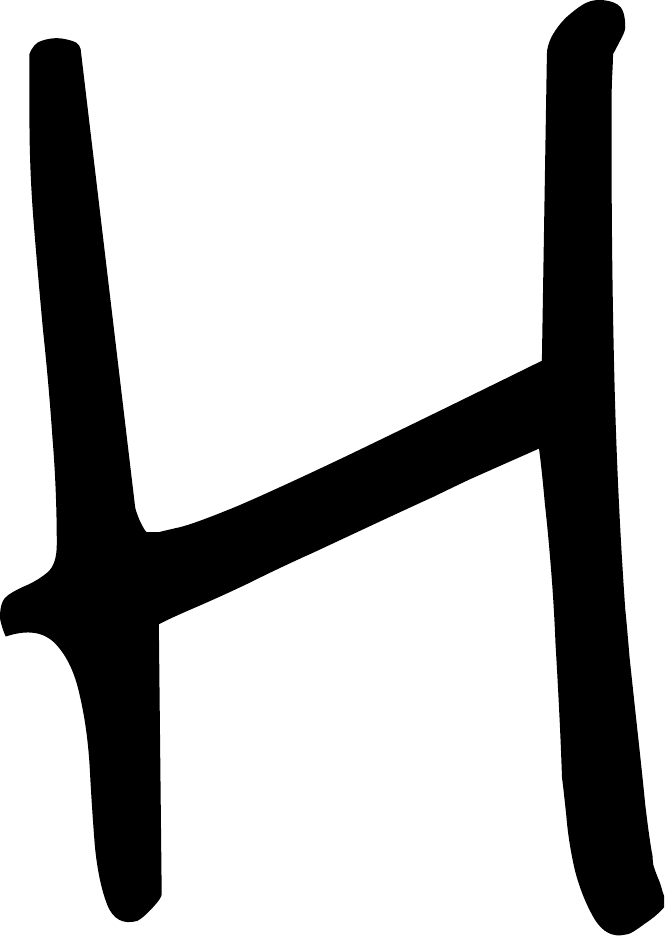} \\
        \midrule
        \raisebox{0.4cm}{Ours} & 
        \includegraphics[height=0.1\linewidth]{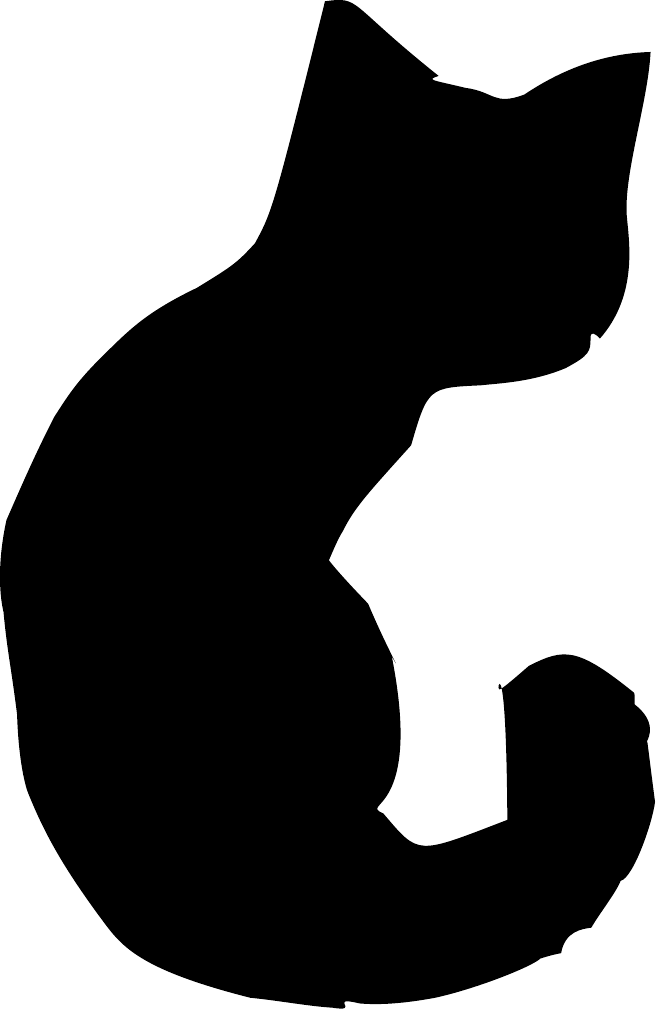} & 
        \includegraphics[height=0.1\linewidth]{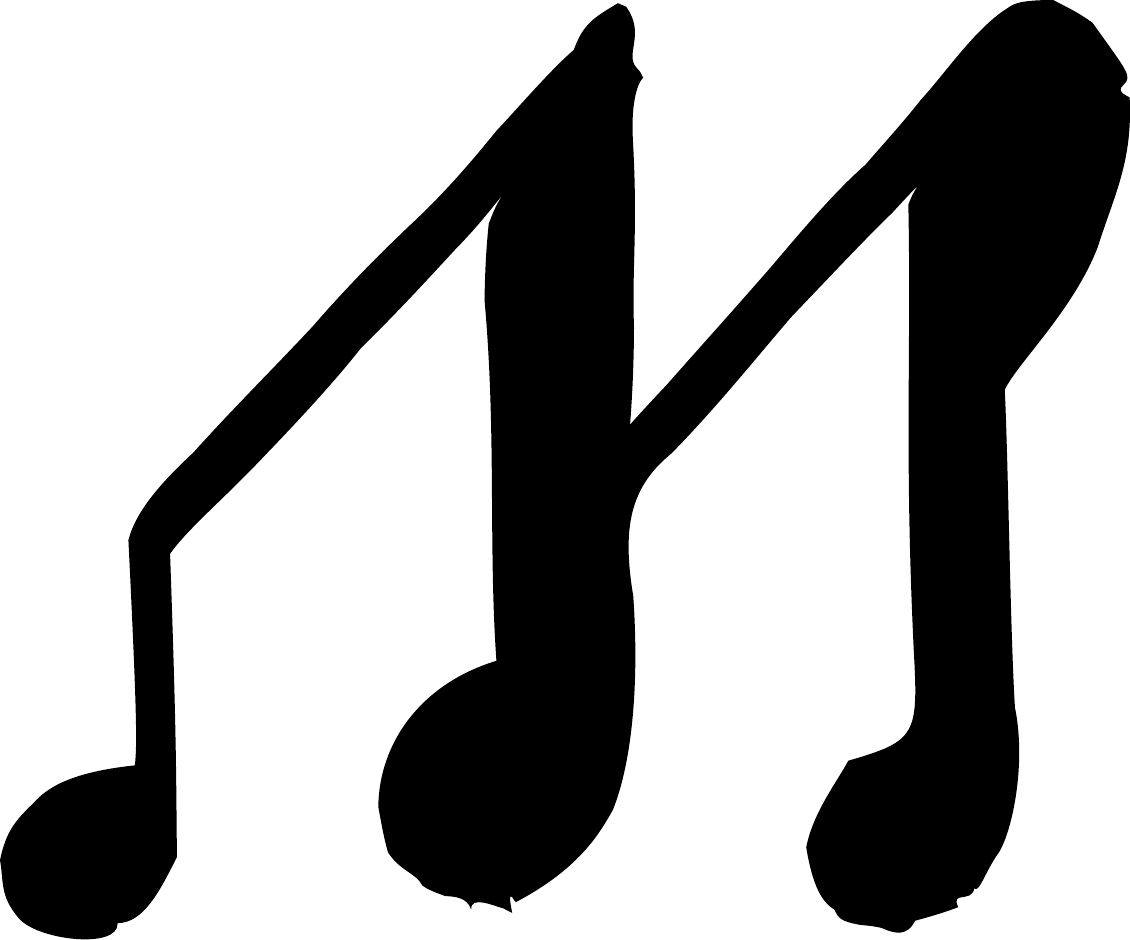} &
        \includegraphics[height=0.1\linewidth]{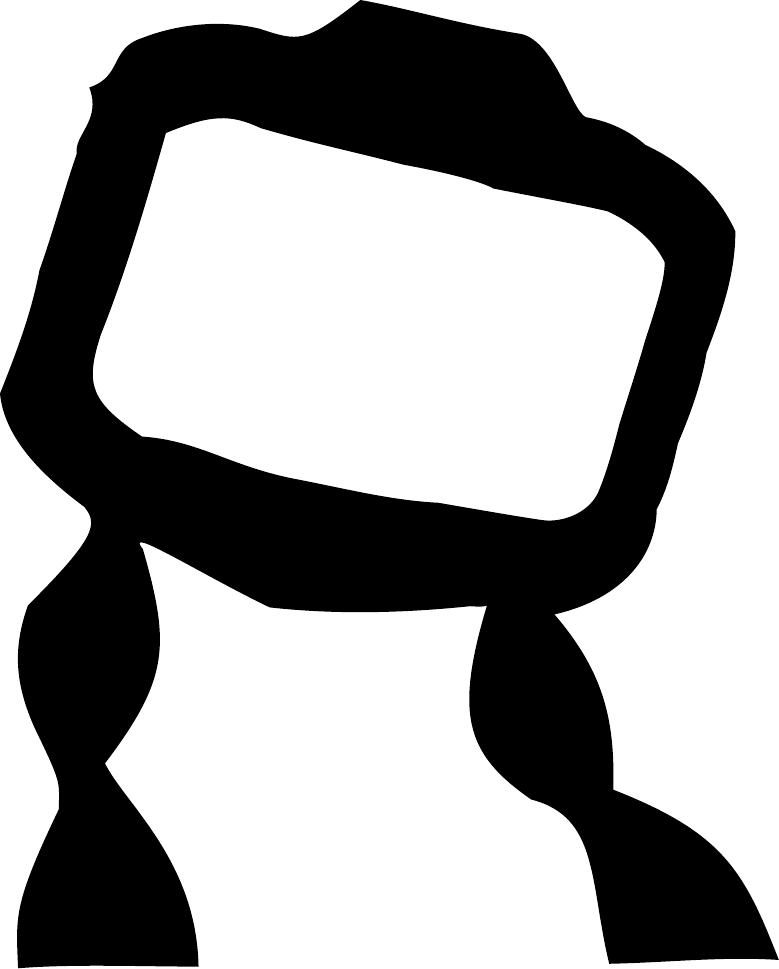} &
        \includegraphics[height=0.1\linewidth]{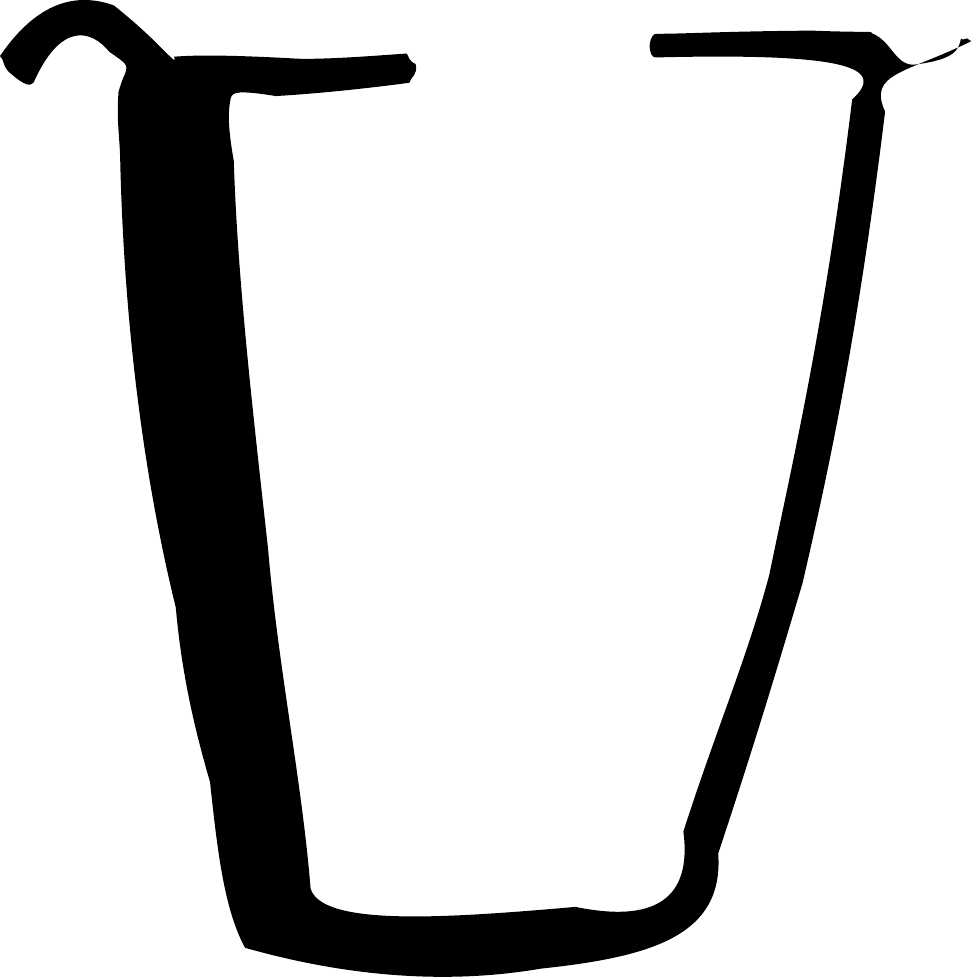} &
        \includegraphics[height=0.1\linewidth]{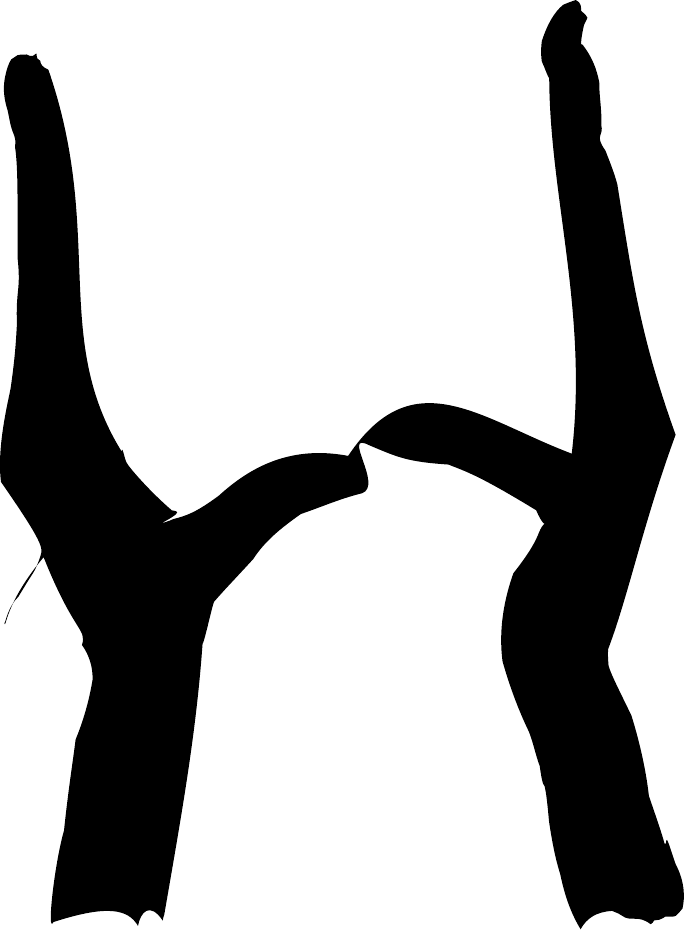} \\
        \raisebox{0.4cm}{\makecell[l]{Only \\ SDS}}  & 
        \includegraphics[height=0.1\linewidth]{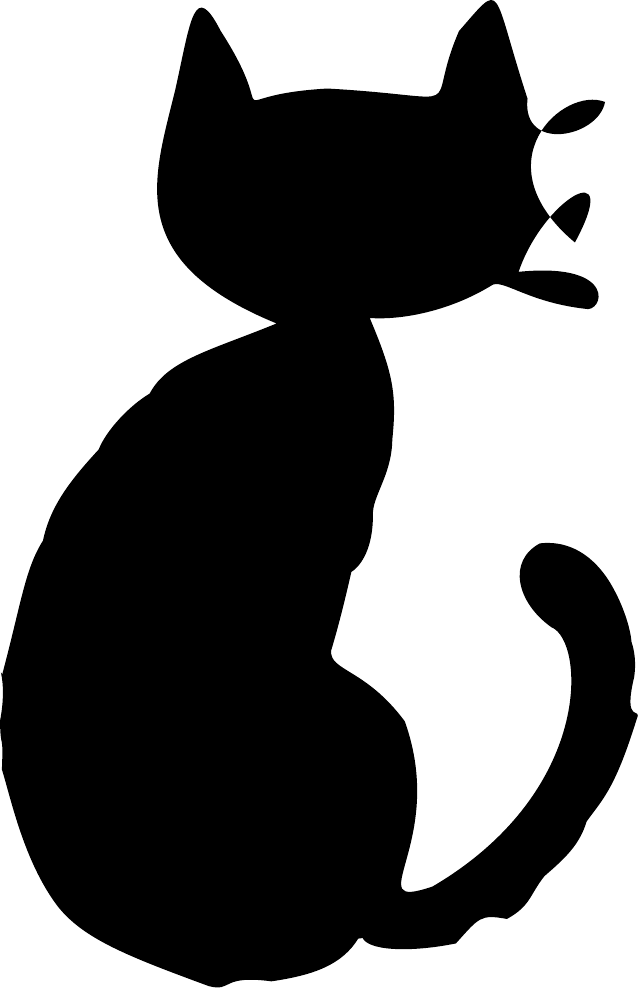} & 
        \includegraphics[height=0.1\linewidth]{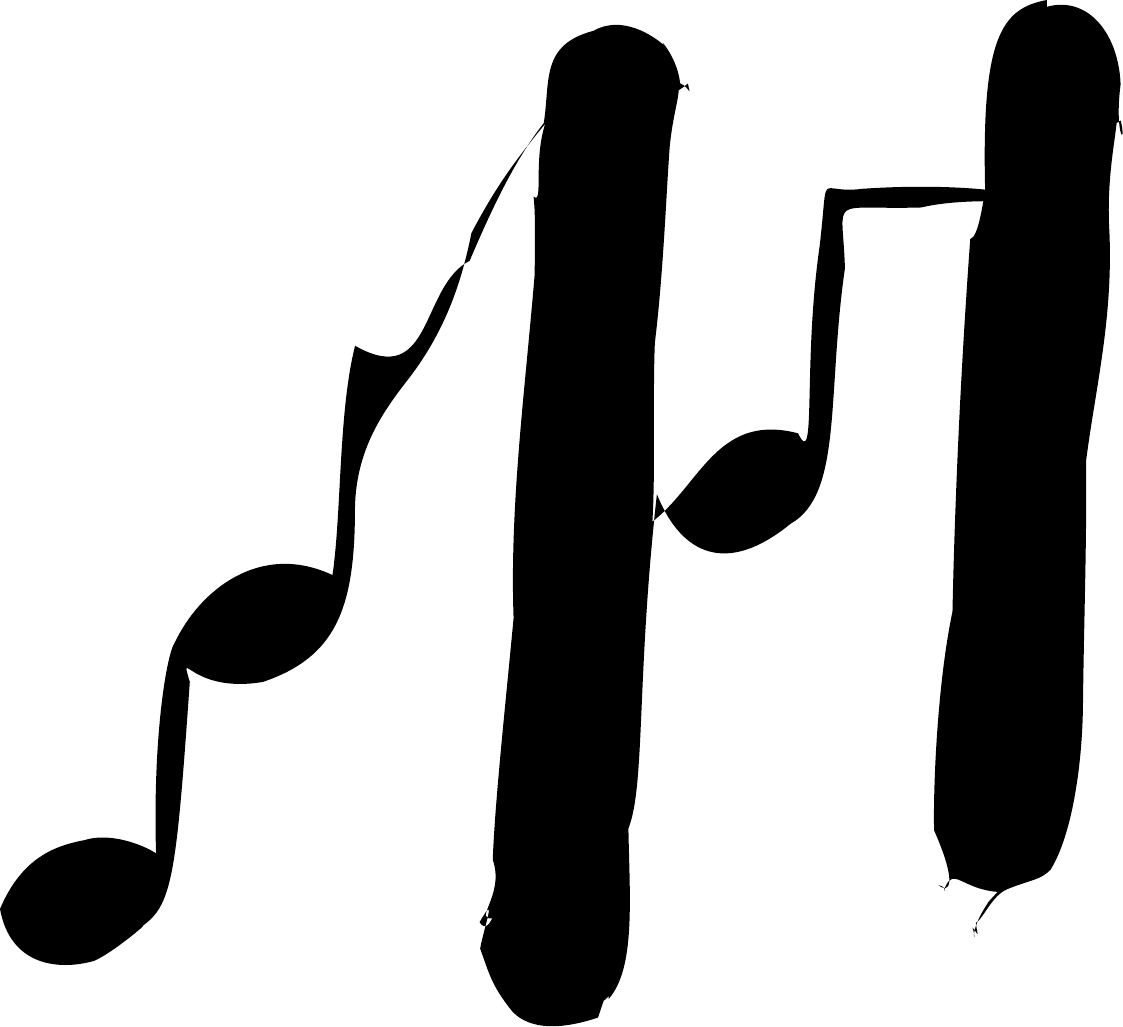} & 
        \includegraphics[height=0.1\linewidth]{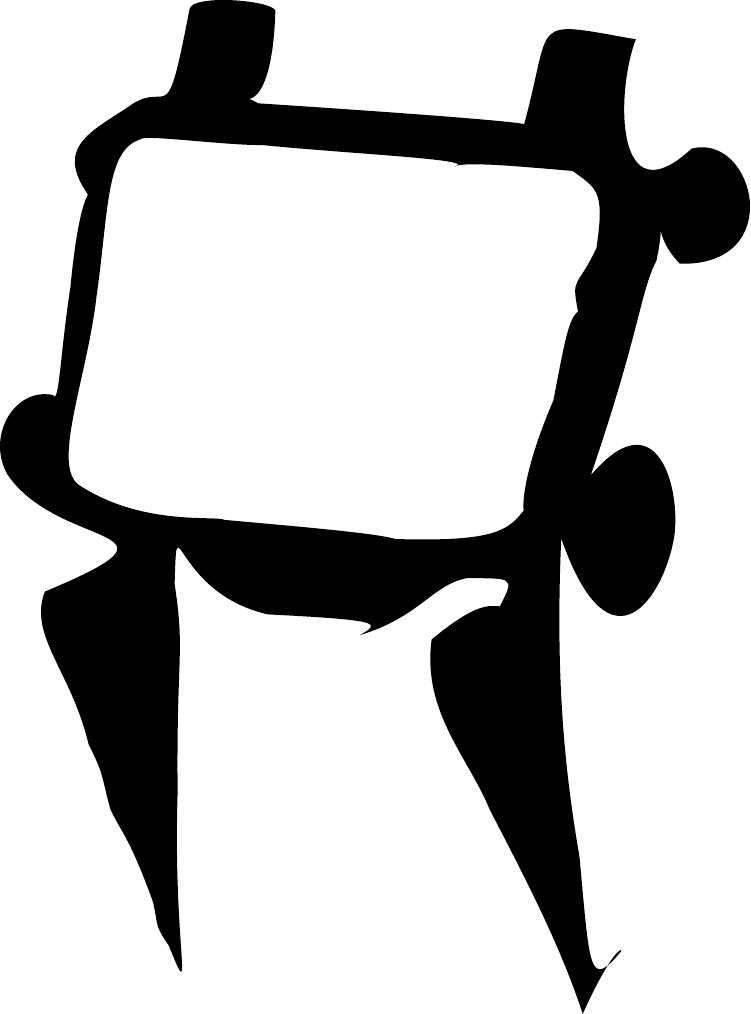} &
        \includegraphics[height=0.1\linewidth]{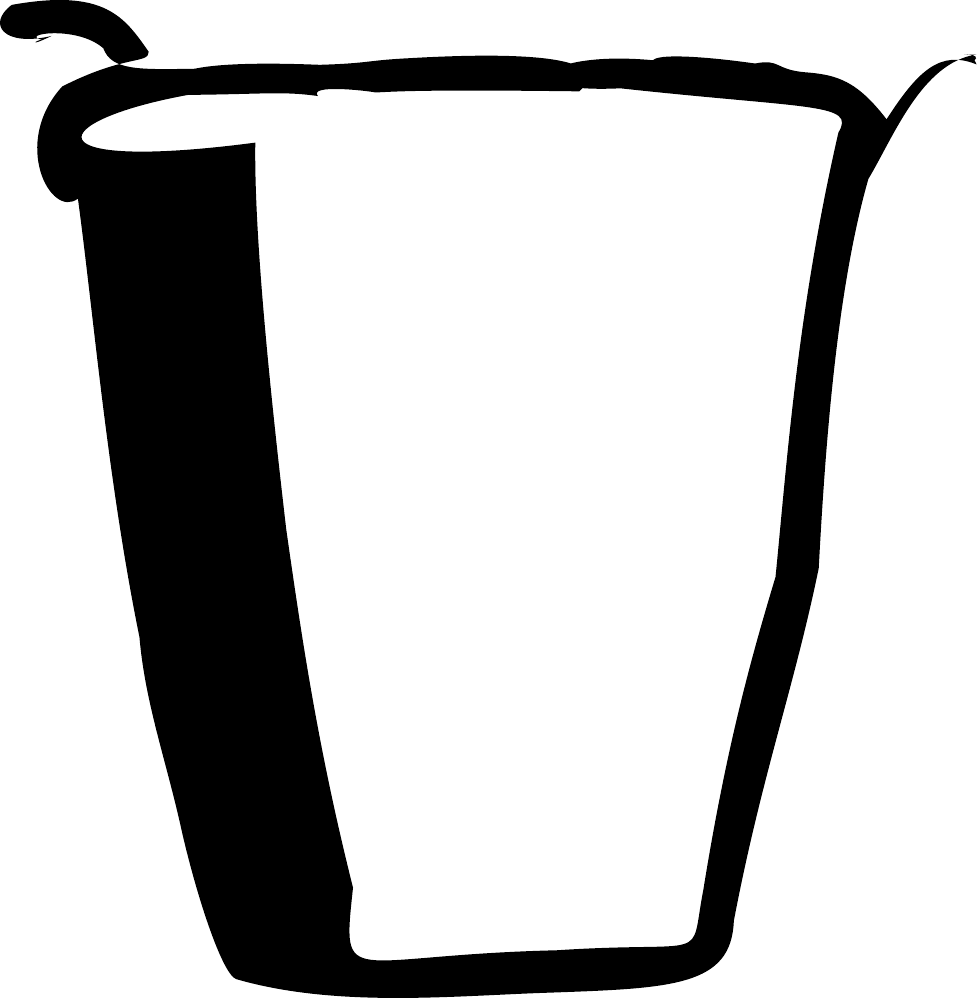} &
        \includegraphics[height=0.1\linewidth]{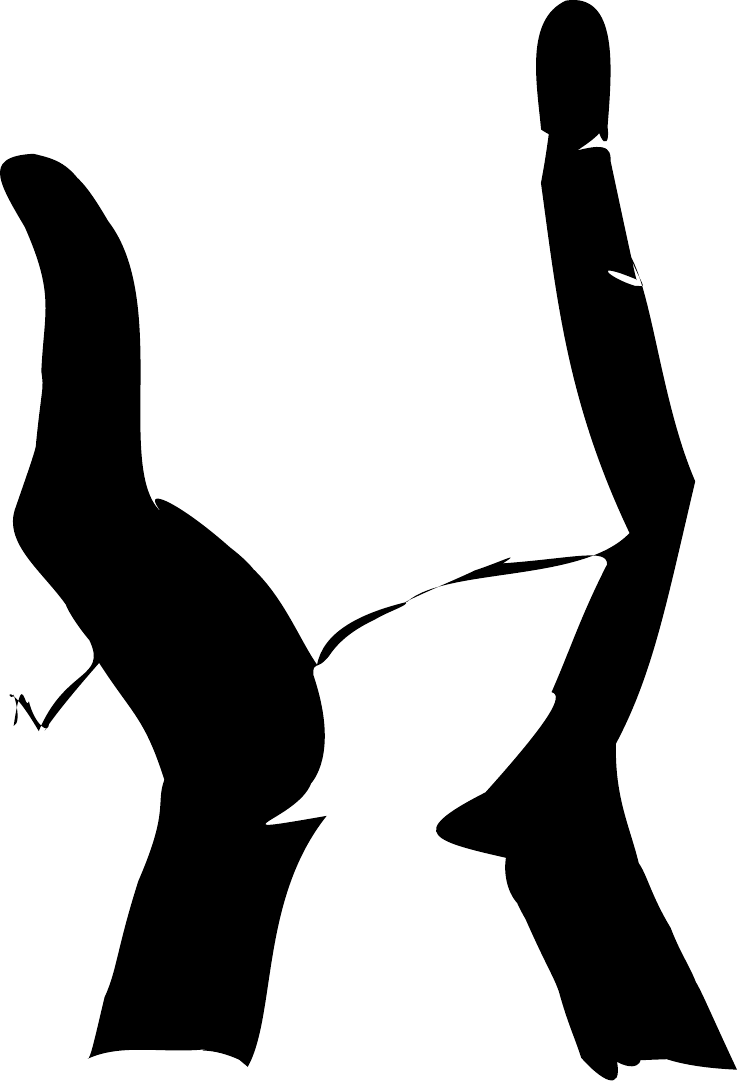} \\
        & "Cat" & "Music" & "Robot" & "Cup" & "Hands" \\
    \end{tabular}
    \caption{The effect of using only the SDS loss: note how the third row simply looks like icon illustrations, while the second row still resembles legible letters. }
    \label{fig:sds_lr}
\end{figure}

\begin{figure}[ht]
\centering
\setlength{\tabcolsep}{5pt}
\renewcommand{\arraystretch}{1} 
\begin{tabular}{c c | c c c c c}
    \raisebox{0.4cm}{"Bear"} &
    \raisebox{0.01cm}{\includegraphics[height=0.09\linewidth]{images/weight_effect/Bell_MT_B_scaled.pdf}} &
    \hspace{0.2cm}
    \includegraphics[height=0.1\linewidth]{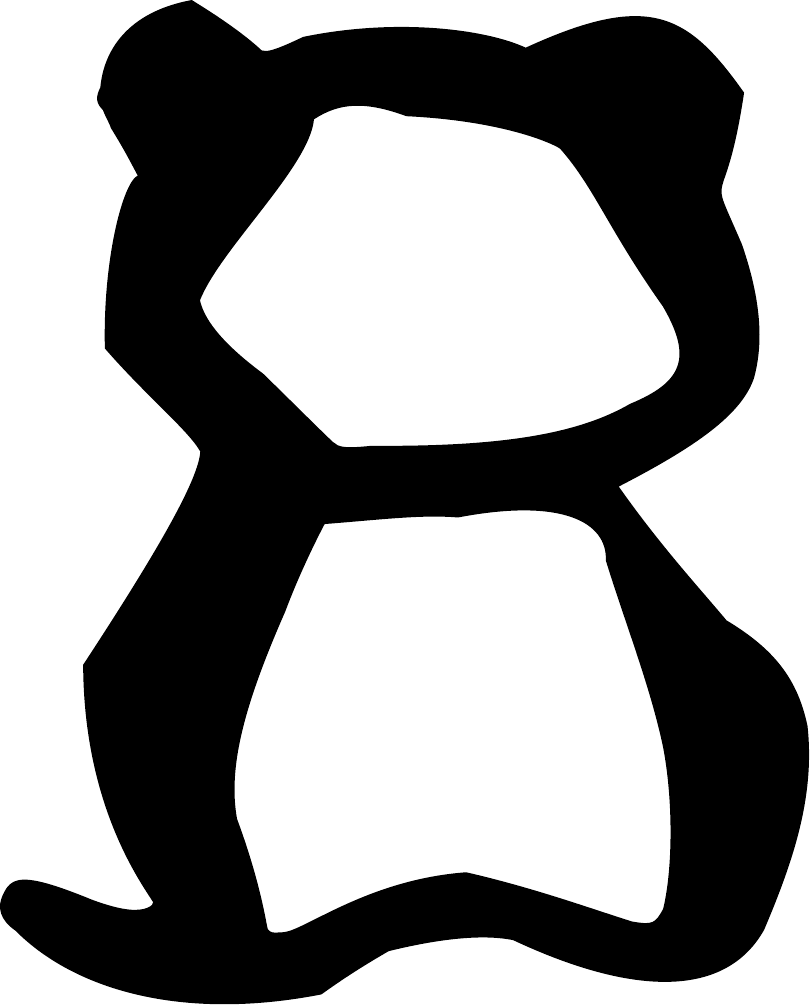} &
    \includegraphics[height=0.1\linewidth]{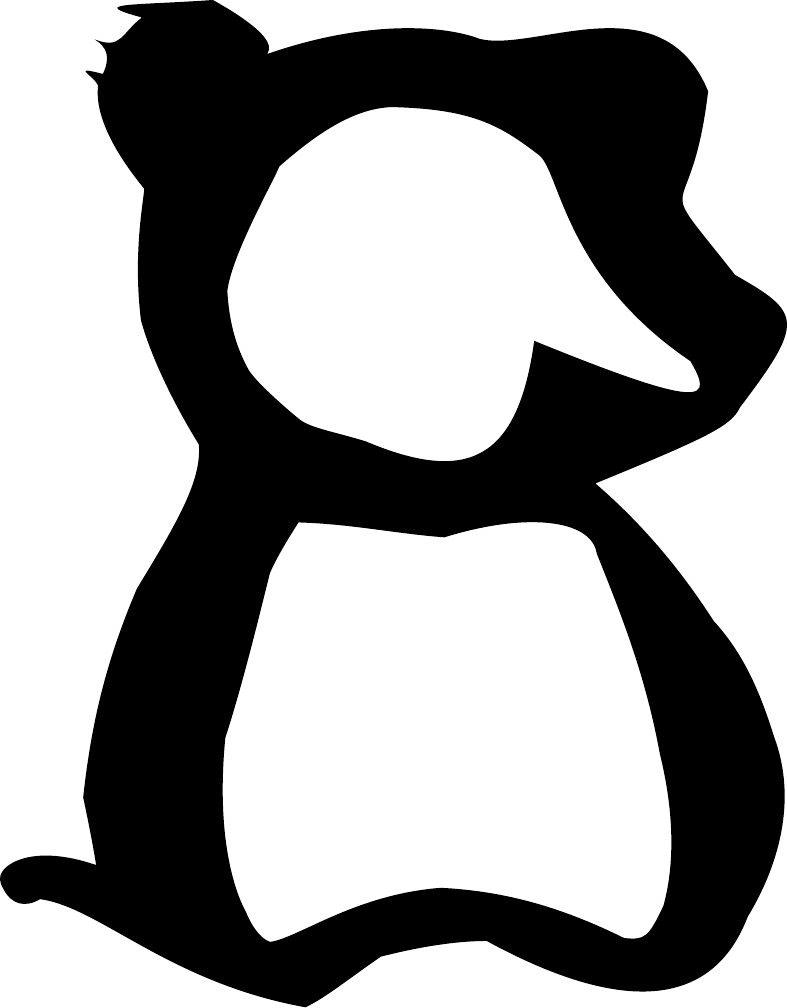} &
    \includegraphics[height=0.1\linewidth]{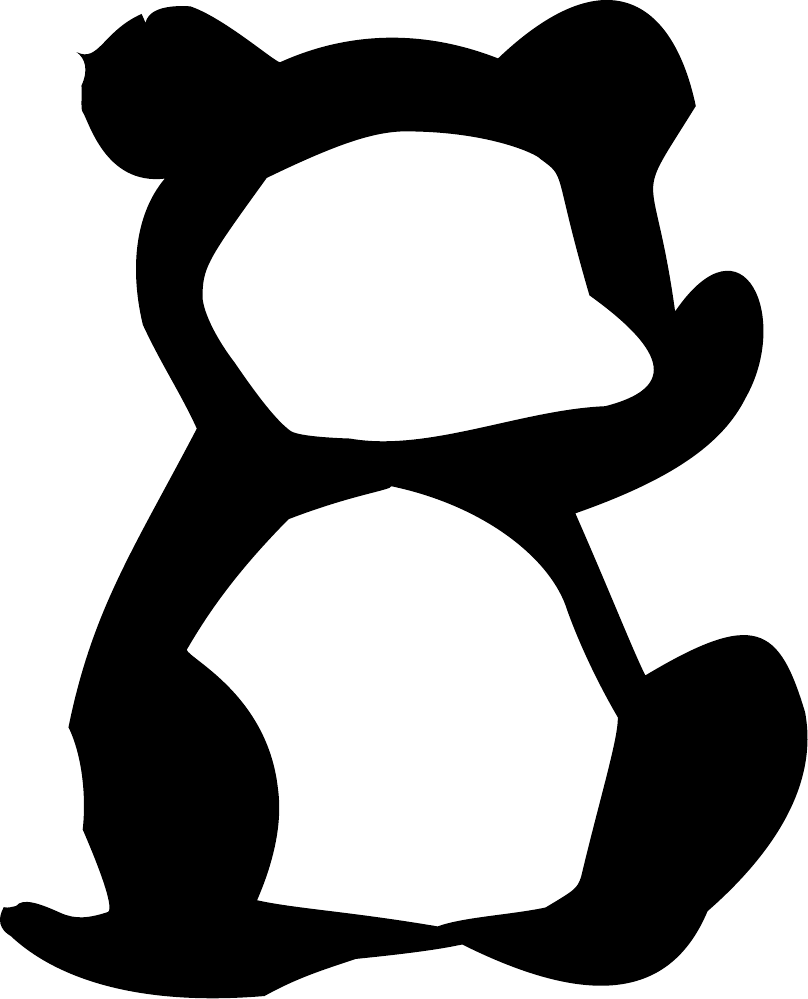} &
    \includegraphics[height=0.1\linewidth]{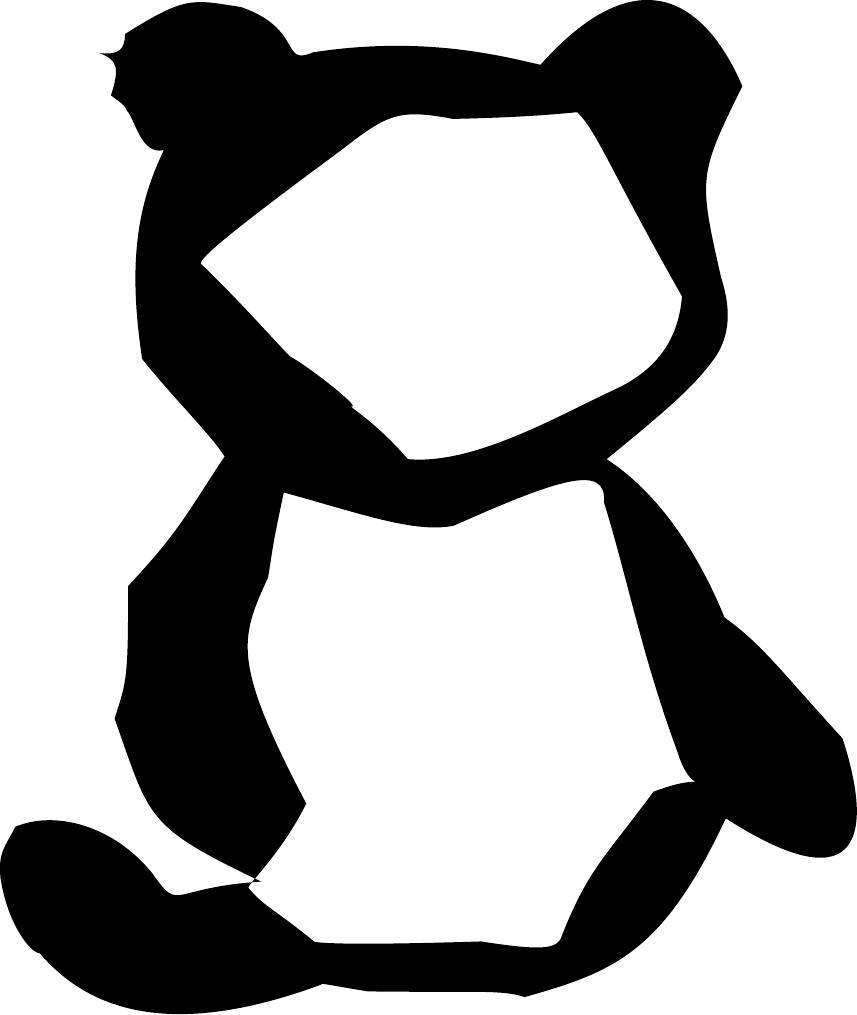} &
    \includegraphics[height=0.1\linewidth]{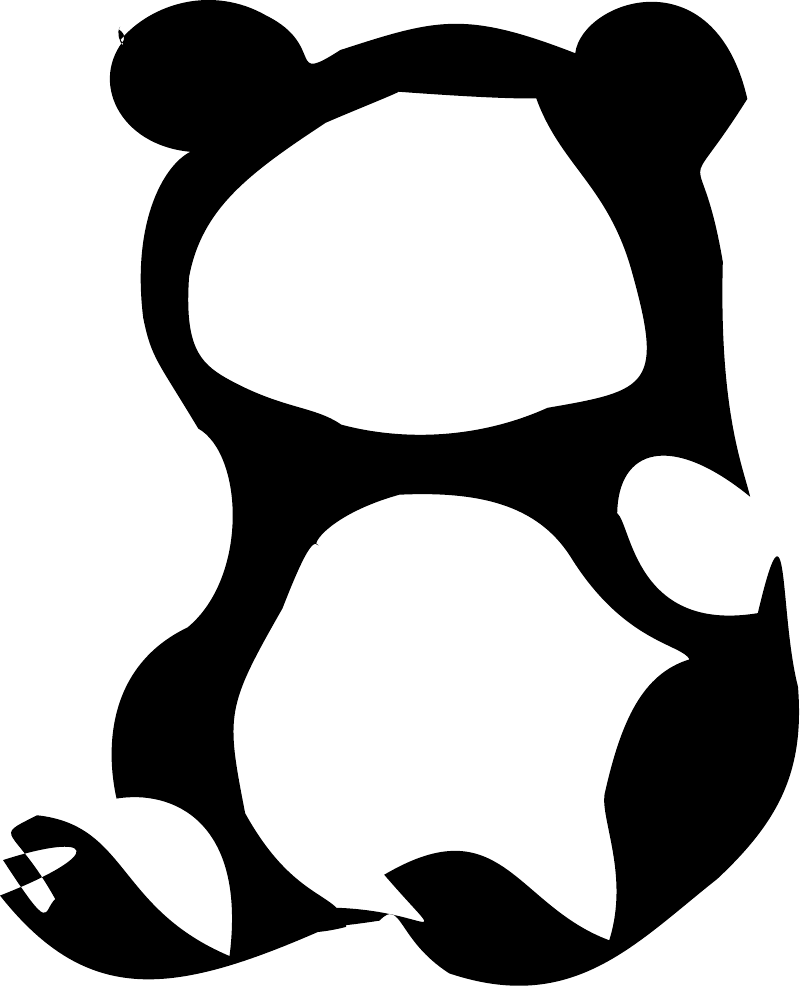} \\

    \raisebox{0.4cm}{"Singer"} &
    \includegraphics[height=0.09\linewidth]{images/weight_effect/Noteworthy-Bold_N_scaled.pdf} &
    \hspace{0.2cm}
    \includegraphics[height=0.1\linewidth]{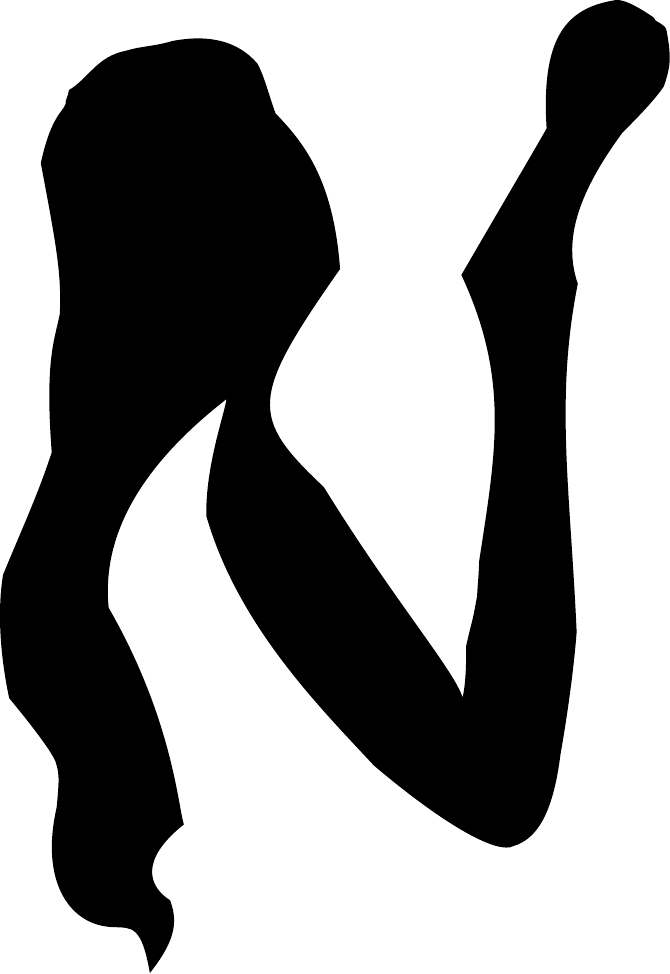} &
    \includegraphics[height=0.1\linewidth]{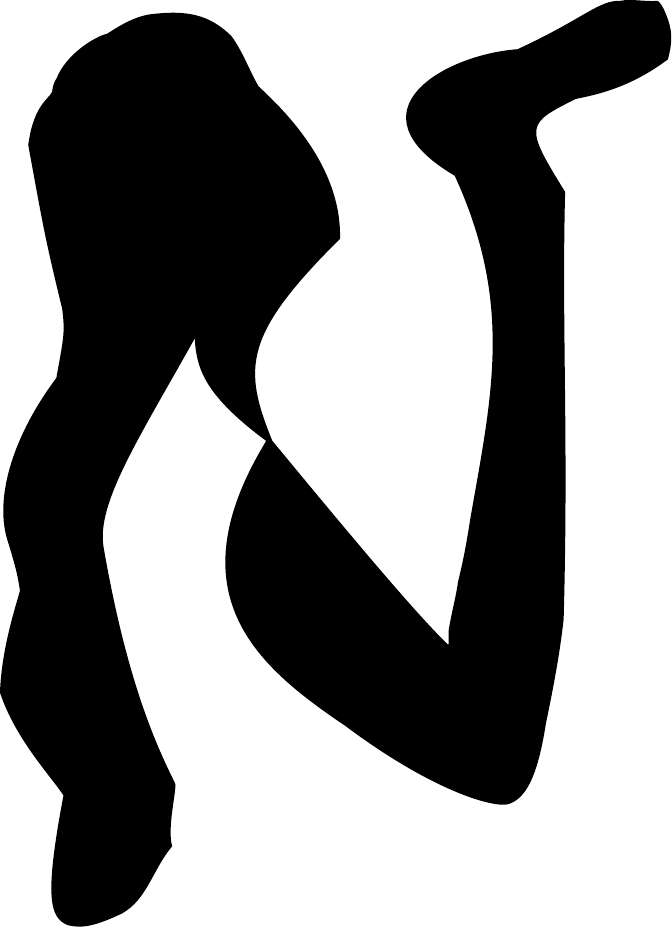} &
    \includegraphics[height=0.1\linewidth]{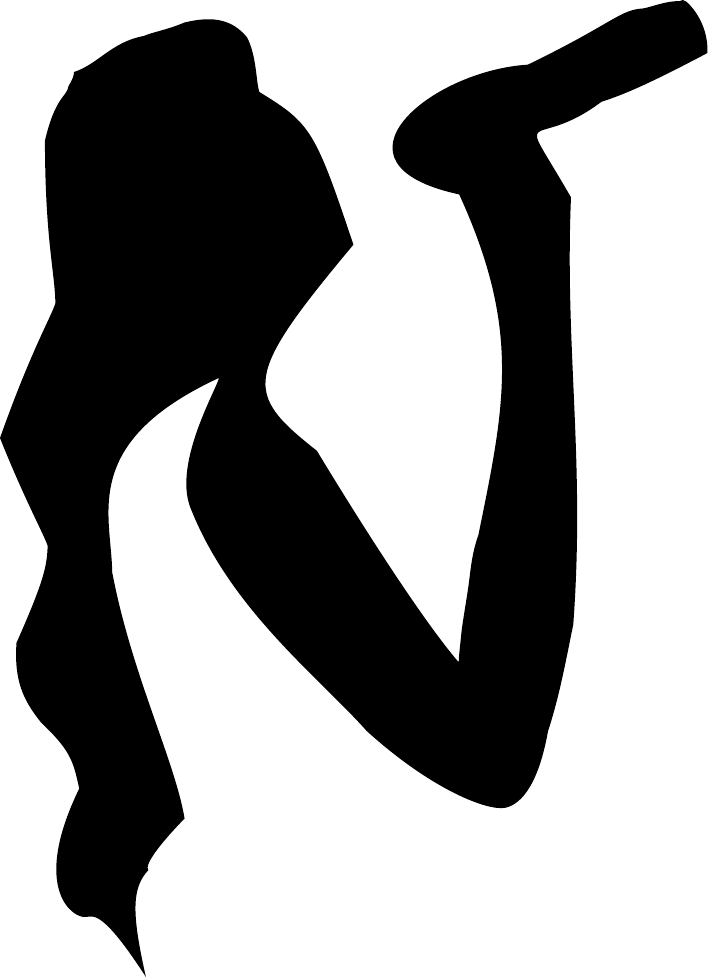} &
    \includegraphics[height=0.1\linewidth]{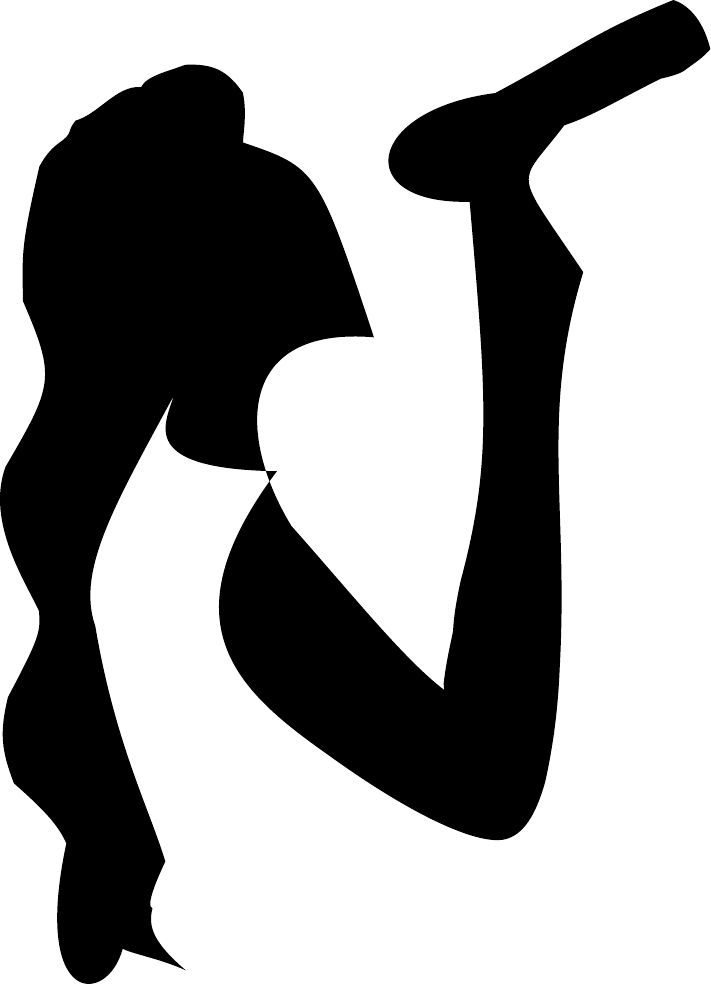} &
    \includegraphics[height=0.1\linewidth]{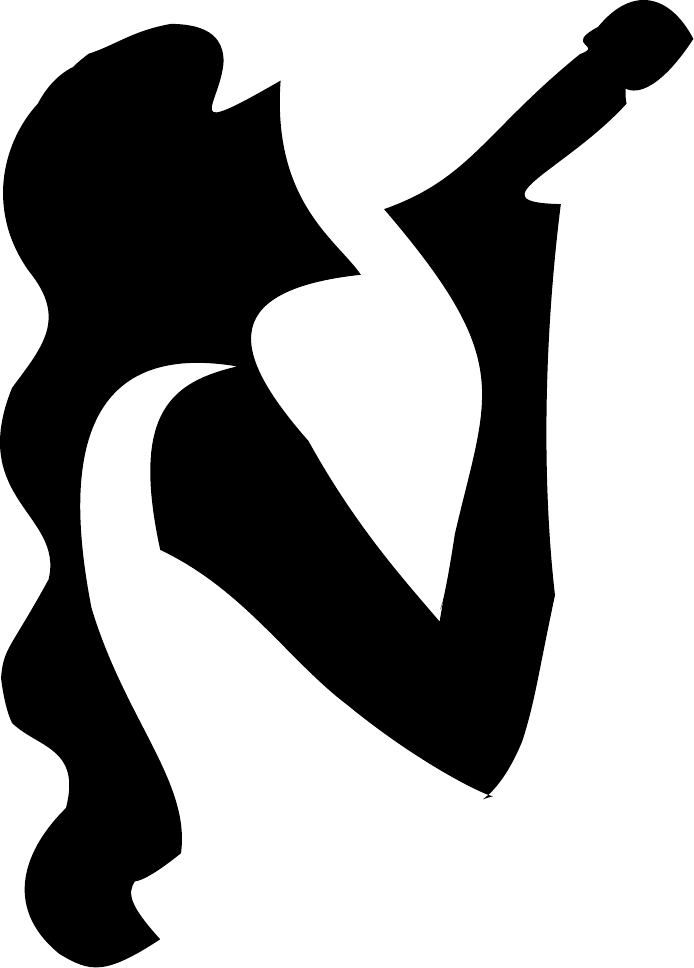} \\

    \raisebox{0.4cm}{"Giraffe"} &
    \includegraphics[height=0.09\linewidth]{images/weight_effect/Bell_MT_R_scaled.pdf} &
    \hspace{0.2cm}
    \includegraphics[height=0.1\linewidth]{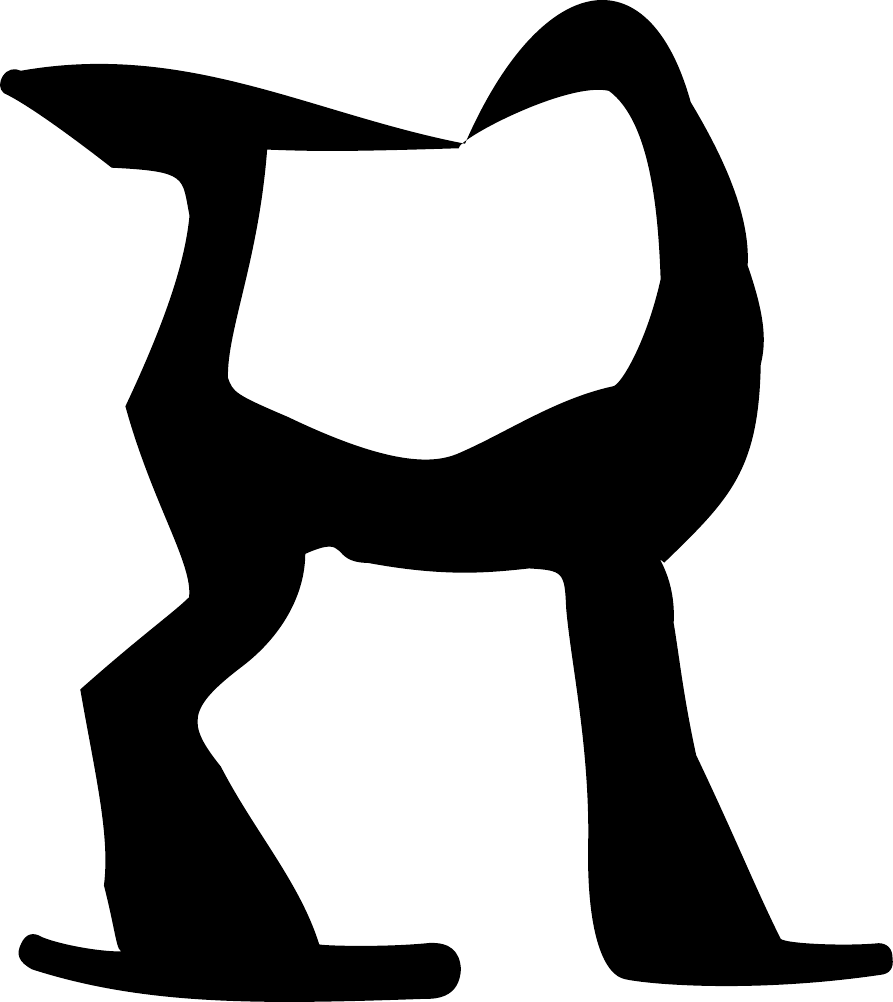} &
    \includegraphics[height=0.1\linewidth]{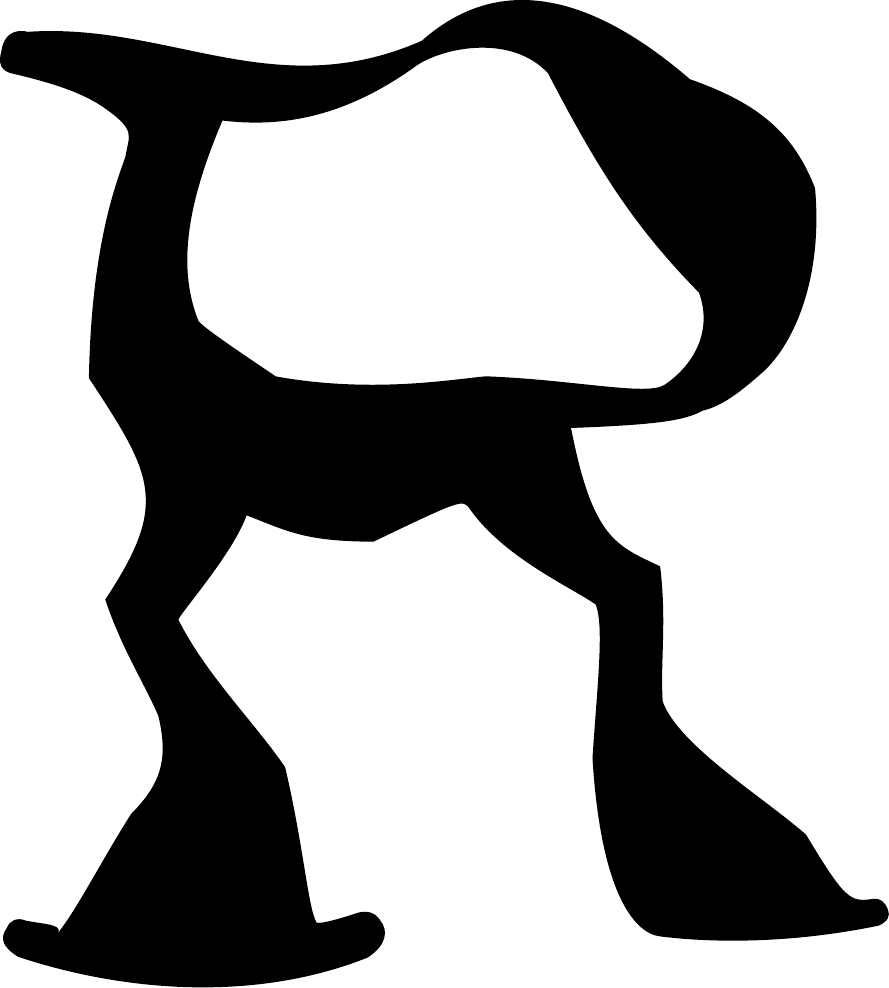} &
    \includegraphics[height=0.1\linewidth]{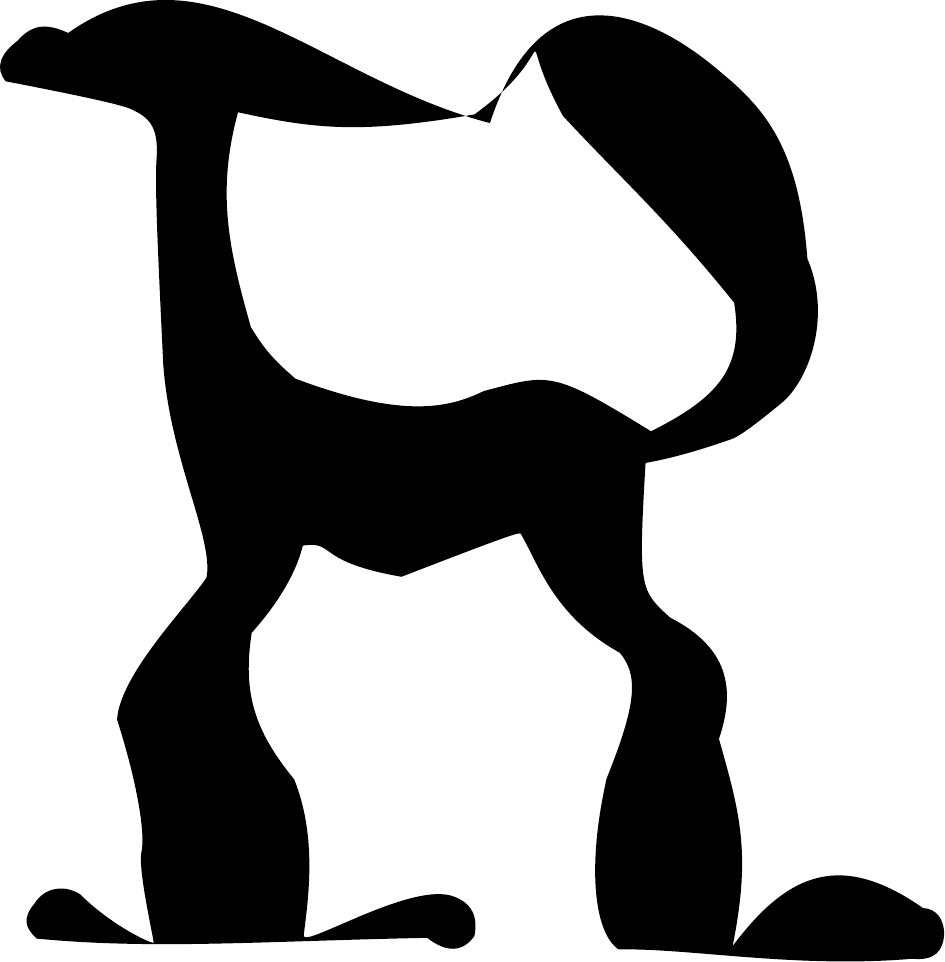} &
    \includegraphics[height=0.1\linewidth]{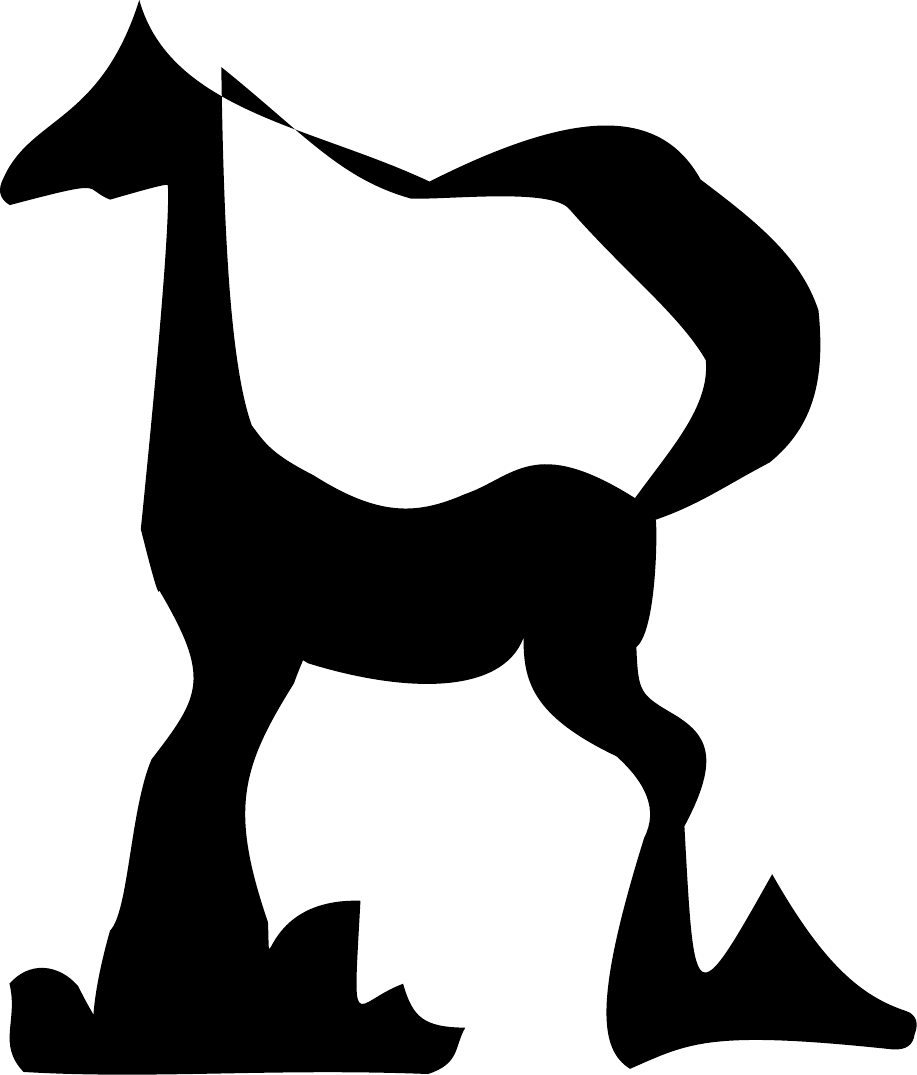} &
    \includegraphics[height=0.1\linewidth]{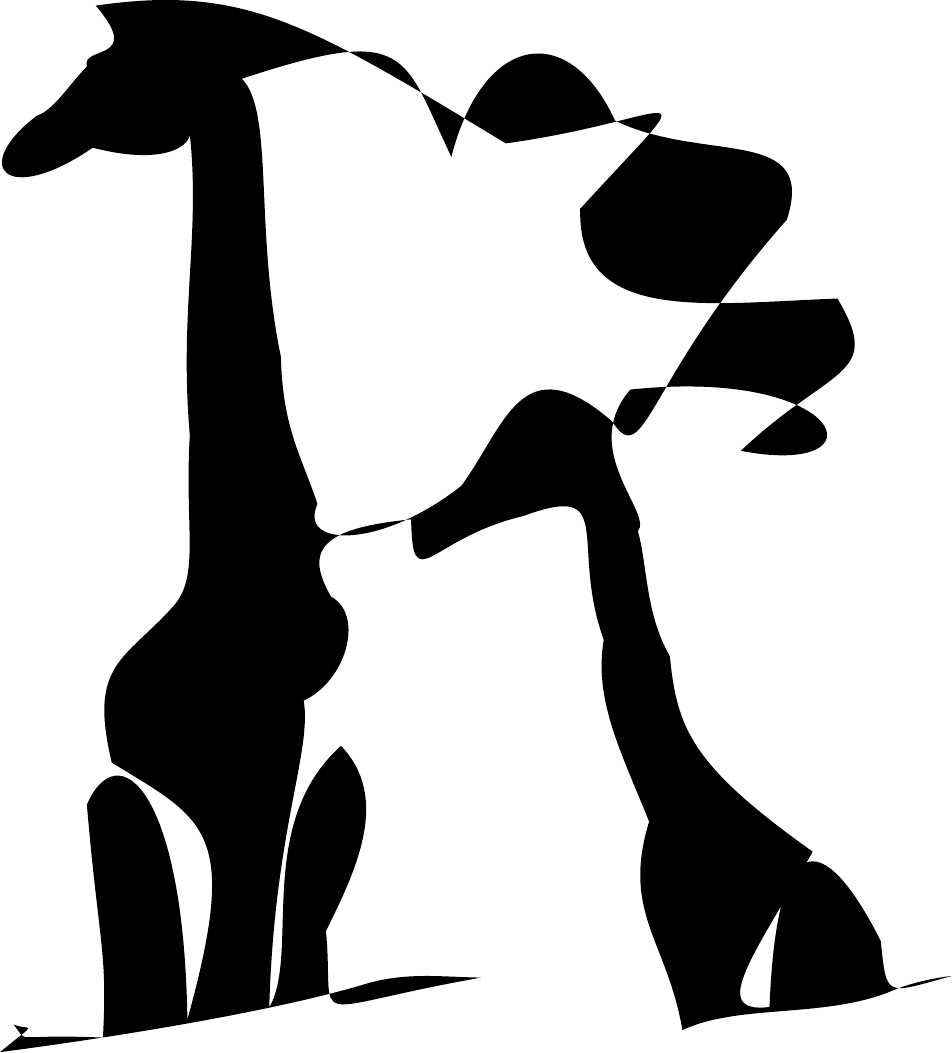} \\
    \multicolumn{2}{c}{Input} & 1 & 0.75 & 0.5 & 0.25 & \makecell[c]{Without \\ $\mathcal{L}_{acap}$} \\ 
\end{tabular}
\caption{Altering the weight $\alpha$ of the $\mathcal{L}_{acap}$ loss. On the leftmost column are the original letters and concepts used, then from left to right are the results obtained when using $\alpha\in\{1, 0.75, 0.5, 0.25, 0\}$.}
\label{fig:weights_conformal}
\end{figure}

Figure \ref{fig:cc_effect} illustrates the impact of the letter's initial number of control points. 
When less control points are used ($P_o$ is the original number of control points), we may get insufficient variations, such as for the gorilla. However, this can also result in more abstract depictions, such as the ballerina.
As we add control points, we get more graphic results, with the tradeoff that it often deviate from the original letter. 
In Figure \ref{fig:sds_lr} we show the results of using only the $\nabla_{\hat{P}} \mathcal{L}_\text{LSDS}$ loss.
As can be seen, in that case the illustrations strongly convey the semantic concept, however at the cost of legibility.
In Figure \ref{fig:weights_conformal} we analyze the effect of the weight $\alpha$ applied to $\mathcal{L}_{acap}$. Ranging from $1$ to $0$.
When $\mathcal{L}_{acap}$ is too dominant, the results may not enough reflect the semantic concept, while the opposite case harms legibility.
Figure \ref{fig:sigma_L2} illustrates a change in the $\sigma$ parameter of the low pass filter. When $\sigma=1$ almost no blur is applied, resulting in a shape constraint that is too strong.

\begin{figure}[ht]
    \centering
    \setlength{\tabcolsep}{5pt}
    \renewcommand{\arraystretch}{1}
    \begin{tabular}{l@{\hspace{0.2cm}} | @{\hspace{0.2cm}}c c c c cl}

        \raisebox{0.4cm}{\makecell[l]{Input \\ Letter }} &
        \hspace{0.2cm}
        \raisebox{0.25cm}{\includegraphics[height=0.1\linewidth]{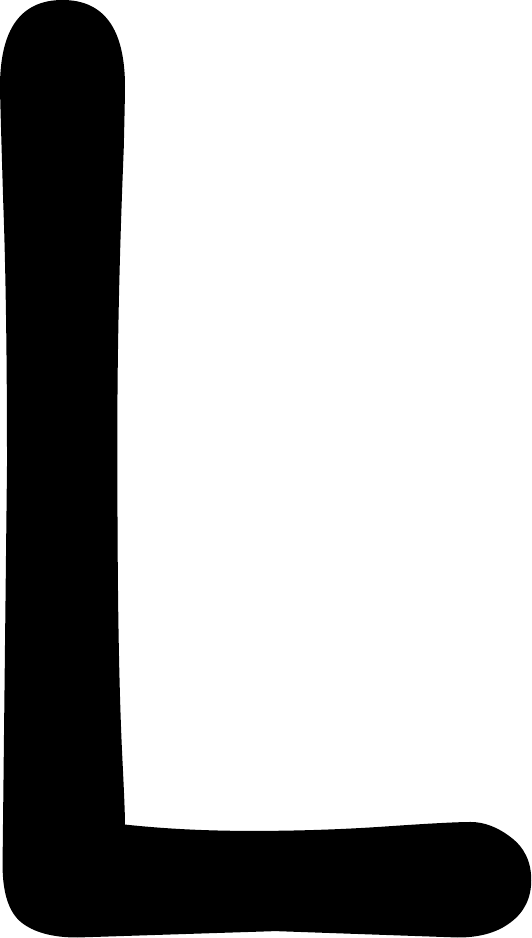}} &
        \raisebox{0.25cm}{\includegraphics[height=0.1\linewidth]{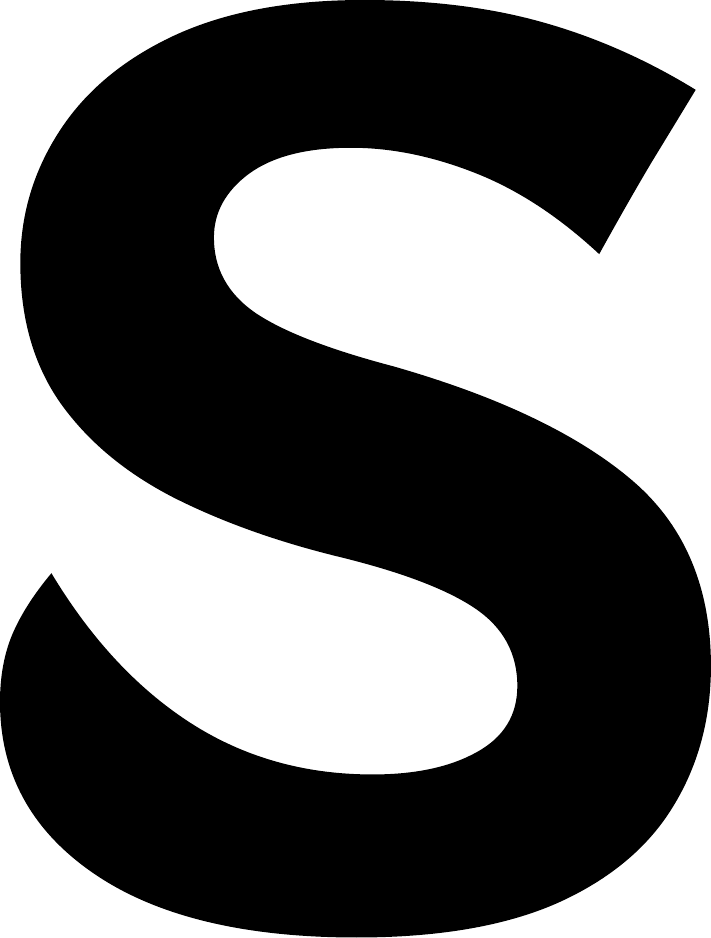}} &
        \raisebox{0.25cm}{\includegraphics[height=0.1\linewidth]{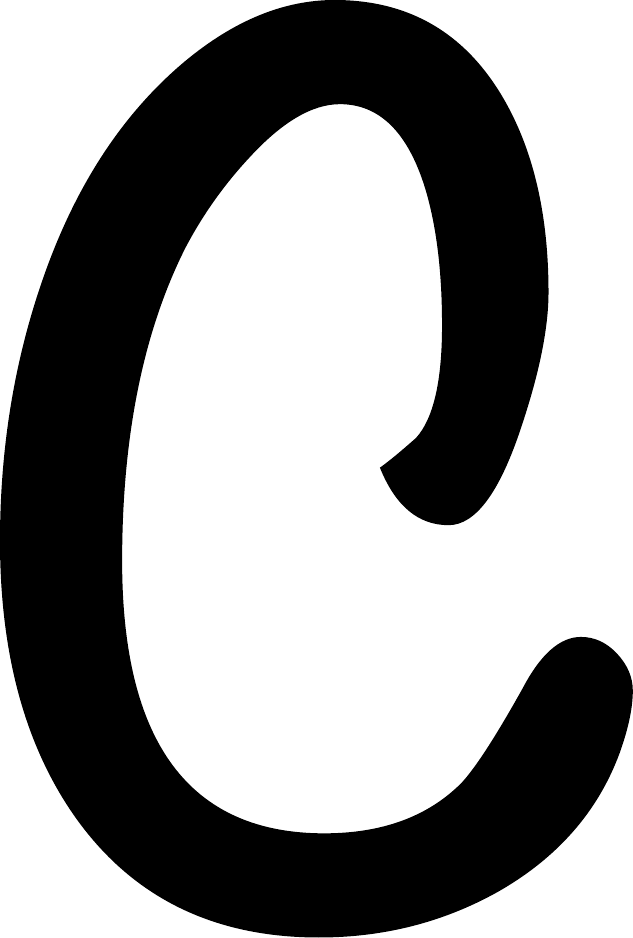}} &
        \raisebox{0.25cm}{\includegraphics[height=0.1\linewidth]{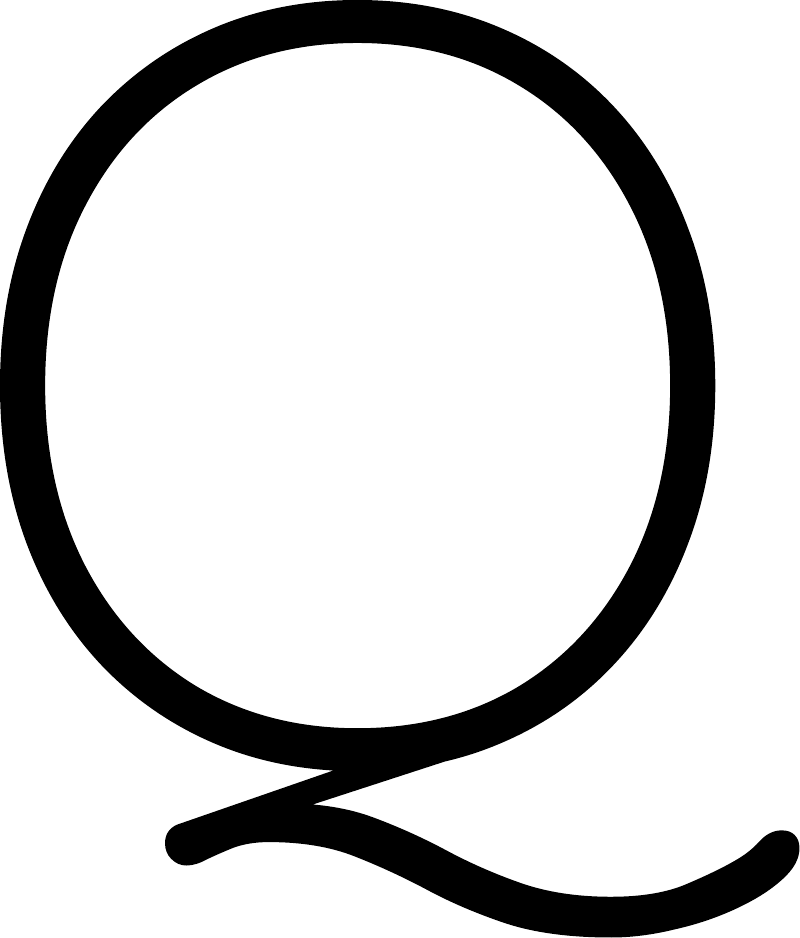}} &
        \raisebox{0.25cm}{\includegraphics[height=0.1\linewidth]{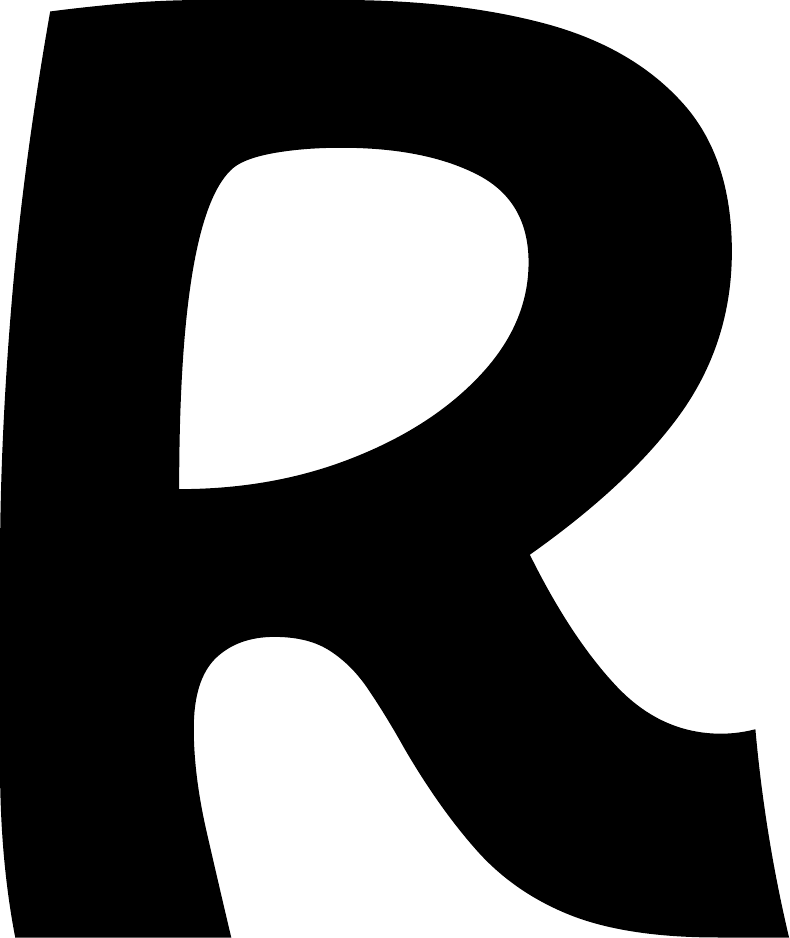}} \\

        \raisebox{0.4cm}{\makecell[l]{CLIP \\ loss}} &
        \hspace{0.2cm}
        \raisebox{0.25cm}{\includegraphics[height=0.1\linewidth]{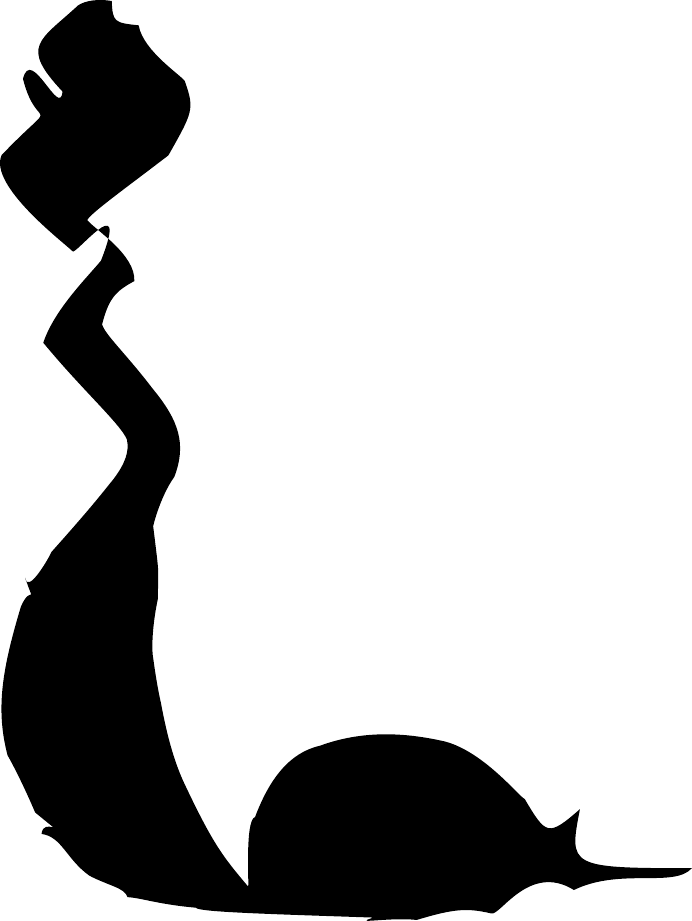}} &
        \raisebox{0.25cm}{\includegraphics[height=0.1\linewidth]{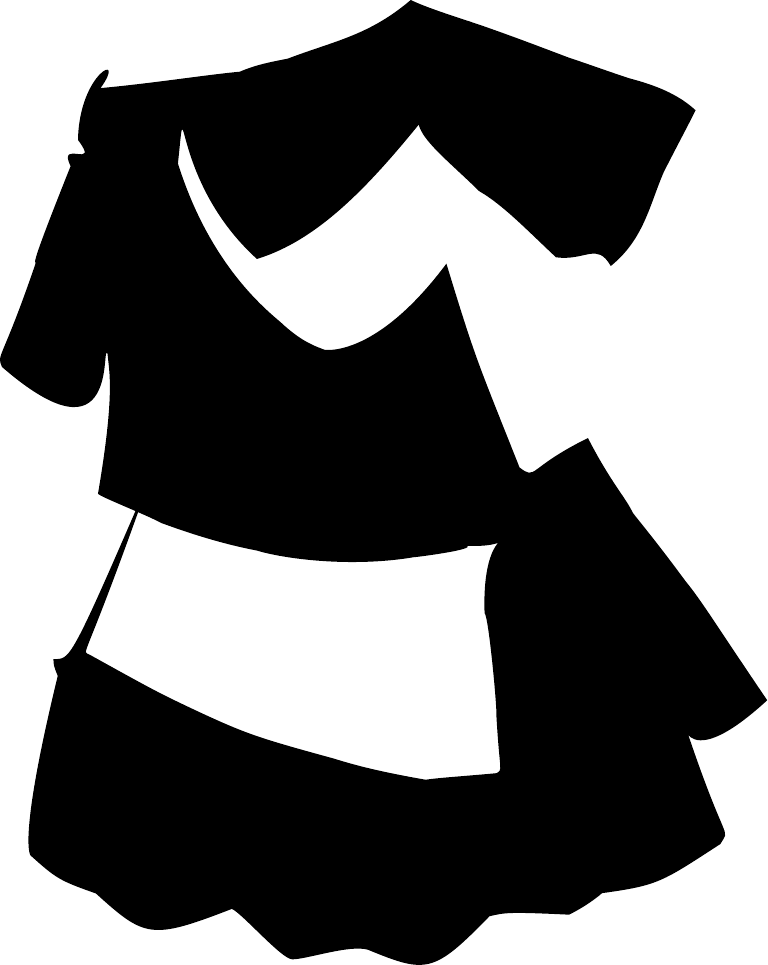}} &
        \raisebox{0.25cm}{\includegraphics[height=0.1\linewidth]{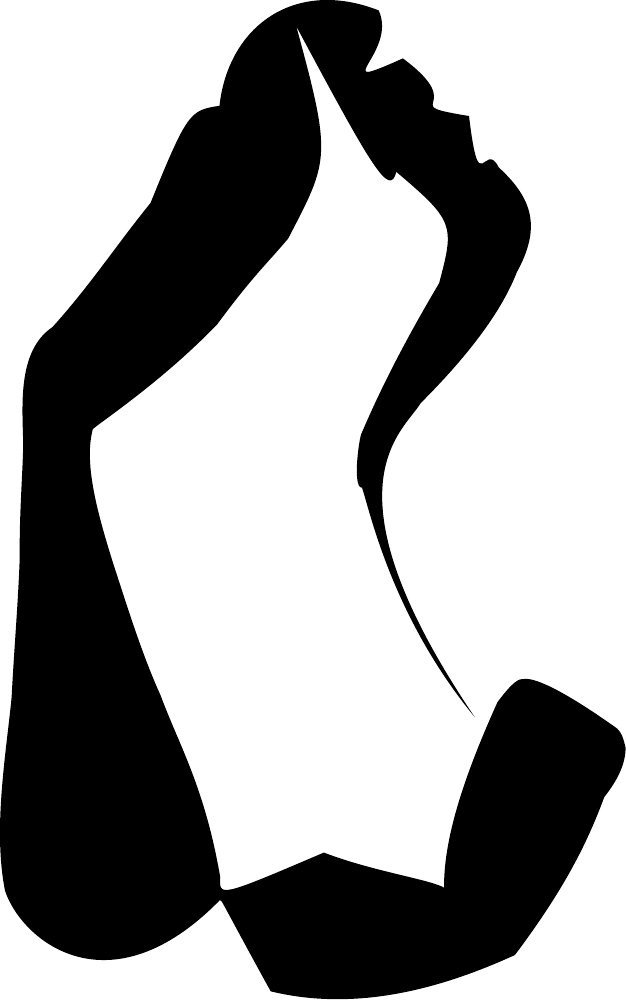}} &
        \raisebox{0.25cm}{\includegraphics[height=0.1\linewidth]{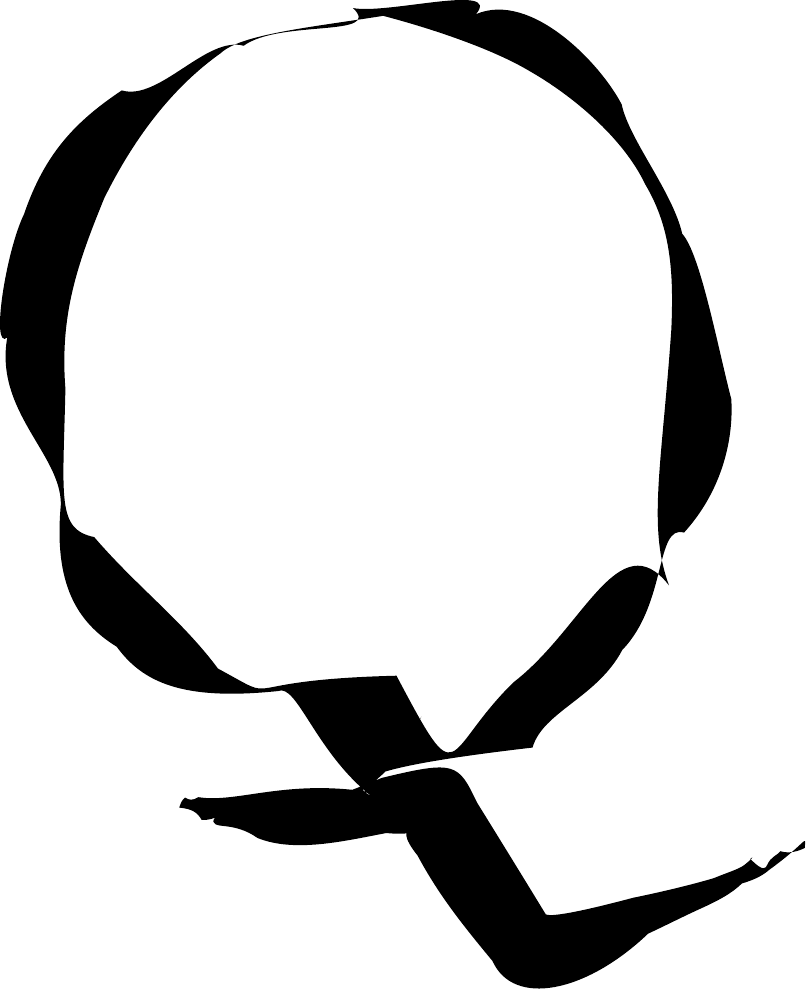}} &
        \raisebox{0.25cm}{\includegraphics[height=0.1\linewidth]{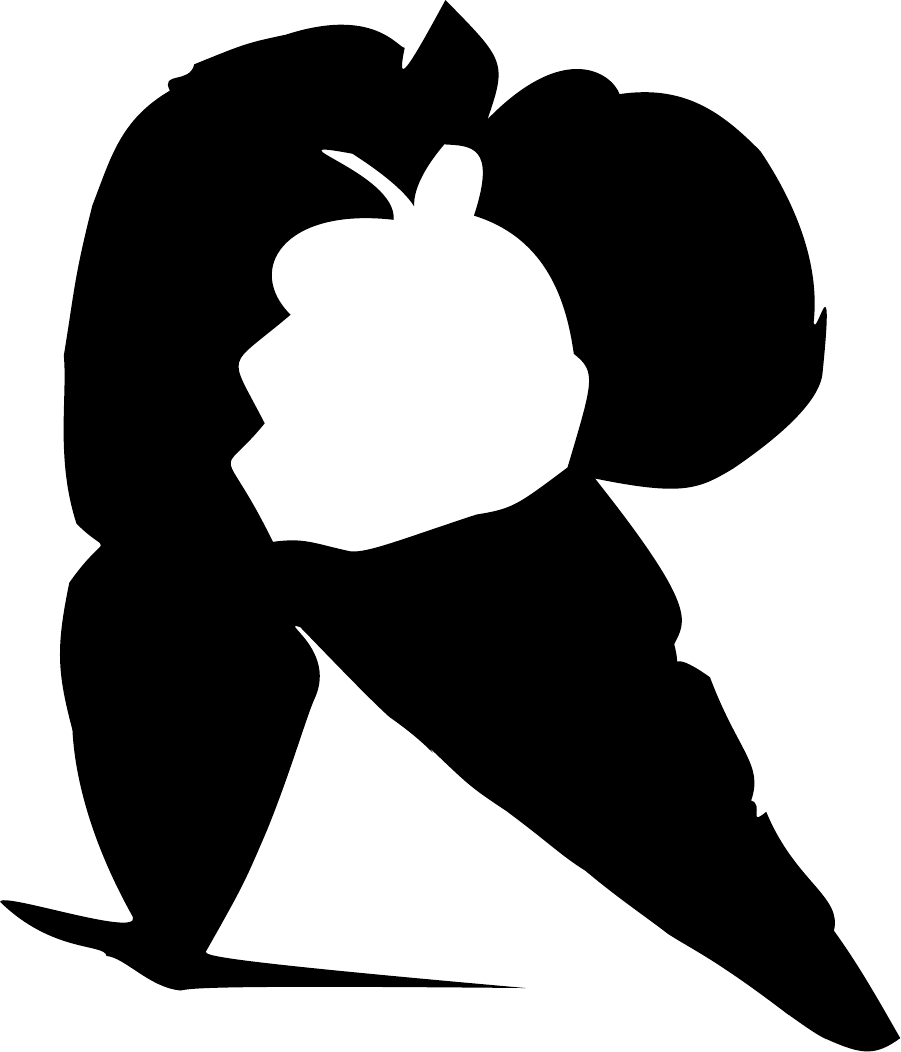}} \\

        \raisebox{0.4cm}{\makecell[l]{SDS \\ loss}} &
        \hspace{0.2cm}
        \raisebox{0.25cm}{\includegraphics[height=0.1\linewidth]{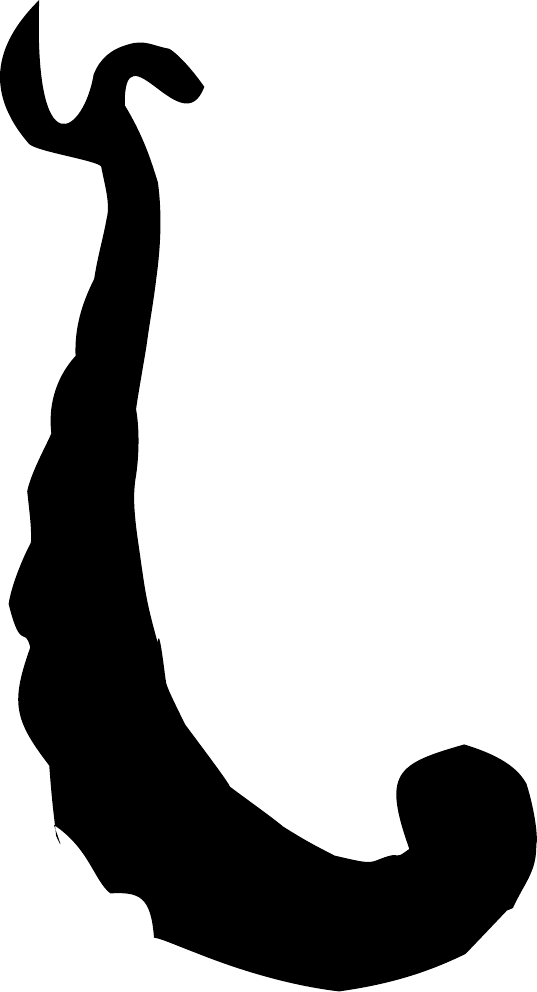}} &
        \raisebox{0.25cm}{\includegraphics[height=0.1\linewidth]{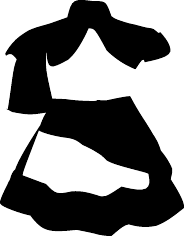}} &
        \raisebox{0.25cm}{\includegraphics[height=0.1\linewidth]{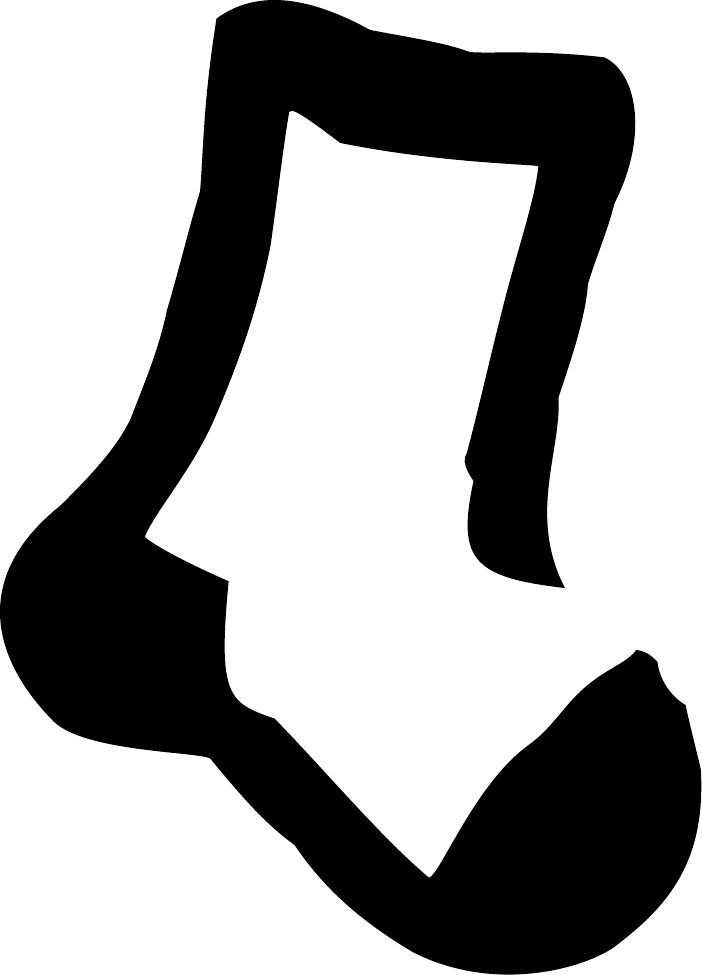}} &
        \raisebox{0.25cm}{\includegraphics[height=0.1\linewidth]{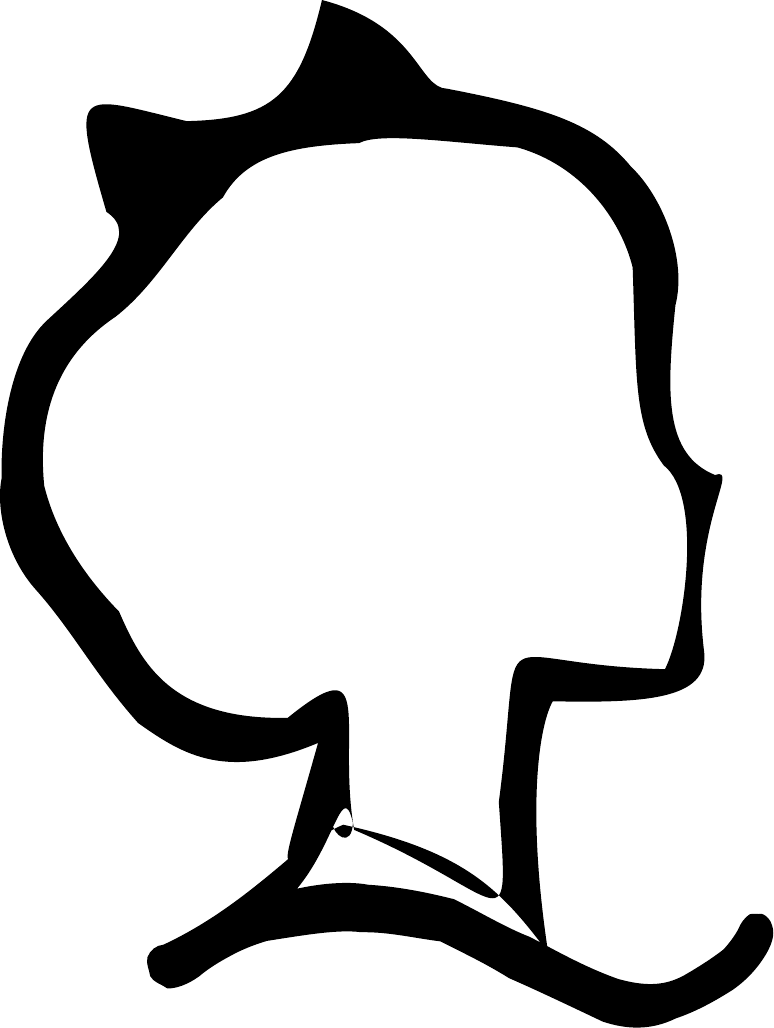}} &
        \raisebox{0.25cm}{\includegraphics[height=0.1\linewidth]{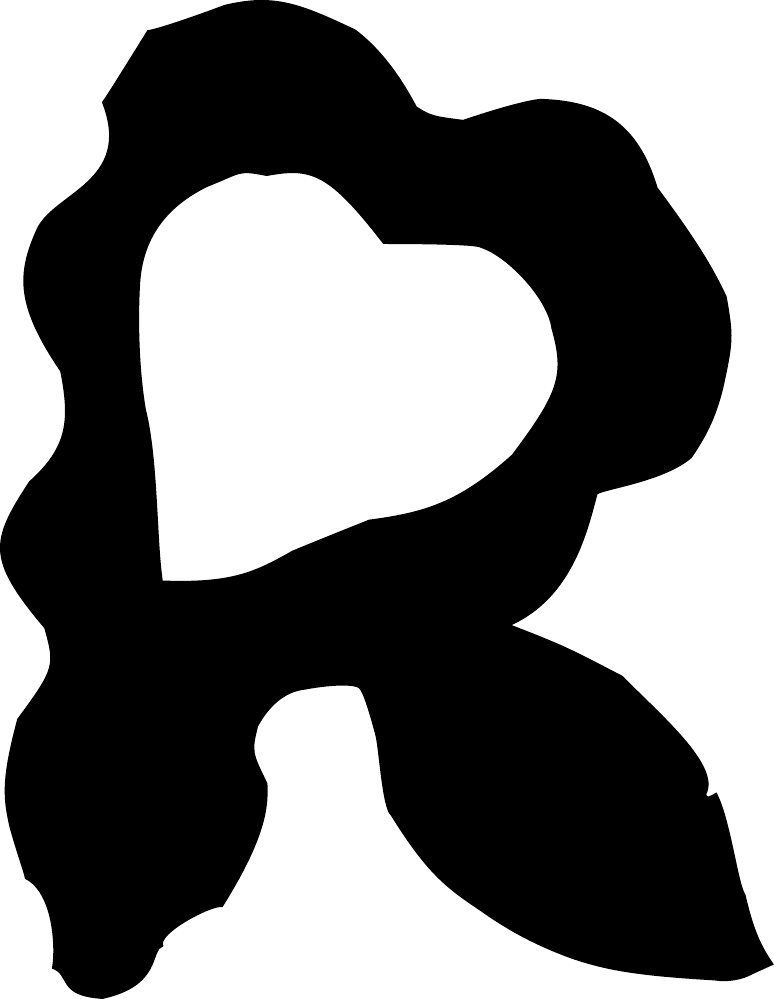}} \\

        &\hspace{0.1cm} 
        "Snail" & "Skirt" & "Socks" & "Queen" & "Strawberry" \\
    \end{tabular}
    \caption{Replacing the SDS loss with a CLIP-based loss.}
    \label{fig:comp_clip}
\end{figure}

In Figure \ref{fig:comp_clip} we show the results of replacing the $\nabla_{\hat{P}} \mathcal{L}_\text{LSDS}$ loss with a CLIP based loss, while using our proposed shape preservation terms. Although the results obtained with CLIP often depict the desired visual concept, we find that using Stable Diffusion leads to smoother illustrations, that capture a wider range of semantic concepts.

By using the hyperparameters described in the paper, we are able to achieve a reasonable balance between semantics and legibility. The parameters were determined manually based on visual assessments, but can be adjusted as needed based on the user's personal taste and goals.

\vspace{-0.35cm}
\section{Conclusions}
We presented a method for the automatic creation of vector-format word-as-image illustrations. Our method can handle a large variety of semantic concepts and use any font, while preserving the legibility of the text and the font's style.

There are limitations to our method. First, our method works letter by letter, and therefore, it cannot deform the shape of the entire word. In the future we can try to optimize the shape of several letters. Second, the approach works best on concrete visual concepts, and may fail with more abstract ones. This can be alleviated by optimizing the shape of letters using different concepts than the word itself. Third, the layout of letters can also be automated for example, using methods such as \cite{Wang_2022_CVPR}.

Our word-as-image illustrations demonstrate visual creativity and open the possibility for the use of large vision-language models for semantic typography, possibly also adding human-in-the-loop to arrive at more synergistic design methods of ML models and humans.

\begin{figure*}[!b]
\centering
    \includegraphics[width=0.8\textwidth]{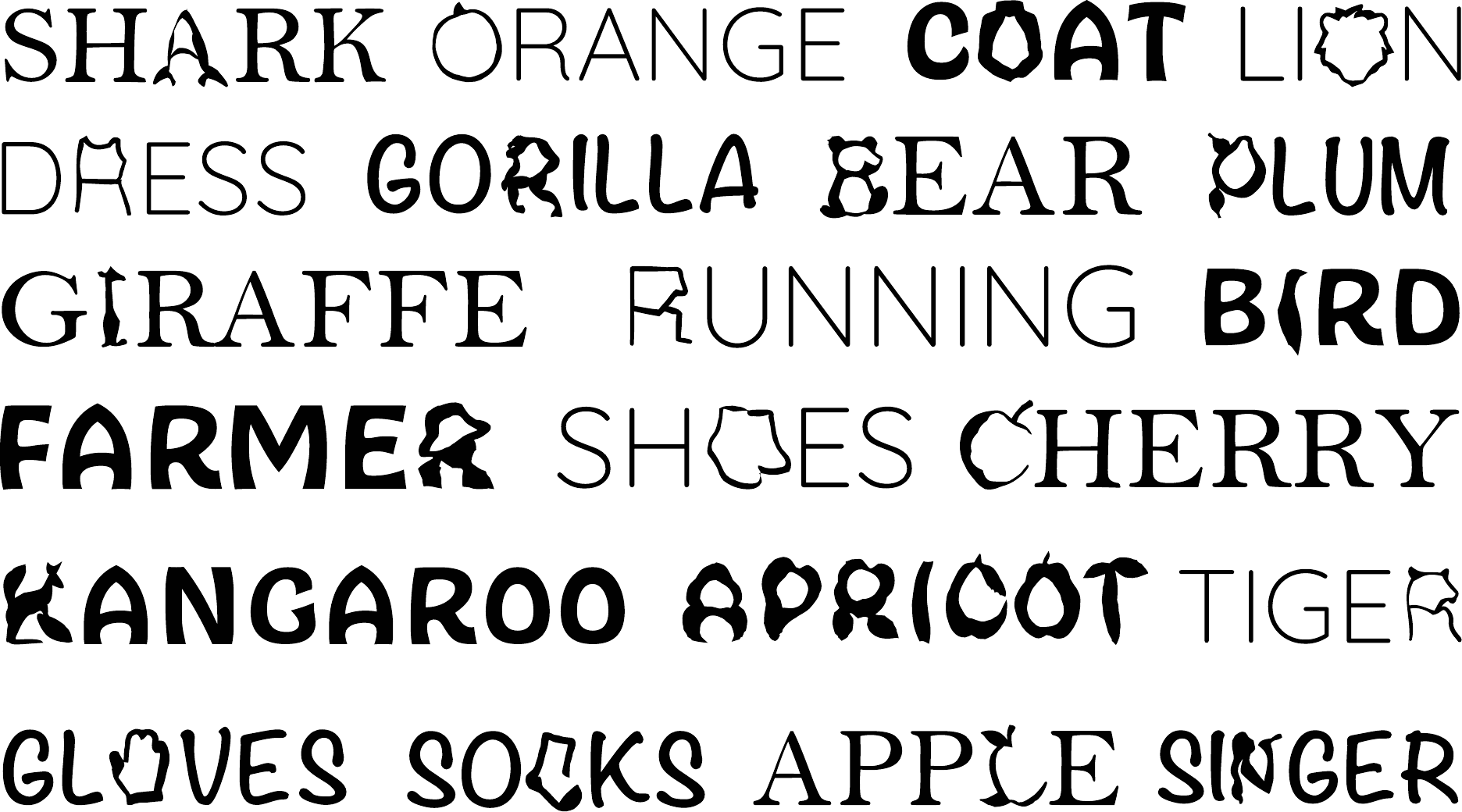}
    \caption{Word-as-images produced by our method. This subset was chosen from the random set of words.}
    \label{fig:all_res1}
\end{figure*}

\clearpage
\newpage
\begin{figure*}[t]
    \includegraphics[width=0.8\textwidth]{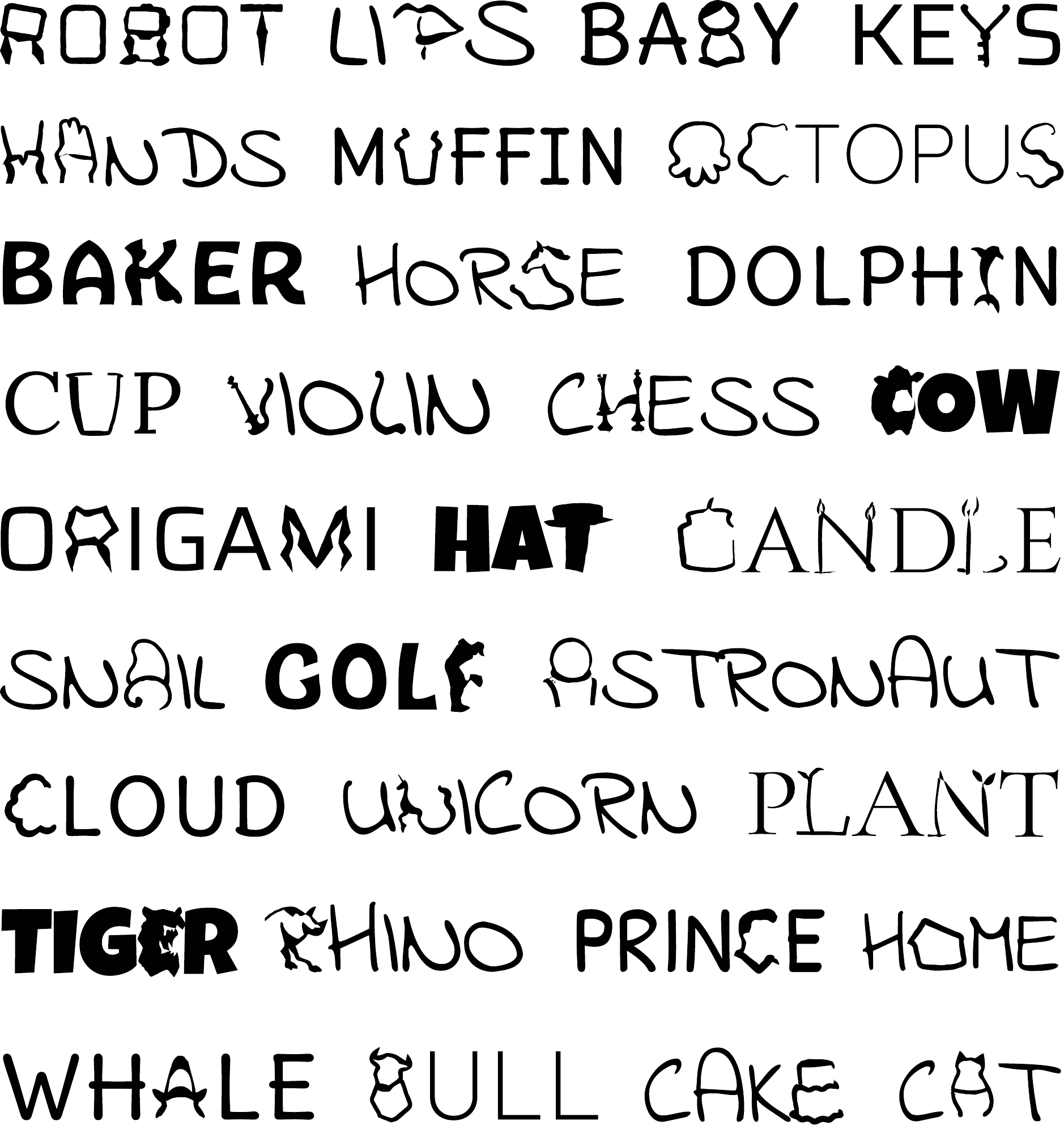}
    \caption{Additional results produced by our method.}
    \label{fig:mayne_res}
    \vspace{-1cm}
\end{figure*}

\clearpage
\newpage
\section{Acknowledgments}
We are grateful to Richard Hao Zhang for the early discussion of the text-as-image problem. Ali Mahdavi-Amiri and Oren Katzir for reviewing earlier versions of the manuscript and to Anran Qi for assisting in evaluating the Chinese words. This research was supported in part by the Israel Science Foundation (grants no. 2492/20 and 3441/21), Len Blavatnik and the Blavatnik family foundation, and the Tel Aviv University Innovation Laboratories (TILabs).

\bibliographystyle{ACM-Reference-Format}
\bibliography{main}

\appendix
\clearpage
\newpage
\section*{\centering{Supplementary Material}}
\vspace{8mm}
\section{Implementation Details}
In this section we provide further implementation details. We intend to release the code to promote future research in this domain.

Our method is based on the pre-trained $v1-5$ Stable Diffusion model \cite{stableDiffusion}, which we use through the diffusers \cite{von-platen-etal-2022-diffusers} Python package.
We optimize only the control points' coordinates (i.e. we do not modify the color, width, and other parameters of the shape). We use the Adam optimizer with $\beta_{1} = 0.9$, $\beta_{2} = 0.9$, $\epsilon = 10^{-6}$. We use learning rate warm-up from $0.1$ to $0.8$ over $100$ iterations and exponential decay from $0.8$ to $0.4$ over the rest $400$ iterations, $500$ iteration in total.
The optimization process requires at least 10GB memory and approximately 5 minutes to produce a single letter illustration on RTX2080 GPU.

Before we feed the rasterized $600x600$ letter image into the Stable Diffusion model, we apply random augmentations as proposed in CLIPDraw \cite{frans2021clipdraw}. Specifically, perspective transform with a distortion scale of $0.5$, with probability $0.7$, and a random $512x512$ crop. 
We add the suffix "a {[\textcolor{red}{\textit{word}}]}. minimal flat 2d vector. lineal color. trending on artstation." to the target word $W$, before feeding it into the text encoder of a pretrained CLIP model.

\vspace{-0.6cm}
\section{Comparisons}
As described in Section 5.2 we define five baselines to compare with.
In this section we provide more details about the evaluation and more qualitative results.
For \textbf{(1) SD}, we run Stable Diffusion \cite{stableDiffusion} with the default hyper parameters of $50$ inference steps and a guidance scale of $7.5$.
We use the prompt ``Word as image of the word [\textcolor{red}{\textit{word}}]. [\textcolor{blue}{\textit{font}}] font. minimal flat 2d vector. lineal color. black and white style''.

For \textbf{(2) SDEdit} \cite{meng2022sdedit}, we utilized the diffusers \cite{von-platen-etal-2022-diffusers} implementation, using the prompt ``A [\textcolor{red}{\textit{word}}]. minimal flat 2d vector. lineal color. black and white style'', and the rasterized input letter as the reference image. We use the default values of $50$ inference steps and a guidance scale of $7.5$. We use a strength value of $0.85$. The strength value determines the quantity of noise added to the input image -- a value close to $1.0$ results in higher degree of variation in the output, and vice versa.

We use the official website of OpenAI to run \textbf{(3) DallE2} \cite{ramesh2022hierarchical}, using the prompt ``Word as image of the word [\textcolor{red}{\textit{word}}]. Where the letter [\textcolor{olive}{\textit{letter}}] looks like a [\textcolor{red}{\textit{word}}]. [\textcolor{blue}{\textit{font}}] font. minimal flat 2d vector. lineal color. black and white style''.
To encourage the manipulation of a specific letter, for \textbf{(4) DallE2+letter} we use the prompt ``The letter [\textcolor{olive}{\textit{letter}}] in the shape of a [\textcolor{red}{\textit{word}}]. [\textcolor{blue}{\textit{font}}] font. minimal flat 2d vector. lineal color. black and white style''. 
For \textbf{(5) CLIPDraw} \cite{frans2021clipdraw}, we use the author's official implementation with the recommended hyper-parameters. Instead of using randomly initialized strokes, we use our vectorized letter as input, along with the prompt ``A [\textcolor{red}{\textit{word}}]. [\textcolor{blue}{\textit{font}}] font. minimal flat 2d vector. lineal color. black and white style''.
We provide more comparisons to the methods described above in Figure \ref{fig:supp_comp_diffusion}.

\begin{figure}[ht!]
\centering
    \includegraphics[width=1\linewidth]{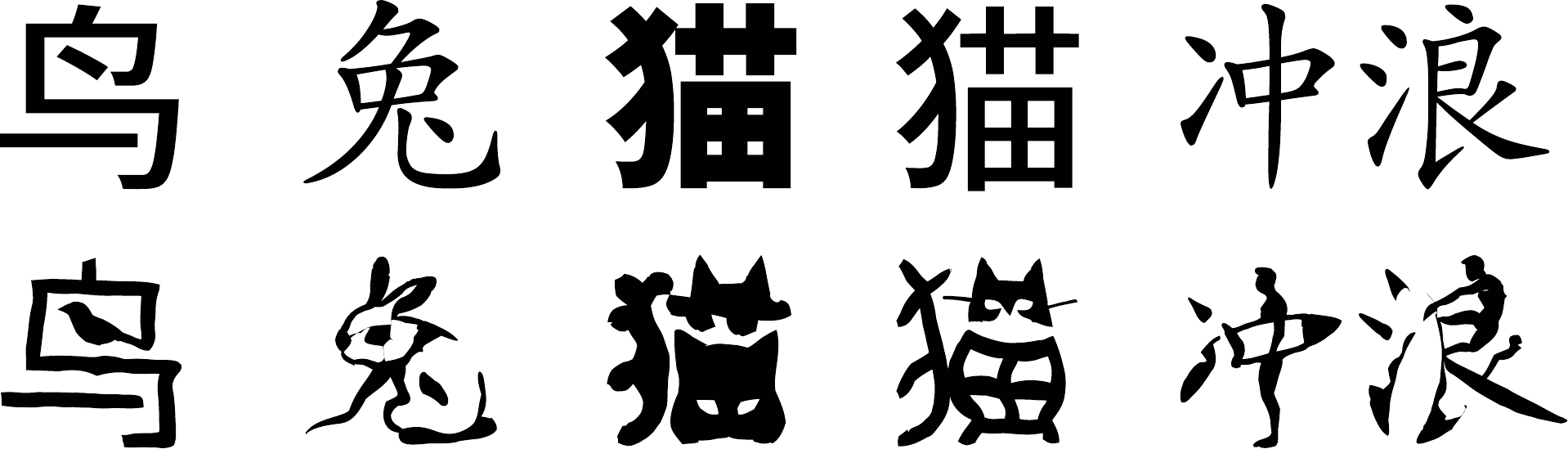}
    \caption{Some additional examples of word-as-image applied on Chinese characters. In Chinese, a whole word can be represented by one character. Here we show from left: bird, rabbit, cat and surfing (two last characters together). The complexity of characters imposes an additional challenge for our method. This could be alleviated in the future for example by dividing the characters to radicals and applying the method only on parts of the character.}
    \label{fig:china}
\end{figure}

\section{Perceptual Study}
\begin{figure}[H]
\centering
    \setlength{\tabcolsep}{6pt}
    \begin{tabular}{ l c c }

        & Ours & Only SDS \\
        \raisebox{0.8cm}{"Coat"} &
        \includegraphics[height=0.09\textwidth]{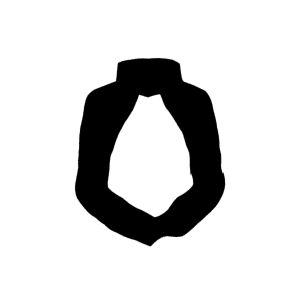} &
        \includegraphics[height=0.09\textwidth]{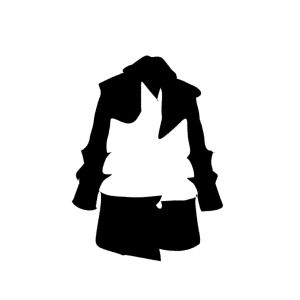} \\

        \raisebox{0.8cm}{"Soccer"} &
        \includegraphics[height=0.09\textwidth]{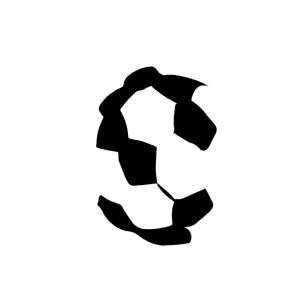} &
        \includegraphics[height=0.09\textwidth]{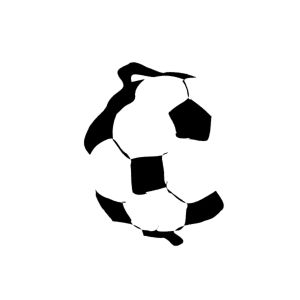} \\

        \raisebox{0.8cm}{"Shirt"} &
        \includegraphics[height=0.09\textwidth]{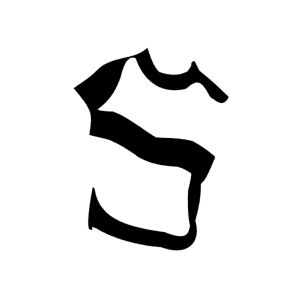} &
        \includegraphics[height=0.09\textwidth]{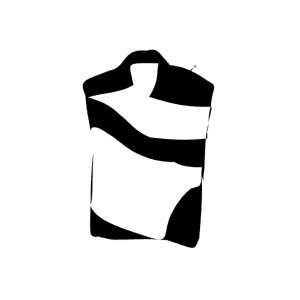} \\

        \raisebox{0.8cm}{"Rugby"} &
        \includegraphics[height=0.09\textwidth]{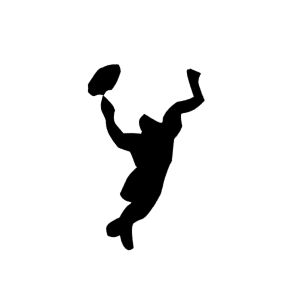} &
        \includegraphics[height=0.09\textwidth]{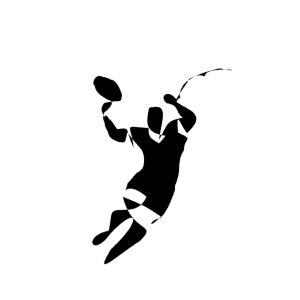} \\
        \midrule
        \raisebox{0.8cm}{\makecell[l]{Font \\ Rec.}} & \multicolumn{2}{c}{\includegraphics[height=0.092\textwidth]{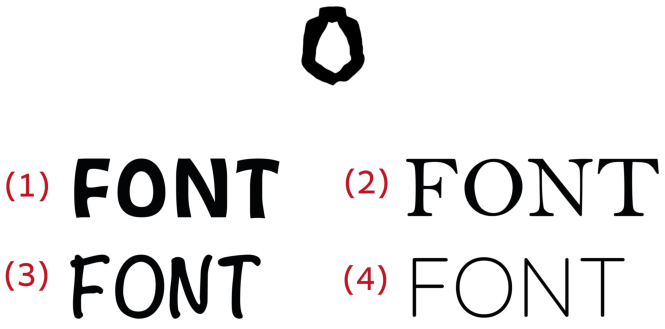}}
    \end{tabular}
    
    \caption{Examples of illustrations presented in the perceptual study. Each pair in the top part shows illustrations obtained using our proposed method (left) and using only SDS loss (right). On the bottom is an example of an illustration presented for the font recognition questions.}
    \label{fig:user_study_ex}
\end{figure}

In this section, we provide more details about the perceptual study described in Section 5.1.
The randomly chosen objects, fonts, and letters are shown in Table \ref{tb:user_study_letters}.
A few visual examples are shown in Figure \ref{fig:user_study_ex}. 

\begin{table}[h]
    \small
    \centering
    \setlength{\tabcolsep}{2pt}
    \caption{Randomly chosen objects, letters, and fonts for the perceptual study.} 
    \begin{tabular}{l c c c} 
    \toprule
    Object & \begin{tabular}{c} Letter \end{tabular} & \begin{tabular}{c} Font \end{tabular} \\
    \midrule
    Pineapple    & P & Noteworthy-Bold \\
    Orange    & O & Quicksand \\
    Rugby    & Y & Noteworthy-Bold \\
    Soccer    & S & Noteworthy-Bold \\
    Bear    & B & Bell MT \\
    Lion    & O & Quicksand \\
    Singer    & N & Noteworthy-Bold \\
    Pilot    & P & Noteworthy-Bold \\
    Coat    & O & HobeauxRococeaux-Sherman \\
    Shirt    & S & Bell MT \\    
    \bottomrule
    \end{tabular}
    \label{tb:user_study_letters}
\end{table}

\begin{figure*}[ht!]
    \centering
    \setlength{\tabcolsep}{12pt}
    {\small
    \begin{tabular}{ c c | c c | c c | c | c }

        \raisebox{0.4cm}{\makecell[l]{"Muffin"}} &
        \hspace{0.1cm}
        \raisebox{0.2cm}{\includegraphics[height=0.04\textwidth]{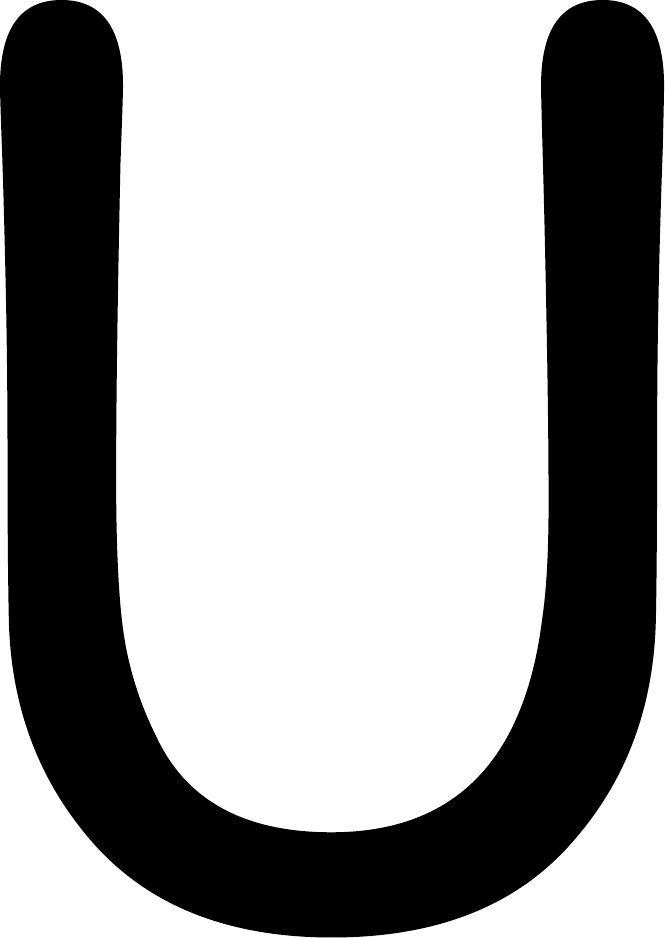}} &
        \hspace{0.1cm}
        \includegraphics[height=0.07\textwidth]{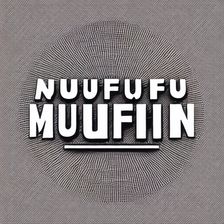} &
        \includegraphics[height=0.07\textwidth]{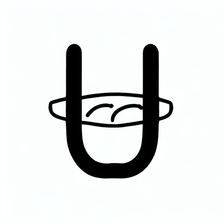} &
        \hspace{0.1cm}
        \includegraphics[height=0.07\textwidth]{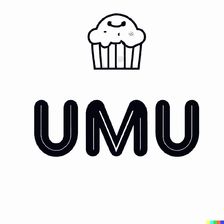} &
        \includegraphics[height=0.07\textwidth]{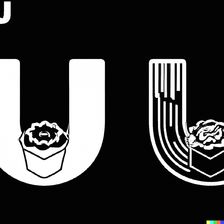} &
        \hspace{0.1cm}
        \includegraphics[height=0.07\textwidth]{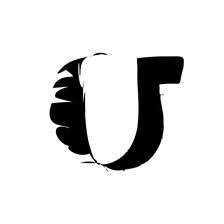} &
        \raisebox{0.2cm}{\includegraphics[height=0.048\textwidth]{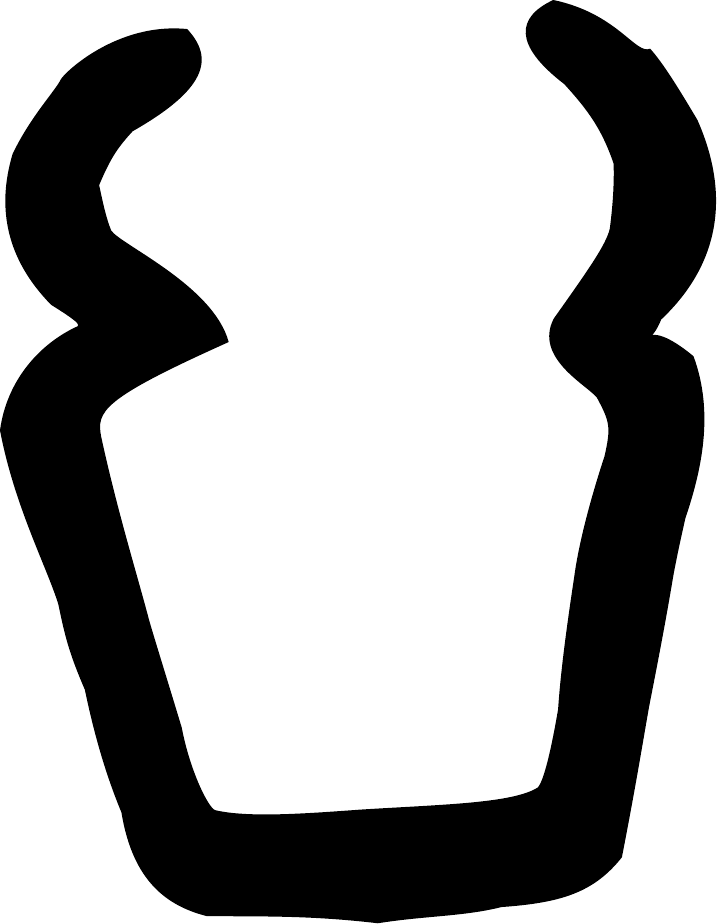}} \\

        \raisebox{0.4cm}{\makecell[l]{"Tiger"}} &
        \hspace{0.1cm}
        \raisebox{0.2cm}{\includegraphics[height=0.04\textwidth]{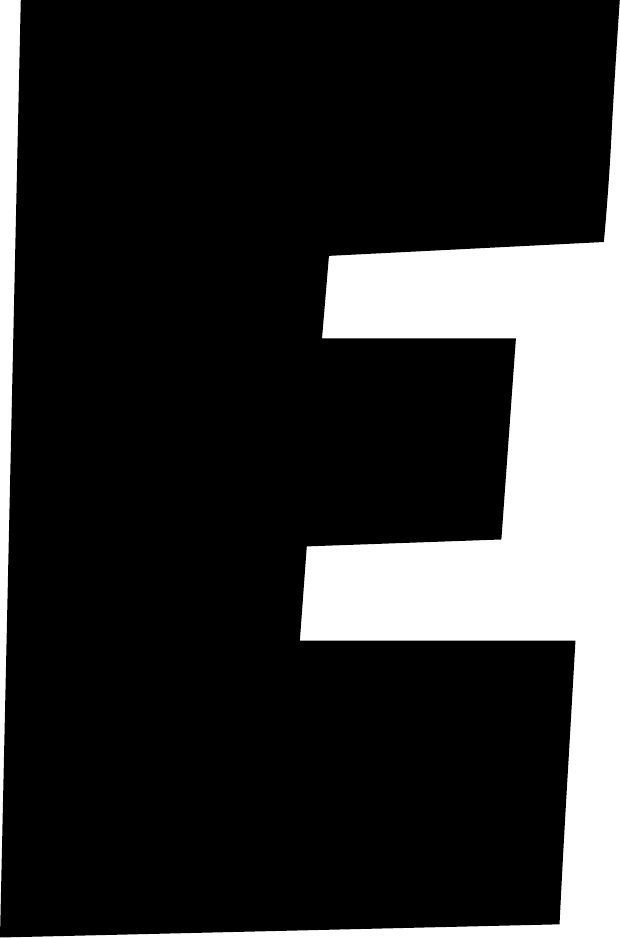}} &
        \hspace{0.1cm}
        \includegraphics[height=0.07\textwidth]{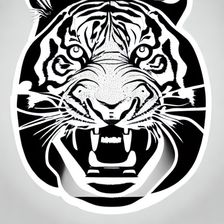} &
        \includegraphics[height=0.07\textwidth]{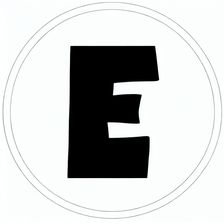} &
        \hspace{0.1cm}
        \includegraphics[height=0.07\textwidth]{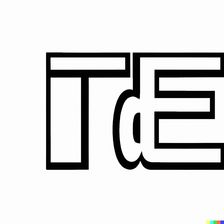} &
        \includegraphics[height=0.07\textwidth]{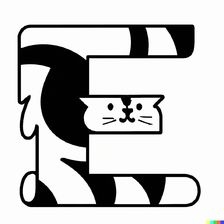} &
        \hspace{0.1cm}
        \includegraphics[height=0.07\textwidth]{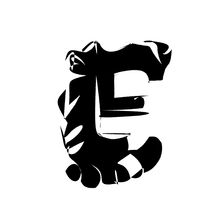} &
        \raisebox{0.2cm}{\includegraphics[height=0.048\textwidth]{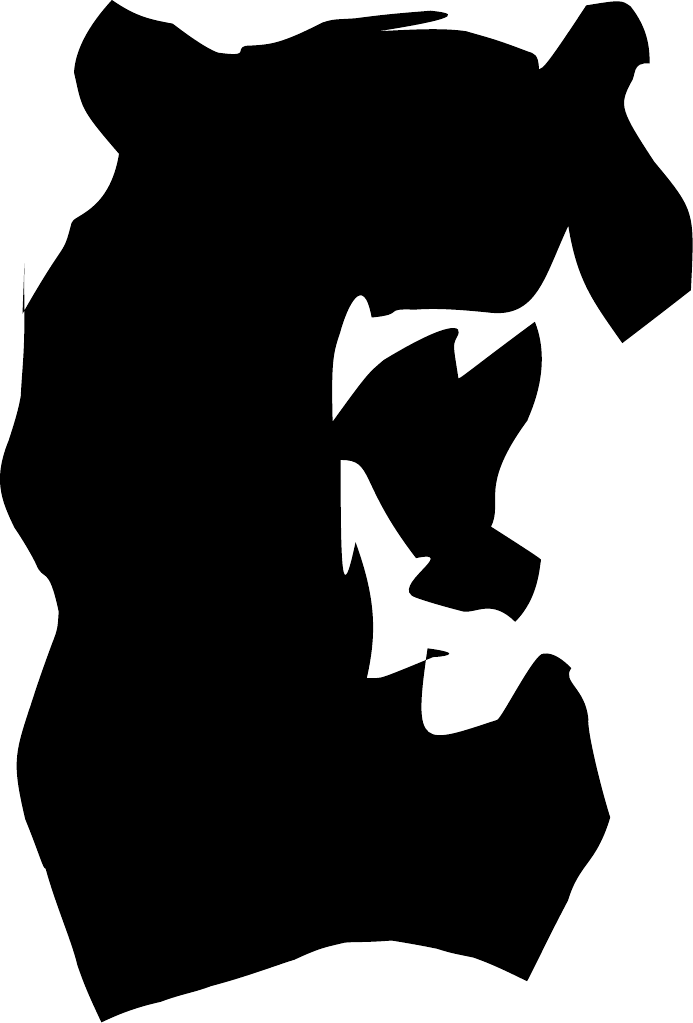}} \\

        \raisebox{0.4cm}{\makecell[l]{"Octopus"}} &
        \hspace{0.1cm}
        \raisebox{0.2cm}{\includegraphics[height=0.04\textwidth]{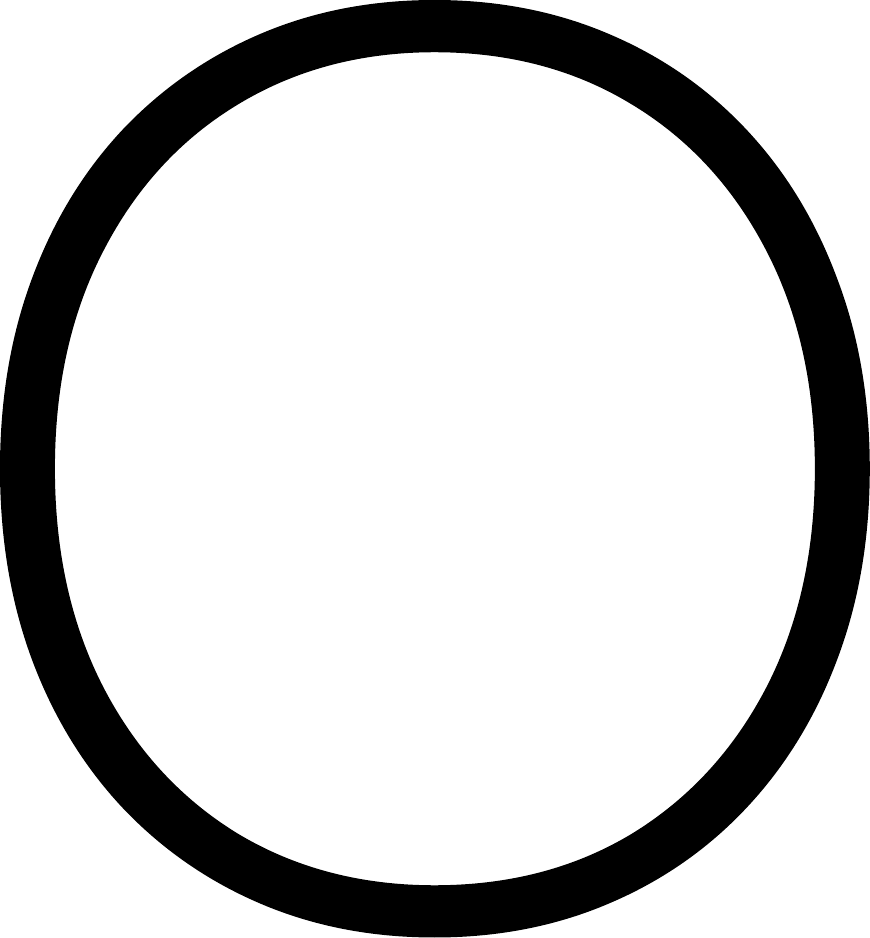}} &
        \hspace{0.1cm}
        \includegraphics[height=0.07\textwidth]{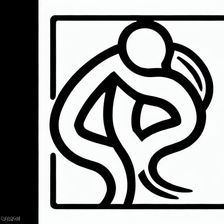} &
        \includegraphics[height=0.07\textwidth]{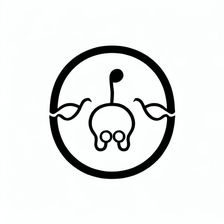} &
        \hspace{0.1cm}
        \includegraphics[height=0.07\textwidth]{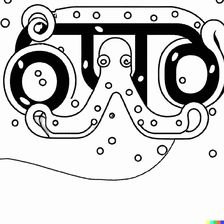} &
        \includegraphics[height=0.07\textwidth]{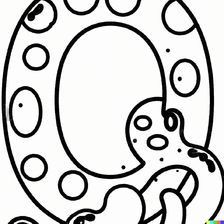} &
        \hspace{0.1cm}
        \includegraphics[height=0.07\textwidth]{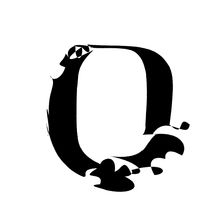} &
        \raisebox{0.2cm}{\includegraphics[height=0.048\textwidth]{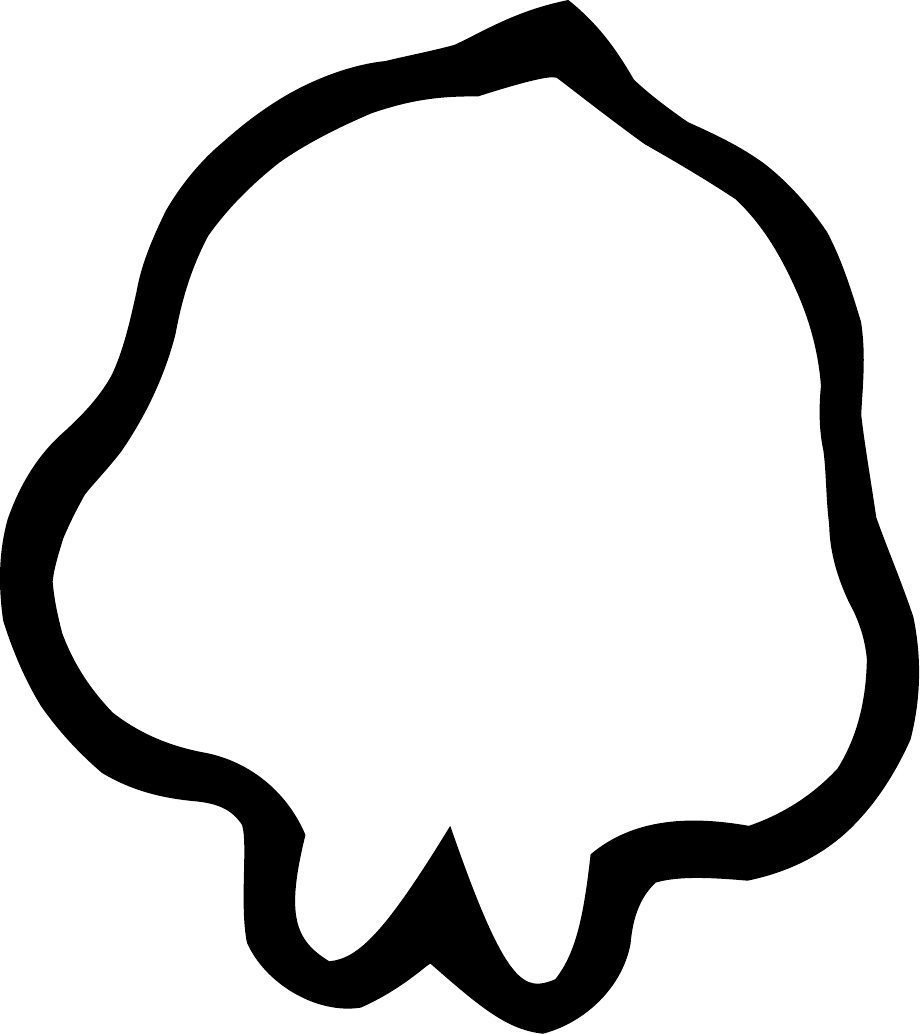}} \\

        \raisebox{0.4cm}{\makecell[l]{"Plant"}} &
        \hspace{0.1cm}
        \raisebox{0.2cm}{\includegraphics[height=0.04\textwidth]{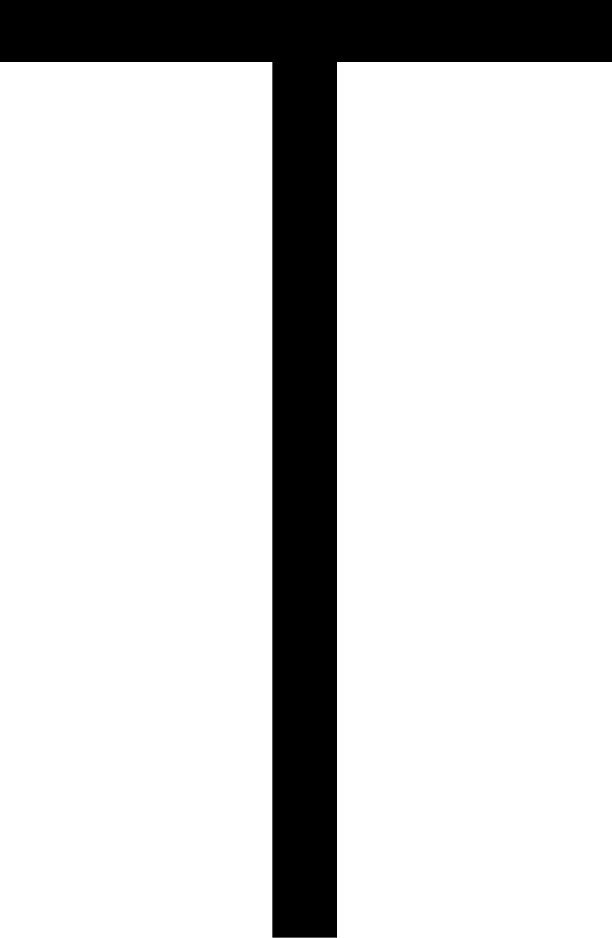}} &
        \hspace{0.1cm}
        \includegraphics[height=0.07\textwidth]{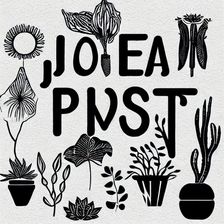} &
        \includegraphics[height=0.07\textwidth]{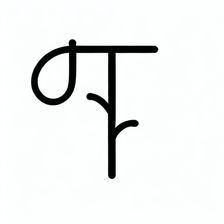} &
        \hspace{0.1cm}
        \includegraphics[height=0.07\textwidth]{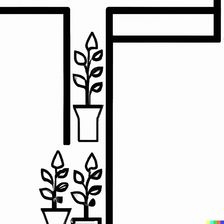} &
        \includegraphics[height=0.07\textwidth]{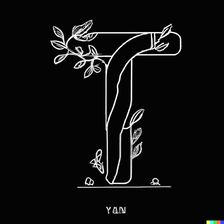} &
        \hspace{0.1cm}
        \includegraphics[height=0.07\textwidth]{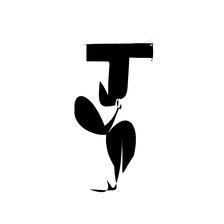} &
        \raisebox{0.2cm}{\includegraphics[height=0.048\textwidth]{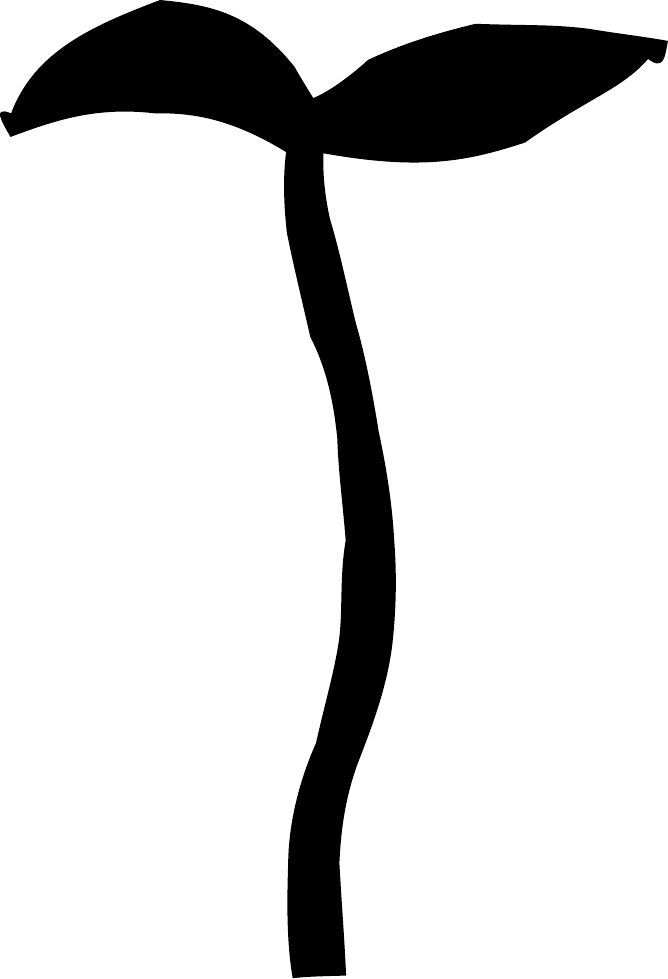}} \\

        \raisebox{0.4cm}{\makecell[l]{"Astronaut"}} &
        \hspace{0.1cm}
        \raisebox{0.2cm}{\includegraphics[height=0.04\textwidth]{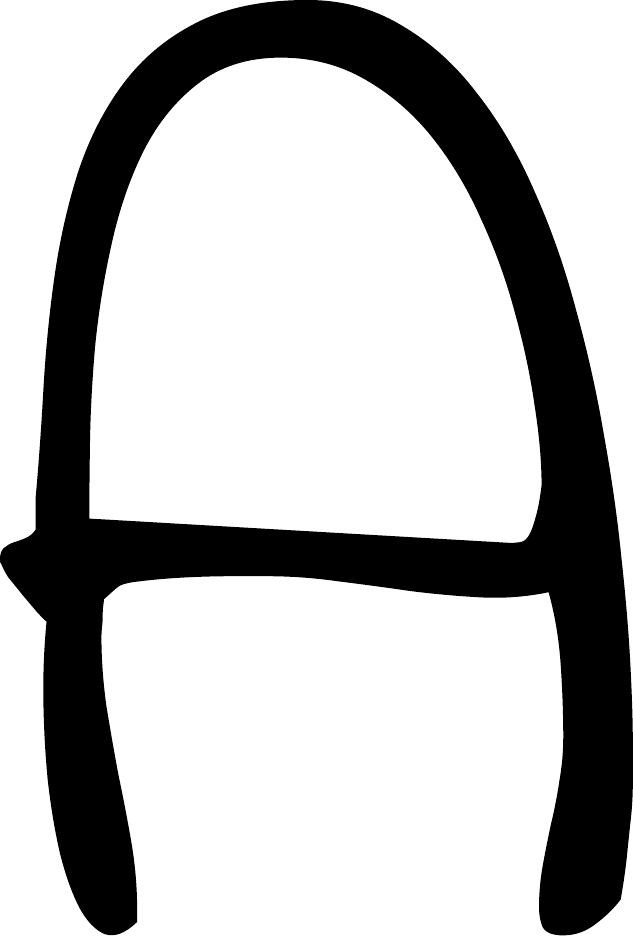}} &
        \hspace{0.1cm}
        \includegraphics[height=0.07\textwidth]{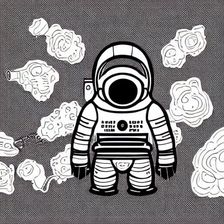} &
        \includegraphics[height=0.07\textwidth]{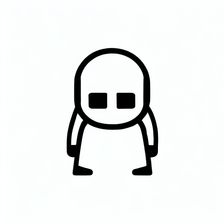} &
        \hspace{0.1cm}
        \includegraphics[height=0.07\textwidth]{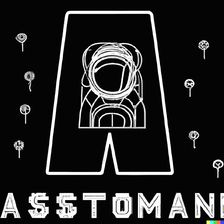} &
        \includegraphics[height=0.07\textwidth]{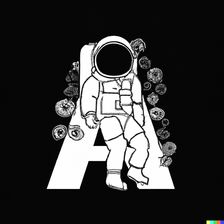} &
        \hspace{0.1cm}
        \includegraphics[height=0.07\textwidth]{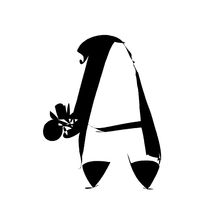} &
        \raisebox{0.2cm}{\includegraphics[height=0.048\textwidth]{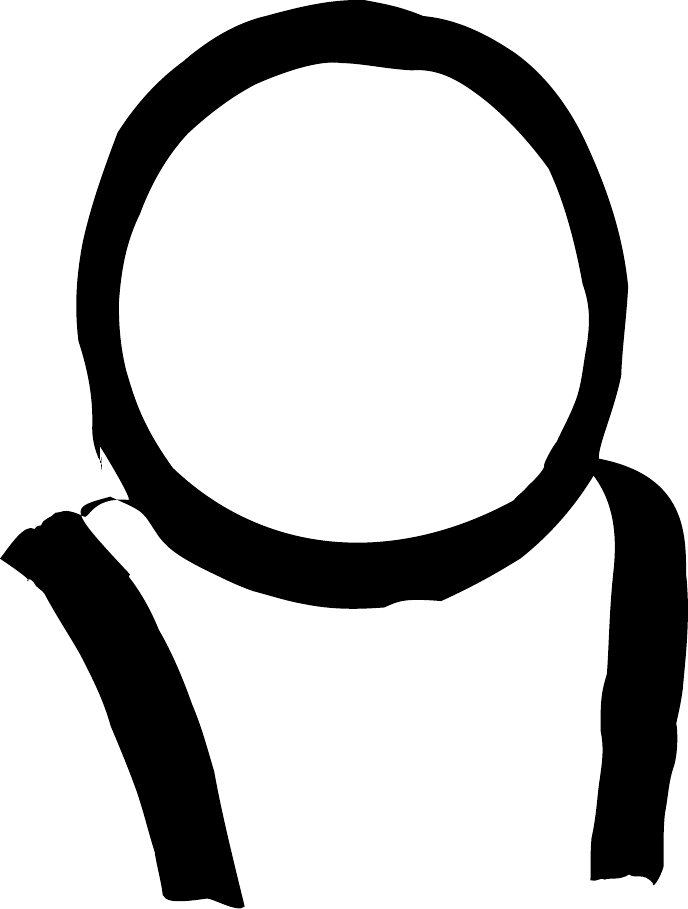}} \\

        \raisebox{0.4cm}{\makecell[l]{"Robot"}} &
        \hspace{0.1cm}
        \raisebox{0.2cm}{\includegraphics[height=0.04\textwidth]{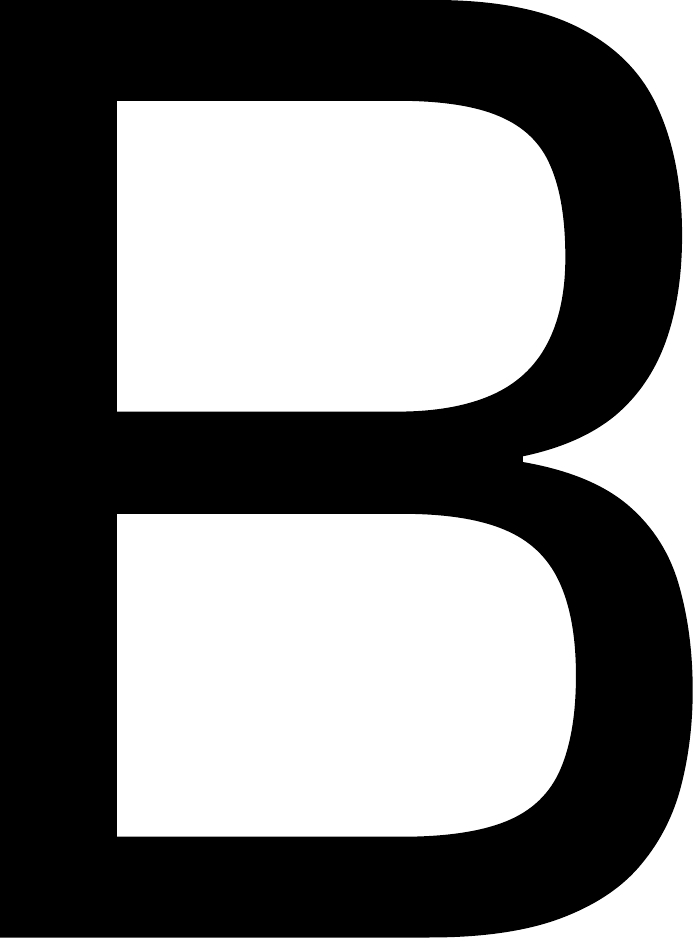}} &
        \hspace{0.1cm}
        \includegraphics[height=0.07\textwidth]{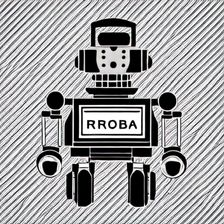} &
        \includegraphics[height=0.07\textwidth]{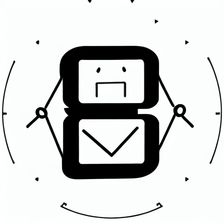} &
        \hspace{0.1cm}
        \includegraphics[height=0.07\textwidth]{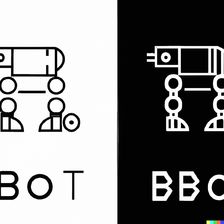} &
        \includegraphics[height=0.07\textwidth]{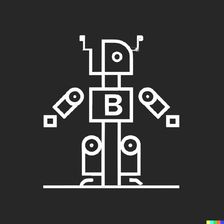} &
        \hspace{0.1cm}
        \includegraphics[height=0.07\textwidth]{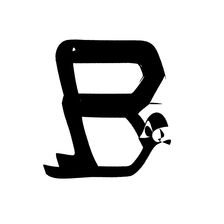} &
        \raisebox{0.2cm}{\includegraphics[height=0.048\textwidth]{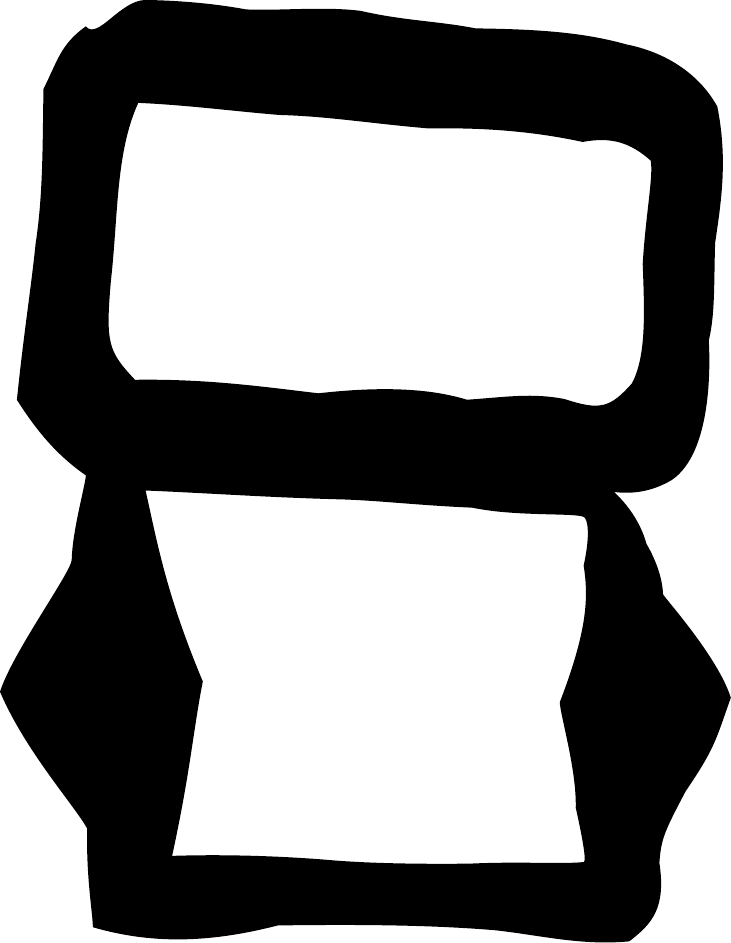}} \\

        \raisebox{0.4cm}{\makecell[l]{"Bunny"}} &
        \hspace{0.1cm}
        \raisebox{0.2cm}{\includegraphics[height=0.04\textwidth]{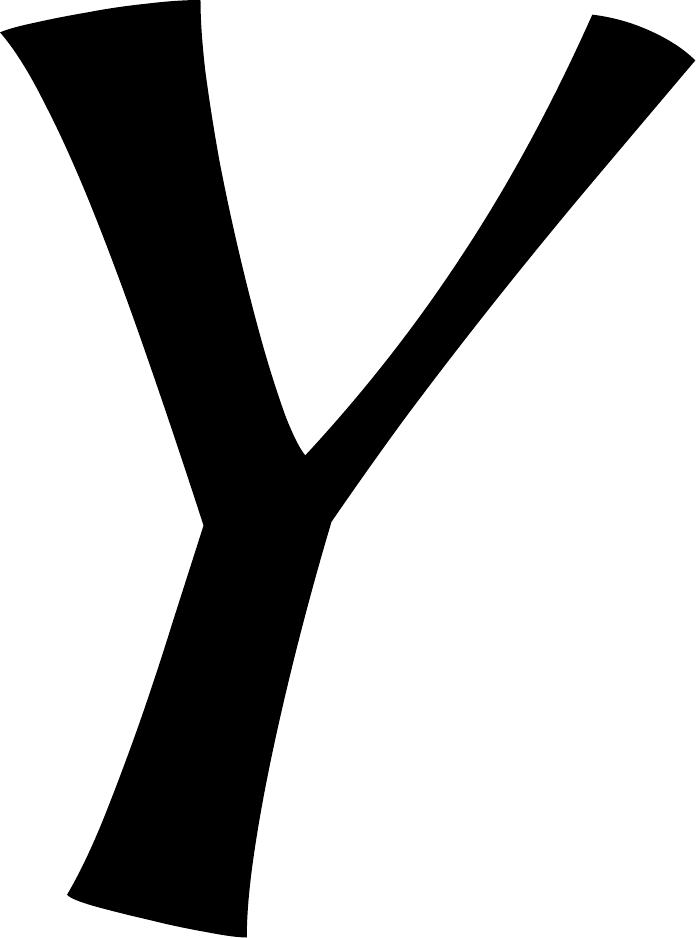}} &
        \hspace{0.1cm}
        \includegraphics[height=0.07\textwidth]{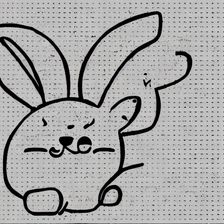} &
        \includegraphics[height=0.07\textwidth]{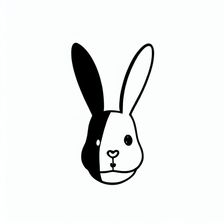} &
        \hspace{0.1cm}
        \includegraphics[height=0.07\textwidth]{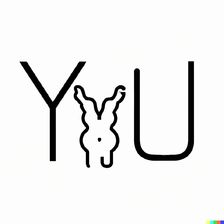} &
        \includegraphics[height=0.07\textwidth]{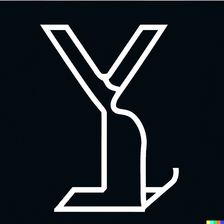} &
        \hspace{0.1cm}
        \includegraphics[height=0.07\textwidth]{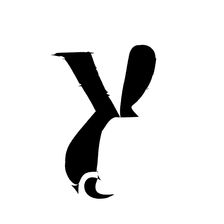} &
        \raisebox{0.2cm}{\includegraphics[height=0.048\textwidth]{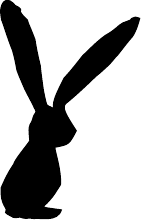}} \\

        \raisebox{0.4cm}{\makecell[l]{"Flamingo"}} &
        \hspace{0.1cm}
        \raisebox{0.2cm}{\includegraphics[height=0.04\textwidth]{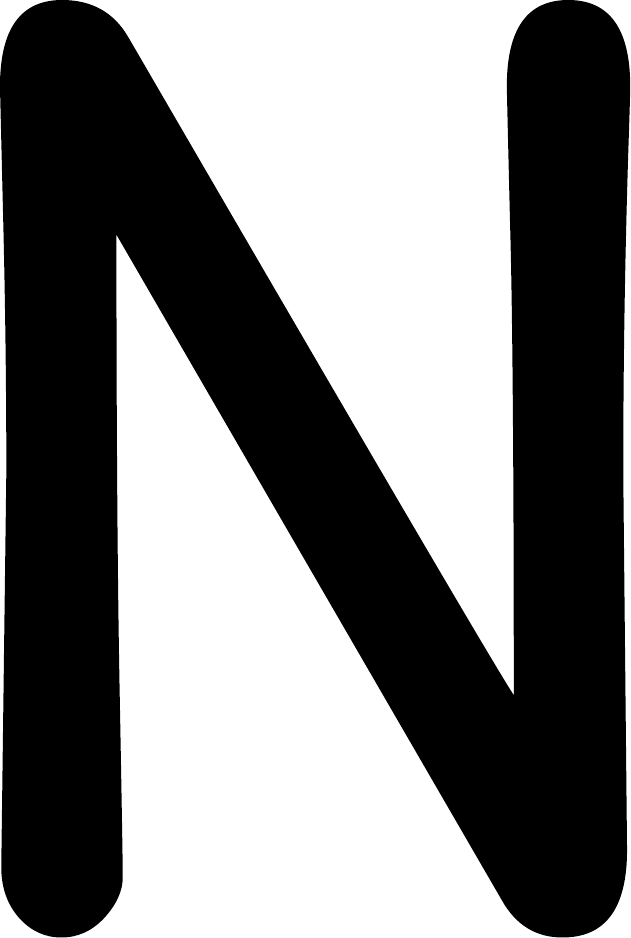}} &
        \hspace{0.1cm}
        \includegraphics[height=0.07\textwidth]{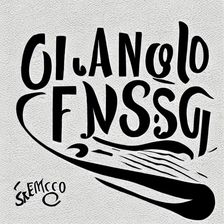} &
        \includegraphics[height=0.07\textwidth]{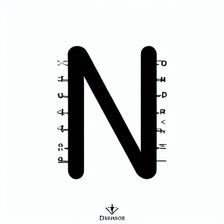} &
        \hspace{0.1cm}
        \includegraphics[height=0.07\textwidth]{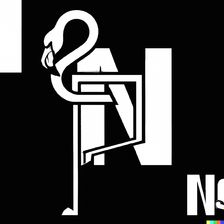} &
        \includegraphics[height=0.07\textwidth]{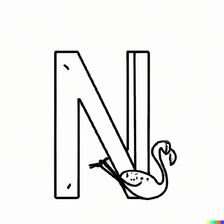} &
        \hspace{0.1cm}
        \includegraphics[height=0.07\textwidth]{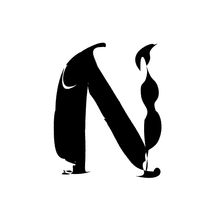} &
        \raisebox{0.2cm}{\includegraphics[height=0.048\textwidth]{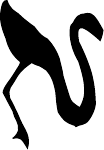}} \\

        \raisebox{0.4cm}{\makecell[l]{"Paris"}} &
        \hspace{0.1cm}
        \raisebox{0.2cm}{\includegraphics[height=0.04\textwidth]{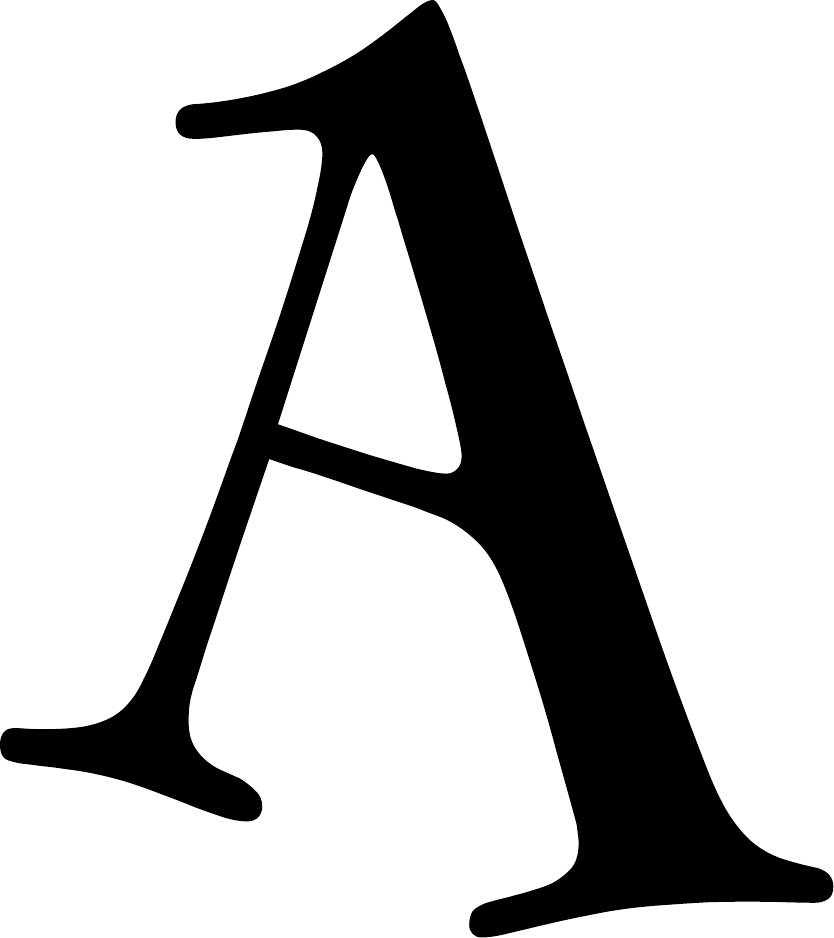}} &
        \hspace{0.1cm}
        \includegraphics[height=0.07\textwidth]{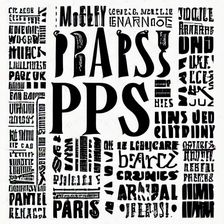} &
        \includegraphics[height=0.07\textwidth]{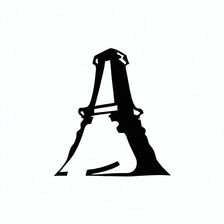} &
        \hspace{0.1cm}
        \includegraphics[height=0.07\textwidth]{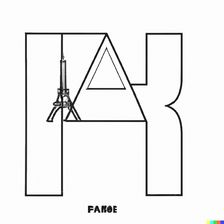} &
        \includegraphics[height=0.07\textwidth]{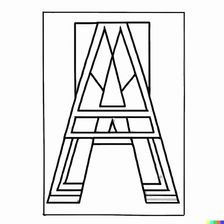} &
        \hspace{0.1cm}
        \includegraphics[height=0.07\textwidth]{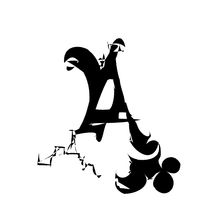} &
        \raisebox{0.2cm}{\includegraphics[height=0.048\textwidth]{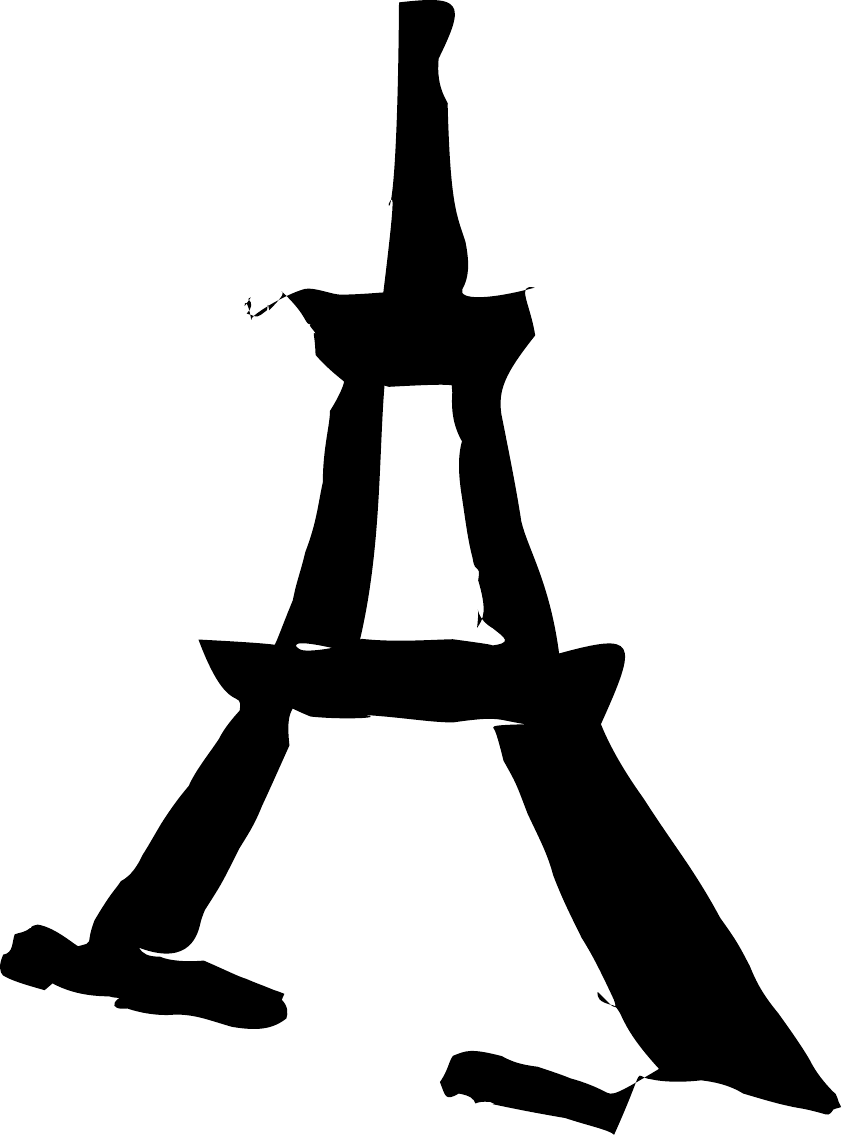}} \\

        \raisebox{0.4cm}{\makecell[l]{"Owl"}} &
        \hspace{0.1cm}
        \raisebox{0.2cm}{\includegraphics[height=0.04\textwidth]{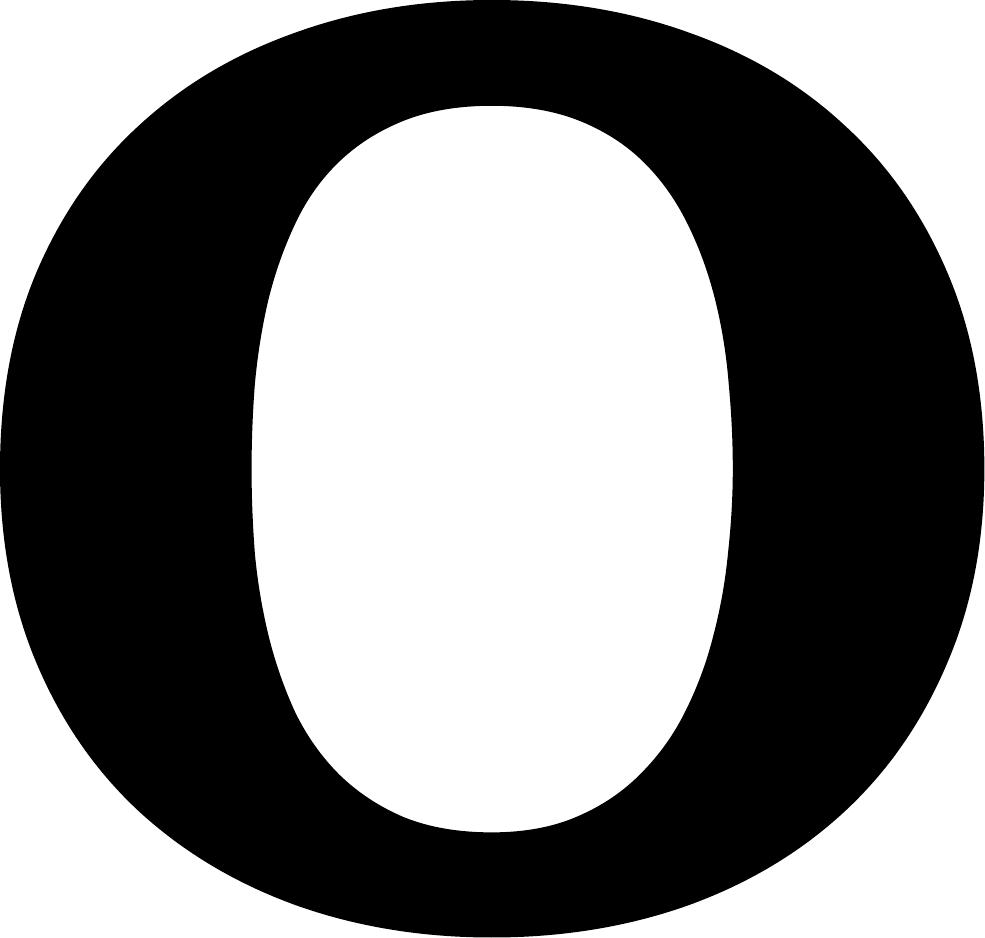}} &
        \hspace{0.1cm}
        \includegraphics[height=0.07\textwidth]{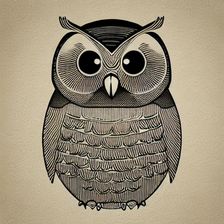} &
        \includegraphics[height=0.07\textwidth]{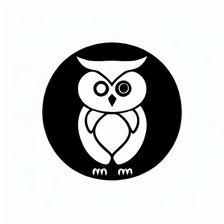} &
        \hspace{0.1cm}
        \includegraphics[height=0.07\textwidth]{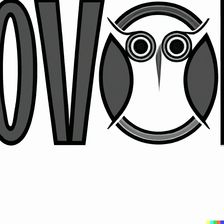} &
        \includegraphics[height=0.07\textwidth]{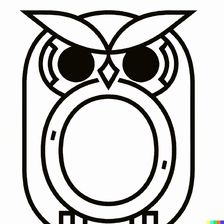} &
        \hspace{0.1cm}
        \includegraphics[height=0.07\textwidth]{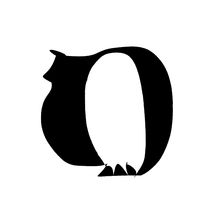} &
        \raisebox{0.2cm}{\includegraphics[height=0.048\textwidth]{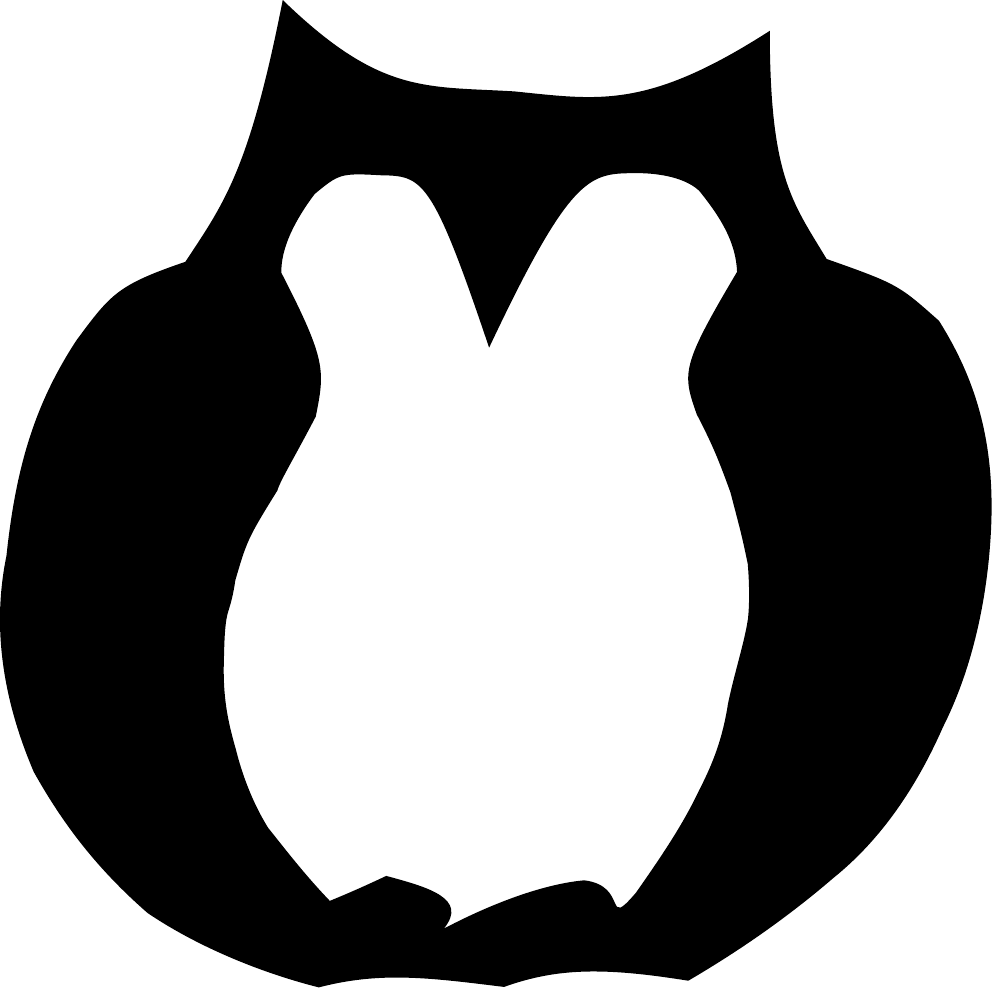}} \\

        \raisebox{0.4cm}{\makecell[l]{"Swan"}} &
        \hspace{0.1cm}
        \raisebox{0.2cm}{\includegraphics[height=0.04\textwidth]{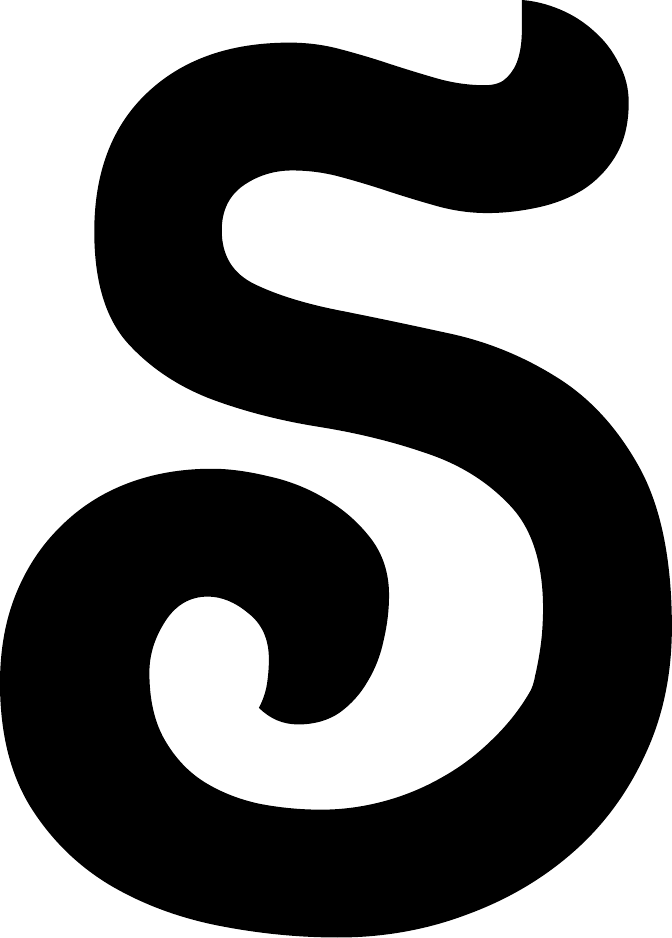}} &
        \hspace{0.1cm}
        \includegraphics[height=0.07\textwidth]{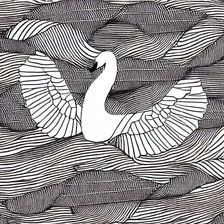} &
        \includegraphics[height=0.07\textwidth]{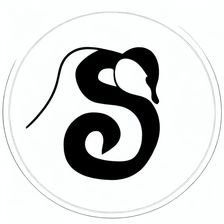} &
        \hspace{0.1cm}
        \includegraphics[height=0.07\textwidth]{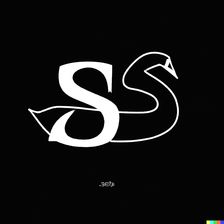} &
        \includegraphics[height=0.07\textwidth]{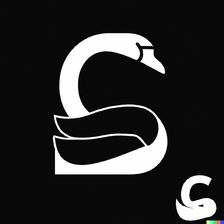} &
        \hspace{0.1cm}
        \includegraphics[height=0.07\textwidth]{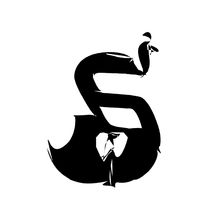} &
        \raisebox{0.2cm}{\includegraphics[height=0.048\textwidth]{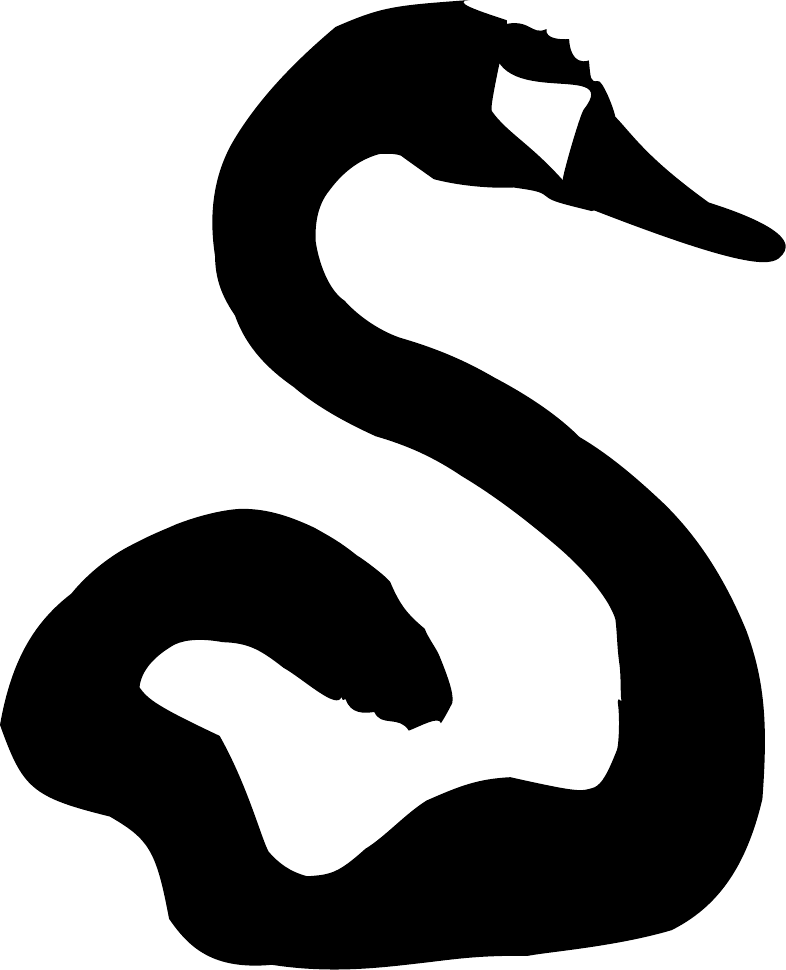}} \\

        \raisebox{0.4cm}{\makecell[l]{"Mermaid"}} &
        \hspace{0.1cm}
        \raisebox{0.2cm}{\includegraphics[height=0.04\textwidth]{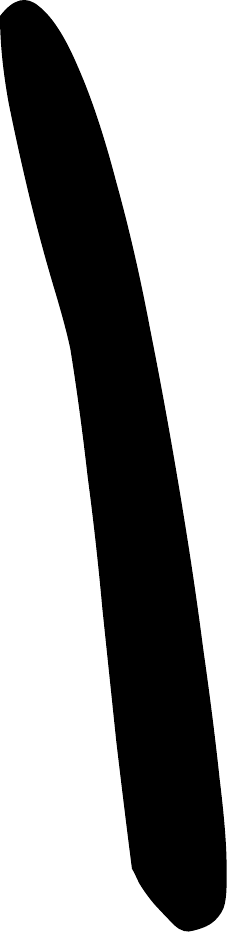}} &
        \hspace{0.1cm}
        \includegraphics[height=0.07\textwidth]{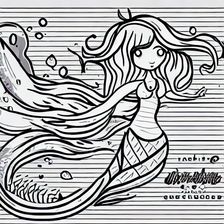} &
        \includegraphics[height=0.07\textwidth]{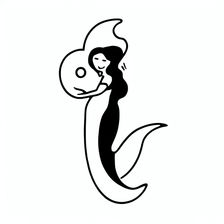} &
        \hspace{0.1cm}
        \includegraphics[height=0.07\textwidth]{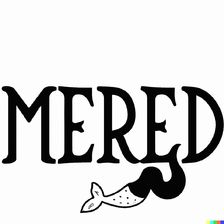} &
        \includegraphics[height=0.07\textwidth]{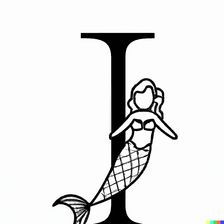} &
        \hspace{0.1cm}
        \includegraphics[height=0.07\textwidth]{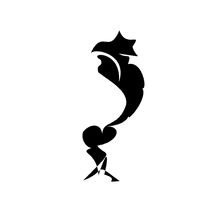} &
        \raisebox{0.2cm}{\includegraphics[height=0.048\textwidth]{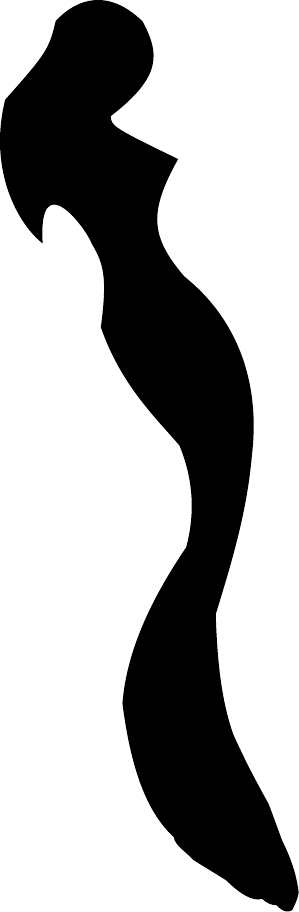}} \\

        \multicolumn{2}{c}{Input}& SD & SDEdit & DallE2 & DallE2+letter & CLIPDraw & Ours

    \end{tabular}
    }
    \caption{Comparison to alternative methods based on large scale text-to-image models. On the left are the letters used as input (only for SDEdit, CLIPDraw, and ours), as well as the desired object of interest. The results from left to right obtained using Stable Diffusion \cite{stableDiffusion}, SDEdit \cite{meng2022sdedit}, DallE2 \cite{ramesh2022hierarchical}, DallE2 with a letter specific prompt, CLIPDraw \cite{frans2021clipdraw}, and our single-letter results.}
    \label{fig:supp_comp_diffusion}
\end{figure*}

\section{Additional Results}
\begin{figure*}[b] 
 \centering 
 \setlength{\tabcolsep}{12pt} 
 \renewcommand{\arraystretch}{3} 
	 \includegraphics[height=0.64\linewidth]{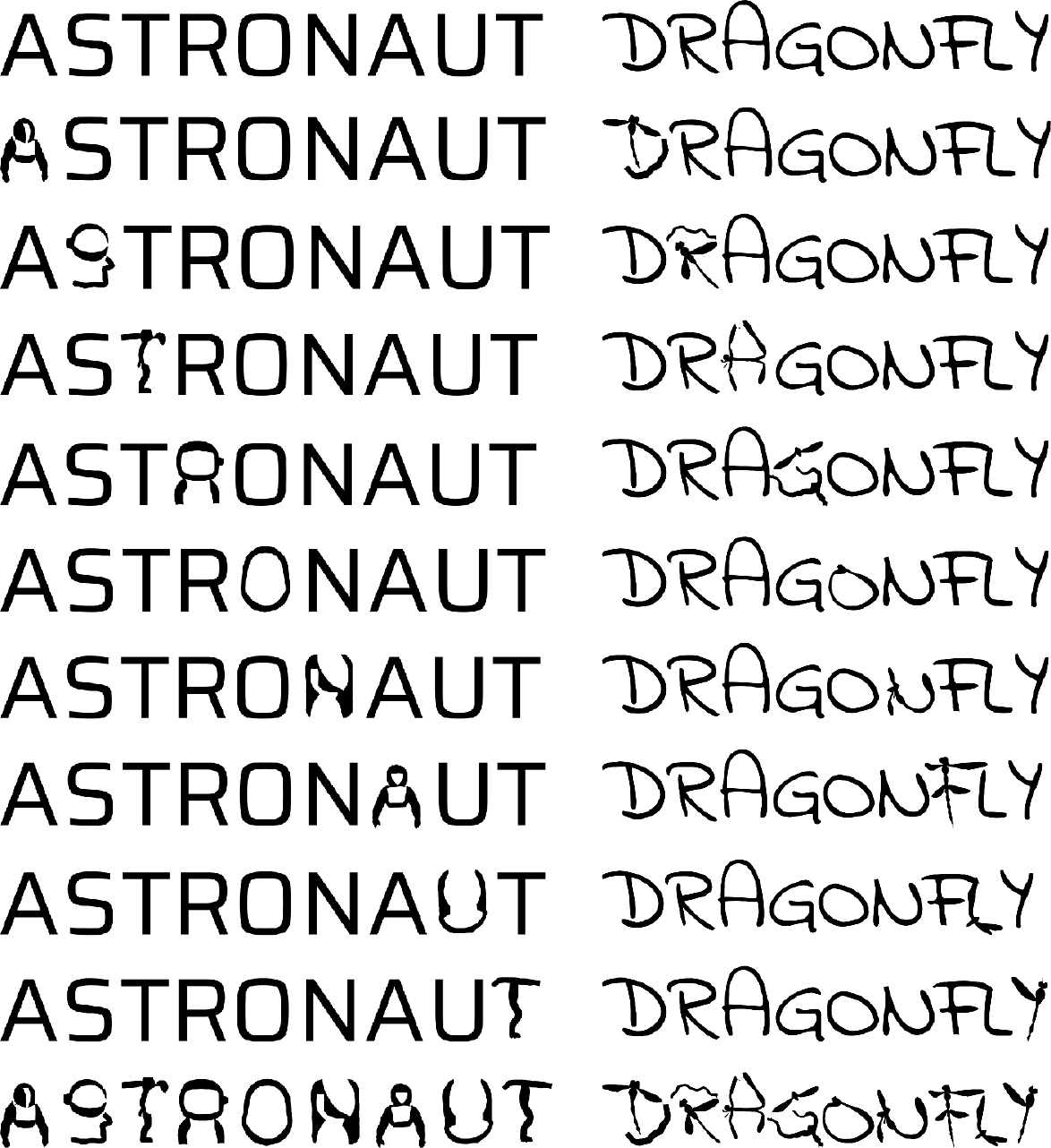} \\
\caption{Word-as-image illustrations created by our method.}
 \label{fig:all_art1} 
\end{figure*}

\begin{figure*}[ht] 
 \centering 
 \setlength{\tabcolsep}{12pt} 
 \renewcommand{\arraystretch}{3} 
	 \includegraphics[height=0.29\linewidth]{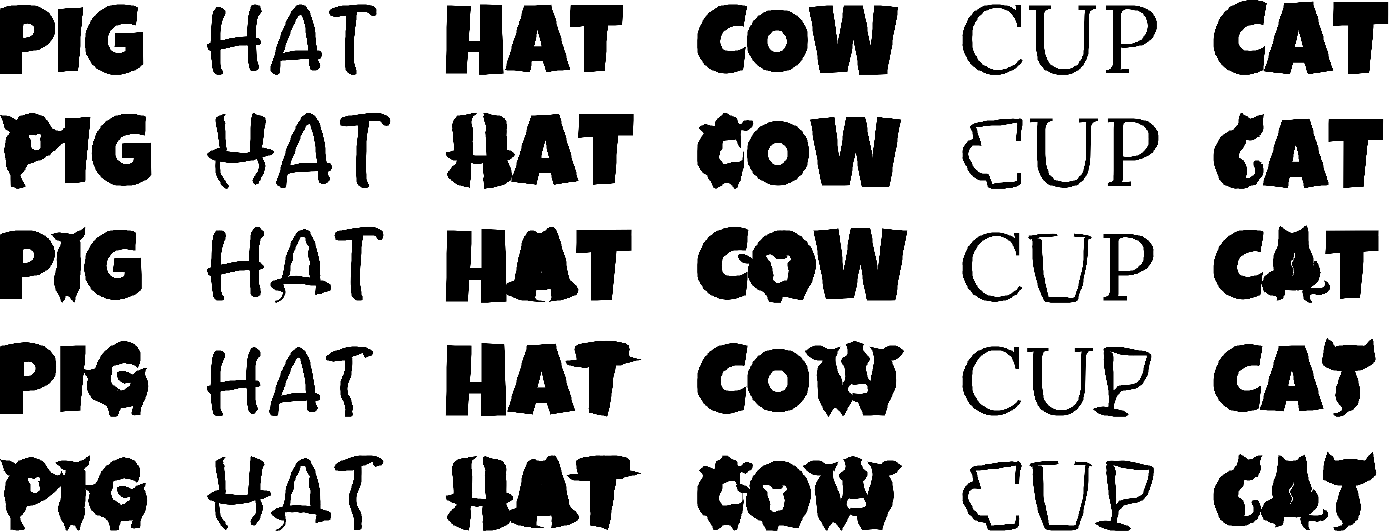} \\
\caption{Word-as-image illustrations created by our method.}
 \label{fig:all_art2} 
 \end{figure*}

\begin{figure*}[ht] 
 \centering 
 \setlength{\tabcolsep}{12pt} 
 \renewcommand{\arraystretch}{3} 
	 \includegraphics[height=0.35\linewidth]{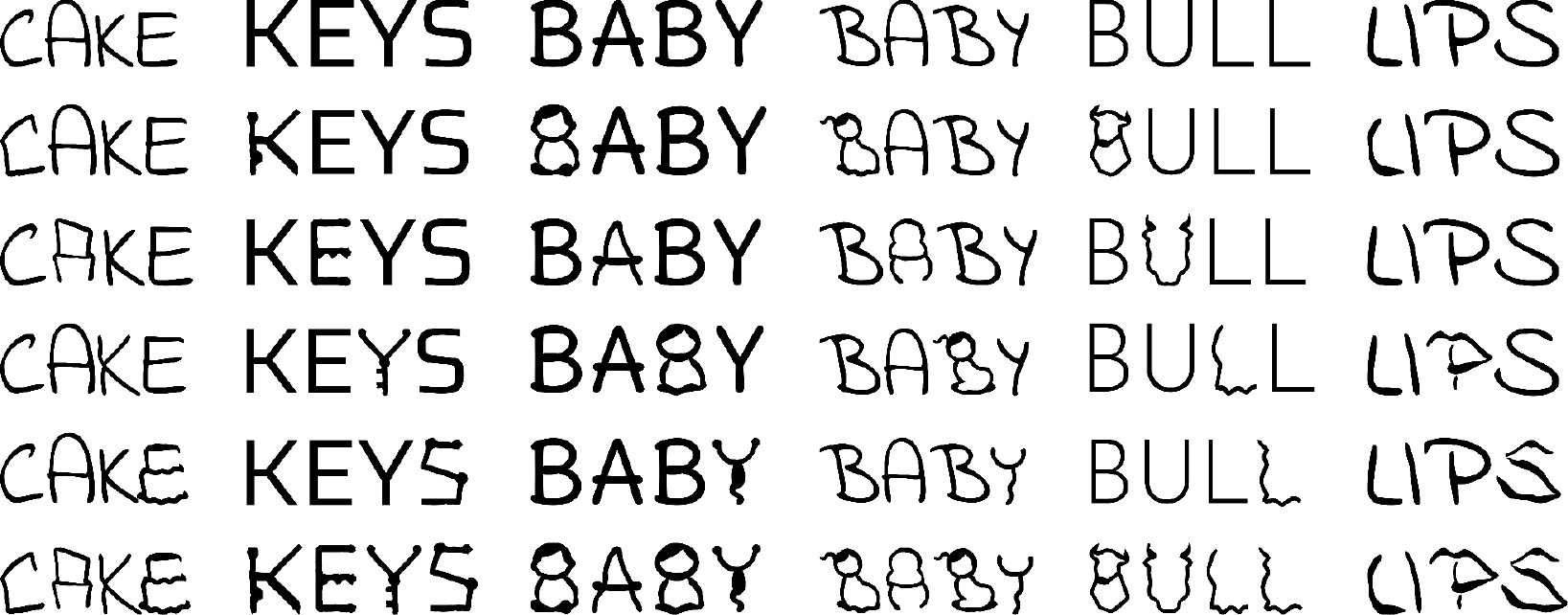} \\
\caption{Word-as-image illustrations created by our method.}
 \label{fig:all_art3} 
 \end{figure*}

\begin{figure*}[ht] 
 \centering 
 \setlength{\tabcolsep}{12pt} 
 \renewcommand{\arraystretch}{3} 
	 \includegraphics[height=0.35\linewidth]{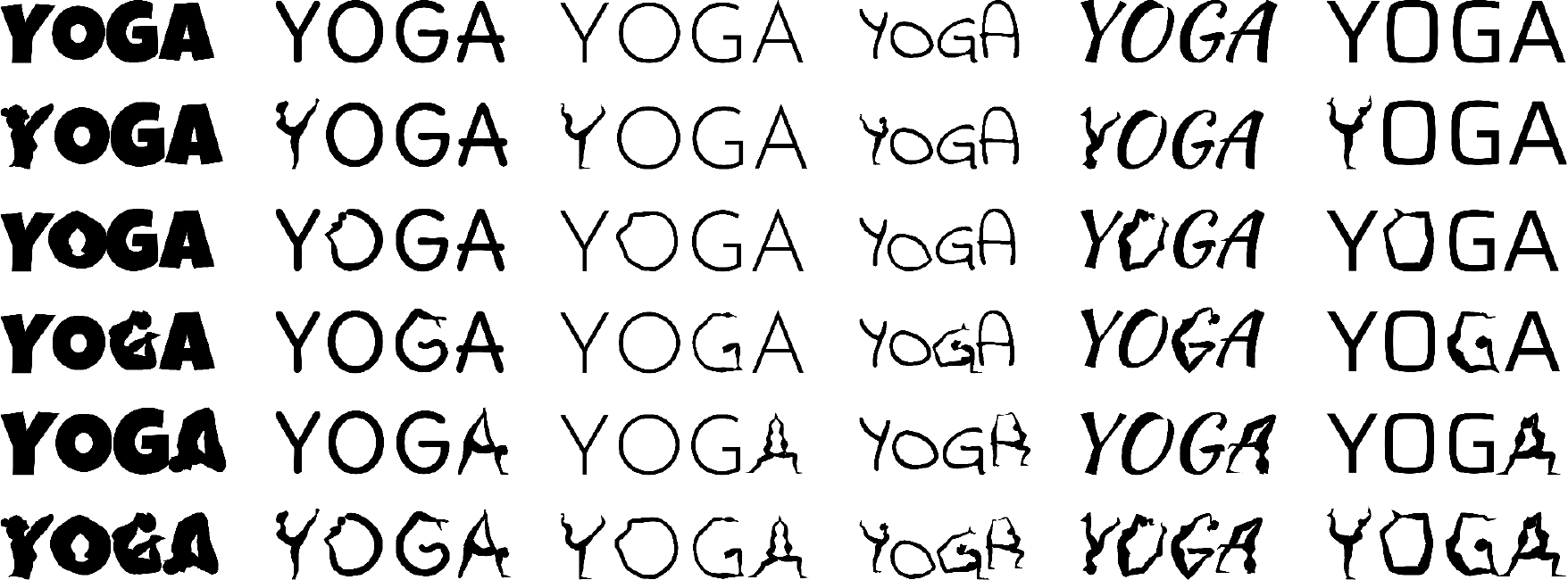} \\
\caption{Word-as-image illustrations created by our method.}
 \label{fig:all_art4} 
 \end{figure*}

\begin{figure*}[ht] 
 \centering 
 \setlength{\tabcolsep}{12pt} 
 \renewcommand{\arraystretch}{3} 
	 \includegraphics[height=0.41\linewidth]{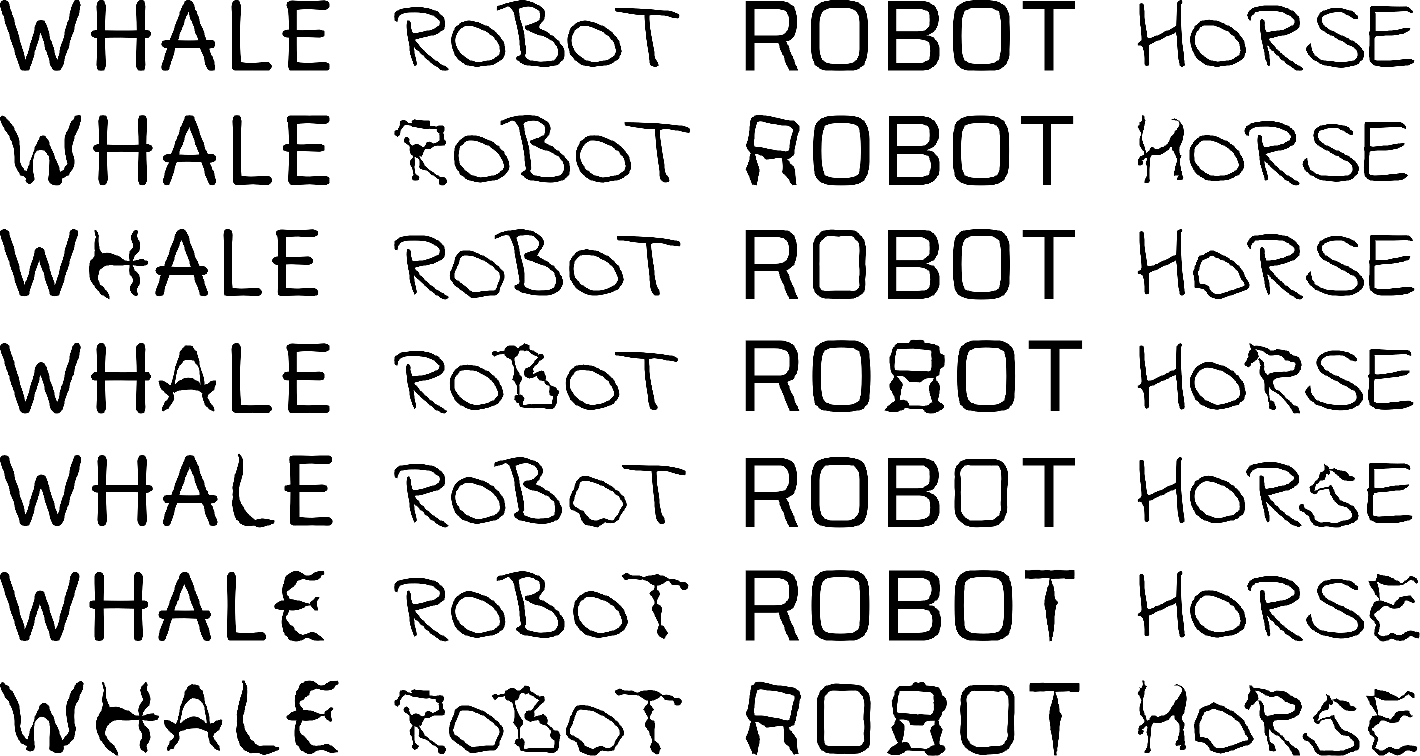} \\
\caption{Word-as-image illustrations created by our method.}
 \label{fig:all_art5} 
 \end{figure*}

\begin{figure*}[ht] 
 \centering 
 \setlength{\tabcolsep}{12pt} 
 \renewcommand{\arraystretch}{3} 
	 \includegraphics[height=0.41\linewidth]{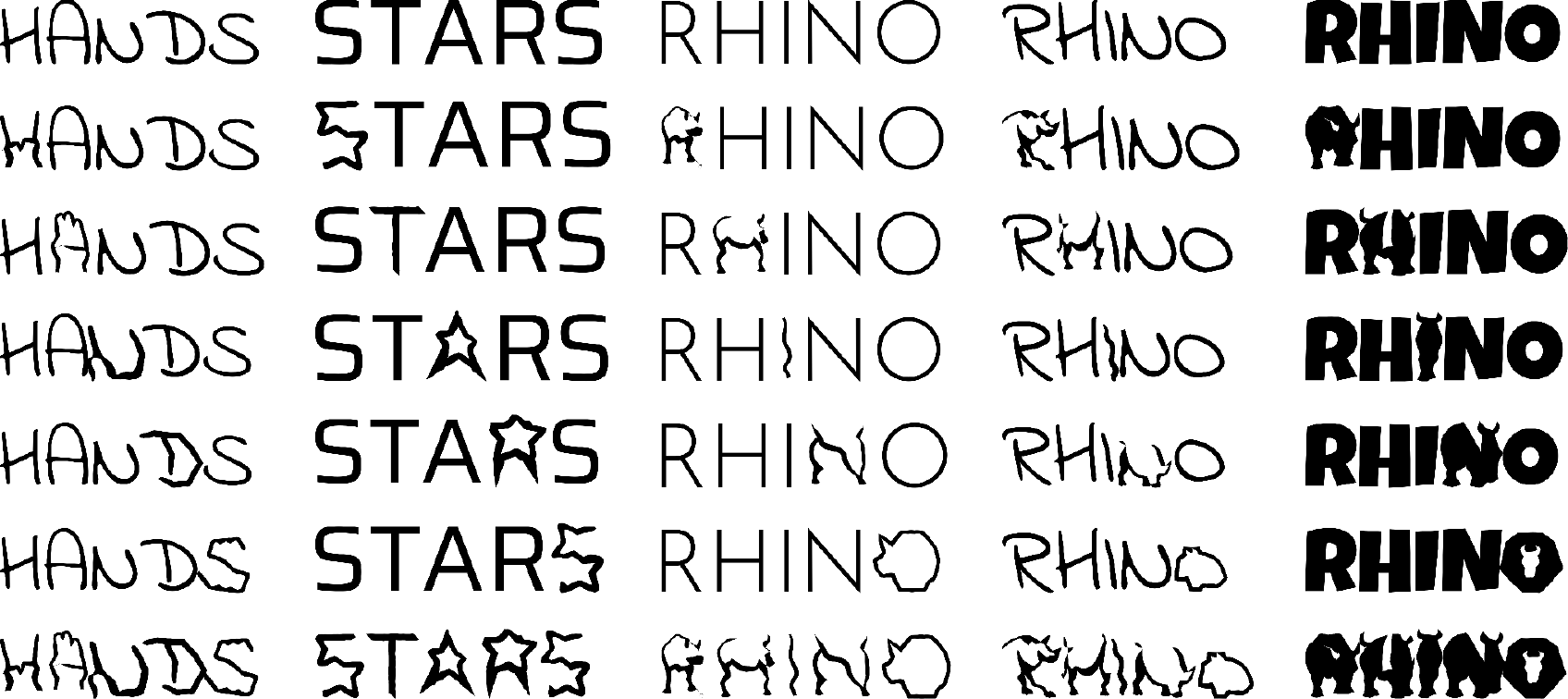} \\
\caption{Word-as-image illustrations created by our method.}
 \label{fig:all_art6} 
 \end{figure*}

\begin{figure*}[ht] 
 \centering 
 \setlength{\tabcolsep}{12pt} 
 \renewcommand{\arraystretch}{3} 
	 \includegraphics[height=0.41\linewidth]{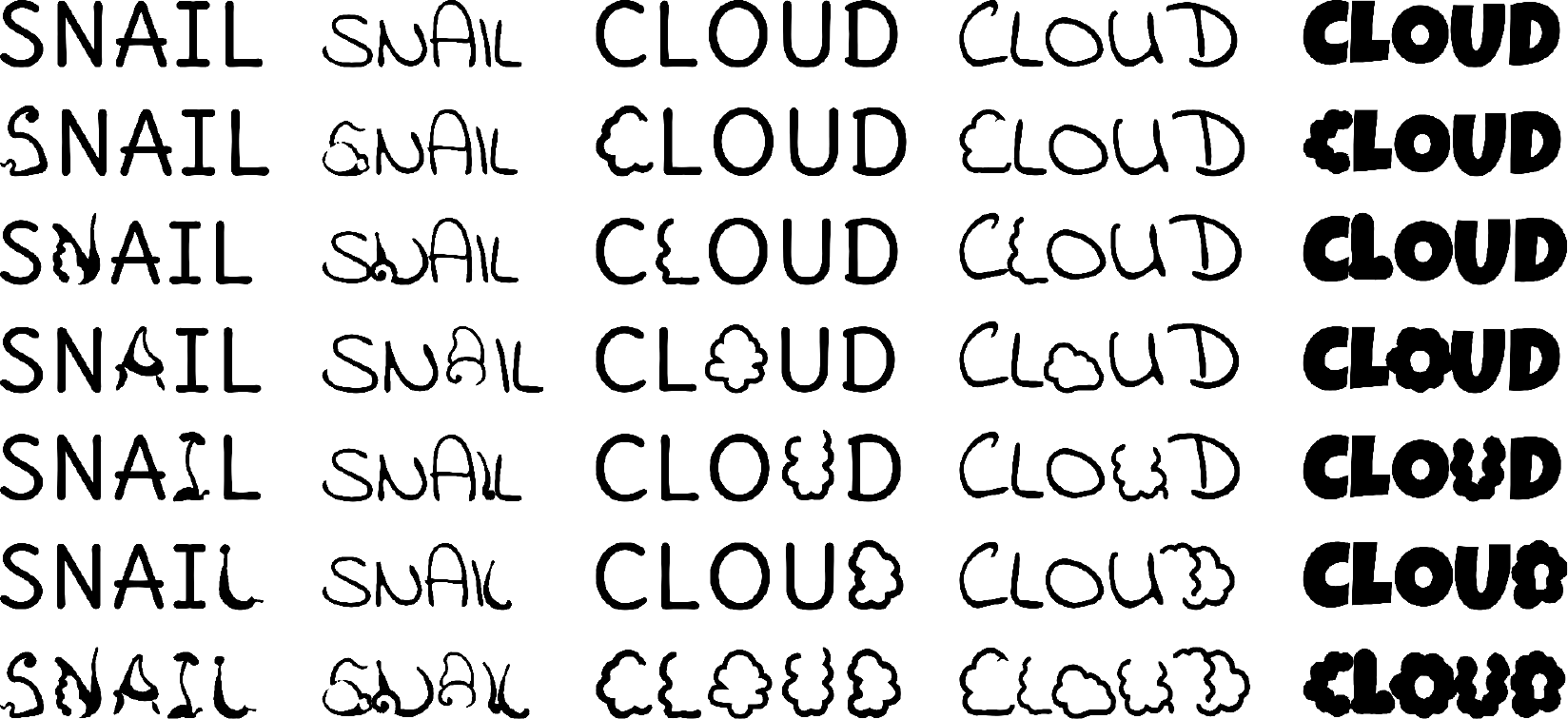} \\
\caption{Word-as-image illustrations created by our method.}
 \label{fig:all_art7} 
 \end{figure*}

\begin{figure*}[ht] 
 \centering 
 \setlength{\tabcolsep}{12pt} 
 \renewcommand{\arraystretch}{3} 
	 \includegraphics[height=0.41\linewidth]{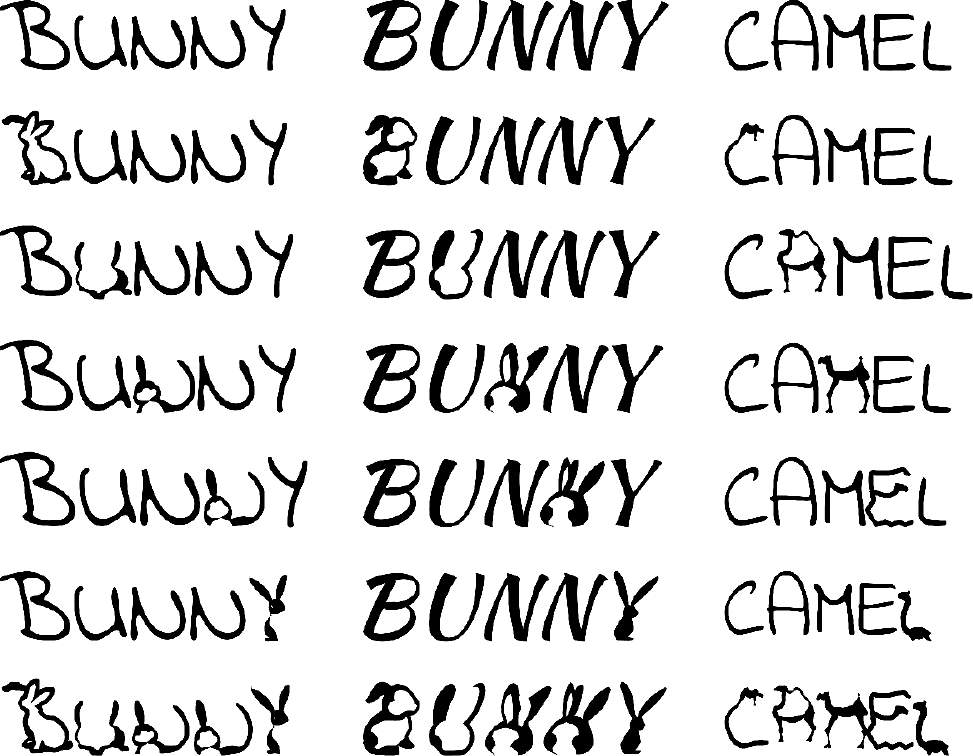} \\
\caption{Word-as-image illustrations created by our method.}
 \label{fig:all_art8} 
 \end{figure*}

\begin{figure*}[ht] 
 \centering 
 \setlength{\tabcolsep}{12pt} 
 \renewcommand{\arraystretch}{3} 
	 \includegraphics[height=0.47\linewidth]{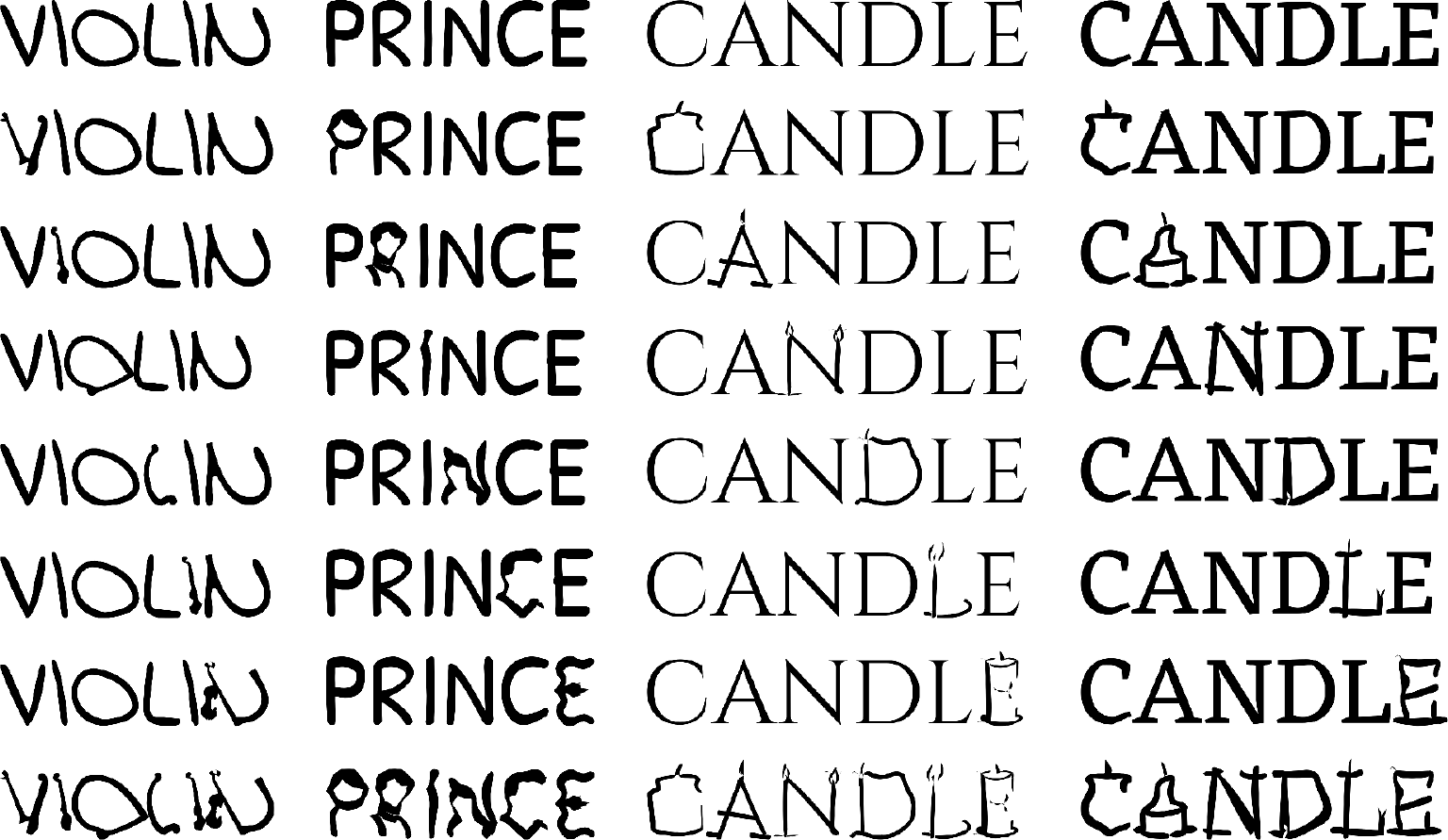} \\
\caption{Word-as-image illustrations created by our method.}
 \label{fig:all_art9} 
 \end{figure*}

\begin{figure*}[ht] 
 \centering 
 \setlength{\tabcolsep}{12pt} 
 \renewcommand{\arraystretch}{3} 
	 \includegraphics[height=0.47\linewidth]{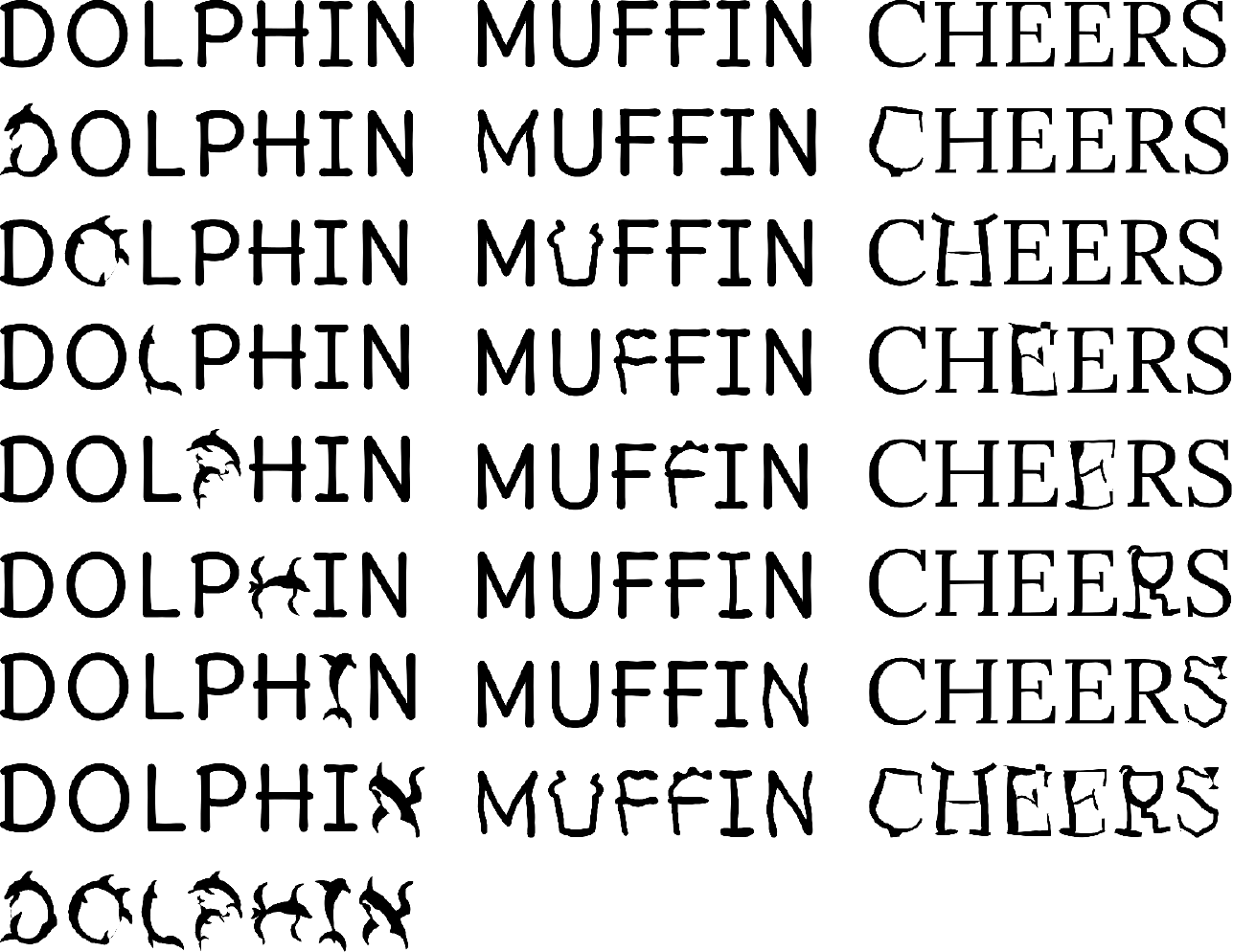} \\
\caption{Word-as-image illustrations created by our method.}
 \label{fig:all_art10} 
 \end{figure*}

\begin{figure*}[ht] 
 \centering 
 \setlength{\tabcolsep}{12pt} 
 \renewcommand{\arraystretch}{3} 
	 \includegraphics[height=0.47\linewidth]{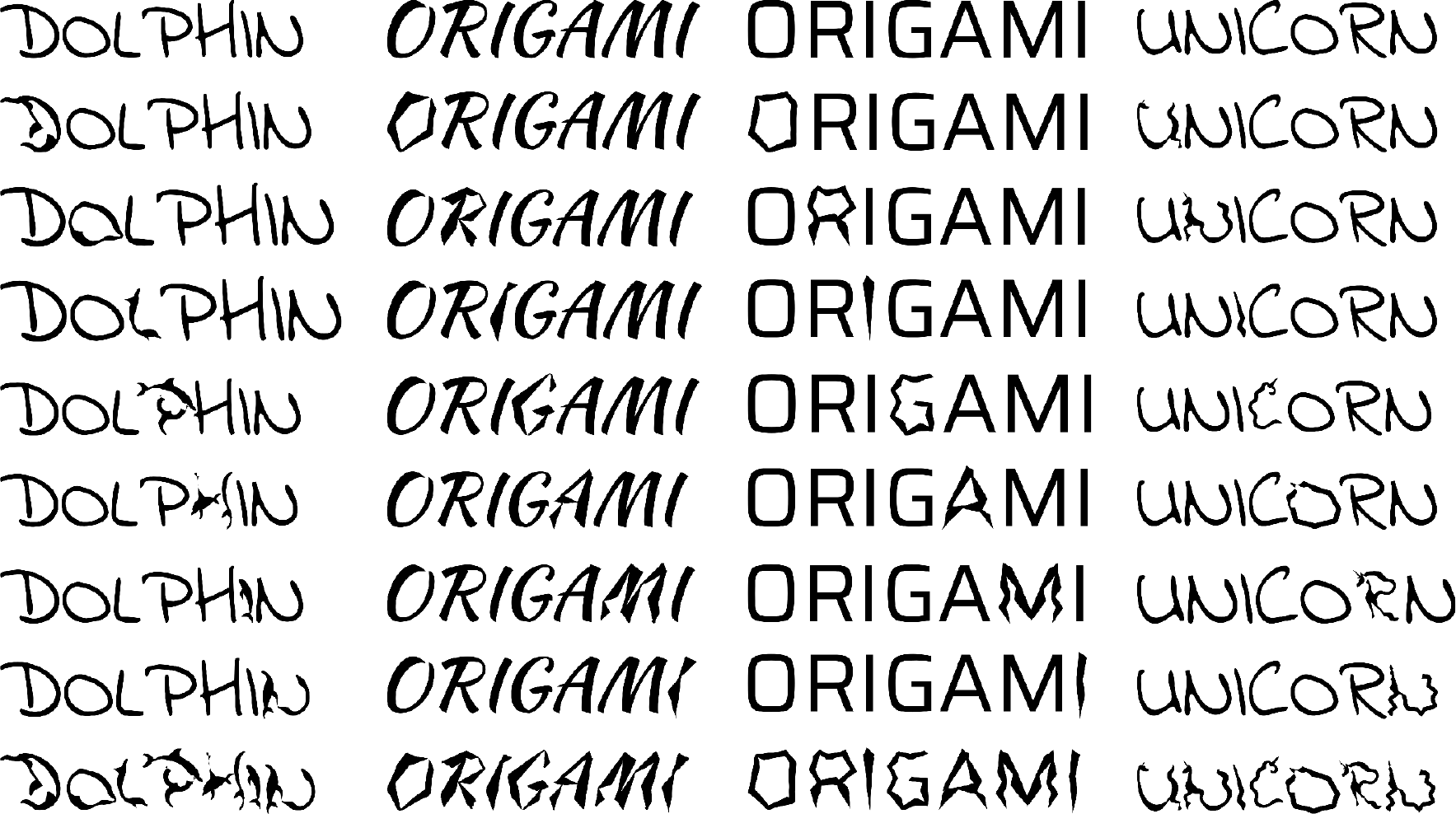} \\
\caption{Word-as-image illustrations created by our method.}
 \label{fig:all_art11} 
 \end{figure*}

\begin{figure*}[ht] 
 \centering 
 \setlength{\tabcolsep}{12pt} 
 \renewcommand{\arraystretch}{3} 
	 \includegraphics[height=0.58\linewidth]{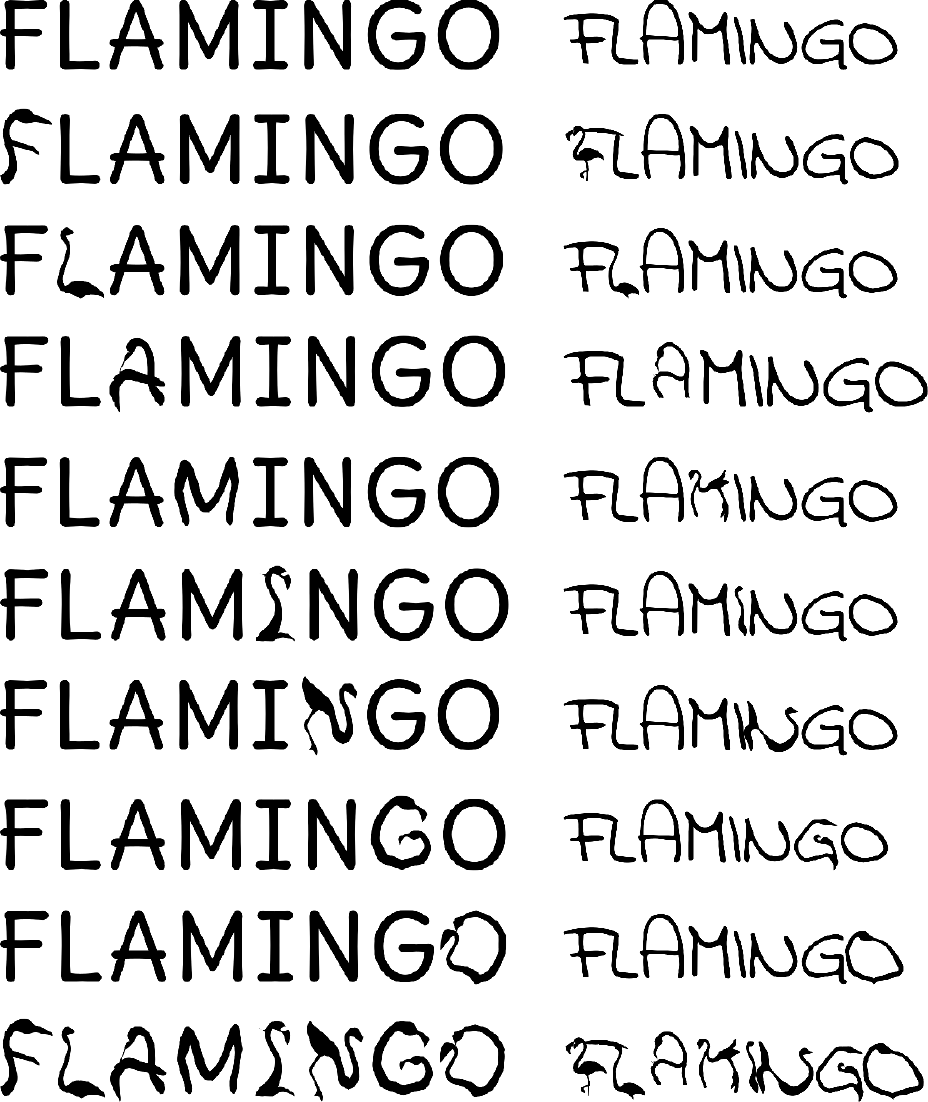} \\
\caption{Word-as-image illustrations created by our method.}
 \label{fig:all_art12} 
 \end{figure*}

\begin{figure*}[ht] 
 \centering 
 \setlength{\tabcolsep}{12pt} 
 \renewcommand{\arraystretch}{3} 
	 \includegraphics[height=0.7\linewidth]{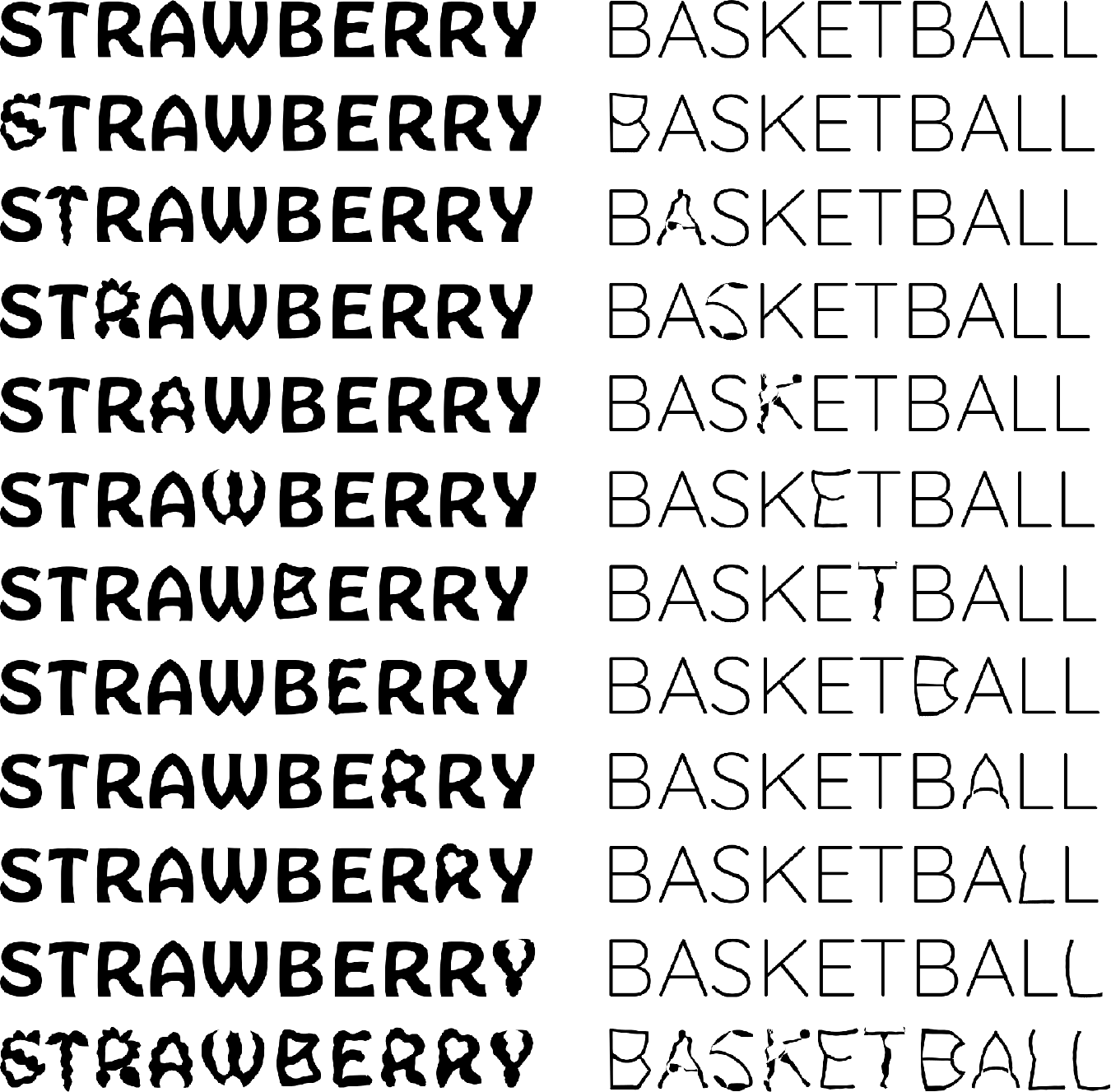} \\
 \caption{Word-as-image illustrations created by our method for randomly chosen words.}  
 \label{fig:all_res17} 
 \end{figure*}

We provide additional results of our generated word-as-images.
In Figures \ref{fig:all_art1}-\ref{fig:all_art12} we show results of selected words and unique fonts.
In Figures \ref{fig:all_res1}-\ref{fig:all_res17} we show the results obtained for the random set of words.

\clearpage
\newpage

\begin{figure*}[ht] 
 \centering 
 \setlength{\tabcolsep}{12pt} 
 \renewcommand{\arraystretch}{3} 
	 \includegraphics[height=0.35\linewidth]{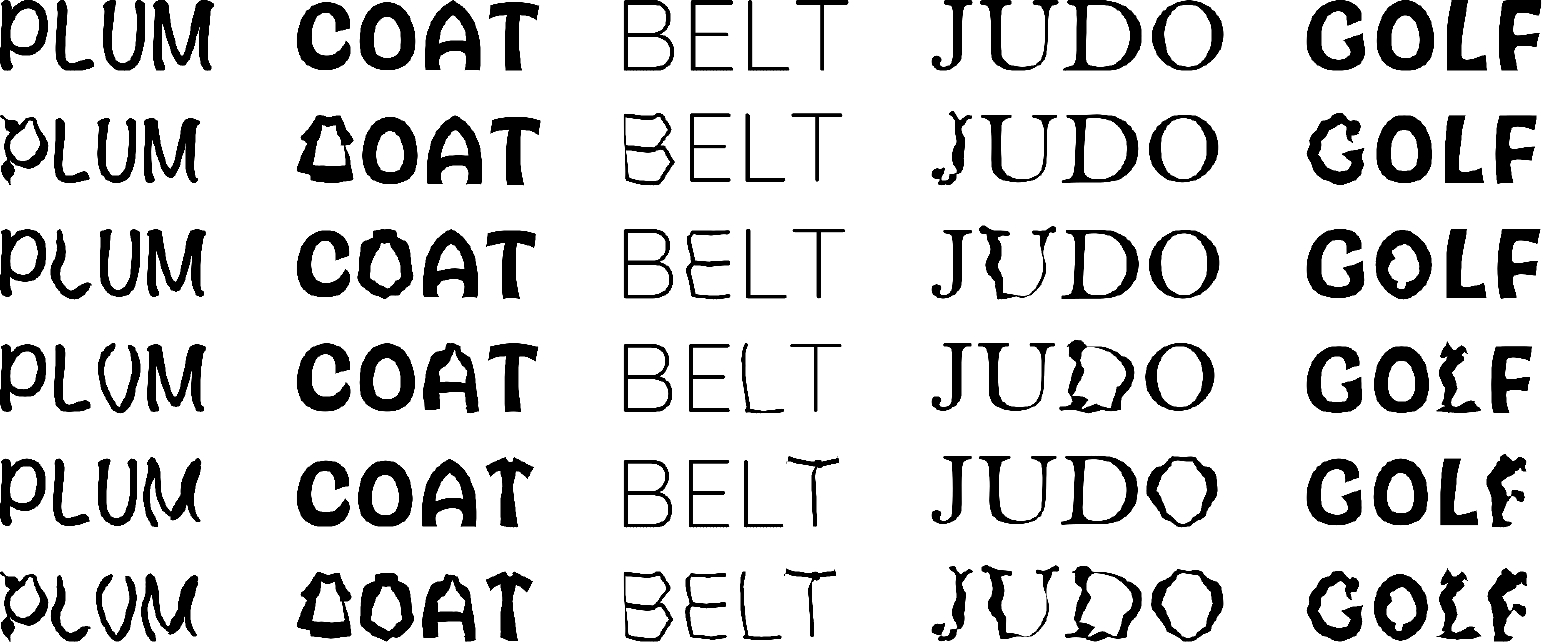} \\
 \caption{Word-as-image illustrations created by our method for randomly chosen words.} 
 \label{fig:all_res1} 
 \end{figure*}

 \begin{figure*}[ht] 
 \centering 
 \setlength{\tabcolsep}{12pt} 
 \renewcommand{\arraystretch}{3} 
	 \includegraphics[height=0.35\linewidth]{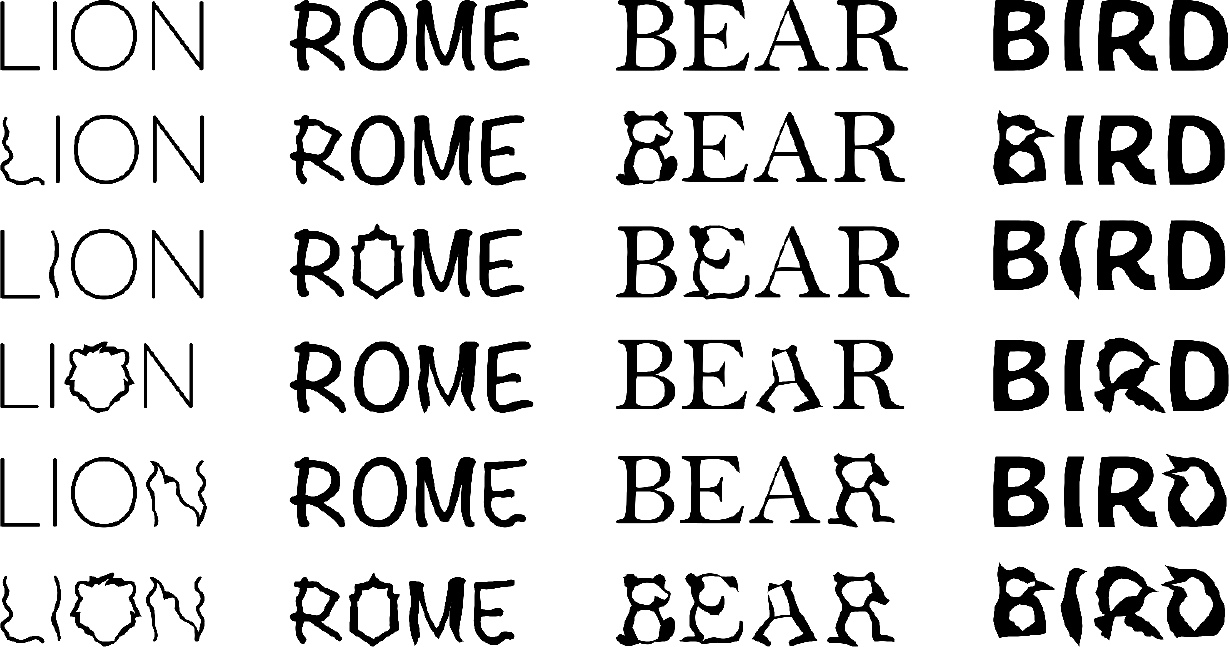} \\
 \caption{Word-as-image illustrations created by our method for randomly chosen words.} 
 \label{fig:all_res2} 
 \end{figure*}

 \begin{figure*}[ht] 
 \centering 
 \setlength{\tabcolsep}{12pt} 
 \renewcommand{\arraystretch}{3} 
	 \includegraphics[height=0.41\linewidth]{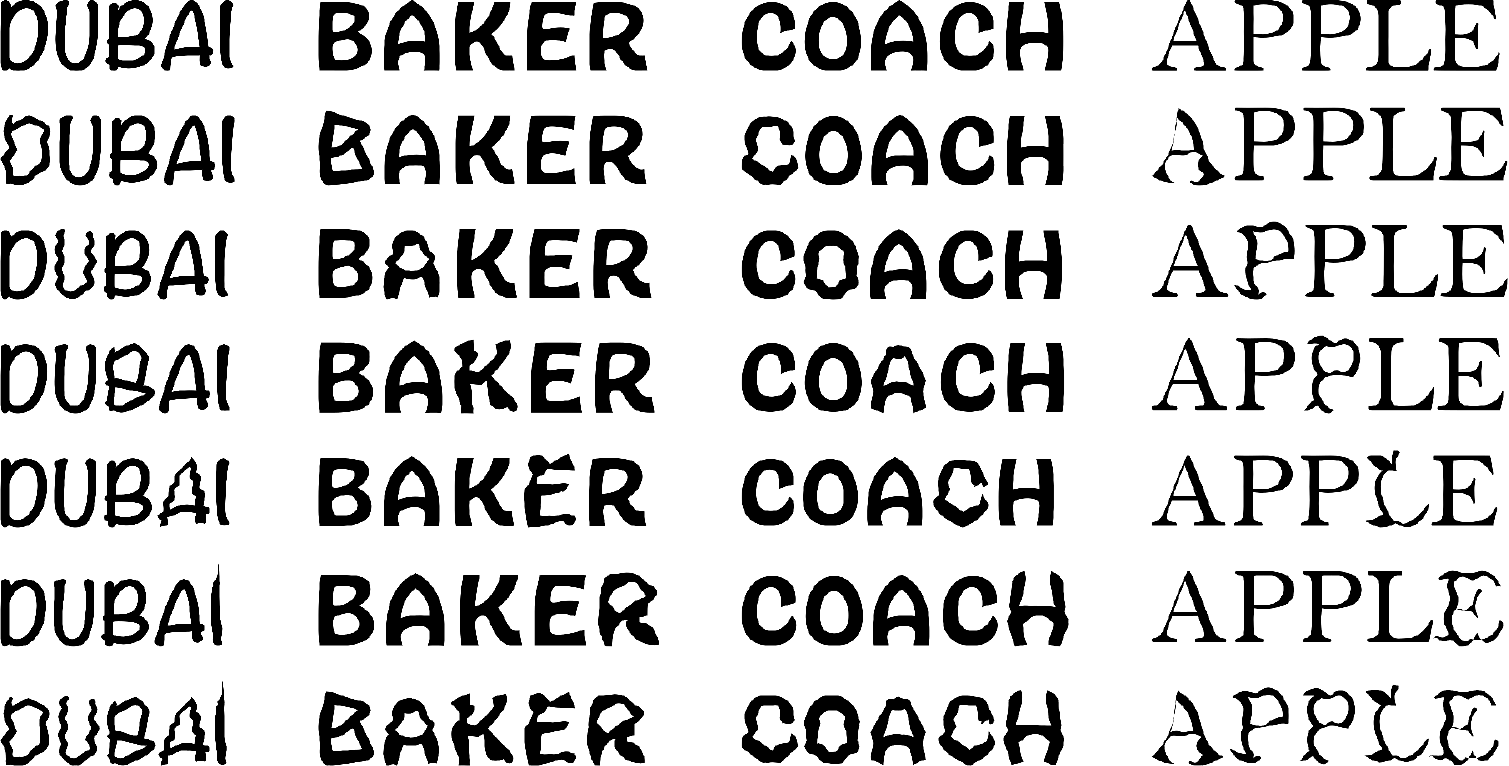} \\
 \caption{Word-as-image illustrations created by our method for randomly chosen words.}  
 \label{fig:all_res3} 
 \end{figure*}

 \begin{figure*}[ht] 
 \centering 
 \setlength{\tabcolsep}{12pt} 
 \renewcommand{\arraystretch}{3} 
	 \includegraphics[height=0.41\linewidth]{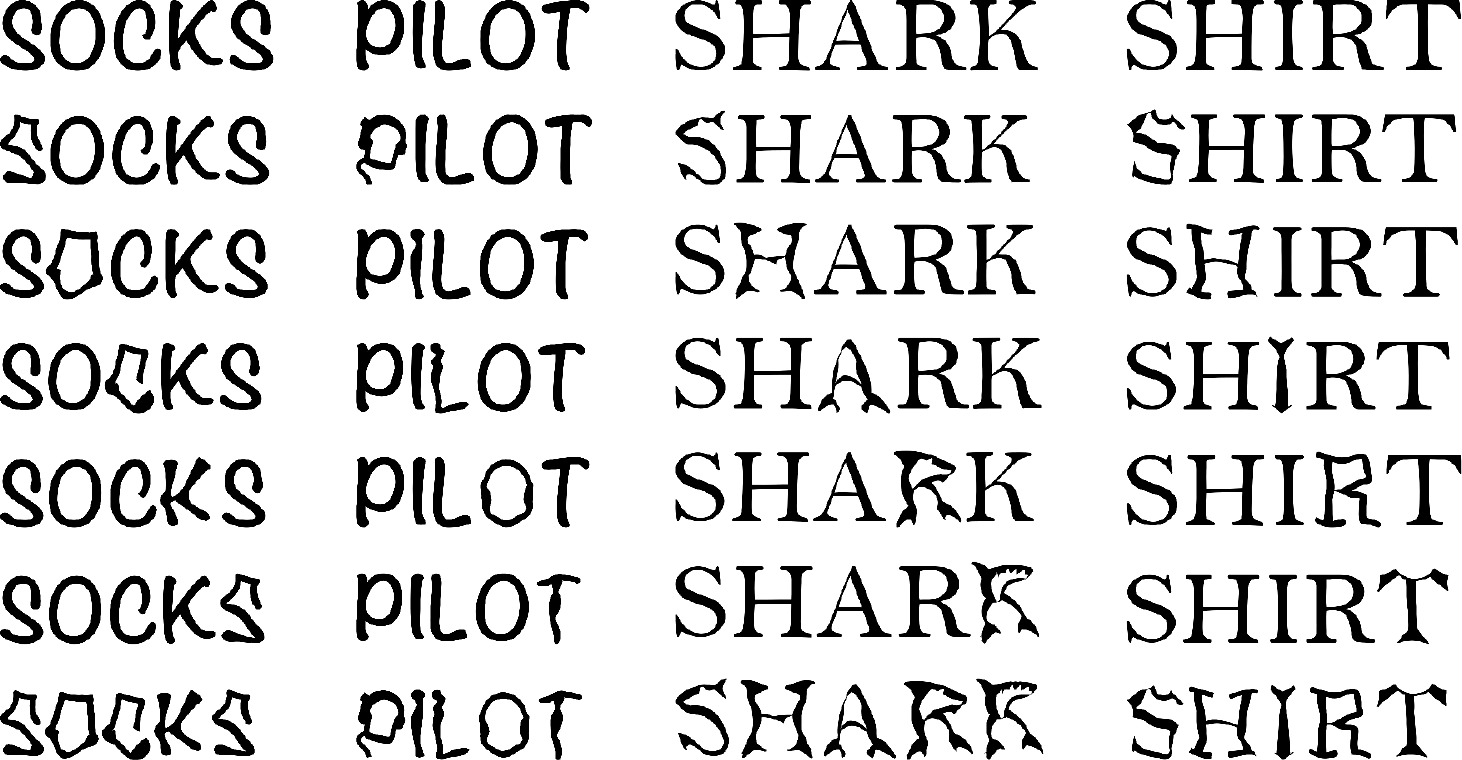} \\
 \caption{Word-as-image illustrations created by our method for randomly chosen words.}  
 \label{fig:all_res4} 
 \end{figure*}

 \begin{figure*}[ht] 
 \centering 
 \setlength{\tabcolsep}{12pt} 
 \renewcommand{\arraystretch}{3} 
	 \includegraphics[height=0.41\linewidth]{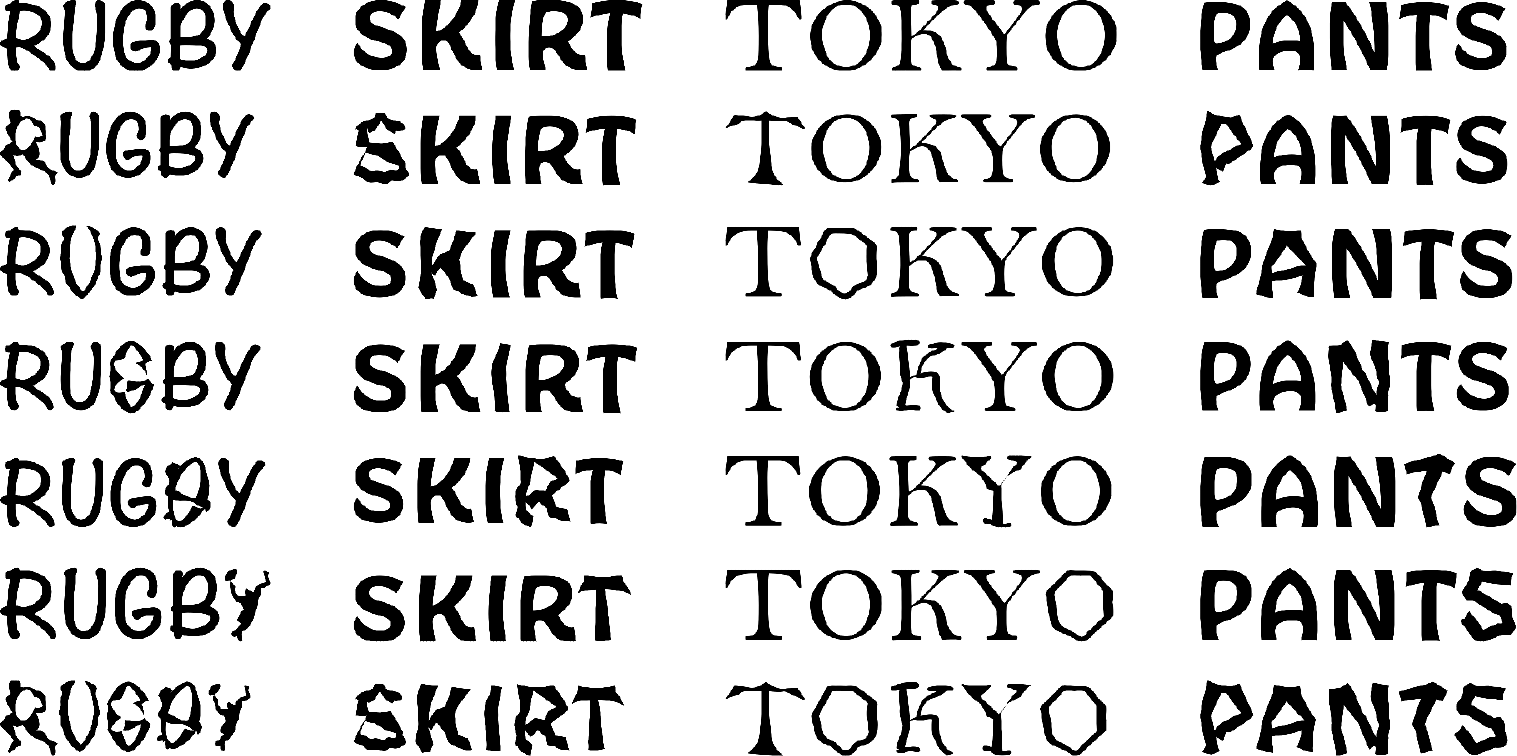} \\
 \caption{Word-as-image illustrations created by our method for randomly chosen words.}   
 \label{fig:all_res5} 
 \end{figure*}

 \begin{figure*}[ht] 
 \centering 
 \setlength{\tabcolsep}{12pt} 
 \renewcommand{\arraystretch}{3} 
	 \includegraphics[height=0.41\linewidth]{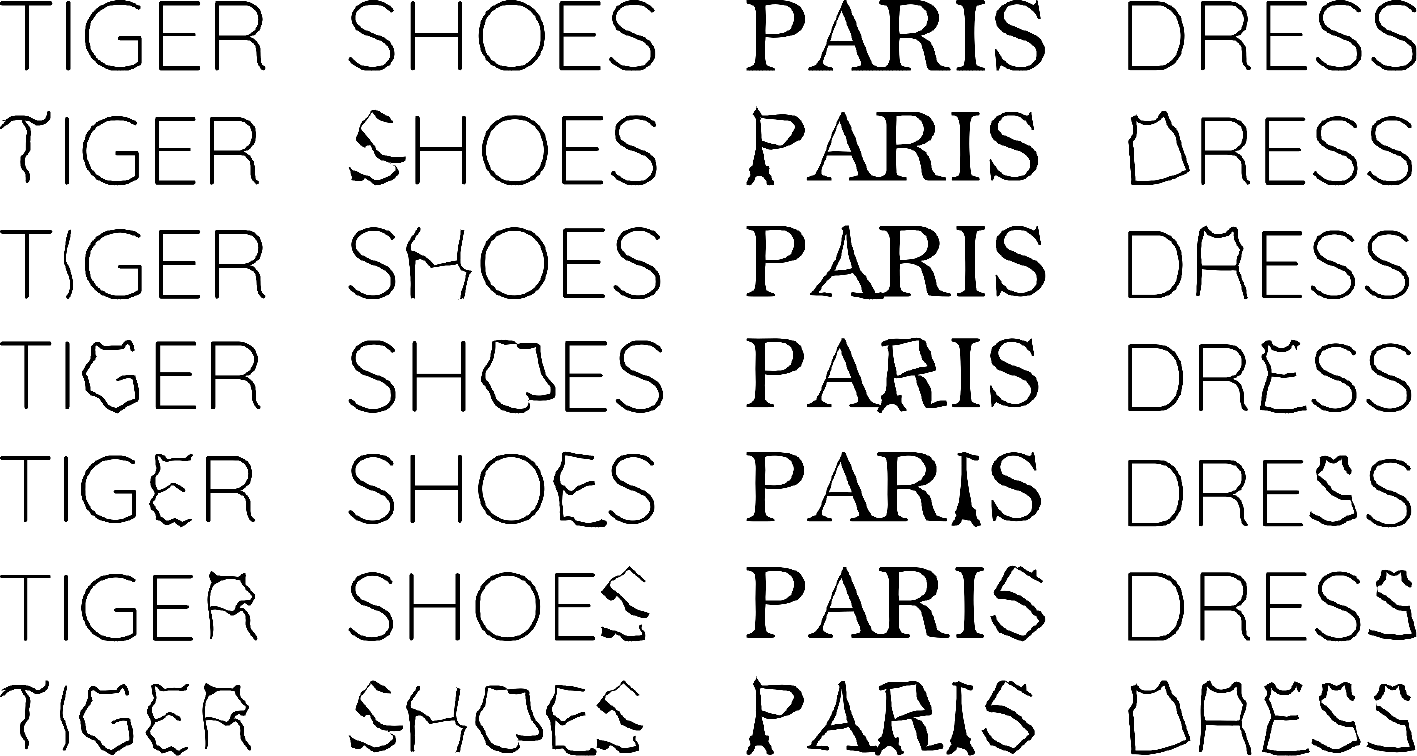} \\
 \caption{Word-as-image illustrations created by our method for randomly chosen words.}    
 \label{fig:all_res6} 
 \end{figure*}
 
 \begin{figure*}[ht] 
 \centering 
 \setlength{\tabcolsep}{12pt} 
 \renewcommand{\arraystretch}{3} 
	 \includegraphics[height=0.47\linewidth]{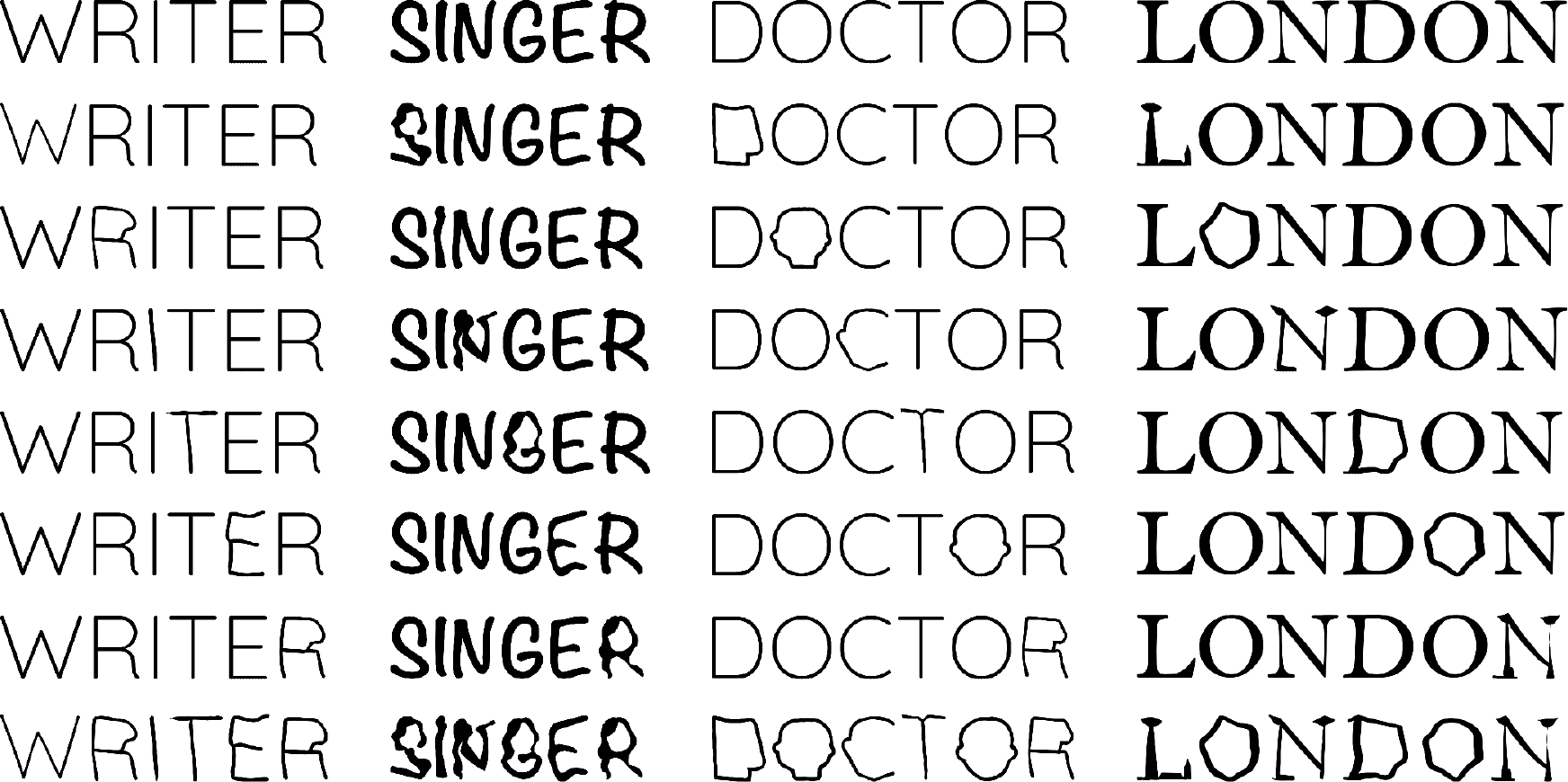} \\
 \caption{Word-as-image illustrations created by our method for randomly chosen words.}    
 \label{fig:all_res7} 
 \end{figure*}

 \begin{figure*}[ht] 
 \centering 
 \setlength{\tabcolsep}{12pt} 
 \renewcommand{\arraystretch}{3} 
	 \includegraphics[height=0.47\linewidth]{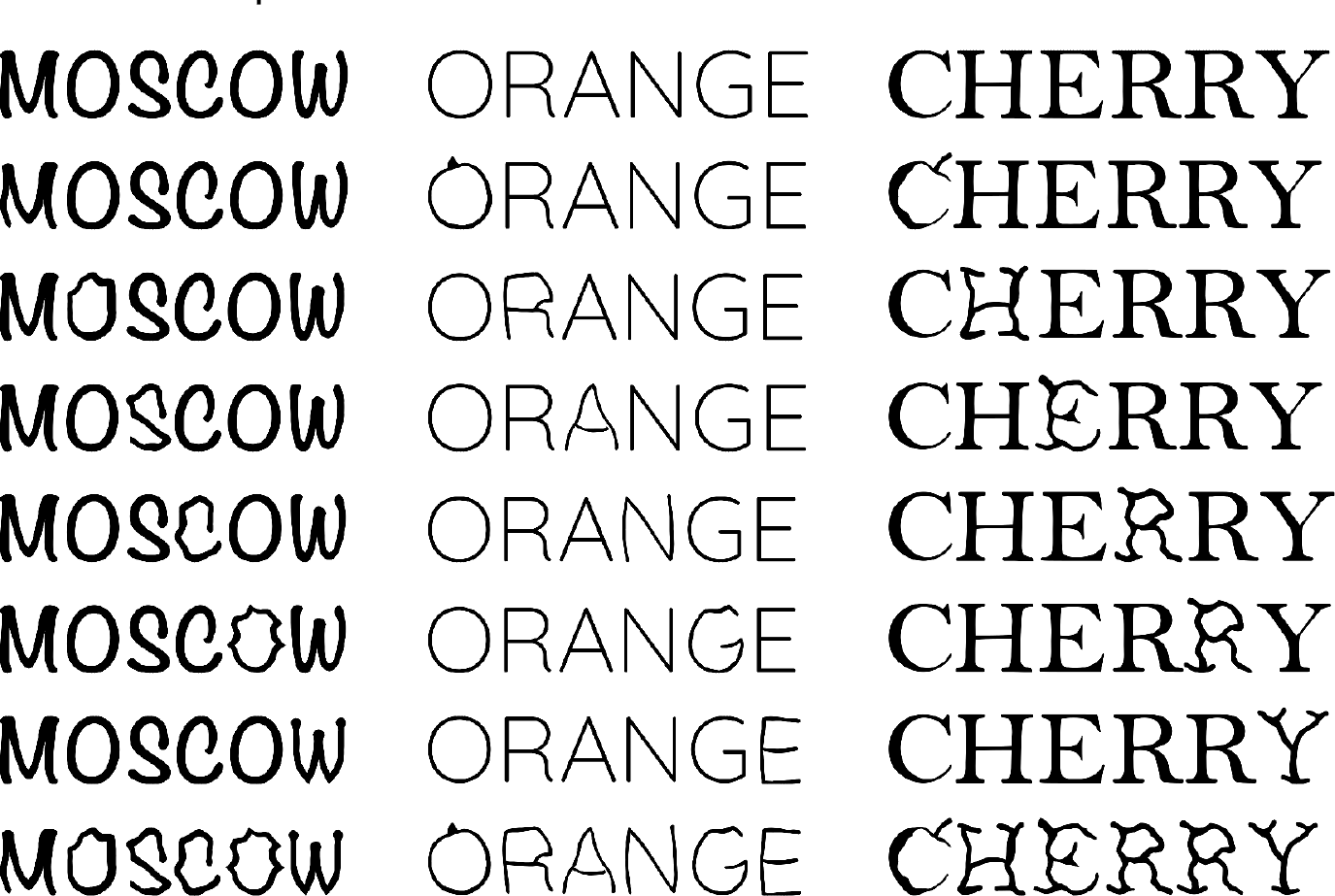} \\
 \caption{Word-as-image illustrations created by our method for randomly chosen words.}   
 \label{fig:all_res8} 
 \end{figure*}

 \begin{figure*}[ht] 
 \centering 
 \setlength{\tabcolsep}{12pt} 
 \renewcommand{\arraystretch}{3} 
	 \includegraphics[height=0.47\linewidth]{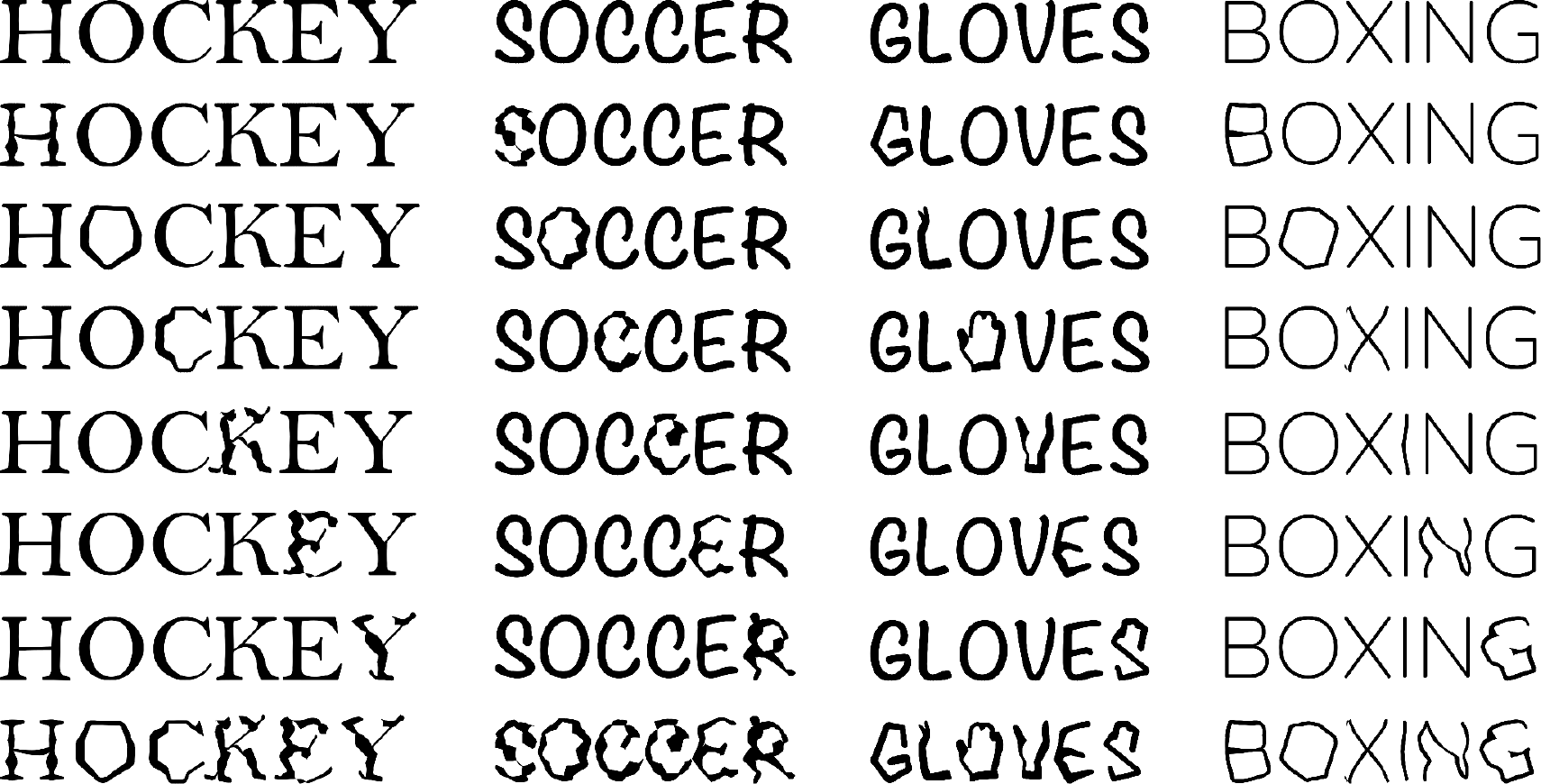} \\
 \caption{Word-as-image illustrations created by our method for randomly chosen words.}  
 \label{fig:all_res9} 
 \end{figure*}

 \begin{figure*}[ht] 
 \centering 
 \setlength{\tabcolsep}{12pt} 
 \renewcommand{\arraystretch}{3} 
	 \includegraphics[height=0.47\linewidth]{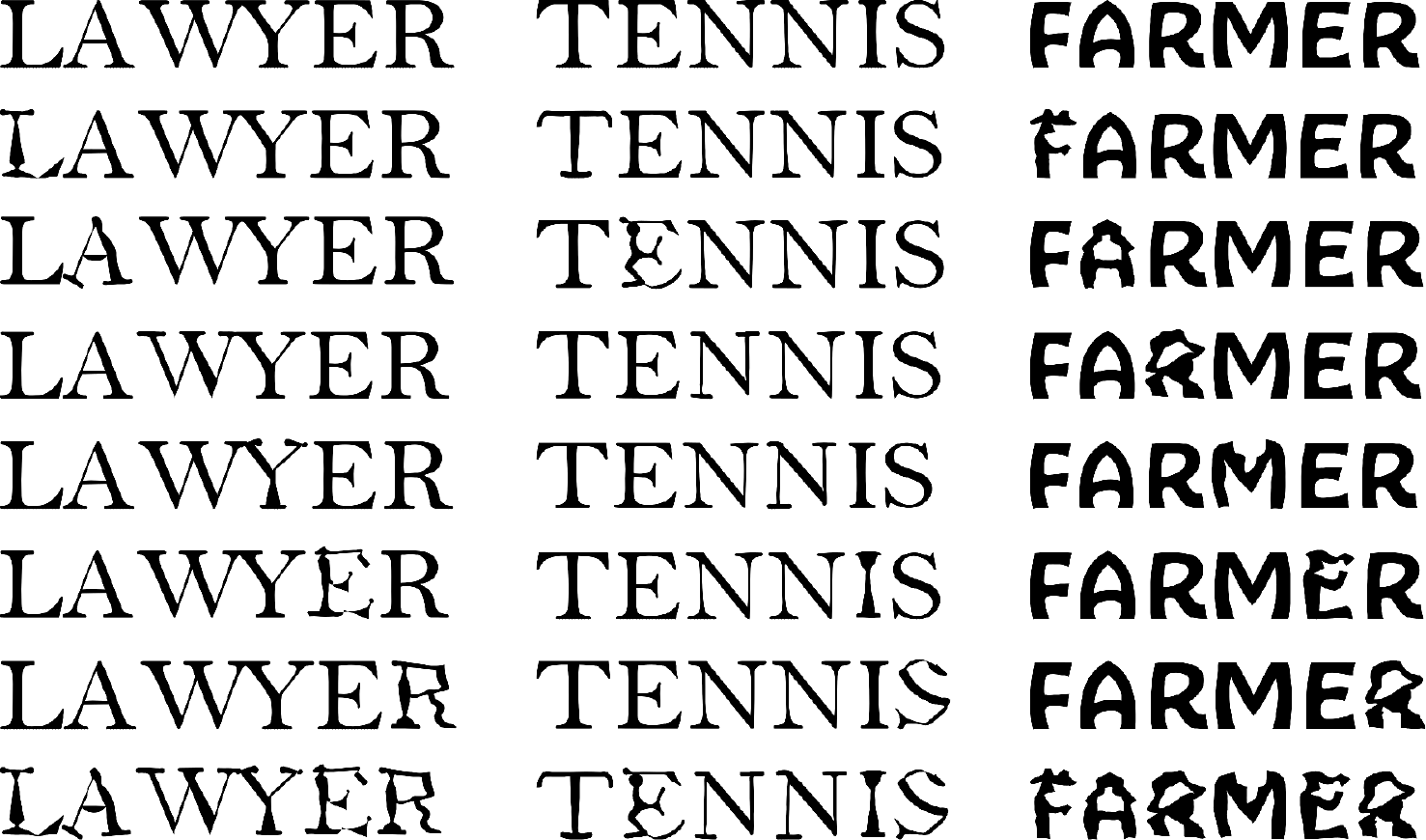} \\
 \caption{Word-as-image illustrations created by our method for randomly chosen words.}    
 \label{fig:all_res10} 
 \end{figure*}

 \begin{figure*}[ht] 
 \centering 
 \setlength{\tabcolsep}{12pt} 
 \renewcommand{\arraystretch}{3} 
	 \includegraphics[height=0.52\linewidth]{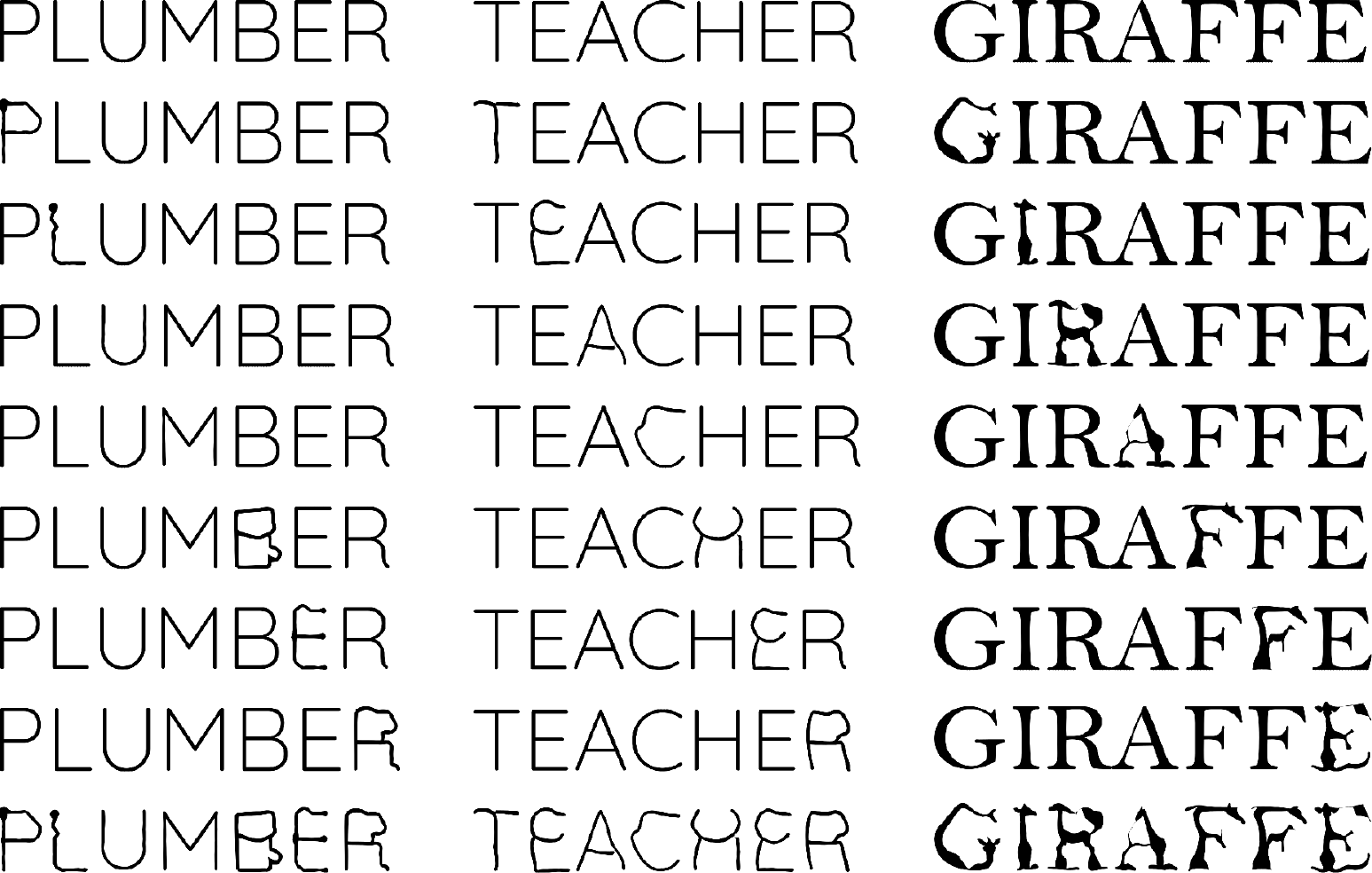} \\
 \caption{Word-as-image illustrations created by our method for randomly chosen words.}   
 \label{fig:all_res11} 
 \end{figure*}

 \begin{figure*}[ht] 
 \centering 
 \setlength{\tabcolsep}{12pt} 
 \renewcommand{\arraystretch}{3} 
	 \includegraphics[height=0.52\linewidth]{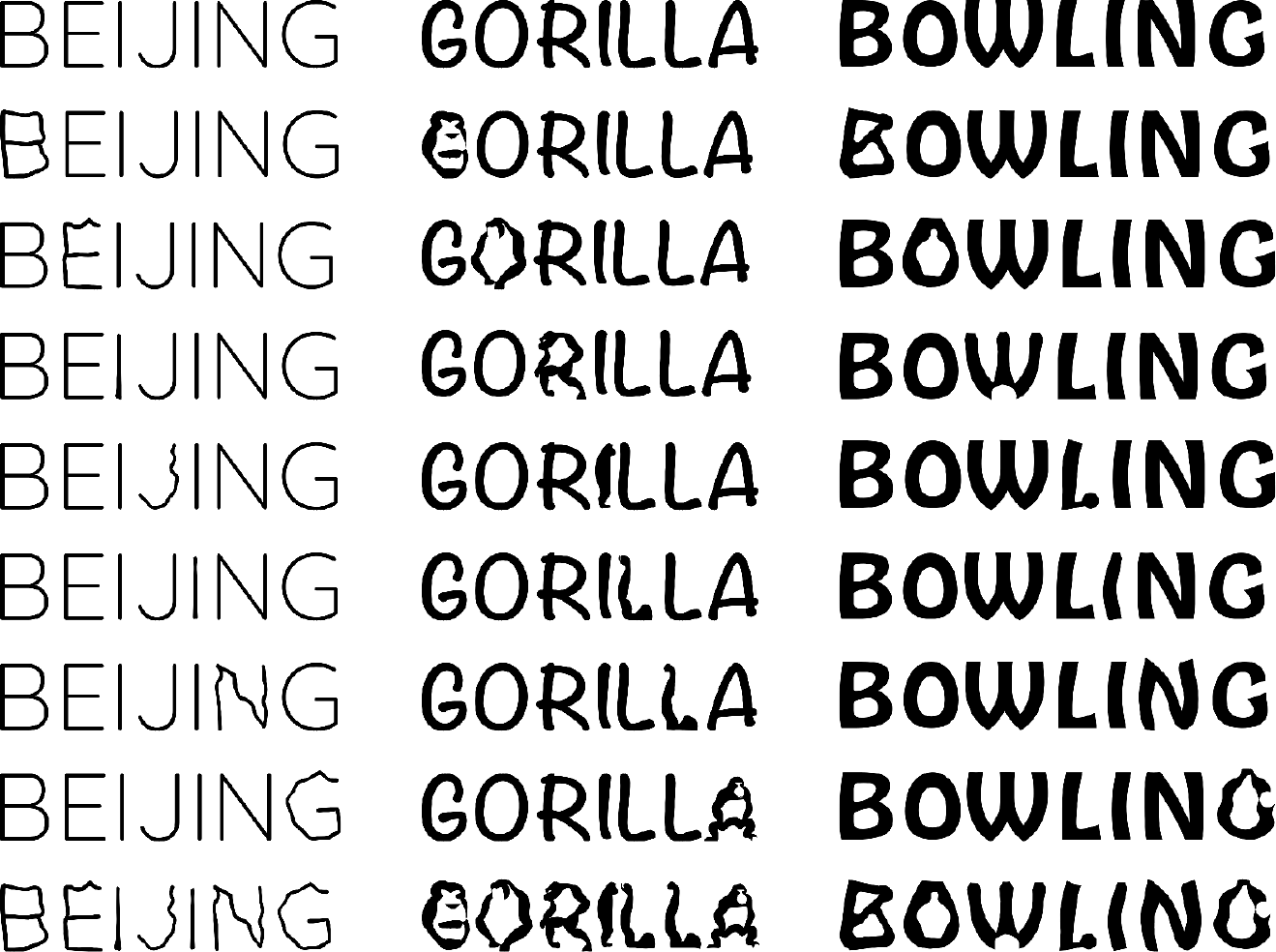} \\
 \caption{Word-as-image illustrations created by our method for randomly chosen words.}    
 \label{fig:all_res12} 
 \end{figure*}

 \begin{figure*}[ht] 
 \centering 
 \setlength{\tabcolsep}{12pt} 
 \renewcommand{\arraystretch}{3} 
	 \includegraphics[height=0.52\linewidth]{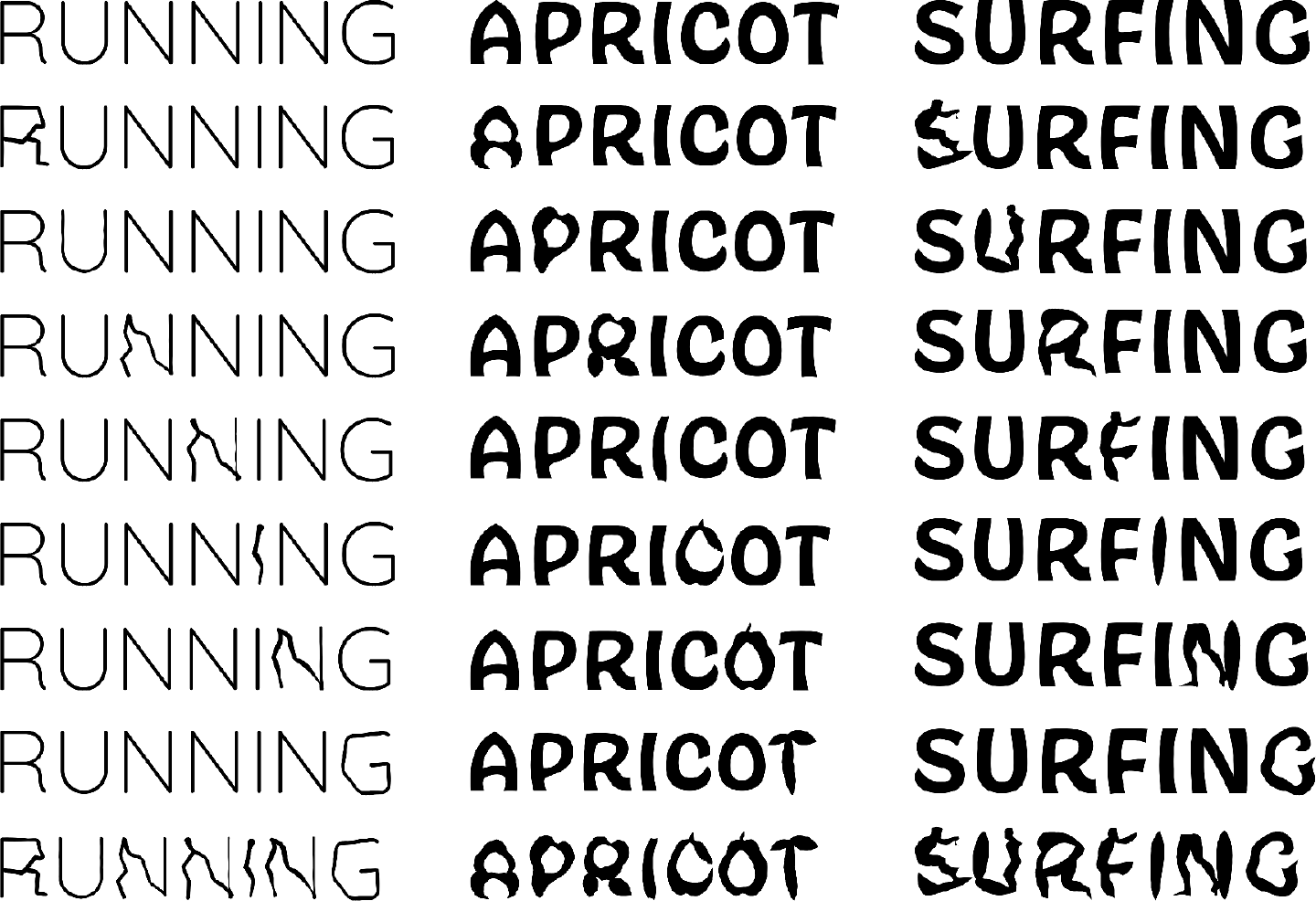} \\
 \caption{Word-as-image illustrations created by our method for randomly chosen words.}   
 \label{fig:all_res13} 
 \end{figure*}

 \begin{figure*}[ht] 
 \centering 
 \setlength{\tabcolsep}{12pt} 
 \renewcommand{\arraystretch}{3} 
	 \includegraphics[height=0.52\linewidth]{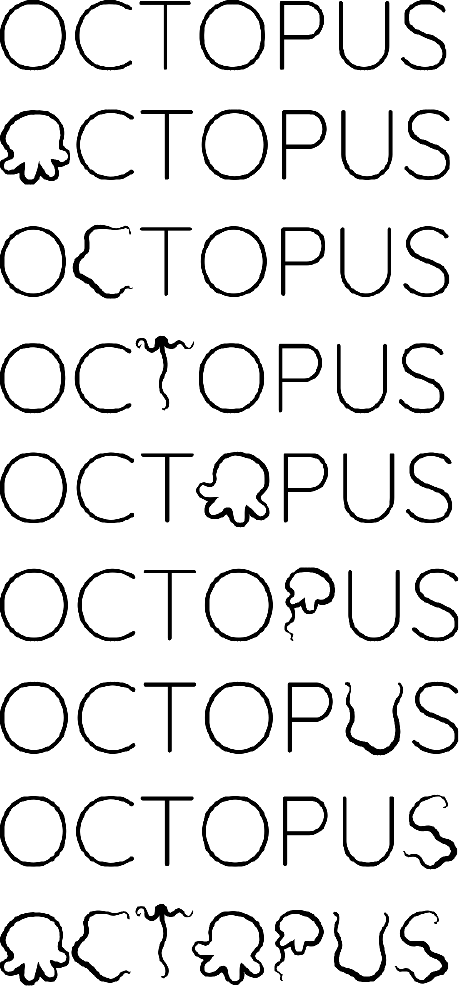} \\
 \caption{Word-as-image illustrations created by our method for randomly chosen words.}    
 \label{fig:all_res14} 
 \end{figure*}

 \begin{figure*}[ht] 
 \centering 
 \setlength{\tabcolsep}{12pt} 
 \renewcommand{\arraystretch}{3} 
	 \includegraphics[height=0.58\linewidth]{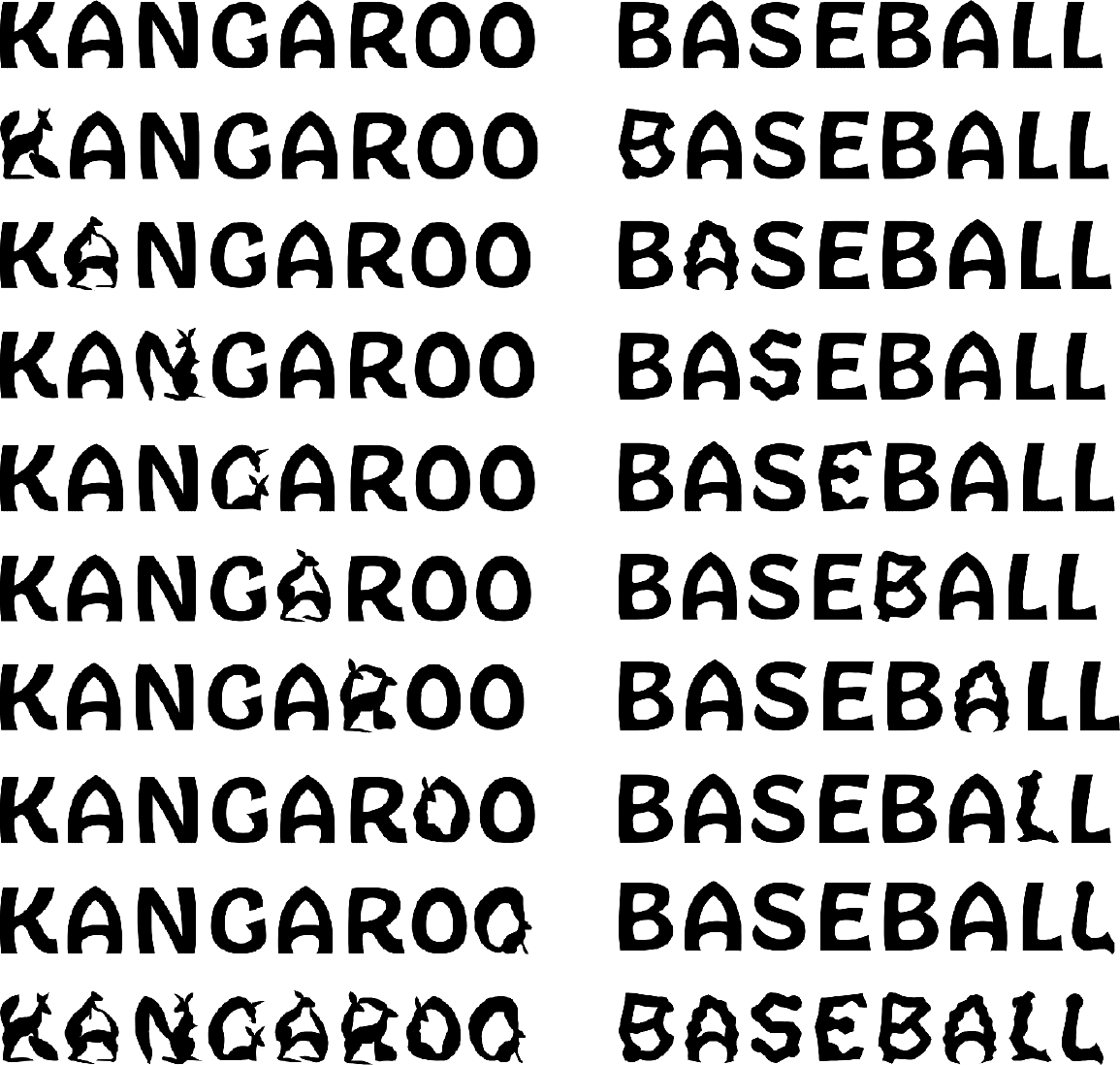} \\
 \caption{Word-as-image illustrations created by our method for randomly chosen words.}    
 \label{fig:all_res15} 
 \end{figure*}

\end{document}